\ifdefined\DoubleSided
  \documentclass[twoside,openright,a4paper, hidelinks]{nusthesis}
\else
  \documentclass[a4paper, hidelinks]{nusthesis}
\fi

\dsp 

\usepackage{indentfirst} 

\usepackage{silence} 
\usepackage{bookmark}

\usepackage{microtype} 
\usepackage{tabulary} 
\usepackage{hhline}

\usepackage{enumitem}

\usepackage{hyperref}

\renewcommand{\chaptermark}[1]{}

\usepackage{listings}

\usepackage{amsthm}
\theoremstyle{definition}
 
\theoremstyle{plain}

\usepackage{mathtools}

\DeclareMathOperator*{\argmax}{arg\,max}

\usepackage{interval}
\intervalconfig{soft open fences}

\usepackage{array}
\newcolumntype{L}[1]{>{\raggedright\let\newline\\\arraybackslash\hspace{0pt}}m{#1}}
\newcolumntype{C}[1]{>{\centering\let\newline\\\arraybackslash\hspace{0pt}}m{#1}}
\newcolumntype{R}[1]{>{\raggedleft\let\newline\\\arraybackslash\hspace{0pt}}m{#1}}

\usepackage{color}
\usepackage[usenames,dvipsnames]{xcolor}
\usepackage{soul}
\soulregister\cite7
\soulregister\ref7
\soulregister\pageref7
\soulregister\autoref7
\soulregister\eqref7

\renewcommand{\eqref}{Equation~\ref}

\newcommand*{\RootPicDir}{pic}

\usepackage{lipsum} 

\usepackage{natbib}

\usepackage{caption}
\usepackage{subcaption}
\usepackage{multicol}
\usepackage{multirow}
\usepackage{algorithm}
\usepackage{algpseudocode}
\newcommand{\sig}[0]{$^\ddagger$}
\usepackage{xcolor}
\usepackage{makecell}
\usepackage{booktabs}
\usepackage{siunitx}
\usepackage{color, colortbl}
\definecolor{Green}{rgb}{0.93,1,0.93}
\definecolor{Gray}{rgb}{0.85,0.85,0.85}
\definecolor{Gray}{gray}{0.9}
\definecolor{LightBlue}{rgb}{0.88, 0.94, 0.97}
\definecolor{Best}{rgb}{0.85, 0.92, 0.83}
\definecolor{HeaderBlue}{rgb}{0.88, 0.94, 0.97}
\definecolor{GrayRow}{gray}{0.96}
\definecolor{BestRow}{rgb}{0.85, 0.92, 0.83}

\usepackage{bbm}

\newtheorem{theorem}{Theorem}

\begin{document}

\title{Video Understanding: Through A Temporal Lens}

\author{NGUYEN THANH THONG}
\degree{Doctor of Philosophy}
\field{Institute of Data Science}
\degreeyear{2025}
\supervisor{Professor See-Kiong Ng}

\examiners{%
  Associate Professor Yao Yingjie Angela \\
  Assistant Professor Bryan Hooi Kuen-Yew}

\maketitle

\declaredate{02 June 2025}
\declaresign{\RootPicDir /signature.png} 







\newpage
    \begin{acknowledgments}
First and foremost, I am deeply indebted to my supervisor,  Prof. See-Kiong Ng, for his guidance, advice and encouragement, as well as giving me extraordinary experiences throughout my doctoral program. Above all and the most needed, he has provided me with unflinching encouragement and support in a wide variety of ways. Without his guidance, this thesis would not have been completed or written.

I am deeply grateful to Prof. Luu Anh Tuan, who is my external supervisor and collaborator ever since my undergraduate years. He has inspired me to take on this PhD journey, and I have continued to benefit from his advice during my PhD years. Thanks to his detailed comments and advice, my research journey has become extremely fruitful and enjoyable.

I also wish to express my gratitude to Prof. Angela Yao, Prof. Min-Yen Kan, and Prof. Bryan Hooi for their priceless advice. I thank them for their thorough comments and suggestions towards my research works and research direction. Their critical comments contribute to the backbone of this research. I also want to thank Prof. Bryan Low for his valuable advice towards my research career.

I would like to thank AI Singapore and Google Inc. for providing me with the scholarship to pursue this research, as well as supporting the finance for me to present papers in several overseas conferences.

Besides, I am thankful to my friends, colleagues, and collaborators. I am grateful for their friendship throughout my study, and I really enjoy my time with these brilliant people.

Finally, I would like to sincerely and deeply thank my parents, who have taken care of me with great love in these years, and my sister for all of her jokes. Most importantly, my heartfelt thanks go to my spouse. It is her constant support and encouragement that gives me strength to persevere throughout my research journey.

\end{acknowledgments}

    \tableofcontents


\newpage

\begin{abstract}

\noindent The fundamental difference between an image and a video is that a video introduces a temporal dimension. Such dimension triggers our awareness of temporal relations present among video elements, such as object positions, frames, and events. For example, when an object changes position, we perceive its positional changes as a movement. When we watch a video, we experience motion in the flow of frames. Additionally, one can recall the way we connect the events in a movie, seeking to construct a coherent narrative. These examples demonstrate that a video is indicative of temporal relations among its elements, requiring any video-understanding model to possess an awareness of temporality. Thus, in this thesis we aim to address the following question: \\

\noindent ``\textit{How to utilize the existence of temporal relations among video elements to advance video understanding?}''\\
 
\noindent First, video understanding performance has been constrained by training data of limited scale and quality, a bottleneck caused by the labor-intensiveness and expensive cost of obtaining high-quality annotated video data. To overcome this, we introduce an automatic annotation approach that harnesses the large vision-language model’s capacity to form a temporal progression among video frames to generate language annotations for videos. Because such automatically generated annotations may introduce noise to training, we propose a contrastive learning objective with a subtractive angular margin to regularize a model against perfect similarity between representations of a video and its language description. Our approach is theoretically guaranteed based on the intuition that a video and its caption often convey overlapping but not identical information.

Second, in resource-constrained scenarios, both manual and automatic approaches are unable to generate sufficient supervised training data for video understanding models. In these cases, fully updating model parameters can lead to overfitting and degraded performance. To address this issue, we employ a parameter-efficient training approach to finetune only a finite subset of modules, which we call adapters. We equip our adapters with recurrent computation layers to construct recurrent adapters that specifically adapt to modeling temporal relations among video frames. Furthermore, we integrate a partial video-language alignment objective into training, ensuring that our model effectively captures temporal semantics relevant to the task. Our approach demonstrates improved performance in multi-task scenarios with limited training data.

Third, existing works focus on short videos with independent or simple events. Most of them seek to infer temporal relations among coarsely sampled video elements, such as visual patches and video frames. However, for long videos, coarsely sampled elements hardly fully capture video content, while densely sampled elements will result in excessive computational cost. To cope with this issue, we introduce a state space layer (SSL) to our video understanding model to efficiently model temporal relations among densely sampled video elements for long-term video understanding. Moreover, recognizing that there is a lack of  benchmarks that authentically evaluate long-term video understanding, we introduce two new long-term video understanding benchmarks: one based on egocentric videos averaging 18 minutes and another based on movies spanning up to 2 hours. We carefully construct detailed prompts to instruct GPT-4 to generate highly difficult questions that require a model to simultaneously summarize, compare, and compress information across a long video. Experiments show that our method demonstrates superiority over self-attention and recurrent layers for long-term video understanding. 

Fourth, despite the existence of temporal relations among video elements, the attention to such relations still remains scarce compared to the attention to cross-modal relations between a video and its language description. To address this gap, we first propose a novel contrastive learning framework to drive a model to focus on relations between two motions in videos. Then, we extend our framework to capture relations between video moments. Extensive experiments indicate that our explicit temporal modeling approach significantly improves video understanding by making representations more aware of fine-grained temporal relations.

Finally, with the advent of large language models (LLMs) featuring outstanding reasoning capabilities, several works have sought to employ LLMs by constructing large vision-language models (LVLM) for video understanding. Unfortunately, rather than explicitly modeling temporal relationships in a video, they often rely on spatial inductive biases, assuming that spatial knowledge can smoothly extend to temporal comprehension. Such reliance has prevented modern video understanding models from fully utilizing their understanding capability. To overcome this limitation, we conduct a thorough empirical study to demystify crucial components that influence temporal understanding in LVLMs. Our empirical study reveals that the crucial component lies in the intermediate interface between the visual encoder and the LLM. Building on such insight, we propose a temporal-oriented recipe that encompasses temporal-oriented training scheme and an upscaled interface. Our final model developed using our recipe significantly enhances previous LVLMs across standard video understanding benchmarks.
\end{abstract}

\chapter*{Authorship Attribution Statement}

\noindent This thesis contains material from the following paper(s) published in the respective peer-reviewed conferences in which I am listed as an author. 

\noindent \textbf{Chapter \ref{ch0:literature_review}} is published with material from the following paper(s):
\begin{enumerate}
    \item \underline{Thong Nguyen}, Yi Bin, Junbin Xiao, Leigang Qu, Yicong Li, Jay Zhangjie Wu, Cong-Duy T Nguyen, See-Kiong Ng,  Luu Anh Tuan. \emph{Video-Language Understanding: A Survey from Model Architecture, Model Training, and Data Perspectives}. Annual Meeting of the Association for Computational Linguistics (\textbf{ACL}), 2024 (Findings)
\end{enumerate}

\noindent \textbf{Chapter \ref{ch1:mama}} is published with material from:
\begin{enumerate}
    \item \underline{Thong Nguyen}, Yi Bin, Xiaobao Wu, Xinshuai Dong, Zhiyuan Hu, Khoi Le, Cong-Duy Nguyen, See-Kiong Ng, Luu Anh Tuan. \emph{MAMA: A Meta-optimized Angular Margin Contrastive Framework for Video-Language Representation Learning}. The European Conference on Computer Vision (\textbf{ECCV}), 2024 
\end{enumerate}

\noindent \textbf{Chapter \ref{ch2:read}} is published with material from:
\begin{enumerate}
   \item \underline{Thong Nguyen}, Xiaobao Wu, Xinshuai Dong, Khoi Le, Zhiyuan Hu, Cong-Duy Nguyen, See-Kiong Ng, Luu Anh Tuan. \emph{READ: Recurrent Adapter with Partial Video-Language Alignment for Parameter-Efficient Transfer Learning in Low-Resource Video-Language Modeling}. Proceedings of AAAI Conference on Artificial Intelligence (\textbf{AAAI}), 2024
\end{enumerate}

\noindent \textbf{Chapter \ref{ch3:global}} is published with material from:
\begin{enumerate}
    \item \underline{Thong Nguyen}, Zhiyuan Hu, Xiaobao Wu, Cong-Duy Nguyen, See-Kiong Ng, Luu Anh Tuan. \emph{Encoding and Controlling Global Semantics for Long-form Video Question Answering}. Proceedings of Empirical Methods in Natural Language Processing (\textbf{EMNLP}), 2024
\end{enumerate}

\noindent \textbf{Chapter \ref{ch4:motionpvsg}} is published with material from:
\begin{enumerate}
    \item \underline{Thong Nguyen}, Xiaobao Wu, Yi Bin, Cong-Duy Nguyen, See-Kiong Ng, Luu Anh Tuan. \emph{Motion-aware Contrastive Learning for Temporal Panoptic Scene Graph Generation}. Proceedings of AAAI Conference on Artificial Intelligence (\textbf{AAAI}), 2025
\end{enumerate}

\noindent \textbf{Chapter \ref{ch5:mstg}} is published with material from:
\begin{enumerate}
    \item \underline{Thong Nguyen}, Yi Bin, Xiaobao Wu, Zhiyuan Hu, Cong-Duy Nguyen, See-Kiong Ng, Luu Anh Tuan. \emph{Multi-Scale Contrastive Learning for Video Temporal Grounding}. Proceedings of AAAI Conference on Artificial Intelligence (\textbf{AAAI}), 2025
\end{enumerate}

\noindent \textbf{Chapter \ref{ch6:recipe}} is published with material from:
\begin{enumerate}
    \item \underline{Thong Nguyen}, Zhiyuan Hu, Xu Lin, Cong-Duy Nguyen, See-Kiong Ng, Luu Anh Tuan. \emph{Temporal-Oriented Recipe for Transferring Large Vision-Language Model to Video Understanding}. In submission to Conference on Neural Information Processing (\textbf{NeurIPS}), 2025
\end{enumerate}

\noindent Other publications during my PhD study:
\begin{enumerate}
    \item Cong-Duy Nguyen, Xiaobao Wu, \underline{Thong Nguyen}, Shuai Zhao, Khoi Le, Viet-Anh Nguyen, Feng Yichao, Luu Anh Tuan. \emph{Enhancing Multimodal Entity Linking with Jaccard Distance-based Conditional Contrastive Learning and Contextual Visual Augmentation}. Proceedings of Annual Conference of the North American Chapter of the Association for Computational Linguistics (\textbf{NAACL}), 2025
    
    \item \underline{Thong Nguyen}, Xiaobao Wu, Xinshuai Dong, Cong-Duy T Nguyen, See-Kiong Ng, Luu Anh Tuan. \emph{Topic Modeling as Multi-Objective Contrastive Optimization}. Proceedings of International Conference on Learning Representations (\textbf{ICLR}), 2024
    
    \item Xiaobao Wu, Fengjun Pan, \underline{Thong Nguyen}, Yichao Feng, Chaoqun Liu, Cong-Duy Nguyen, Luu Anh Tuan. \emph{On the affinity, rationality, and diversity of hierarchical topic modeling}. Proceedings of AAAI Conference on Artificial Intelligence (\textbf{AAAI}), 2024

    \item Xiaobao Wu, \underline{Thong Nguyen}, Luu Anh Tuan. \emph{A survey on neural topic models: methods, applications, and challenges}. Artificial Intelligence Review, 2024
    
    \item Xiaobao Wu, \underline{Thong Nguyen}, Delvin Ce Zhang, William Yang Wang, Luu Anh Tuan. \emph{FASTopic: A Fast, Adaptive, Stable, and Transferable Topic Modeling Paradigm}. Proceedings of Advances in Neural Information Processing Systems (\textbf{NeurIPS}), 2024

    \item Cong-Duy Nguyen, \underline{Thong Nguyen}, Duc Anh Vu, Luu Anh Tuan. \emph{KDMCSE: Knowledge distillation multimodal sentence embeddings with adaptive angular margin contrastive learning}. Proceedings of Annual Conference of the North American Chapter of the Association for Computational Linguistics (\textbf{NAACL}), 2024

    \item \underline{Thong Nguyen}, Xiaobao Wu, Xinshuai Dong, Cong-Duy Nguyen, See-Kiong Ng, Luu Anh Tuan. \emph{DemaFormer: Damped Exponential Moving Average Transformer with Energy-Based Modeling for Temporal Language Grounding}. Proceedings of Empirical Methods in Natural Language Processing (\textbf{EMNLP}), 2023 (Findings)

    \item Cong-Duy Nguyen, \underline{Thong Nguyen}, Anh Vu, Luu Anh Tuan. \emph{Improving multimodal sentiment analysis: Supervised angular margin-based contrastive learning for enhanced fusion representation}. Proceedings of Empirical Methods in Natural Language Processing (\textbf{EMNLP}), 2023 (Findings)

    \item Xiaobao Wu, Xinshuai Dong, \underline{Thong Nguyen}, Luu Anh Tuan. \emph{Effective Neural Topic Modeling with Embedding Clustering Regularization}. Proceedings of International Conference on Machine Learning (\textbf{ICML}), 2023

    \item \underline{Thong Nguyen}, Xiaobao Wu, Xinshuai Dong, Luu Anh Tuan, Cong-Duy Nguyen, Zhen Hai, Lidong Bing. \emph{Gradient-Boosted Decision Tree for Listwise Context Model in Multimodal Review Helpfulness Prediction}. Annual Meeting of the Association for Computational Linguistics (\textbf{ACL}), 2023 (Findings)

    \item Xiaobao Wu, Xinshuai Dong, \underline{Thong Nguyen}, Chaoqun Liu, Liangming Pan, Luu Anh Tuan. \emph{InfoCTM: A Mutual Information Maximization Perspective of Cross-lingual Topic Modeling}. Proceedings of AAAI Conference on Artificial Intelligence (\textbf{AAAI}), 2023 

    \item \underline{Thong Nguyen}, Xiaobao Wu, Tuan Luu, Cong-Duy Nguyen, Zhen Hai, Lidong Bing. \emph{Adaptive Contrastive Learning on Multimodal Transformer for Review Helpfulness Predictions}. Proceedings of Empirical Methods in Natural Language Processing (\textbf{EMNLP}), 2022
\end{enumerate}
\chapter{Introduction}
\label{chapter:introduction}
\bigskip
\bigskip
\noindent Video data is among the richest and most rapidly growing forms of digital information in the world. Every day, millions of videos, spanning hundreds of thousands of hours, are uploaded to online platforms. This steadily increasing data treasure is highly diverse, reflecting real-world complexity not only through static visual cues but also through temporal dynamics. There are two fundamental video types, \textit{i.e.} natural footage and edited videos. Natural footage captures events as they unfold, whereas edited videos are shaped by human interventions such as cuts, splits, and transitions. 

In this work, we aim to develop artificial intelligence capable of understanding video content by constructing video understanding models. Different from image understanding, which primarily captures static visual information, video understanding must model both spatial and temporal aspects to determine what happens, when it happens, and how events evolve over time.

Both natural and edited video understanding have extensive applications. For instance, analyzing natural footage enables tasks such as temporal grounding to localize video moments of anomalous activities, such as arson, shoplifting, and fighting. Similarly, video summarization can distill long meeting recordings into concise highlights to support business decision-making. Edited video understanding also offers practical applications. For example, text-video retrieval models can help users locate relevant movies based on queries, enhancing entertainment experiences.

Unlike images, videos inherently encompass a temporal dimension, marking events as they unfold. Nevertheless, this temporal aspect is difficult to represent explicitly. To robustly capture this temporal aspect of videos, we develop strong temporal understanding capabilities for video understanding models. Humans naturally perceive temporal context, such as when we see a friend waving, we infer their movement; when a ball is suspended mid-air, we understand that it is moving. This awareness of change, sequence, and continuity stems from our lifelong exposure to dynamic events, and arguably from evolutionary experiences encoded in our genes. In contrast, instilling such temporal awareness in neural models is computationally demanding. Videos consist of numerous high-resolution frames, and processing hundreds of them often already exceeds the memory capacity of modern GPUs. Moreover, beyond frame-level analysis, effectively extracting and leveraging meaningful temporal cues remains a major challenge. As a result, video understanding has not reached the same level of success as achieved in image or language understanding, primarily due to the complexity of modeling temporal dynamics. 

Therefore, this thesis argues that explicitly modeling temporal signals for video understanding can substantially enhance video understanding. In particular, we concentrate on relations that only emerge when time is considered, \textit{e.g.} relations between object motions or relations between video moments, and investigate techniques to capture these temporal cues across three dimensions, \textit{i.e.} data, model, and training. While the datasets used in this dissertation are predominantly edited videos, our methods are designed to generalize to both natural footage and edited videos. We believe that tailoring approaches to either natural or edited videos offers a promising future research avenue.

The following chapters of this thesis delve into the specifics of our temporal-oriented approach, demonstrate its advantages, and discuss its implications for advancing the field of video understanding.
\section{Challenges of Video Understanding}
\noindent\textbf{Performance Limits In Video Understanding.} Prior research has established scaling laws in deep learning, showing that performance tends to improve as both model size and training data increase \citep{hestness2017deep, kaplan2020scaling}. This implies that, regardless of advancements in architecture or training techniques, meaningful performance gains will eventually stall without a corresponding increase in scale. Video understanding faces unique challenges in this regard: a single video can comprise hundreds or thousands of high-resolution frames, making processing computationally expensive. Moreover, videos demand significantly more storage than images due to their multi-frame nature. As a result, video models and datasets have lagged behind those in image and language domains in terms of scale. Building on established scaling laws, this thesis aims to push the boundaries of video understanding by proposing strategies to scale models and training datasets to enable the effective capture of implicit temporal relations among video elements for video understanding.

\noindent\textbf{Inefficient capture of temporal relations in video understanding models.} Most video understanding architectures leverage self-attention-based modules originally designed for language understanding tasks. These modules require large volumes of training samples to capture useful inductive biases required for accurate semantic interpretation of text or video. Prior work has shown inferior performance under data scarcity \citep{dosovitskiy2020image}. Unfortunately, as mentioned, video datasets are generally far smaller than their image or language counterparts---a gap that widens in privacy-sensitive or long-tail scenarios. As a consequence, we must develop efficient techniques that enable models to quickly extract useful cues from limited training data.

\noindent\textbf{Ineffective capture of temporal relations in video understanding models.} A key challenge addressed in this thesis is the under-utilization of temporal cues in video understanding. Since many prior works focus on static frame-level features, they overlook deeper semantics that arise from temporal phenomena such as motion continuity, event formation, and long-range temporal dependencies. These elements derive their semantics only when we consider core temporal properties: change (\textit{e.g.} detecting motion through variation across object masks), duration (\textit{e.g.} identifying the span of an event), and succession (\textit{e.g.} understanding how moments build into narratives). For instance, when an object shifts position across time---such as hand waving---we instinctively infer the motion, not just a sequence of static poses. Similarly, we interpret a stream of frames not as isolated images, but as a cohesive event, often organizing them into structured stories that unfold over time. In this thesis, we focus on emphasizing these semantics in representations generated by video understanding models by forming internal relations for motions, events, and long-term context. 

\section{Research Objectives}
\noindent Given the challenges, we seek to advance video understanding by addressing them. As such, we tackle the following objectives which respectively correspond to the aforementioned challenges. 

\begin{itemize}
    \item \textbf{Research Objective 1 (RO1): Pushing the Frontier of Video Understanding.} Video understanding remains constrained by both architecture and dataset scale. We will expand these dimensions by (a) scaling up the model components most critical for temporal understanding, and (b) introducing an automatic annotation approach that automatically annotates a video by temporally linking sampled frames into a coherent video caption. 
    
    \item \textbf{Research Objective 2 (RO2): Efficient Capture of Temporal Relations in Video Understanding Models.} In certain scenarios---such as privacy-sensitive environments or settings with limited computational and storage resources---acquiring large-scale video training data, whether supervised or non-supervised, is often infeasible. In such cases, models may lack access to rich temporal cues, particularly if their architectures do not inherently target video data. Furthermore, these low-resource contexts might also demand high adaptability, requiring support for multiple downstream tasks such as video summarization or temporal grounding. Therefore, beyond simply improving effectiveness, we also design modular components or training strategies that facilitate efficient temporal modeling under resource constraints. Our proposed solution offers a lightweight, task-adaptable framework that maintains strong performance across diverse video understanding tasks while minimizing training and storage costs. 

    \item \textbf{Research Objective 3 (RO3): Effective Capture of Temporal Relations in Video Understanding Models.} We seek to integrate our inductive biases that emphasize temporal semantics in videos. These biases reflect common temporal relations among video elements: \textit{(i) Relation between Motions:} When an object changes positions over time, it exhibits motion. Actions belonging to the same semantic category often share similar motion patterns. We encourage the model to learn representations that are close in feature space for actions of the same type, thereby reinforcing motion-level consistency. \textit{(ii) Relation between Events:} As frames progress sequentially, temporal properties such as duration, succession, and order emerge, forming a coherent event. To encode the content of an event, models generate compact representations for temporally ordered frame sequences that capture event-level temporal dynamics.  In our work, we encourage the model to map events of similar semantics into similar embeddings, while separating dissimilar events. \textit{(iii) Long-Term Context:} Videos typically consist of temporally linked that form a coherent narrative or context. In longer videos, such as movies, modeling this long-term temporal context poses both computational and representational challenges. We enhance the capacity of video models to capture and utilize long-range temporal dependencies, and we validate our approach on datasets featuring extended video content.

\end{itemize}

\section{Contributions and Outline of This Thesis}
\begin{figure*}[h!]
    \centering
    \includegraphics[width=\linewidth]{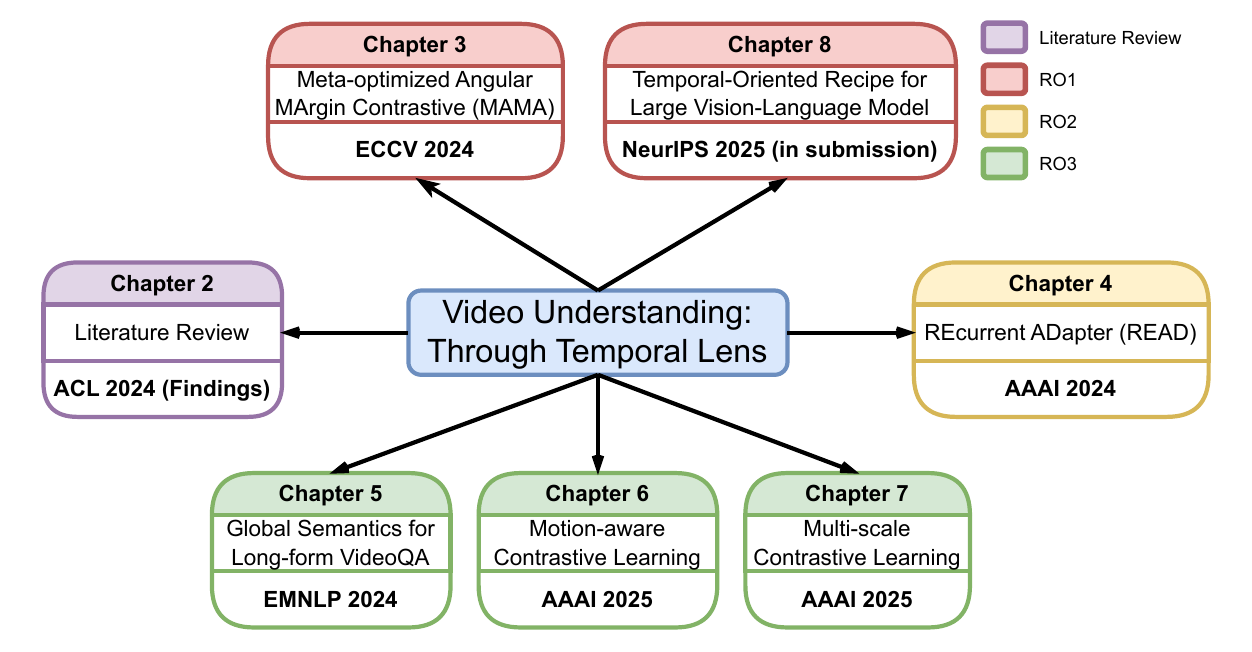}
    \caption{Outline of the thesis.}
    \label{fig:thesis_structure}
\end{figure*}

\noindent In this section, we describe our contributions of this thesis with the according research objectives. For each contribution, we provide reference to the respective chapter in this thesis. The outline of the thesis is illustrated in Figure \ref{fig:thesis_structure}.

\begin{itemize}
    \item \textbf{Contribution 1 - RO1 - Chapter \ref{ch1:mama}: Meta-optimized Angular MArgin (MAMA) contrastive framework} for pushing the limit of video understanding through escalating training data. Particularly, MAMA utilizes a pretrained large vision-language model to annotate a vast set of video data. Because automatically annotated data can contain noise, MAMA uses regularization training which consist of a subtractive angular margin integrated in the contrastive objective along with a multi-layer perceptron (MLP) to dynamically adjust model's attention to training samples. This can alleviate the noisy effect of synthetic data upon video understanding model training.
    
     \item \textbf{Contribution 2 - RO2 - Chapter \ref{ch2:read}: REcurrent ADapter (READ)} for efficient capture of temporal relations in low-resource video understanding. In low-resource scenarios, there exists limited video data to provide sufficient temporal cues for video understanding models to learn from. To remedy this problem, we incorporate recurrent computation into low-dimensional adapters to encourage temporal modeling capability. We further implement a video-text alignment training objective to motivate the model to focus upon task-related information during video encoding. 
 
     \item \textbf{Contribution 3 - RO3 - Chapter \ref{ch3:global}: Encoding Global Video Semantics} for effective capture of temporal long-term context in long videos. Prior works cope with excessive computational cost of long-form video understanding by selecting a small number of frames for video understanding model, which leads to significant information loss. We mitigate this problem by incorporating SSLs into model architectures to encode all densely sampled frames which can absorb video's global semantics with reduced complexity. We target our SSL-augmented architecture for long-form VideoQA, and validate its superiority on our two collected benchmarks involving movies and egocentric videos.
     
    \item \textbf{Contribution 3 - RO3 - Chapter \ref{ch4:motionpvsg}: Motion-aware Contrastive Learning (MCL)} for effective capture of  relations between object motions. MCL employs a novel contrastive training objective to pull closer representations for pairs of object masks possess similar actions. We explicitly target dynamic actions such as ``\textit{run}'', ``\textit{walk}'', and ``\textit{throw}'' with strong motion cues. Experiments demonstrate the effectiveness of our method for video panoptic scene graph generation, a comprehensive video understanding task whose aim is to segment a video into object masks and predict their relations.
    
     \item \textbf{Contribution 4 - RO3 - Chapter \ref{ch5:mstg}: Multi-Scale Contrastive Learning (MSCL)} for effective capture of relations between video moments. Our MSCL iteratively learns close representations for short-range and long-range video moments that are commonly described by an event description. By modeling the relations between short-range and long-range moments, we are able to deconfound the effect of surface properties of video elements, \textit{e.g.} video moment length. Moreover, we can adopt a query-based sampling strategy for our framework to significantly reduce computational and memory cost, which are significant burdens of previous contrastive learning frameworks.
     
    \item \textbf{Contribution 5 - RO1 - Chapter \ref{ch6:recipe}: Temporal-Oriented Recipe for Transferring Large Vision-Language Model to Video Understanding} for pushing the limit of video understanding through advancing model architecture. With the advent of large language models (LLMs), a popular research avenue is to design LLM-oriented systems to inherit LLM's impressive reasoning capability for video understanding. In our work, to further push the frontier of these systems, we first conduct extensive experiments to demystify essential components that significantly affect video understanding performance. As we discover that the vision-language interface is the most influential part, we devise a recipe centered on the interface. In our recipe, we progressively choose Q-Former as the vision-language interface, add temporal-oriented training schemes, incorporate temporal memory bank to store video representations, and augment mixture-of-experts (MoE) into the interface. These steps result in notable improvement for large vision-language models, as demonstrated by their strong performance on standard video understanding benchmarks.
    \end{itemize}

\chapter{Literature Review}
\label{ch0:literature_review}

\section{Video Understanding Tasks}

\begin{figure*}[h!]
    \centering
    \includegraphics[width=\linewidth]{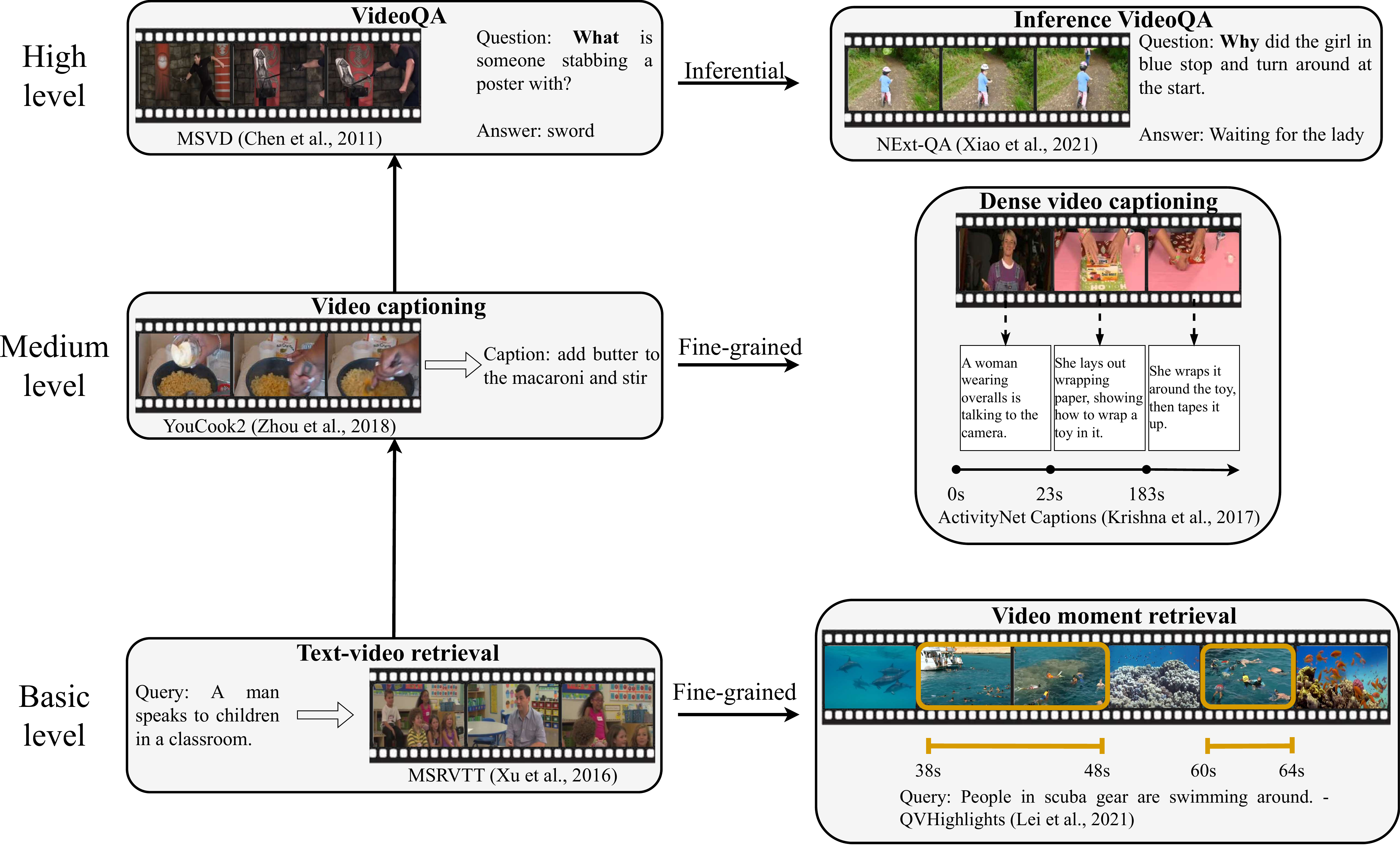}
    \caption{Level hierarchy of video understanding tasks.}
    \label{fig:task_hierarchy}
\end{figure*}

\noindent\textbf{Text-video retrieval.} Text-video retrieval is the task to search for the corresponding video given a language query (text-to-video), or oppositely search for the language description given a video (video-to-text).

In practical applications, returning an entire video may not be desirable. Hence, video moment retrieval (VMR) has emerged with an aim to accurately locating relevant moments within a video based on user queries. VMR examines more nuanced and fine-grained understanding to capture different concepts and events in a video in order to pinpoint specific moments rather than capturing the overall theme in standard text-video retrieval.

\noindent\textbf{Video captioning.}
Video captioning is the task to generate a concise language description for a video. A video captioning model receives as input a video and optionally a language transcript transcribed from the audio in the video. Typically, a model produces a sentence-level caption for the whole video, or might also generate a paragraph as a more detailed summary.

\noindent\textbf{Video question answering (videoQA). }Video question answering is the task to predict the correct answer based on a question $q$ and a video $v$. There are two fundamental types of VideoQA, \textit{i.e.} \textbf{multi-choice} VideoQA and \textbf{open-ended} VideoQA. In multi-choice VideoQA, a model is presented with a certain number of candidate answers and it will choose the correct answer among them. Open-ended VideoQA can be formulated as a classification problem, a generation problem, or a regression problem. Classification-based VideoQA associates a video-question pair with an answer from a pre-defined vocabulary set. Generation-based VideoQA is not restricted to a vocabulary set, in which a model can generate a sequence of tokens that represent the answer to a question. Regression-based VideoQA is often used for counting questions, \textit{e.g.} counting the repetitions of an action or counting the number of an object in a video.  

\begin{figure*}[h!]
    \centering
    \begin{subfigure}[t]{0.3\linewidth}
        \centering
        \includegraphics[width=\linewidth]{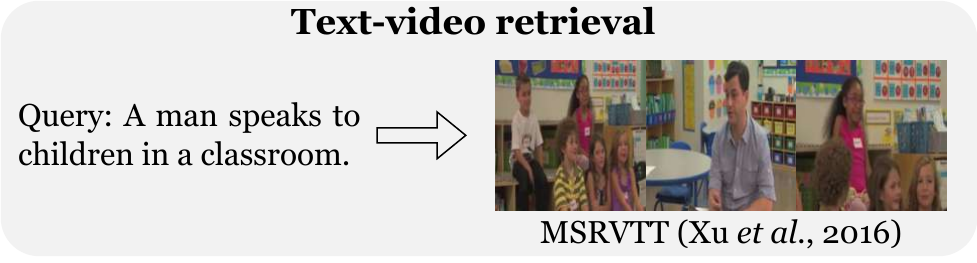}
    \end{subfigure}
    \begin{subfigure}[t]{0.3\linewidth}
        \centering
        \includegraphics[width=\linewidth]{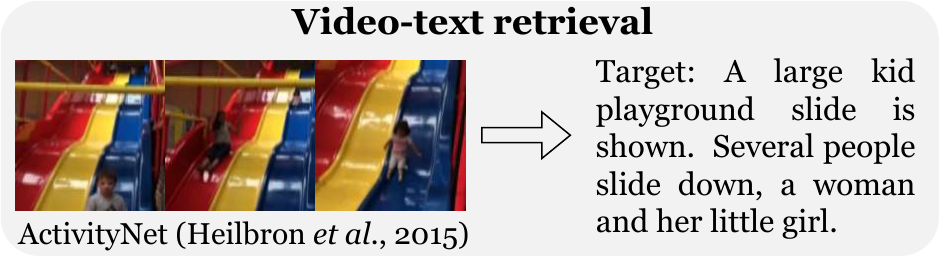}
    \end{subfigure}
    \begin{subfigure}[t]{0.3\linewidth}
        \centering
        \includegraphics[width=\linewidth]{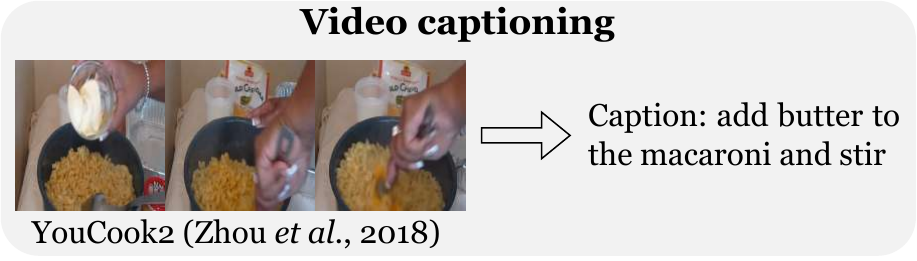}
    \end{subfigure}\\
    \begin{subfigure}[t]{0.3\linewidth}
        \centering
        \includegraphics[width=\linewidth]{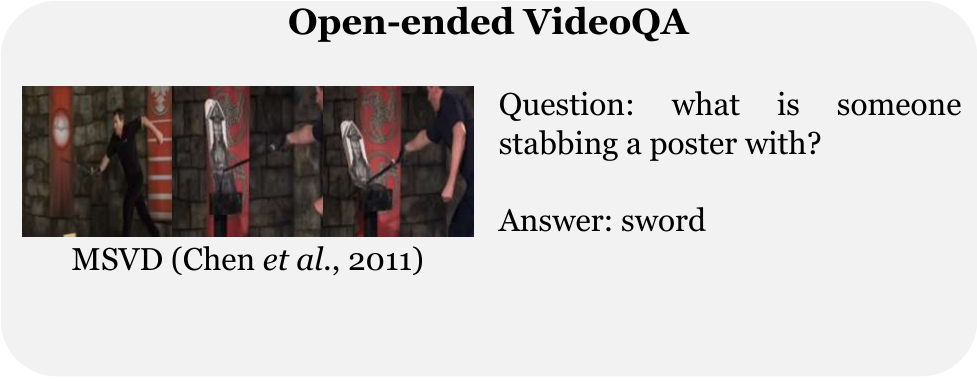}
    \end{subfigure}
    \begin{subfigure}[t]{0.3\linewidth}
        \centering
        \includegraphics[width=\linewidth]{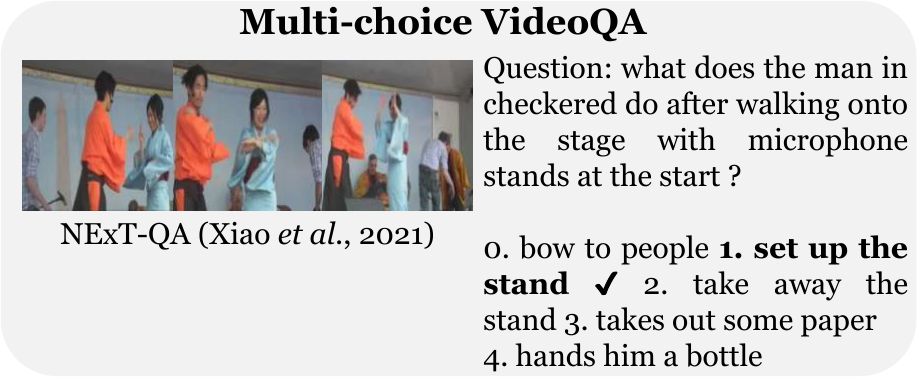}
    \end{subfigure}
    \begin{subfigure}[t]{0.3\linewidth}
        \centering
        \includegraphics[width=\linewidth]{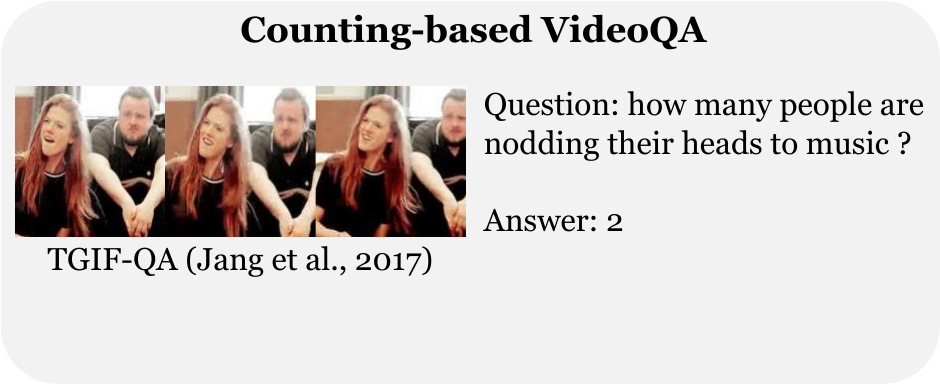}
    \end{subfigure}
    \caption{Illustration of text-video retrieval, video captioning, and video question answer (videoQA) tasks. 
    }
    \label{fig:illustration_video_language_understanding_tasks}
\end{figure*}

\begin{figure*}[h!]
    \centering
    \begin{subfigure}[t]{0.32\linewidth}
        \centering
        \includegraphics[width=\linewidth]{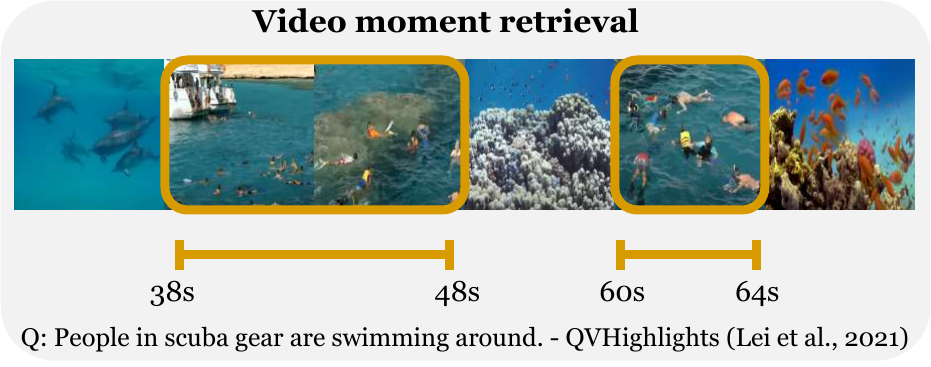}
    \end{subfigure}
    \begin{subfigure}[t]{0.29\linewidth}
        \centering
        \includegraphics[width=\linewidth]{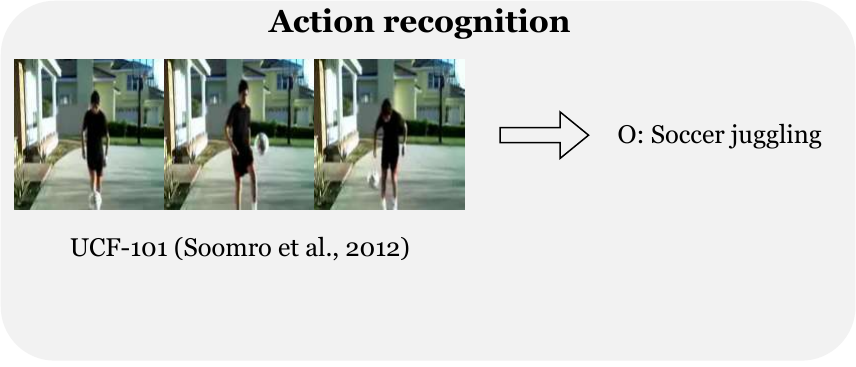}
    \end{subfigure}
    \begin{subfigure}[t]{0.32\linewidth}
        \centering
        \includegraphics[width=\linewidth]{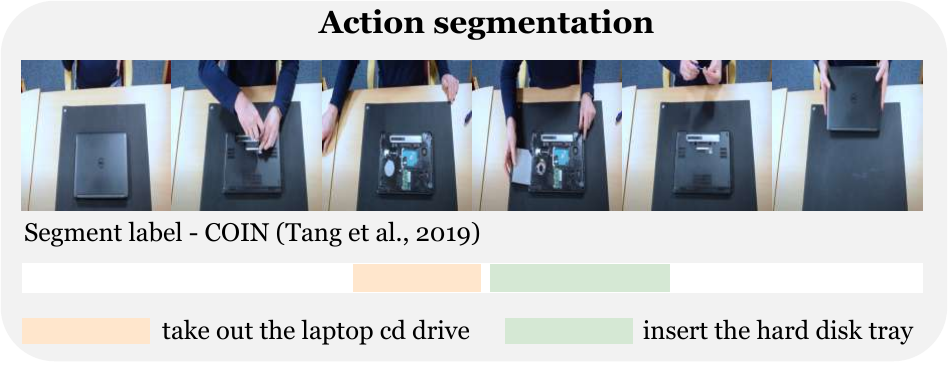}
    \end{subfigure} \\
    \begin{subfigure}[t]{0.17\linewidth}
        \centering
        \includegraphics[width=\linewidth]{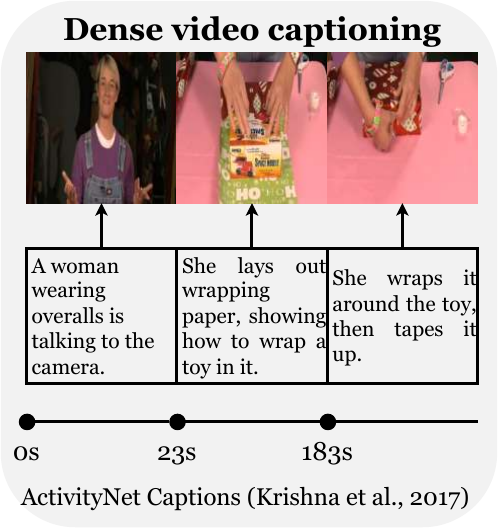}
    \end{subfigure}
    \begin{subfigure}[t]{0.33\linewidth}
        \centering
        \includegraphics[width=\linewidth]{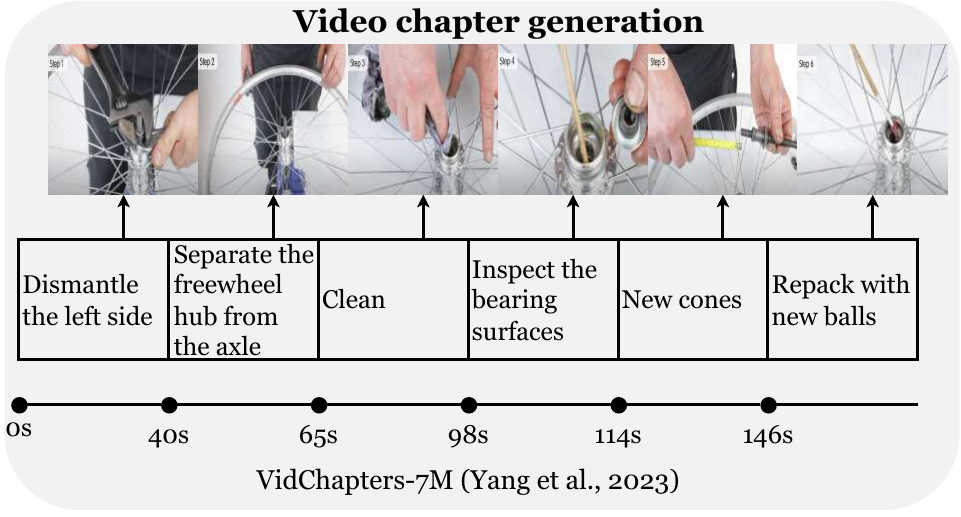}
    \end{subfigure}
    \begin{subfigure}[t]{0.45\linewidth}
        \centering
        \includegraphics[width=\linewidth]{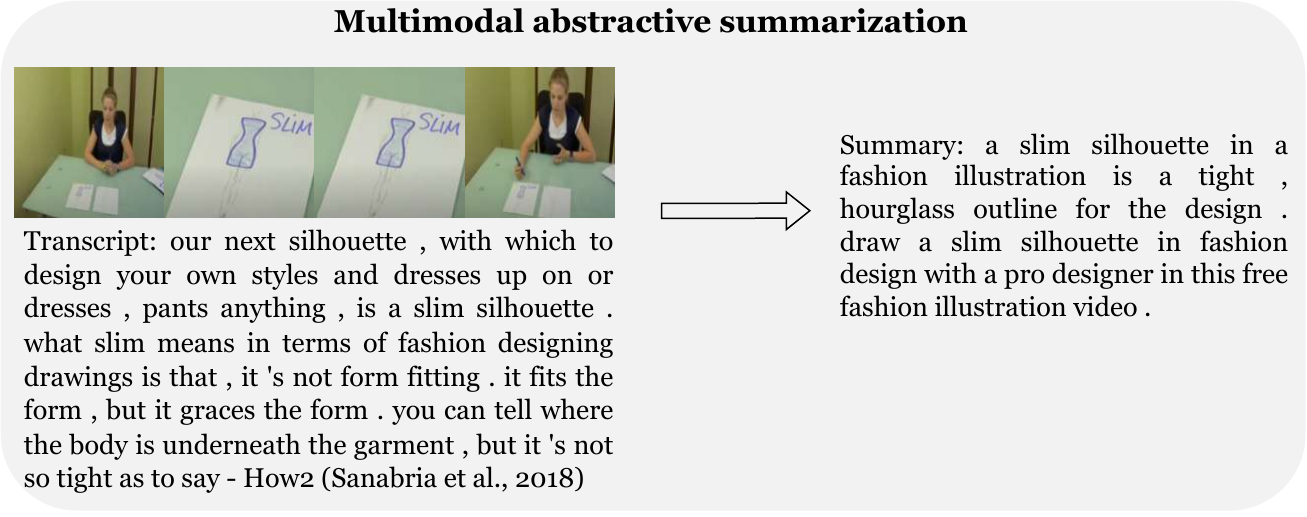}
    \end{subfigure}
    \caption{More illustration of video moment retrieval, action recognition, action segmentation, dense video captioning, video chapter generation, and multimodal abstractive summarization tasks.}
    \label{fig:more_illustration_video_language_understanding_tasks}
\end{figure*}

\noindent\textbf{Connections among video understanding tasks.}
These tasks constitute the three fundamental testbeds for evaluating video understanding (see Figures \ref{fig:illustration_video_language_understanding_tasks} and \ref{fig:more_illustration_video_language_understanding_tasks} for their examples). Despite their operational differences, they can be unified under a common framework. Formally, let $V=\{f_1, f_2, \dots, f_N\}$ denote a video with $N$ frames, and let $T$ denote a textual input, such as a query, question, or prompt. A video-language model $M$ learns a mapping from the video $V$ and text input $T$ into an output $Y$, i.e. $f: (V, T) \rightarrow Y$. In text-video retrieval, $Y$ is a relevance score between $V$ and $T$. In video captioning, $Y$ is a sequence of words describing $V$. In VideoQA, $Y$ is the predicted answer conditioned on $V$ and $T$. This unified formulation highlights that video understanding tasks share a foundation of joint video-language representation learning.

These video understanding tasks can also be organized into a hierarchy of understanding levels, as illustrated in Figure \ref{fig:task_hierarchy}. At the basic level, text-video retrieval globally aligns a video with its textual counterpart. At the intermediate level, video captioning requires mapping salient entities and events in a video to a coherent language description, make it a more difficult task than text-video retrieval. At the highest level, VideoQA demands reasoning over both modalities to generate accurate answers. Each level further extends to more fine-grained or inferential variants. In particular, video moment retrieval, or temporal grounding, extends text-video retrieval; dense video captioning \citep{zhou2018end} and video chapter generation \citep{yang2023vid2seq} extend capitoning; and inference-focused VideoQA \citep{xiao2021next, li2022representation} extends VideoQA. These advanced tasks pose greater challenges and are increasingly central to advancing video research towards human-level intelligence \citep{fei2022searching}.

\begin{figure*}[t]
    \centering
    \includegraphics[width=\linewidth]{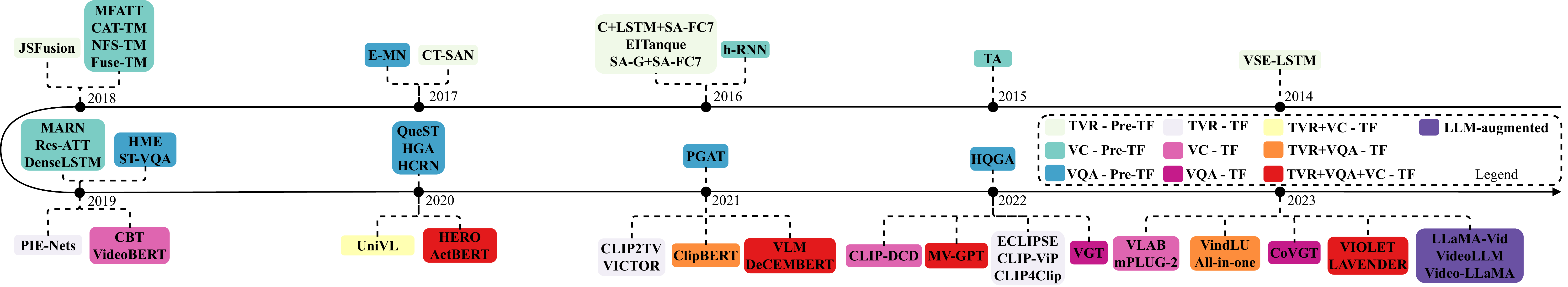}
    \caption{Timeline of the established video understanding methods (TVR: Text-video retrieval, VC: video captioning, VQA: video question answering, TF: Transformer, LLM: large language model). From left to right, our legend table follows the order: pre-Transformer (Pre-TF), task-specific Transformer, multi-task Transformer, and LLM-augmented architectures.}
    \label{fig:timeline}
\end{figure*}

\section{Challenges of Video Understanding}
\noindent The discussed video understanding tasks present unique challenges compared with image-language understanding, since a video incorporates an additional temporal channel. We summarize their important challenges as follows:

\noindent\textbf{Intra-modal and cross-modal interaction.} While intra-modal interaction modeling within language can be directly taken from image-language understanding, intra-modal interaction modeling within video is different since it jointly consists of spatial interaction and temporal interaction. Spatial interaction delves into the relationships among pixels, patches, regions, or objects within an individual frame, whereas temporal interaction captures sequential dependencies among video frames or video segments. Longer video durations amplify the complexity of temporal modeling by necessitating the recognition of more objects and events in a higher number of video frames \citep{yu2020long, lin2022eclipse}, and reasoning their long-term dependencies \cite{zhao2018open}. Particular video domains, such as egocentric videos, also complicate temporal interaction modeling, as objects undergo drastic appearance and disappearance dynamics over time, posing challenges in capturing their relationships \citep{bansal2022my, tang2023egotracks}.

Given the larger semantic gap for video-language compared to image-language, cross-modal interaction plays a crucial role in video understanding. The interaction between visual and language features is pivotal for aligning the semantics of video and text query to associate them for text-video retrieval, or identifying relevant parts to answer the question and writing the caption in videoQA and video captioning, respectively. In addition, incorporating the interaction of motion and language features can mitigate the extraction of noisy information from videos \citep{ding2022language}. \citet{lin2022eclipse} also discover that the interaction between audio and language features can compactly capture information related to objects, actions, and complex events, compensating for sparsely extracted video frames.

\noindent\textbf{Cross-domain adaptation.} Given the infinitude of online videos, that our video understanding model will encounter testing scenarios which are identically distributed to our training data is an impractical assumption. Moreover, with the advent of LLM-augmented models that can tackle a variety video understanding tasks \citep{li2023videochat, li2023llama}, it is currently more advisable to train a model that can effectively adapt to multiple tasks and domains than to obtain a model which specializes in a specific understanding task. Furthermore, since a video can be considered as a sequence of images, training a model on video-text data is more computationally expensive than image-text data. Combined with the large-scale of recent video understanding models \citep{jiang2022cross, yang2022zero}, there is also a need to devise an efficient fine-tuning strategy to save the computational cost of fine-tuning these models. 

\noindent\textbf{Data preparation.} 
Although \citet{lei2021less} only use image-text data to train models for video understanding tasks, in essence, video-text data are crucial for the effectiveness of these models. In particular, compared with a static image, a video offers richer information with diverse spatial semantics with consistent temporal dynamics \citep{zhuang2023video}. As such, \citet{cheng2023vindlu} find that training on videos outperforms training on images, but jointly training on both data achieves the best performance. As additional evidence, \citet{yuan2023videoglue} shows that video-pretrained models outperform image-pretrained models in classifying motion-rich videos. However, video-text data takes up more storage cost than image-text data since a video comprises multiple images as video frames. Moreover, annotating a video is also more time-consuming and labor-intensive than annotating an image \citep{xing2023svformer}. Therefore, video understanding models have been limited by the small size of clean paired video-text corpora in contrast to billion-scale image-text datasets \citep{zhao2023learning}. Various efforts \citep{zhao2023learning, xing2023svformer} have been put into devising efficient and economical methods to curate and label video-text data.

\noindent\textbf{Addressing challenges.} These identified challenges encompass three critical perspectives: model architecture, model training, and data preparation in the field of video understanding. In general, there should be a synergistic relationship among these components. Specifically, model architecture should be designed to effectively capture video-language interactions. Concurrently, model training should be tailored to enable the architecture to adapt to target domains with their captured video-language interactions. Lastly, data preparation plays a pivotal role in shaping model training, which in turn significantly impacts the development of an efficacious model architecture.
\section{Model Architecture for Video- Language Understanding}
\label{sect:model_architecture}
\begin{figure}[t]
    \centering
    \includegraphics[width=0.7\linewidth]{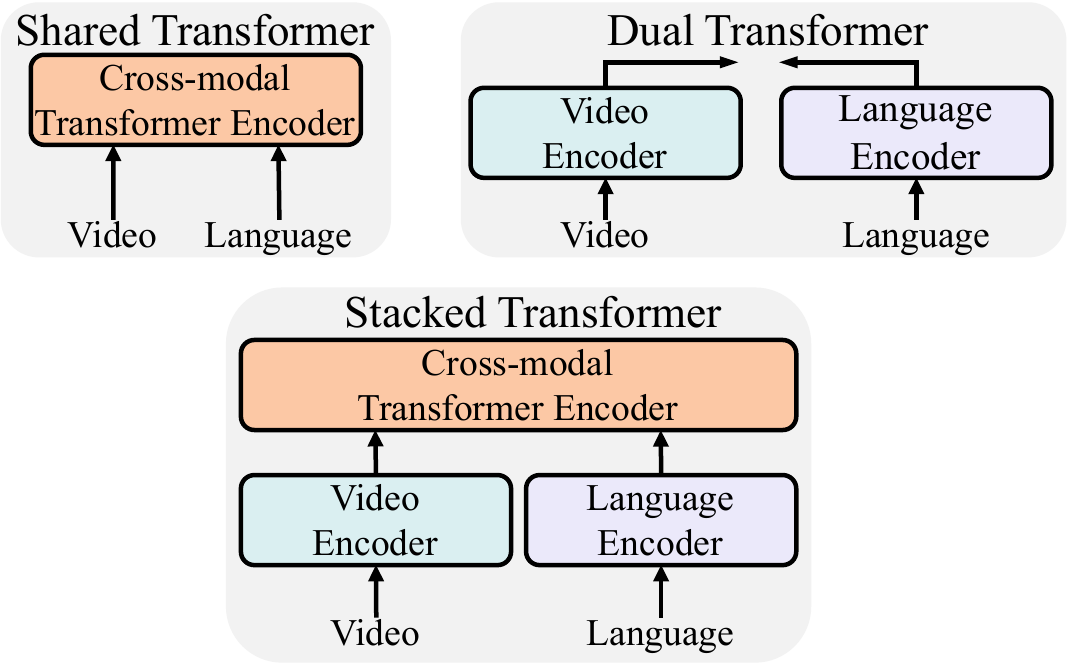}
    \caption{Illustration of Transformer-based architectures.
    }
    \label{fig:transformer_architectures}
\end{figure}

\noindent Addressing the challenge of intra-modal and cross-modal interaction is the key aim in designing video understanding model architectures, which can be divided into \textbf{Pre-transformer} and \textbf{Transformer-based architectures}. The advent of LLMs with remarkable zero-shot capability in addressing multiple tasks led to the design of \textbf{LLM-augmented architectures} that exhibit cross-domain adaptation ability to various video understanding tasks.

\subsection{Pre-Transformer Architecture}
\noindent Pre-transformer architectures typically comprise unimodal video and language encoders for implementing intra-modal interactions and cross-modal encoders for cross-modal interactions. 

\noindent\textbf{Unimodal encoders.} A video encoder often encodes raw videos by extracting frame appearance and clip motion features as spatial and temporal representations, respectively. As each video frame can be considered as a single image, various works have utilized CNNs to extract spatial representations \citep{simonyan_NIPS2014_twoStream, feichtenhofer_CVPR2016_fusion, zhao2017videoa}. For temporal representations, the sequential nature of RNN makes it a popular choice in pre-transformer architectures \citep{yang2017tensor, zhao2017two, venugopalan2015sequence,wang2019holistic}. Furthermore, 3D CNNs with an additional temporal channel inserted to 2D CNN have also demonstrated effectiveness in extracting spatio-temporal representations \citep{tran2017convnet, carreira2017quo}. In addition to CNN and RNN, \citet{chen2018probabilistic}, \citet{gay2019visual}, and \citet{wei2017graph} also build graphs to incorporate intra-modal relationships among video entities such as video segments or visual objects. These graph-structured works emphasize the reasoning ability of the model architecture.

A common framework of language encoder is to extract pre-trained word embeddings such as word2vec \citep{kaufman2016temporal, yu2017end} or GloVe \citep{torabi2016learning, kiros2014unifying}, then proceed with RNN-based modules such as LSTM or GRU. Such framework is taken from language model architectures before the era of Transformer.

\noindent\textbf{Cross-modal encoders.} \citet{gao2017tall} and \citet{zeng2017leveraging} apply element-wise multiplication to fuse the global video and question representations for video question answering. It demonstrates the advantage of a simple operation for video-language fusion. Attention has also been used to model video-language relations, in order to identify salient parts in video and language sentence \citep{yuan2019find}, or to refine the representation of the video based on the language question \citep{xu2017video}. Pre-transformer video-language works have also combined attention with a wide variety of techniques, including hierarchical learning \citep{baraldi2017hierarchical}, memory networks \citep{fan2019heterogeneous}, and graph networks \citep{xiao2022hqga, wei2023multi}. 

\subsection{Transformer-based Architecture}
\noindent Developed based on the self-attention mechanism, which exhaustively correlates every pair of input tokens with each other, Transformer-based architecture has the capacity to capture long-term dependencies and learn from web-scale data. It has demonstrated remarkable performance in many video-language tasks. Similar to the pre-transformer architecture, the Transformer-based framework also comprises unimodal encoders and cross-modal encoders to model intra-modal and cross-modal interactions, respectively. For unimodal encoders, several works find vision transformer for video encoding and BERT encoder for language encoding outperform RNN- and CNN-based encoding \citep{fu2021violet, bain2021frozen, seo2022end}. We then summarize fundamental types of Transformer-based architectures and illustrate them in Figure \ref{fig:transformer_architectures}.

\noindent\textbf{Shared Transformer.} Motivated by the success of Transformer in language modeling \citep{devlin2018bert}, \citet{akbari2021vatt} and \citet{wang2023all} construct a shared Transformer encoder for video understanding. Their encoder architectures receive the concatenation of visual patches and language tokens, then jointly calculate their interactions in a BERT-based manner. \citet{akbari2021vatt} additionally incorporate modality embeddings which comprise three values to denote three kinds of input modalities, \textit{i.e.} (video, audio, text).

\noindent\textbf{Stacked Transformer.} \citet{li2020hero} reveals that a shared Transformer encoder is weak in modeling temporal relations between videos and texts. To address this problem, they introduce a stacked Transformer architecture, with a hierarchical stack consisting of unimodal encoders to encode video and language inputs separately, and then a cross-modal Transformer to compute video-language interactions. A multitude of video understanding works follow such design to stack a cross-modal Transformer-based encoder above unimodal encoders \citep{fu2023empirical, li2023lavender, lei2021less, wei2022audio, luo2022clip4clip, nie2022search, wei2024learning}. To perform video captioning, \citet{seo2022end} and \citet{luo2020univl} further insert a causal Transformer-based decoder that generates language tokens based on the encoded cross-modal representations.

\noindent\textbf{Dual Transformer.} Dual Transformer architectures have been favored for text-video retrieval \citep{luo2022clip4clip, bain2021frozen, bain2022clip, lin2022eclipse, xue2022clip}. These architectures use two Transformer encoders to encode video and language separately, yielding global representations for each input modality, then applying simple operations such as cosine similarity to compute cross-modal interaction. Such a separate encoding scheme enables them to mitigate the computational cost of computing pairwise interactions between every pair of video and language inputs. 
They have accomplished not only efficiency but also effectiveness in text-video retrieval problems.

\subsection{LLM-Augmented Architecture}
\label{sect:llm_augmented_architecture}

\noindent Large language models (LLMs) have achieved impressive results in simultaneously tackling multiple NLP tasks. Recent efforts have sought to apply LLMs for video understanding to extend its cross-domain adaptation ability to video-language settings \citep{chen2023videollm, li2023videochat}. These efforts can be categorized into two approaches. The first approach employs LLM as a controller and video understanding models as helping tools. The controller will call the specific tool according to the language input instruction. The second approach utilizes LLM as the output generator and seeks to align video pre-trained models to the LLM. For video understanding, since the second approach dominates the first one with a long list of recent works \citep{chen2023videollm, li2023videochat, chen2023videollm, li2023llama, zhang2023video, maaz2023video}, we review them as follows:

\noindent\textbf{LLM as Output Generator.} The framework comprises a visual encoder, a semantic translator, and an LLM as the output generator. Regarding visual encoder, LLM-augmented architectures often use vision transformer and CNN models of the pre-Transformer and Transformer-based architectures \citep{chen2023videollm}. Since an LLM has never seen a video during its training, a semantic translator is needed to translate the visual semantics of a video to the LLM’s semantics. For the translator, Video-LLaMA \citep{zhang2023video} and VideoChat \citep{li2023videochat} implement a Q-Former as a Transformer-based module that uses a sequence of query embeddings that interact with visual features of the video to extract informative video information. Instead of Q-Former, VideoLLM \citep{chen2023videollm}, Video-ChatGPT \citep{maaz2023video}, and LLaMA-Vid \citep{li2023llama} find that a simple linear projection that projects visual features into the LLM’s input dimension can achieve effective performance. Subsequently, these visual-based query embeddings or projected visual features are combined with the language instruction to become the input fed to the LLM to produce the final output. 

\subsection{Architecture Analysis}
\label{sect:architecture_analysis}
\noindent In Figure \ref{fig:timeline}, we show the timeline of video understanding methodologies, categorized according to our defined architecture taxonomy and their affiliated downstream tasks. The evolution of pre-transformer models aligns with our hierarchy of video understanding levels, \textit{i.e.} models for video captioning generally appear subsequent to those for text-video retrieval, followed by the development of videoQA models. Owing to their impressive capacity, Transformer-based models capable of addressing multiple tasks have been introduced concurrently with task-specific Transformer frameworks. Recently, large language models (LLMs) have gained prominence for their superior in-context learning ability, enabling them to handle diverse tasks without fine-tuning. Consequently, new LLM-augmented architectures have emerged to utilize this capability to address multiple understanding tasks.

Among Transformer-based architectures, the dual Transformer stands out as the most effective for text-video retrieval, adeptly associating global semantics of video and language modality. On the other hand, the stacked Transformer architecture excels at facilitating intra-modal and cross-modal interactions through its specialized unimodal and cross-modal encoders. These encoders are particularly efficient at correlating video content with the question in videoQA. Additionally, for video captioning, cross-modal encoder plays a crucial role in translating video content into textual descriptions. Recent LLM-augmented models have begun to outperform Transformer-based architectures in videoQA, signalling their potential as the next frontier in video understanding research. We provide full details of performance in text-video retrieval, video captioning, and videoQA tasks in Table \ref{tab:app_exp_text_video_retrieval}, \ref{tab:app_exp_video_captioning}, and \ref{tab:app_exp_videoqa}, respectively.

\begin{table*}[h]
\centering
\resizebox{\linewidth}{!}{
\begin{tabular}{l|c|c|c|ccc}
\hline
\textbf{Methods}       & \textbf{Model architecture}      & \textbf{Video}             & \textbf{Text}          & \textbf{R@1}  & \textbf{R@5}  & \textbf{R@10} \\ \hline
VSE-LSTM   \citep{kiros2014unifying}    &     \multirow{6}{*}{Pre-TF}                    & ConvNet/OxfordNet & GloVe-LSTM    & 3.8  & 12.7 & 17.1 \\
C+LSTM+SA-FC7 \citep{torabi2016learning} &                         & VGG               & GloVe-LSTM    & 4.2  & 12.9 & 19.9 \\
EITanque  \citep{kaufman2016temporal}    &                         & VGG               & word2vec-LSTM & 4.7  & 16.6 & 24.1 \\
SA-G+SA-FC7 \citep{torabi2016learning}   &                         & VGG               & GloVe         & 3.1  & 9.0  & 13.4 \\
CT-SAN    \citep{yu2017end}    &                      & RN                & word2vec-LSTM & 4.4  & 16.6 & 22.3 \\ 
JSFusion  \citep{yu2018joint}    &  & RN                & GloVe-LSTM    & 10.2 & 31.2 & 43.2 \\ \hline
All-in-one  \citep{wang2023all}   & Shared TF               & Linear               & BT            & 37.9 & 68.1 & 77.1 \\
VLM   \citep{xu2021vlm}     & Shared TF               & S3D               & BT            & 28.1 & 55.5 & 67.4 \\
DeCEMBERT   \citep{tang2021decembert}     & Shared TF               & RN              & BT            & 17.5 & 44.3 & 58.6 \\
ActBERT   \citep{zhu2020actbert}     & Stacked TF               & Faster-RCNN              & BT & 16.3 & 42.8 & 56.9 \\
VIOLET   \citep{fu2023empirical}     & Stacked TF               & VS-TF               & BT            & 37.2 & 64.8 & 75.8 \\
VindLU   \citep{cheng2023vindlu}     & Stacked TF               & ViT               & BT            & \underline{48.8} & \underline{72.4} & \underline{82.2} \\
HERO   \citep{li2020hero}       & Stacked TF                & RN+SlowFast       & BT            & 16.8 & 43.4 & 57.7 \\
MV-GPT   \citep{seo2022end}     & Stacked TF                & ViViT             & BT            & 37.3 & 65.5 & 75.1 \\
CLIP2TV  \citep{gao2021clip2tv}    & Dual TF                 & ViT               & CLIP-text     & 32.4	 & 58.2 & 68.6 \\
CLIP-ViP  \citep{xue2022advancing}    & Dual TF                 & ViT               & CLIP-text     & \textbf{49.6} & \textbf{74.5} & \textbf{84.8} \\
CLIP4Clip \citep{luo2022clip4clip}    & Dual TF                 & ViT               & CLIP-text     & 44.5 & 71.4 & 81.6 \\ \hline
\end{tabular}}
\caption{Performance on text-video retrieval. (Pre-TF: Pre-transformer, Shared TF: Shared Transformer, Stack TF: Stack Transformer, Dual TF: Dual Transformer, RN: ResNet/ResNeXt \citep{he_CVPR2016_resnet, xie_CVPR2017_resnext}, ViT: Vision Transformer \citep{dosovitskiy2020image}, BT: BERT \citep{devlin2018bert}, ViViT: Video Vision Transformer \citep{arnab2021vivit}). We report recall at rank 1 (R@1), 5 (R@5), and 10 (R@10). We choose MSRVTT as one of the most popular datasets for text-video retrieval.}
\label{tab:app_exp_text_video_retrieval}
\end{table*}

\begin{table*}[h]
\centering
\resizebox{\linewidth}{!}{
\begin{tabular}{l|c|c|ccc}
\hline
\textbf{Methods} & \textbf{Model architecture }             & \textbf{Video} & \textbf{BLEU-4} & \textbf{METEOR} & \textbf{CIDEr} \\ \hline
TA      \citep{yao2015describing}                    &          \multirow{9}{*}{Pre-TF}                 & Video: 3D-CNN             & 36.5   & 25.7  &  - \\
h-RNN     \citep{yu2016video}                  &                           & Video: VGG                & 36.8   & 25.9 &  - \\
MFATT    \citep{long2018video}                   &    & Video: RN+C3D             & 39.1   & 26.7  & - \\
CAT-TM   \citep{long2018video}      &                           & Video: RN+C3D             & 36.6   & 25.6 &  - \\
NFS-TM     \citep{long2018video}                 &                           & Video: RN+C3D             & 37.0   & 25.9 & - \\
Fuse-TM     \citep{long2018video}                &                           & Video: RN+C3D             & 37.5   & 25.9 & - \\ 
MARN     \citep{pei2019memory}                &                           & Video: RN             & -   & - & 46.8 \\ 
Res-ATT     \citep{li2019residual}                &                           & Video: RN             & 37.0   & 26.9 & 40.7 \\ 
DenseLSTM     \citep{zhu2019attention}                &                           & Video: VGG             & 38.1   & 27.2 & 42.8 \\  \hline
VIOLET   \citep{fu2023empirical}     & \multirow{8}{*}{Stacked TF}               & VS-TF               &             - & - &  58.0 \\
LAVENDER      \citep{li2023lavender}                               &                     & VS-TF                                   & -          & -     & 57.4   \\
VLAB     \citep{he2023vlab}                   &  & EVA-G                     & 54.6   & 33.4 &  74.9 \\
UniVL   \citep{luo2020univl}                    &                           & S3D                       & 41.8   & 28.9  &  50.0 \\
MV-GPT    \citep{seo2022end}                  &                           & ViViT                     & 48.9   & 38.7 &  60.0 \\
CLIP-DCD      \citep{yang2022clip}              &                           & ViT                       & 48.2   & 30.9  & 64.8 \\
DeCEMBERT    \citep{tang2021decembert}               &                           & RN                        & 45.2   & 29.7 &  52.3 \\
mPLUG-2    \citep{xu2023mplug}                 &                           & ViT                       & 57.8   & 34.9  & 80.3 \\ \hline
\end{tabular}}
\caption{Performance on video captioning. (Pre-TF: Pre-transformer, Stacked TF: Stacked Transformer, RN: ResNet/ResNeXt \citep{he_CVPR2016_resnet, xie_CVPR2017_resnext}, ViViT: Video Vision Transformer \citep{arnab2021vivit}, EVA-G: \citet{fang2023eva}). We report BLEU-4 and METEOR, which are two popular metrics for language generation. We choose MSRVTT as one of the most popular datasets for video captioning.}
\label{tab:app_exp_video_captioning}
\end{table*}

\begin{table*}[h]
\centering
\resizebox{\linewidth}{!}{
\begin{tabular}{l|c|c|c|cc}
\hline
\multicolumn{1}{c|}{\multirow{2}{*}{\textbf{Methods}}} & \multirow{2}{*}{\textbf{Architecture}} & \multirow{2}{*}{\textbf{Video}} & \multirow{2}{*}{\textbf{Text}} & \multicolumn{2}{c}{\textbf{Dataset}} \\ 
                         &                               &                        &                       & MSRVTT        & MSVD        \\ \hline
E-MN    \citep{xu2017video}                                    & \multirow{8}{*}{Pre-TF}       & VGG + C3D               & GloVe-LSTM            & 30.4          & 26.7           \\
QueST    \citep{jiang2020divide}                                    &        & RN + C3D               & GloVe-LSTM            & 40.0          & -           \\
HME      \citep{fan2019heterogeneous}                                    &                               & RN/VGG + C3D           & GloVe-GRU             & 34.6          & 36.1        \\
HGA        \citep{jiang2020reasoning}                                  &                               & RN/VGG + C3D           & GloVe-GRU             & 33.0          & 33.7        \\
ST-VQA  \citep{jang2019video}                                     &                               & RN+C3D                 & GloVe-LSTM            & 35.5          & 34.7        \\
PGAT       \citep{peng2021progressive}                                  &                               & Faster-RCNN            & GloVe-LSTM            & 38.1          & 39.0        \\
HCRN          \citep{le2020hierarchical}                               &                               & RN                     & GloVe-LSTM            & 35.6          & 36.1  
\\
HQGA          \citep{xiao2022hqga}                               &                               & Faster-RCNN                     & BERT-LSTM            & 38.6          & 41.2       
\\ \hline
All in one     \citep{wang2023all}                              & Shared TF                     & Linear                    & BT                    & 44.3          & 47.9        \\
LAVENDER      \citep{li2023lavender}                               & Stacked TF                     & VS-TF                  & BT                    & 45.0          & 56.6        \\
DeCEMBERT   \citep{tang2021decembert}     & Stacked TF               & RN               & BT             & 37.4 & - \\
VindLU   \citep{cheng2023vindlu}     & Stacked TF               & ViT               & BT             & 44.6 & - \\
VIOLET        \citep{fu2023empirical}                               & Stacked TF                     & VS-TF                  & BT                    & 44.5          & 54.7        \\
ClipBERT        \citep{lei2021less}                             & Stacked TF                     & CLIP-text              & BT                    & 37.4          & -           \\
VGT    \citep{xiao2022video}                                       & Dual TF                       & Faster-RCNN            & BT                    & 39.7          & -           \\
CoVGT     \citep{xiao2023contrastive}                                   & Dual TF                       & Faster-RCNN            & BT                    & 40.0          & -        \\ \hline
LLaMA-Vid      \citep{li2023llama}                              & LLM-Augmented                 & EVA-G                  & Vicuna                & 58.9          & 70.0    \\ \hline   
\end{tabular}}
\caption{Performance on videoQA. (Pre-TF: Pre-transformer, Dual TF: Dual Transformer, RN: ResNet/ResNeXt \citep{he_CVPR2016_resnet, xie_CVPR2017_resnext}, BT: BERT \citep{devlin2018bert}, VS-TF: Video Swin Transformer \citep{liu2021video}, EVA-G: \citet{fang2023eva}). We report accuracy of the methods. We choose MSRVTT and MSVD as two of the most popular datasets for videoQA. }
\label{tab:app_exp_videoqa}
\end{table*}
\section{Model Training for Video Understanding}
\label{sect:model_training}
\noindent Model training seeks to address the cross-domain adaptation ability of video understanding models. To achieve this goal, pre-training strategies have been devised to gain world knowledge that generalizes across multiple scenarios, then task-specific fine-tuning is conducted to specifically improve downstream task performance.
\subsection{Pre-training for Video Understanding}
\noindent In this section, we mainly summarize pre-training strategies for video understanding models into three groups:

\noindent\textbf{Language-based pre-training.} The most popular language-based pre-training task is masked language modeling (MLM) \citep{lei2021less, sun2019videobert, cheng2023vindlu}, which randomly masks a portion of words in the language input and trains the model to predict the masked words based on unmasked language words and video entities. Instead of masking a portion of words, UniVL \citep{luo2020univl} and VICTOR \citep{lei2021understanding} discover that masking the whole language modality benefits video captioning task. MLM can be combined with other language-based pre-training task, \textit{e.g.} masked sentence order modeling which is to classify the original order of the shuffled language sentences \citep{lei2021understanding}.

\noindent\textbf{Video-based pre-training.} Video-based pre-training tasks help video-language models capture contextual information in the video modality. As a counterpart of MLM, masked video modeling (MVM) trains the model to predict the portion of masked video entities based upon the unmasked entities and language words. The continuous nature of videos leads to different choices of video entities, such as frame patches \citep{li2020hero} or video frames \citep{fu2021violet}. In terms of the training objective, \citet{li2020hero} use L2 regression loss to train the model to predict pre-trained features of the masked video frames extracted by ResNet and SlowFast models, while \citet{fu2021violet} use cross-entropy loss to train the model to predict the masked visual tokens, which are quantized by a variational autoencoder from visual frame patches. 

\noindent\textbf{Video-text pre-training.} Video-text pre-training is crucial for a model to capture video-language relation. \citet{xue2022clip}, \citet{gao2021clip2tv}, and \citet{bain2021frozen} utilize a framework of video-text contrastive learning to produce close representations for semantically similar video and language inputs. These works focus on creating a joint semantic space that aligns separate representations of video and language. Instead of separate representations, \citet{tang2021decembert}, \citet{fu2021violet}, and \citet{li2023lavender} enable video and textual representations to interact with each other and use a single token to represent the cross-modal input, which is forwarded to predict whether the video-text pair is matched or not. In these two pre-training frameworks, not only video-text data but also image-text data are utilized during pre-training, in which an image is considered as a video with a single frame.

Contrastive learning has revealed promising results  \citep{lin2022eclipse, gao2021clip2tv, xue2022clip, nguyen2022adaptive, nguyen2021contrastive, nguyen2024topic, nguyen2024kdmcse, nguyen2023improving, wu2023infoctm, wu2024dynamic, wu2022mitigating}. MLM has contributed to enhancing VideoQA since the task resembles MLM in predicting the language word given a video-language pair (the question is the language input in videoQA). Compared to these pre-training strategies, MVM does provide performance gain for video understanding but its gain is less significant. For more details about pre-training, please refer to \citep{cheng2023vindlu}.

\subsection{Fine-tuning for Video Understanding}
\noindent Task-specific fine-tuning is commonly used by pre-Transformer architectures to train from scratch since these models do not have sufficient parameter capacity to learn generalizable features through pre-training. It is also widely adopted by Transformer-based architectures to improve the performance for a specific downstream task. Moreover, LLM-augmented architectures also utilize instruction tuning as a variant of fine-tuning, to adapt from the visual and audio spaces to the LLM language space.

\noindent\textbf{Fine-tuning strategies.} Normally, all of the model parameters are updated during fine-tuning \citep{gao2017tall, xu2019multilevel, anne2017localizing, nguyen2023demaformer, wu2023effective}. However, in cases computational resources or training data are limited, only adaptation layers such as low-rank adapters \citep{pan2022st, yang2022zero, nguyen2024read} or learnable prompt vectors \citep{ju2022prompting} are fine-tuned to reduce training cost or prevent overfitting. Such risks also apply for LLM-augmented architectures discussed in Section \ref{sect:llm_augmented_architecture}, since LLMs exhibit a billion scale of parameters, thus incurring excessively huge cost if full fine-tuning is conducted. For such models, \citet{zhang2023video} and \citet{li2023llama} design a two-stage instruction tuning strategy which only fine-tunes the semantic translator. The first stage trains the model to generate the textual description based on the combined video and the language instruction, in order to align visual representations extracted by the visual encoder with the language space of LLM. The second stage is often performed on small-scale video-text pairs manually collected by the authors to further tailor the output features of the translator towards the target domains.

\section{Data Perspective for Video Understanding}
\label{sect:data}
\noindent In this section, we analyze data preparation approaches for video understanding models, and provide details of the datasets in Table \ref{tab:datasets}.

\begin{table*}[t]
\centering
\resizebox{\linewidth}{!}{
\begin{tabular}{l|l|c|c|c}
\hline\hline
\multicolumn{1}{c|}{\textbf{Dataset}}            & \multicolumn{1}{c|}{\textbf{Video source}}     & \textbf{Annotation}   & \textbf{Tasks}            & \textbf{\#Videos/\#Routes} \\ \hline
 MSVD   \citep{chen2011collecting}            & YouTube videos         & Manual       & TVR, VC, VideoQA & 1.9K              \\
 MSRVTT     \citep{xu2016msr}        & Web videos         & Manual       & TVR, VC, VideoQA & 7.2K              \\
 ActivityNet   \citep{yu2019activitynet}     &  YouTube videos     & Manual       & AL, TVR, VC, VMR & 5.8K              \\
 FIBER      \citep{castro2022fiber}        & VaTeX \citep{wang2019vatex}         & Manual       & VC, VideoQA      & 28K               \\
 WildQA    \citep{castro2022wild}         & YouTube videos         & Manual       & VideoQA          & 0.4K              \\
 NExT-QA  \citep{xiao2021next}          & VidOR \cite{shang2019annotating} & Manual       & VideoQA          & 5.4K              \\
 CausalVid-QA  \citep{li2022representation}     & Kinetics-700 \citep{carreira2019short} & Manual       & VideoQA          & 26K               \\
 HowTo100M  \citep{miech2019howto100m}        & YouTube videos         & Auto         & PT               & 1.2M              \\
 HD-VILA-100M \citep{xue2022advancing}      & YouTube videos         & Auto         & PT               & 3.3M              \\
 YT-Temporal-180M \citep{zellers2021merlot}  & YouTube videos         & Auto         & PT               & 6M                \\
 TGIF-QA  \citep{jang2017tgif}          & Animated GIFs & Manual       & VideoQA          & 71K               \\
 TGIF-QA-R   \citep{peng2021progressive}       & TGIF-QA \citep{jang2017tgif} & Manual, Auto & VideoQA          & 71K               \\
 DiDeMo    \citep{anne2017localizing}         & YFCC100M \citep{thomee2016yfcc100m} & Manual       & TVR              & 11K               \\
 YouCook2   \citep{zhou2018towards}        & YouTube videos         & Manual       & TVR, VC          & 2K                \\
HMDB-51  \citep{hmdb51}          & Web videos         & Manual       & TVR, AR          & 6.8K              \\
Kinetics-400 \citep{kinetics400}      & YouTube videos         & Manual       & AR               & 306K              \\
Kinetics-600  \citep{carreira2018short}     & Kinetics-400 \citep{kinetics400}         & Manual       & AR, VG           & 480K              \\
Kinetics-700  \citep{carreira2019short}     & Kinetics-600 \citep{carreira2018short}         & Manual       & AR               & 650K              \\
VaTeX    \citep{wang2019vatex}          & Kinetics-600 \citep{carreira2018short} & Manual       & TVR, VC          & 41K               \\
TVR     \citep{lei2020tvr}           & TVQA \citep{lei2018tvqa} & Manual       & VMR              & 22K               \\
How2R  \citep{li2020hero}            & HowTo100M \citep{miech2019howto100m} & Manual       & VMR              & 22K               \\
How2QA  \citep{li2020hero}           & HowTo100M \citep{miech2019howto100m} & Manual       & VideoQA          & 22K               \\
YouTube Highlights \citep{sun2014ranking} & YouTube videos         & Manual       & VMR              & 0.6K              \\
TACoS    \citep{regneri2013grounding}          & MPII Composites \citep{rohrbach2012database} & Manual       & VMR              & 0.1K              \\
QVHighlights  \citep{lei2021detecting}     & YouTube vlogs         & Manual       & VMR              & 10K               \\
TVSum    \citep{song2015tvsum}          & YouTube videos         & Manual       & VMR              & 50                \\
ViTT     \citep{huang2020multimodal}          & YouTube-8M \citep{youtube8m} & Manual       & VMR              & 5.8K              \\
VidChapters-7M  \citep{yang2023vidchapters}   & YT-Temporal-180M \citep{zellers2021merlot}         & Auto         & VC, VMR          & 817K              \\
VideoCC3M   \citep{nagrani2022learning}       & Web videos         & Auto         & PT               & 6.3M              \\
WebVid-10M  \citep{bain2021frozen}       & Web videos         & Auto         & PT               & 10.7M             \\
COIN      \citep{tang2019coin}         & YouTube videos         & Manual       & AS               & 12K               \\
CrossTask   \citep{zhukov2019cross}       & YouTube videos         & Manual       & AR               & 4.7K              \\
Alivol-10M   \citep{lei2021understanding}      & E-commerce videos         & Auto         & PT               & 10M               \\
LSMDC    \citep{rohrbach2015dataset}          & British movies         & Manual       & TVR              & 72                \\
EK-100   \citep{damen2022rescaling}          & Manual           & Manual       & AR, AL           & 7K                \\
SSV1     \citep{sthsth}          & Manual           & Manual       & AR               & 108K              \\
SSV2    \citep{sthsth}           & Manual           & Manual       & AR               & 221K              \\
Moments in Time  \citep{monfort2019moments}  & Web videos         & Manual       & AR               & 1M                \\
InternVid   \citep{wang2023internvid}       & YouTube videos         & Auto         & PT               & 7.1M              \\
How2      \citep{sanabria2018how2}         & YouTube videos         & Auto         & VC               & 13.2K             \\
WTS70M    \citep{stroud2020learning}         & YouTube videos         & Auto         & PT               & 70M               \\
Charades     \citep{gao2017tall}      & Manual           & Manual       & AR, VMR, VideoQA & 10K              \\ \hline
\end{tabular}}
\caption{Video understanding datasets in the literature. (VMR: Video moment retrieval, TVR: text-video retrieval, VC: video captioning, AL: action localization, AR: action recognition, AS: action segmentation, VG: video generation, PT: pre-training).}
\label{tab:datasets}
\end{table*}
\subsection{Data curation}

\noindent\textbf{Manual collection.} To curate video-language data, multiple works search for publicly available online videos, which exhibit a wide diversity of content. Video-language datasets with online videos are mostly aimed for pre-training models to learn generalizable knowledge, \textit{e.g.} HowTo100M \citep{miech2019howto100m} and YT-Temporal-180M \citep{zellers2021merlot}, or they can also be used for fine-tuning, \textit{e.g.} MSRVTT \citep{xu2016msr} and YouCook2 \citep{zhou2018towards}. To satisfy a certain requirement, videos different from the online ones can be inherited from existing datasets, \textit{e.g} \citet{xiao2021next} utilize 6,000 videos from VidOR dataset and \citep{li2022representation} inherit 546,882 videos from Kinetics-700 since they describe scenes of daily life and real world, respectively. Apart from making use of existing datasets' and online videos, videos can also be recorded by human annotators to enable quality control \citep{sthsth, damen2022rescaling}.

\noindent\textbf{Data augmentation.} Rather than manually collecting videos from external sources, \citet{xing2023svformer} and \citet{jiang2022semi} explore data augmentation techniques which are particularly designed for videos. In detail, their TubeTokenMix mixes two videos in which the mixing coefficient is defined upon the temporal dimension, and their temporal shift randomly shifts video frame features backward or forward over the temporal dimension. These techniques outperform standard augmentation approaches for image data, such as CutMix \citep{yun2019cutmix}, Mixup \citep{zhang2017mixup}, and PixMix \citep{hendrycks2022pixmix}.

\subsection{Label annotation}
\noindent\textbf{Manual annotation.} Several works \citep{li2022representation, lei2021detecting, xiao2021next} use human annotators since they provide high-quality labels. However, such approach is expensive, particularly when dealing with video data. For example, annotating QVHighlights dataset \citep{lei2021detecting} costs approximately \$16,000 for 10K videos and 3 months to complete. Similarly, NExT-QA \citep{xiao2021next} needs 100 undergraduate students and 1 year to annotate only 5K videos.

\noindent\textbf{Automatic generation.} Directly taking language transcripts of YouTube videos as textual labels could reduce annotation cost \citep{miech2019howto100m, xue2022advancing, zellers2021merlot}. However, these labels have been shown to be grammatically incorrect and temporally misalign with the video content \citep{tang2021decembert}. Motivated by the success of LLMs, \citet{zhao2023learning} train a system consisting of a TimeSformer-L visual encoder and a GPT-2XL decoder to write dense captions for videos. Moreover, \citet{li2023videochat} use GPT-4 to generate summaries for movie synopses. 

\section{Future Directions}
\label{sect:future_directions}

\noindent\textbf{Fine-grained understanding.} Existing methods excel at video understanding at a coarse-grained level, enabling effective responses to questions like ``\textit{what is}'' or the recognition of global events without significant difficulty \citep{xiao2021next}. Nevertheless, limiting comprehension to this coarse level could hinder practical utility of existing systems. In real-world scenarios, a user might require a precise timestamp and location of an object within a video \citep{jiang2022video}, or request the AI agent to forecast potential alternative events, which is a common need in predictive analytics \citep{xiao2021next, li2022representation}. These tasks necessitate an advanced understanding and inference capability regarding the causal and temporal relationships present in a video. At present, models exhibit a constrained visio-linguistic capacity to engage in temporal reasoning, categorizing them as image-sequence-and-language models rather than video-language models \citep{kesen2023vilma}. Therefore, future research in this direction deserves more attention and exploration. 

\noindent\textbf{Long-form video understanding.} Current understanding systems have demonstrated remarkable performance on short video clips lasting several seconds. However, they tend to struggle when switching to long-form videos which last several minutes or hours. To enhance the applicability of these systems, it is essential to enhance their capability of understanding long-form videos. Current approaches mainly feature reducing computational cost through architectures more efficient than Transformer-based ones such as state space models \citep{yang2024vivim, li2024videomamba}, which can be considered as linear RNN with specifically designed fixed weights, or compensating sparsely extracted video frames with additional information \citep{lin2022eclipse}. In general, how to effectively model long-form videos and adapt them to the joint context with language deserves more attention.

\noindent\textbf{Trustworthiness of video understanding models.} Although modern video understanding systems have demonstrated remarkable performance, their black-box nature undermines our trust to deploy them. In particular, we still do not precisely understand what part of the video a videoQA model looks at to answer the question \citep{li2022equivariant}, or how video and language semantic information flows into the common representation space of the video retrieval model \citep{jia2022adversarial}. Furthermore, adversarial noise sensitivity or hallucination of video understanding models are also open problems. Future trustworthiness benchmarks such as \citep{xiao2023can,wang2021dutrust} for video understanding are of great significance towards practical systems.

\chapter[MAMA: Meta-optimized Angular Margin Contrastive Framework for Video-Language Representation Learning]{MAMA: Meta-optimized Angular Margin Contrastive Framework for Video-Language Representation \\ Learning}
\label{ch1:mama}

\section{Introduction}
\begin{figure*}[t]
    \centering
    \caption{Examples of video inputs and their textual descriptions.}
    \label{fig:example_figure}
    \begin{subfigure}[h!]{0.48\linewidth}
    \includegraphics[width=\linewidth]{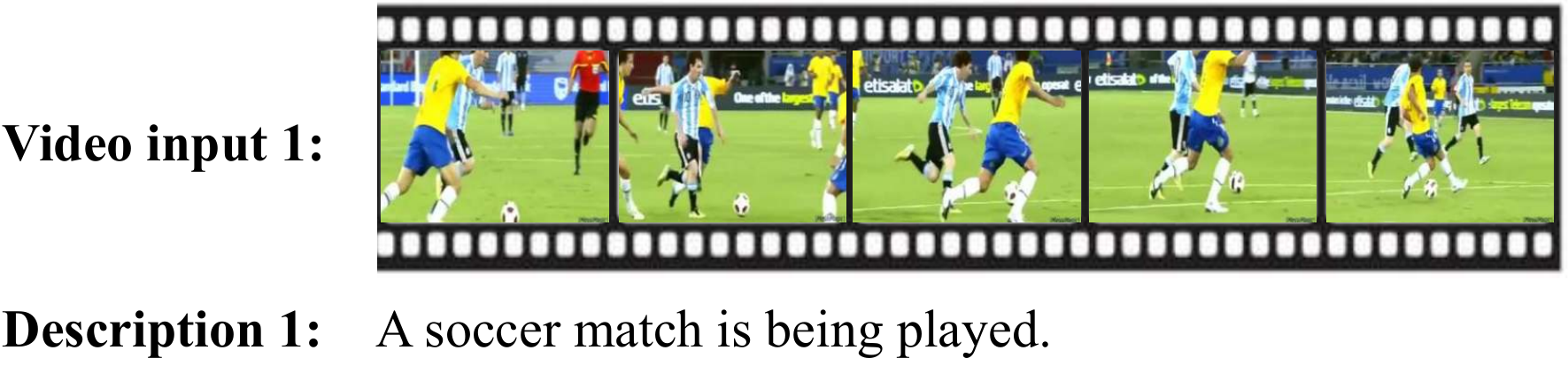}
    \end{subfigure}
    \quad
    \begin{subfigure}[h!]{0.48\linewidth}
    \includegraphics[width=\linewidth]{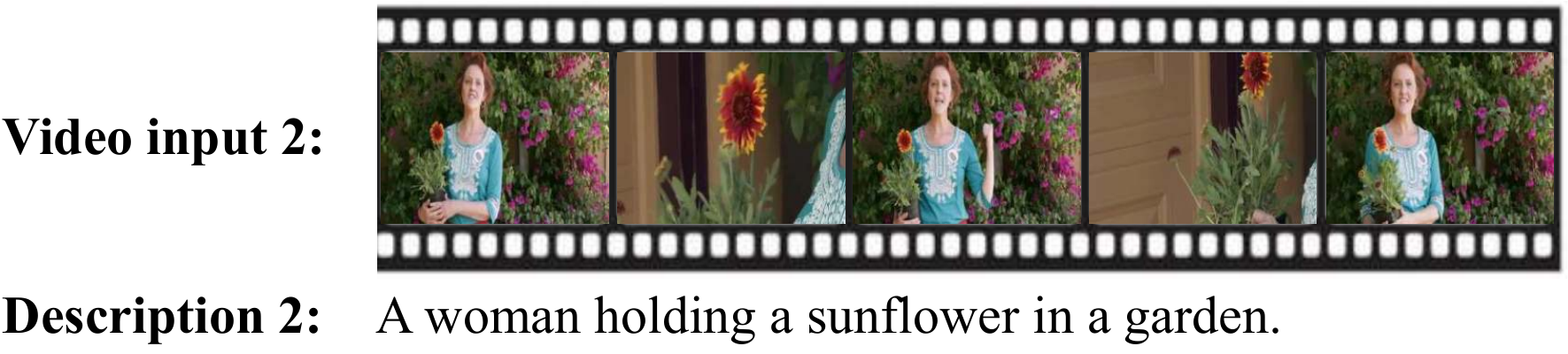}
    \end{subfigure}
\end{figure*}

\begin{figure*}[t]
    \centering
    \caption{Topic distribution of the MSRVTT dataset. We use Latent Dirichlet Allocation (LDA) to extract topics from manually annotated descriptions of videos.}
    \label{fig:distribution_figure}
    \includegraphics[width=0.7\linewidth]{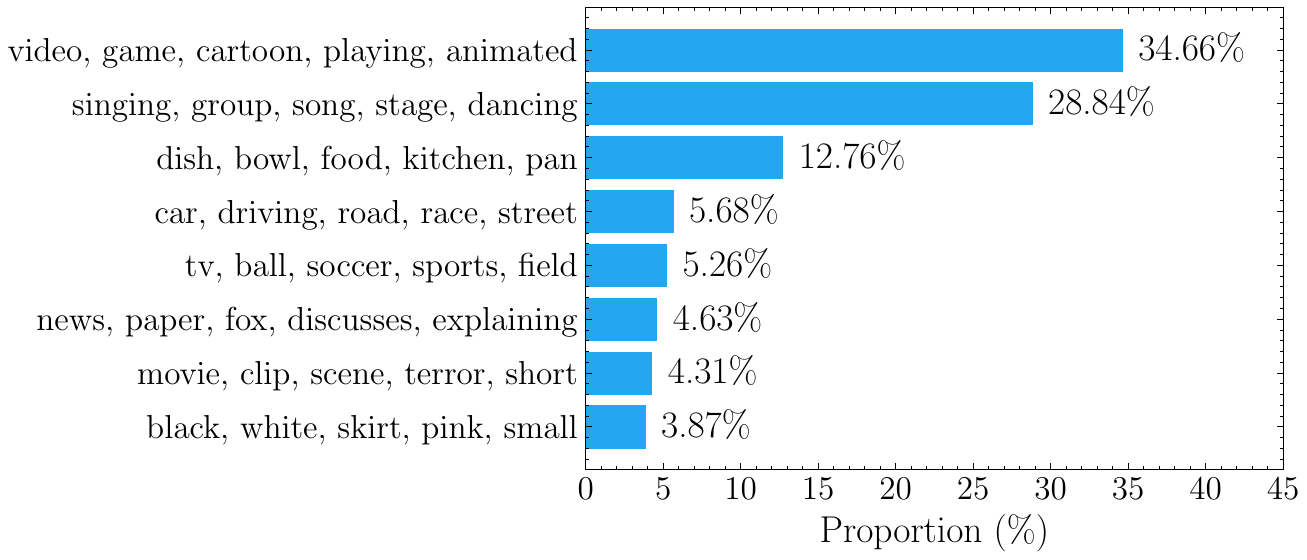}
\end{figure*}

\noindent Learning vision and language representations has advanced state-of-the-art across multiple cross-modal tasks, such as video question answering (VideoQA) \citep{lei2021less, li2020hero, fu2021violet, fu2023empirical, cheng2023vindlu, seo2022end, wang2023all} and text-video retrieval \citep{zhu2020actbert, tang2021decembert, xu2021vlm, lei2022revealing, li2023lavender, wang2022omnivl, buch2022revisiting}. The success of vision-language representation learning mainly results from the effectiveness of contrastive learning, which projects video and text inputs into a common latent space based on their semantic similarities.

Data quality stands at the forefront of influencing the efficacy of video-language representation learning, particularly through its cleanliness and diversity, which are pivotal in optimizing model performance and bolstering generalization capabilities \citep{zhao2023learning}. Within this context, we pinpoint two interrelated issues that significantly impact the \textit{cleanliness} and \textit{diversity} of video-text data: (1) the imperfection in alignment, characterized by a scarcity of fine-grained details, and (2) the imbalance among the concepts of data samples. First, a video and its textual description commonly does not perfectly align with each other, \textit{e.g.} the description may omit certain details in the video, such as the green grass on a soccer field or the pink flowers at the background in two videos of Fig. \ref{fig:example_figure}. As such, aggressively minimizing a contrastive loss to pull video and language representations together might result in distorted video-language representations that do not closely capture the semantic similarity of video-text pairs, thus compromising the interpretive video-language understanding. Second, as an example, Fig. \ref{fig:distribution_figure} shows a majority of video-text pairs in the well-known MSRVTT dataset denote the video game or singing topic (the topic bars), while a minority of them describe the fashion topic (the bottom bar). This might skew the model’s exposure towards certain topics at the expense of others, undermining the model’s ability to perform uniformly across a broad spectrum of subjects \citep{li2022trustworthy}. Addressing these intertwined challenges is crucial for advancing video-language representation learning and facilitating more effective video-language understanding.

Our method, called \textbf{MAMA}: \textbf{M}eta-optimized \textbf{A}ngular \textbf{MA}rgin Contrastive Learning, subtracts a margin between a positive video and text sample in the angular space. Our mathematical derivation shows that the subtracted margin can decay the gradient norm, thus providing a regularization effect to constrain positive but imperfectly aligned samples from reaching perfect similarity. For the imbalance issue, to enable the network to dynamically adjust its focus during training, MAMA introduces a sample reweighting strategy that maps loss values to sample weights. A natural idea is to assign higher weights to larger losses to emphasize minority classes for which the network is likely to make mistakes \citep{sun2007cost, malisiewicz2011ensemble}. However, with respect to the imperfect alignment issue, it is advisable to assign higher weights to smaller losses, as there is a greater chance that these samples better align with each other, hence forming cleaner samples which the network should focus on \cite{zhang2018generalized, wang2017robust}. To avoid the exhaustive effort in manually specifying a weighting function and enhance the function generalization, we parameterize our weighting function as a multi-layer perceptron (MLP) as theoretically a universal approximator for any continuous function \citep{csaji2001approximation}, and use a small unbiased validation set (meta-data) to train the MLP.

Moreover, to further enhance the diversity in video-language representation learning, MAMA utilizes off-the-shelf large vision-language model (LVLM) to augment downstream video-text data. In particular, given an additional video input, we adopt the density peak-based clustering approach to extract its key frames, then concatenate the extracted frames into one grid image, and forward the image with a relevant input prompt to obtain the text pairing. Combined with the LVLM-augmented video-text data, MAMA considerably outperforms previous state-of-the-art video-language representation learning methods on standard MSRVTT, DiDeMo, ActivityNet, TGIF-QA-R, NExT-QA, and Causal-VidQA datasets.

\begin{figure*}[t]
    \centering
\includegraphics[width=\linewidth]{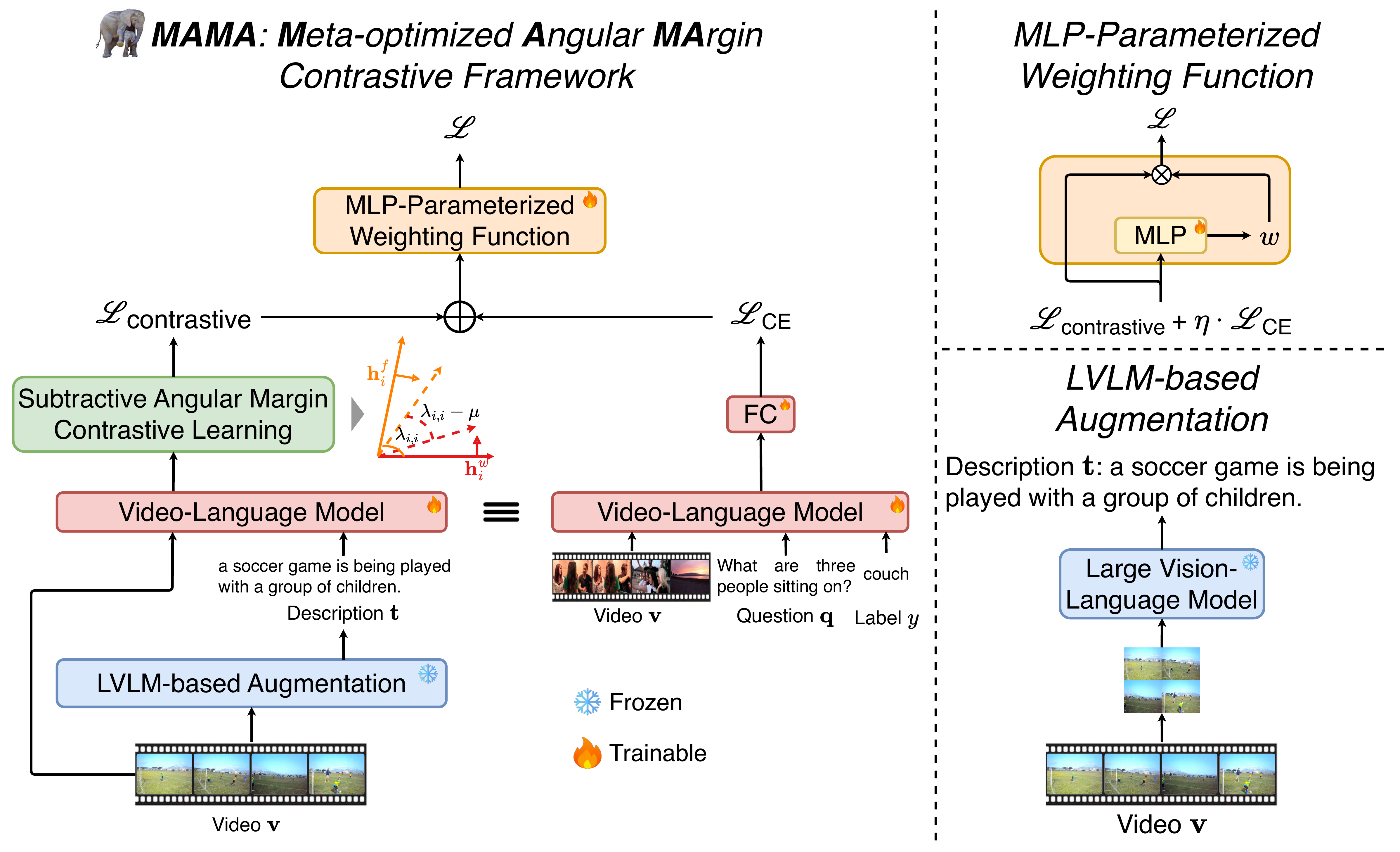}
    \caption{Illustration of the proposed MAMA framework and its components.} 
    \label{fig:mama_overall_illustration}
\end{figure*}
\section{MAMA Framework}
\noindent In this section, we explain MAMA, the meta-optimized angular margin contrastive framework with our augmentation strategy for video-language representation learning. We provide an overall illustration of MAMA in Figure \ref{fig:mama_overall_illustration}.
\subsection{Video-Language Representation Learning}
\label{sect:video_language_representation_learning}

\noindent We are given a corpus of video data, in which each video $\mathbf{v}_{i}$ is attached with a textual description $\mathbf{t}_{i}$ and possibly a label $y_{i}$. In the beginning, we embed the video input $\mathbf{v}_{i}$ into a sequence of visual representations $\mathbf{V}_{i}^{f} = \{\mathbf{h}_{i,j}^{f}\}_{j=1}^{N_{\mathbf{v}_{i}}}$, where $N_{\mathbf{v}_{i}}$ is the number of randomly sampled video frames in $\mathbf{v}_{i}$. We also embed the textual description into a sequence of representations $\mathbf{T}^{w}_{i} = \{\mathbf{h}_{i,j}^{w}\}_{j=1}^{N_{\mathbf{t}_{i}}}$, where $N_{\mathbf{t}_{i}}$ is the number of words in $\mathbf{t}_{i}$.

In this work, we conduct video-language representation learning for two types of video-language model, \textit{i.e.} dual and bidirectional model. For the dual type, we will pool the visual representations $\mathbf{V}^{f}_{i}$ and textual representations $\mathbf{T}^{w}_{i}$ to obtain global representations $\mathbf{h}^{V}_{i}$ and $\mathbf{h}^{T}_{i}$ for the video and text input, respectively. We use cosine similarity to calculate the cross-modal similarity $S_{\mathbf{v}_{i}, \mathbf{t}_{i}}$. Then, we aim to maximize the similarities $S_{\mathbf{v}_{i}, \mathbf{t}_{i}}$ of video-text pairs within the data, relative to those of the unaligned video-text pairs.

The bidirectional model will concatenate the visual and textual representations into a sequence, then feed the sequence to Transformer attention layers to capture the video-text relations. Subsequently, we will forward the [CLS] token which represents the video-text sequence to a fully-connected (FC) layer. The FC layer will calculate the similarity $S_{\mathbf{v}_{i}, \mathbf{t}_{i}}$ if the video input $\mathbf{v}_{i}$ has a text pairing $\mathbf{t}_{i}$, or calculate the log-likelihood $\log q(y_{i} | \mathbf{v}_{i}, \mathbf{t}_{i})$ of the answer if the video input $\mathbf{v}_{i}$ has a label $y_{i}$. In the former case, similar to the dual type we also maximize the similarity of in-distribution video-text pairs, whereas for the latter case we maximize the log-likelihood of the answer label $y_{i}$.

In general, for all video inputs, we maximize the similarity $S_{\mathbf{v}, \mathbf{t}}$ through minimizing the cross-modal contrastive loss as:
\begin{equation}
\hspace{-35pt}
\small
\mathcal{L}^{v,t}_{\text{contrastive}, i} = -\log \frac{e^{S_{\textbf{v}_{i}, \textbf{t}_{i}}/\tau}}{\sum\limits_{j=1}^{B}e^{S_{\textbf{v}_{i}, \textbf{t}_{j}}/\tau}}, \;\;\;\; \mathcal{L}^{t,v}_{\text{contrastive}, i} = - \log \frac{e^{S_{\textbf{t}_{i}, \textbf{v}_{i}}/\tau}}{\sum\limits_{j=1}^{B}e^{S_{\textbf{t}_{i}, \textbf{v}_{j}}}/\tau},
\end{equation}
where $B$ is the batch size and $\tau$ is the temperature hyperparameter. If a video $\mathbf{v}_{i}$ additionally exhibits an answer label $y_{i}$, we jointly combine the cross-entropy loss to maximize the log-likelihood of the label with the above contrastive loss:
\begin{gather}
\small
\mathcal{L}_{\text{CE}, i} = - \sum\limits_{y_{i}} p(y_{i}|\mathbf{v}_{i}, \mathbf{t}_{i}) \log q(y_{i}|\mathbf{v}_{i}, \mathbf{t}_{i}), \\  \mathcal{L}_{\text{train},i} =  \mathcal{L}^{v,t}_{\text{contrastive},i} + \mathcal{L}^{t,v}_{\text{contrastive},i} + \eta \cdot \mathcal{L}_{\text{CE},i},
\label{eq:total_training_objective}
\end{gather}

\noindent where $\eta$ is a hyperparameter to balance both losses. It is worth noting that $\eta = 0$ if a video input $\mathbf{v}_{i}$ is not attached with $y_{i}$.
\subsection{Meta-optimized Angular Margin Contrastive Framework}
\begin{algorithm}[t]
\footnotesize	
    \renewcommand{\algorithmicrequire}{\textbf{Input:}}
    \renewcommand{\algorithmicensure}{\textbf{Output:}}
    \caption{Our meta-optimized learning framework}
    \label{alg:example}
    \begin{algorithmic}[1]  \small
        \Require  Training data $\mathcal{D}$, meta-data $\widehat{\mathcal{D}}$, training batch size $B$, meta-data batch size $M$, initialized video-language model parameter $\Theta^{(0)}$ and MLP network parameter $\theta^{(0)}$
        \Ensure  Video-language model parameter $\Theta^{(K)}$
        \For{$k=0$ {\bfseries to} $K-1$}
        \State Sample a training minibatch $\{\textbf{v},\textbf{t},y\}$.
        \State Sample a meta-data minibatch $\{\textbf{v}^{\text{meta}},\textbf{t}^{\text{meta}},y^{\text{meta}}\}$.
        \State Estimate $\hat{\Theta}^{(t)}$ by Eq. (\ref{eq:estimate_video_language_model}).
        \State Update $\theta^{(k+1)}$ by Eq. (\ref{eq:update_mlp_network}).
        \State Update $\Theta^{(k+1)}$ by Eq. (\ref{eq:update_video_language_model}).
        \EndFor
    \end{algorithmic}
    \label{algo:meta_optimized_framework}
\end{algorithm}
\noindent\textbf{Subtractive angular margin contrastive learning.} 
As cosine similarity is used to calculate $S_{\mathbf{v}_{i}, \mathbf{t}_{j}} \in [-1, 1]$, we denote the angle between the representation of video $i$ and text $j$ as:
\begin{equation}
\small
\lambda_{i,j} = \arccos \left(S_{\mathbf{v}_{i}, \mathbf{t}_{j}}\right).
\label{eq:subtractive_angular_margin_contrastive}
\end{equation}
The original video-language representation learning minimizes $\lambda_{i,i}$ to approach a 0-degree angle. However, since the video and its textual description commonly does not perfectly align with each other, we want the gradient to be regularized when the similarity of positive pair $\textbf{v}_{i}$ and $\textbf{t}_{i}$ becomes small. Particularly, we replace $\mathcal{L}_{\text{contrastive},i}^{v,t}$ by a new training objective:
\[
\mathcal{L}_{\text{angular},i}^{v,t} =
\begin{cases}  
-\log\frac{e^{\cos\left([\lambda_{i,i}-\mu]_{+}\right)/\tau}}{e^{\cos\left([\lambda_{i,i}-\mu]_{+}\right)/\tau} + \sum\limits_{j \neq i} e^{\cos(\lambda_{i,j})/\tau}}, & \text{if } \lambda_{i,i} \leq \frac{\pi}{2} \\
-\log\frac{e^{\cos(\lambda_{i,i})/\tau}}{e^{\cos(\lambda_{i,i})/\tau} + \sum\limits_{j \neq i} e^{\cos(\lambda_{i,j})/\tau}}, & \text{otherwise},
\end{cases}
\]

\noindent A similar objective $\mathcal{L}_{\text{angular},i}^{t,v}$ is formulated between $\mathbf{t}_i$ and $\mathbf{v}_i$. As long as the angular difference $\lambda_{i,i}$ is smaller than $\mu$, the similarity score will become 1. On the other hand, if $\lambda_{i,i}$ is larger but starts to turn small, \textit{i.e.} $\mu \leq \lambda_{i,i} \leq \frac{\pi}{2}$, the subtractive margin $\mu$ will temporarily pull the positive video and text closer to regularize the gradient of the positive samples, thus restraining them from reaching perfect similarity. This intuition is formalized by the following theorem:

\begin{theorem}
Let $\lambda_{i,j}$ denote the angle between the representation of two samples $i, j$, $\mathcal{L}^{v,t}_{\textup{angular}, i}$ and $\mathcal{L}^{v,t}_{\textup{contrastive}, i}$ denote the training objectives with and without the angular margin, respectively. Then, if $\lambda_{i,i} \leq \frac{\pi}{2}$, the following inequality holds:
\begin{equation}
\small
\left|\frac{\partial \mathcal{L}_{\textup{angular},i}^{v,t}}{\partial \lambda_{i,i}}\right| \leq \left|\frac{\partial \mathcal{L}_{\textup{contrastive}, i}^{v,t}}{\partial \lambda_{i,i}}\right| 
\end{equation}
\end{theorem}
In order to speed up the training in the beginning and constrain the update of positive pairs in the latter stage, we adopt an adaptive strategy that gradually increases the margin $\mu$ towards a limit:
\begin{equation}
\small
\mu^{(k)} = \frac{a_{0}}{a_{1} + e^{-a_{2} \cdot k}},
\end{equation}
where $k$ denotes the training step, and $a_{0}$, $a_{1}$, and $a_{2}$ denote hyperparameters.

\noindent\textbf{MLP-parameterized weighting function.} To control the effect of the data imbalance issue, we  construct a weighting function as a $\theta$-parameterized MLP to map each training loss to a sample weight:
\begin{equation}
\small
w(\theta, \mathcal{L}_{i}^{\text{train}}(\Theta)) = \text{MLP}_{\theta}(\mathcal{L}_{i}^{\text{train}}(\Theta)), \;\; i \in \{1, 2, …, B\}, 
\end{equation}
where $\mathcal{L}_{i}^{\text{train}}(\Theta)$ denotes the training loss of sample $i$ in a batch of size $B$, calculated using the $\Theta$-parameterized video-language model.

Then, the final training objective is the sum of the training losses of data samples weighted by $w$:
\begin{equation}
\small
\mathcal{L}(\Theta, \theta) = \frac{1}{B}\sum\limits_{i=1}^{B} w(\theta, \mathcal{L}_{i}^{\text{train}}(\Theta)) \cdot \mathcal{L}_{i}^{\text{train}}(\Theta).
\end{equation}
\noindent\textbf{Meta-optimized learning framework.} At present, we have a $\theta$-parameterized MLP network and a $\Theta$-parameterized video-language model to train. Since jointly training these two models might be unstable \citep{shu2019meta}, we instead develop a meta-learning approach. In our procedure, we manually extract a small amount of unbiased meta-data set $\{\textbf{v}_{i}^{\text{meta}}, \textbf{t}_{i}^{\text{meta}}, y_{i}^{\text{meta}}\}_{i=1}^{M}$, i.e. with semantically aligned textual description and balanced label distribution, representing the meta-knowledge of the groundtruth sample-label distribution, where $M$ is the number of the meta-samples and $M \ll N$. 

We first estimate the update of the video-language model parameters $\Theta$:
\begin{equation}
\small
\hat{\Theta}^{(k)} = \Theta^{(k)} - \frac{\alpha}{B} \sum\limits_{j=1}^{B} w_{i}\left(\theta^{(k)}, \mathcal{L}_{j}^{\text{train}}\left(\Theta^{(k)}\right)\right) \cdot \nabla_{\Theta} \mathcal{L}_{j}^{\text{train}}\left(\Theta^{(k)}\right),
\label{eq:estimate_video_language_model}
\end{equation}
where $\alpha$ is the learning rate for the video-language model. Subsequently, we update the MLP network using the estimated $\hat{\Theta}$:
\begin{equation}
\small
\theta^{(k+1)} = \theta^{(k)} - \frac{\eta}{M} \sum\limits_{i=1}^{M} \nabla_{\theta} \mathcal{L}^{\text{meta}}_{i} \left(\hat{\Theta}^{(k)}\right),
\label{eq:update_mlp_network}
\end{equation}
where $\beta$ is the learning rate for the MLP network. Lastly, we obtain the new video-language model’s parameters as:
\begin{equation}
\begin{split}
\Theta^{(k+1)} = \Theta^{(k)} - \frac{\alpha}{B}\sum\limits_{j=1}^{B} w\left(\theta^{(k+1)}, \mathcal{L}_{j}^{\text{train}}\left(\Theta^{(k)}\right)\right) \cdot \nabla_{\Theta}\mathcal{L}^{\text{train}}_{j}\left(\Theta\right)\Big|_{\Theta^{(k)}}.
\end{split}
\label{eq:update_video_language_model}
\end{equation}
We summarize our algorithm in Algorithm \ref{algo:meta_optimized_framework}. From an empirical perspective, we compare the effectiveness of joint learning and our meta-optimized framework in Section \ref{subsect:ablation_study}. From a theoretical perspective, we conduct further derivation for Eq. (\ref{eq:update_mlp_network}), resulting in the following formulation:
\begin{gather}
\theta^{(k+1)} = \theta^{(k)} + \frac{\alpha\beta}{BM} \sum_{i=1}^{M}\sum_{j=1}^{B} G_{ij} \frac{\partial w\left(\theta^{(k)}, \mathcal{L}_{j}^{\text{train}}\left(\Theta^{(k)}\right)\right)}{\partial\theta}\Big|_{\theta^{(k)}}, \\
G_{ij} = \nabla_{\hat{\Theta}} \mathcal{L}_{i}^{\text{meta}} \left(\hat{\Theta}\right)\Big|_{\hat{\Theta}^{(k)}} \cdot \nabla_{\Theta} \mathcal{L}_{j}^{\text{train}} \left(\Theta\right)\Big|_{\Theta^{(k)}},
\end{gather}
where $G_{ij}$ denotes the coefficient of the gradient. The coefficient will increase for training samples whose gradient is in the same direction with the gradient of the meta-data (meta gradient), as this is more likely to be a clean learning direction. In contrast, the coefficient will decrease for those whose gradient is opposite from the meta gradient, since the learning direction is likely to be noisy.
\subsection{Large Vision-Language Model for Augmentation}
\noindent To further enhance our video-language representation learning, we devise a strategy to utilize large vision-language model to augment additional video-text data. Our strategy consists of two stages, \textit{i.e.} extracting key frames and generating textual descriptions, which we delineate as follows:

\noindent\textbf{Key frame extraction.} Inspired by \citep{jin2023video}, we adopt a density peak-based clustering approach to extract key video frames. Starting with a sequence of visual frame features $\textbf{V}_{i}^{f} = \{\mathbf{h}^{f}_{i,j}\}_{j=1}^{N_{\textbf{v}_{i}}}$, we calculate the locality density $d_{i,j}$ of each feature $\mathbf{h}^{f}_{i,j}$ based on its $K$-nearest neighbors:
\begin{equation}
\small
d_{i,j} = \exp\left(-\frac{1}{K} \sum\limits_{\textbf{h}^{f}_{i,l} \in \text{KNN}(\mathbf{h}^{f}_{i,j})} \left|\left|\textbf{h}^{f}_{i,l} - \textbf{h}^{f}_{i,j}\right|\right|^{2}\right),
\end{equation}
where $\text{KNN}(\mathbf{h}^{f}_{i,j})$ denotes the $K$-nearest neighbors of $\mathbf{h}^{f}_{i,j}$.

Subsequently, we estimate the distance index $\gamma_{i,j}$ of each frame $\mathbf{h}_{i,j}^{f}$:
\begin{align}
\small
\gamma_{i,j}=
\begin{cases}
\min\limits_{l: d_{i,j} > d_{i,l}} ||\mathbf{h}^{f}_{i,l} - \mathbf{h}^{f}_{i,j}||^{2},& \text{if } \exists l \; \text{s.t.} \; d_{i,l} > d_{i,j} \\
\max\limits_{l} ||\mathbf{h}^{f}_{i,l} - \mathbf{h}^{f}_{i,j}||,& \text{otherwise}
\end{cases}.
\end{align}

Our intuition is that $d_{i,j}$ denotes the local density of a video frame, and $\gamma_{i,j}$ denotes the distance from other frames of high density. We proceed to extract the top $Q$ video frames of the highest $d_{i,j} \times \gamma_{i,j}$ values as key video frames.

\noindent\textbf{Textual description generation.} 
Since large vision-language models have established impressive results on natural instruction tuning and visual reasoning capabilities, we leverage a LVLM to augment textual descriptions for video data \citep{dai2305instructblip, li2023blip, liu2023visual}. Particularly, given the top $Q$ video frames, we concatenate them into a single image $I$ of the $W \times H$ grid. We then forward the image $I$ along with the textual prompt ``\textit{Write a short caption sentence for the video in order from left to right, top to bottom}’’ to a large vision-language model to generate the sentence-level textual description $\textbf{t}$, which we combine with the video $\textbf{v}$ to form a video-text pair $(\textbf{v},\textbf{t})$. 

\section{Experiments}
\subsection{Experimental Settings}
\noindent\textbf{Downstream tasks.} Following previous works \citep{jin2023video, sun2019videobert, zhao2023learning, man2023tevl}, we evaluate our framework on two popular video question answering and text$\leftrightarrow$video retrieval tasks. 
\begin{itemize}
    \item Video question answering (VideoQA): We experiment with two videoQA settings, \textit{i.e.} open-ended and multi-choice videoQA. Open-ended videoQA classifies a pair of video and question into one of the pre-defined set of answer labels. Multi-choice videoQA chooses the correct answer from five choices given the video and the language question. We assess our video-language representations by the VideoQA task using the following five datasets,  MSRVTT \citep{xu2016msr}, MSVD \citep{chen2011collecting}, TGIF-QA-R \citep{peng2021progressive}, NExT-QA \citep{xiao2021next}, and Causal-VidQA \citep{li2022representation}. 
    \item Text$\leftrightarrow$video retrieval: The retrieval task is to extract the corresponding video given the textual query, or extract the textual description given the video. We evaluate the retrieval ability of our video-language representations using three datasets: MSRVTT \citep{xu2016msr}, DiDeMo \citep{anne2017localizing}, ActivityNet \citep{krishna2017dense}.
\end{itemize}
\noindent\textbf{Evaluation metrics.} For the videoQA task, we use the answer accuracy as the evaluation metric. For the retrieval task, we employ Recall at rank $K$ ($R@K$), with $K \in \{1, 5, 10\}$ to evaluate the performance.

\noindent\textbf{Video-language backbones.} To extensively validate the effectiveness of our framework, we conduct experiments on various models of bidirectional and dual architectures. Particularly, for the dual architecture, we experiment with CLIP-ViP \citep{xue2022clip} and video graph transformer (VGT) \citep{xiao2023contrastive}, while for the bidirectional architecture, we experiment with the well-known VIOLET architecture \citep{fu2023empirical}.

\noindent\textbf{Baseline methods.} We compare our method with a comprehensive list of video-language understanding models, along with the approaches that use LVLM to augment video-text data: (i) \textbf{CLIP4Clip} \citep{luo2022clip4clip}, a model that transfers image-text CLIP model \citep{radford2021learning} to text$\leftrightarrow$video retrieval tasks; (ii) \textbf{MERLOT} \citep{zellers2021merlot}, a method that trains on both spatial and temporal objectives to learn video-language representations; (iii) \textbf{LAVENDER} \citep{li2023lavender}, a model that learns video-language representations using a unified framework of masked language modeling; (iv) \textbf{Singularity} \citep{lei2022revealing}, curates a pre-training dataset and uses an early fusion strategy to improve single-frame video-language representation learning; (v) \textbf{OmniVL} \citep{wang2022omnivl}: a model that is trained on both image-language and video-language data to enhance video-language representation learning; (vi) \textbf{VindLU} \citep{cheng2023vindlu}, follows a recipe to select pre-training objectives, pre-training data, and model scale for effective video-language representation learning; (vii) \textbf{All-in-one} \citep{wang2023all}, a model that embeds raw video and textual signals into hidden representations without using pre-trained unimodal encoders; (viii) \textbf{DRL} \citep{wang2022disentangled}, a disentangled representation method that decouples sequential from hierarchical representations to specifically improve both of them for cross-modal retrieval; (ix) \textbf{CLIP2Video} \citep{bogolin2022cross}, consists of two normalization methods to improve the effectiveness and robustness of cross-modal retrieval models; (x) \textbf{LaViLa} \citep{zhao2023learning}, an approach that fine-tunes LLM to be conditioned on visual frames to generate additional textual descriptions, which are used to improve video-language representation learning; (xi) \textbf{Vid2Seq} \citep{yang2023vid2seq}, a pre-trained dense video captioning model which is used to generate additional textual descriptions for video data, which are employed to enhance video-language representation learning.

\noindent\textbf{Implementation details.} We utilize LLaVA model as the LVLM \citep{liu2023visual} to augment video inputs of the HowTo100M dataset \citep{miech2019howto100m}, which is a large-scale dataset for video-language representation learning but has been known for its weakly aligned video-text pairings \citep{han2022temporal}. To construct the input for the LVLM, we use $K = 6$ and $Q = 12$, then concatenate key frames into an image as a grid of $3\times4$ frames. Based upon validation, we adopt $\tau = 1$, $a_{0} = 0.2, a_{1} = 10$, and $a_{2} = -0.1$ for angular margin contrastive learning, and $\eta = 1$ for our optimization objective. For videoQA, to fairly compare with previous methods, we fine-tune our VIOLET-based model using AdamW with an initial learning rate of $\alpha = \beta = 2e-5$, betas of (0.9, 0.98), weight decay of 1e-3, and batch size $B$ of 4 for 10 epochs. Similarly, for the VGT-based model, we use Adam optimizer with an initial learning rate of 1e-5 of a cosine annealing schedule, and batch size $B$ of 4 for 30 epochs. For the text$\leftrightarrow$video retrieval task, we fine-tune the model with an initial learning rate of 5e-6 and a fixed weight decay of 5e-2, using a batch size $B$ of 4 for 5, 20, and 20 epochs on MSRVTT, DiDeMo, and ActivityNet datasets, respectively. 

\subsection{Main Results}
\begin{table}[t]
\centering
\caption{VideoQA results on MSRVTT, MSVD, and TGIF-QA-R. Open-ended videoQA is evaluated on MSRVTT, MSVD, and TGIF-Frame datasets. Multi-choice videoQA is evaluated upon TGIF-Action and TGIF-Transition.}
\label{tab:videoqa_msrvtt_msvd_tgif_qa}
\resizebox{0.8\linewidth}{!}{
\begin{tabular}{l|c|c|ccc}
\hline
\multicolumn{1}{c|}{\multirow{2}{*}{\textbf{Method}}} & \multirow{2}{*}{\textbf{MSRVTT-QA}} & \multirow{2}{*}{\textbf{MSVD-QA}} & \multicolumn{3}{c}{\textbf{TGIF-QA-R}}                 \\
                               &                                     &                                   & \textbf{Action} & \textbf{Frame} & \textbf{Transition} \\ \hline
MERLOT                                              & 43.1                                &   51.9	                              &      61.4	            &         69.3	             &              84.0  \\
LAVENDER                                    & 44.2                                &             52.4                      &   57.1              &          66.9           &     84.0           \\
Singularity                                          & 43.5                                &         49.6                          &   53.1	              &       65.1              &         	81.5       \\
OmniVL                                            & 44.1                                &          51.0                         &     62.0	            &        69.5             &       85.5         \\
VindLU                                              & 44.6                                &         51.0                          &    59.5             &      65.8               &     87.2           \\
VGT                                                & 40.0                                &          36.4                         & 61.0            &       61.7              & 73.2           \\
VIOLET                                        &         44.5                            &         52.5                          &          62.6       &       70.0              &          86.3      \\
All-in-one                                          &         44.3                            &       47.9                            &       34.9          &     65.0                &       53.8         \\
LaViLa-VIOLET                                               &             44.9
                     &             53.7                      &     64.3		            &   70.7                  &          87.0      \\
Vid2Seq-VIOLET                                            &           44.8                             &  53.1                                 &     64.2		            &       70.1              &         86.6       \\ \hline
MAMA-VGT                               &    41.6	                                  &       37.1                            &      61.7		           &              62.7       &      74.0          \\
MAMA-VIOLET                            & \textbf{46.4}                                & \textbf{55.8}                             & \textbf{66.5}            & \textbf{71.7}                & \textbf{89.5}           \\ \hline
\end{tabular}}
\end{table}

\begin{table}[t]
\centering
\caption{VideoQA results on NExT-QA and Causal-VidQA. All of the datasets target multi-choice videoQA.}
\label{tab:videoqa_nextqa_causalvid_qa}
\resizebox{\linewidth}{!}{
\begin{tabular}{l|cccc|cccc}
\hline
\multirow{2}{*}{\textbf{Method}} & \multicolumn{4}{c|}{\textbf{NExT-QA}}                                                                                                  & \multicolumn{4}{c}{\textbf{Causal-VidQA}}                                                                                                                                   \\ 
                              & \textbf{Descriptive} & \textbf{Temporal} & \textbf{Causal} & \textbf{All} & \textbf{Descriptive} & \textbf{Explanatory} & \textbf{Predictive} & \textbf{Counterfactual} \\ \hline
MERLOT                                             &        66.6			              &                        62.5	               &                              58.2	       &           59.4                       &            67.2                              &                   65.7	                       &                         57.2	                &          57.0                                   \\
LAVENDER                                           &     63.4                 &  56.5                                      &     54.6                                &          56.7                        &         62.0                                 &          61.6                                &     46.3                                    &     50.4                                        \\
Singularity                                          &      60.0						                &        61.1                               &    50.9                                 &          54.8                        &   54.3                                       &                  49.6                        &            41.3                             &                  46.9                           \\
OmniVL                                                &          67.1	            &       63.8                             &         55.9                             &             59.9                     &             67.0                             &                  66.0                        &                       56.5                  &        57.1                                     \\
VindLU                                               &          68.4       &      59.7                                 &                55.7                      &   59.8                             &        63.6			                                  &           54.8                               &       57.7                                  &                 54.4                            \\
VGT                                              &     69.6                &     65.4            & 56.2            &      61.8      & 74.4                 & 75.6                 & 60.7                & 65.6                    \\
VIOLET                                         &          67.7           &    58.0                                   &       50.7                              &              58.5                    &      67.6			                                    &        66.6                                  &         57.1                                &         57.6                                    \\
All-in-one                                        &   	64.8							                   &       63.9                               &           50.9                          &          57.9                      &             60.4                          &                        51.3                   &                       51.2                     &                         50.7                     \\
LaViLa-VGT                                              &          		70.9     				            &                69.2                    &                            59.3       &                        64.8          &                            74.7	              &       76.1                                   &        61.6                                &               65.7                              \\
Vid2Seq-VGT                                             &   70.3					                   &                        67.0	               &          58.5                           &             64.2                     &                         73.7                 &          76.0                                &                    61.3	                     &                   65.6                          \\ \hline
MAMA-VGT                               &  \textbf{72.7}				                 &    \textbf{70.6}	              &        \textbf{62.2}         &     \textbf{66.3}       &                \textbf{75.3}     &     \textbf{77.1}	        &    \textbf{62.2}	                  &         \textbf{68.2}               \\
MAMA-VIOLET                            & 71.2                 & 68.6              & 61.1            & 65.8         & 72.7                 & 68.0                 & 60.5                & 59.5   \\ \hline
\end{tabular}}
\end{table}

\begin{table}[t]
\centering
\caption{Text$\rightarrow$video retrieval results on MSRVTT, DiDeMo, and ActivityNet.}
\label{tab:text_video_retrieval_msrvtt_didemo_activitynet}
\resizebox{\linewidth}{!}{
\begin{tabular}{l|ccc|ccc|ccc}
\hline
\multirow{2}{*}{\textbf{Method}} & \multicolumn{3}{c|}{\textbf{MSRVTT}}         & \multicolumn{3}{c|}{\textbf{DiDeMo}}         & \multicolumn{3}{c}{\textbf{ActivityNet}}    \\
                                 & \textbf{R@1} & \textbf{R@5} & \textbf{R@10} & \textbf{R@1} & \textbf{R@5} & \textbf{R@10} & \textbf{R@1} & \textbf{R@5} & \textbf{R@10} \\ \hline
VIOLET                          & 37.2         & 64.8         & 75.8          & 47.9         & 76.5         & 84.1          &   18.1           &      43.1        &    56.7           \\
VindLU                          & 48.8         & 72.4         & 82.2          & 59.8         & 86.6         & 91.5          & 55.9         & 82.3         & 90.9          \\
CLIP4Clip                        & 44.5         & 71.4         & 81.6          & 42.8         & 68.5         & 79.2          & 40.5         & 72.4         & 83.4          \\
CLIP2Video                     & 47.2         & 73.0         & 83.0          &     -         &        -      &      -         & -            & -            & -             \\
DRL                             & 47.4         & 74.6         & 83.8          & 47.9         & 73.8         & 82.7          & 44.2         & 74.5         & 86.1          \\
CLIP-ViP                         & 55.7         & 77.7         & 86.6          & 55.7         & 78.1         & 86.1          & 59.1         & 83.9         & 91.3          \\ 
LaViLa - CLIP-ViP                      &   56.0       &     79.8     &      87.2     &      56.6    &    79.8      &   87.1        &      58.7    &    82.8      &     90.5      \\ 
Vid2Seq - CLIP-ViP                      &      55.3	     &    77.9      &    86.1       &   57.6       &   79.9       &    88.4       &       55.1    &     79.0     &     87.4       \\ \hline
MAMA - CLIP-ViP      & \textbf{60.0}         & \textbf{82.2}         & \textbf{89.2}          &    \textbf{62.7}          &       \textbf{89.9}       &     \textbf{96.0}          & \textbf{60.0}         & \textbf{85.0}         & \textbf{91.7}       \\ \hline
\end{tabular}}
\end{table}

\begin{table}[h!]
\centering
\caption{Video$\rightarrow$text retrieval results on MSRVTT, DiDeMo, and ActivityNet.}
\label{tab:video_text_retrieval_msrvtt_didemo_activitynet}
\resizebox{\linewidth}{!}{
\begin{tabular}{l|ccc|ccc|ccc}
\hline
\multirow{2}{*}{\textbf{Method}} & \multicolumn{3}{c|}{\textbf{MSRVTT}}         & \multicolumn{3}{c|}{\textbf{DiDeMo}}         & \multicolumn{3}{c}{\textbf{ActivityNet}}    \\
                                 & \textbf{R@1} & \textbf{R@5} & \textbf{R@10} & \textbf{R@1} & \textbf{R@5} & \textbf{R@10} & \textbf{R@1} & \textbf{R@5} & \textbf{R@10} \\ \hline
VIOLET                          &     36.6		         &        64.1      &       75.1         &     44.8    &      72.9        &        82.4       &      15.8        &      39.8        &      54.8         \\
VindLU                          &     46.0	         &   71.8           &        80.2       &         57.2		     &       83.1       &        88.9       &     47.1	        &    77.5          &      86.6         \\
CLIP4Clip                        & 43.1         & 70.5         & 81.2          &   17.5           &    37.5          &       49.4        & 42.6         & 73.4         & 85.6          \\
CLIP2Video                     &  43.5		            &      72.3        &         82.1       &       -      &        -      &     -          &         -     &       -       &       -        \\
DRL                          &       45.1		       &     72.9            &   83.5         &   45.4     &  72.6         &   82.1           &      42.2          &           74.0   &    86.2           \\
CLIP-ViP                         & 48.0         & 72.4         & 82.9          & 46.3         & 73.2         & 81.9          & 50.2         & 78.3         & 87.5          \\ 
LaViLa - CLIP-ViP                     &     49.0     &   74.5       &  83.4         &     47.1     &      74.2    &    83.0       &   50.3       &  78.7        &     88.4      \\ 
Vid2Seq - CLIP-ViP                        &  49.1			       &   74.1       & 82.5          &    47.2      &    74.4      &   83.1        &    50.9		     & 79.5         &      89.6    \\  \hline
Our method - CLIP-ViP      &   \textbf{50.1}         &   \textbf{76.8}         &   \textbf{84.8}          &   \textbf{48.1}         &   \textbf{78.1}         &   \textbf{85.5}          &   \textbf{52.3}         &   \textbf{80.8}         &   \textbf{90.1}        \\ \hline 
\end{tabular}}
\end{table}

\noindent We denote the results for videoQA in Table \ref{tab:videoqa_msrvtt_msvd_tgif_qa} and \ref{tab:videoqa_nextqa_causalvid_qa}, for text$\rightarrow$video retrieval in Table \ref{tab:text_video_retrieval_msrvtt_didemo_activitynet}, and for video$\rightarrow$text retrieval in Table \ref{tab:video_text_retrieval_msrvtt_didemo_activitynet}.

\noindent\textbf{VideoQA.} In terms of open-ended videoQA, we substantially outperform the LVLM approaches, \textit{e.g.} improve upon LaViLa with improvements of 1.5\% on MSRVTT, 2.1\% on MSVD, and 3.7\% on TGIF-Frame. We also surpass previous video-language understanding models, \textit{e.g.} VindLU by 1.8\% on MSRVTT, VIOLET by 3.3\% on MSVD, and OmniVL by 2.2\% on TGIF-Frame. Moreover, for multi-choice videoQA, we consistently surpass both LVLM and video-language understanding models, \textit{e.g.} achieving an overall gain of 1.5\% on NExT-QA and 1.2\% on Causal-VidQA over LaViLa, while the gains over VGT are 4.5\% and 1.6\%, respectively on these two datasets.

\noindent\textbf{Text$\leftrightarrow$video retrieval.} Our observation in videoQA applies for the text$\leftrightarrow$video retrieval task. In the text$\rightarrow$video direction, we enhance upon CLIP-ViP by 2.6 R@10 points on MSRVTT, 5.5 R@1 points on DiDeMo, and 1.1 R@5 points on ActivityNet. We also significantly outperform the LVLM approaches, \textit{e.g.} by 4.0 R@1 points and by 2.4 R@5 points on MSRVTT over LaViLa. Our superiority on video$\rightarrow$text retrieval is analogous to the text$\rightarrow$video case. 

These results substantiate that our framework is applicable to various video-language understanding tasks and model architectures. We hypothesize that we can better control the alignment between the positive video and language, and their effect upon the model training, thus leading to more reasonable video-language representations.

\subsection{Ablation Study}
\label{subsect:ablation_study}
\noindent We ablate our meta-optimized framework to investigate which factor results in the demonstrated effectiveness and explore intriguing properties. We conduct all experiments of our ablation study on the MSRVTT and MSVD for videoQA, and MSRVTT and DiDeMo for the text$\leftrightarrow$video retrieval task.

\begin{figure}[t]
\centering
\caption{(Left) Validation videoQA accuracy on MSVD with respect to $\mu$; (Middle) Validation videoQA accuracy on MSVD with respect to $a_0$; (Right) Relationship between loss values and weight values generated by our MLP-parameterized weighting function.}
\label{fig:analysis_mu_a0_weight}
\begin{subfigure}{0.30\linewidth}
    \centering
  \includegraphics[width=\linewidth]{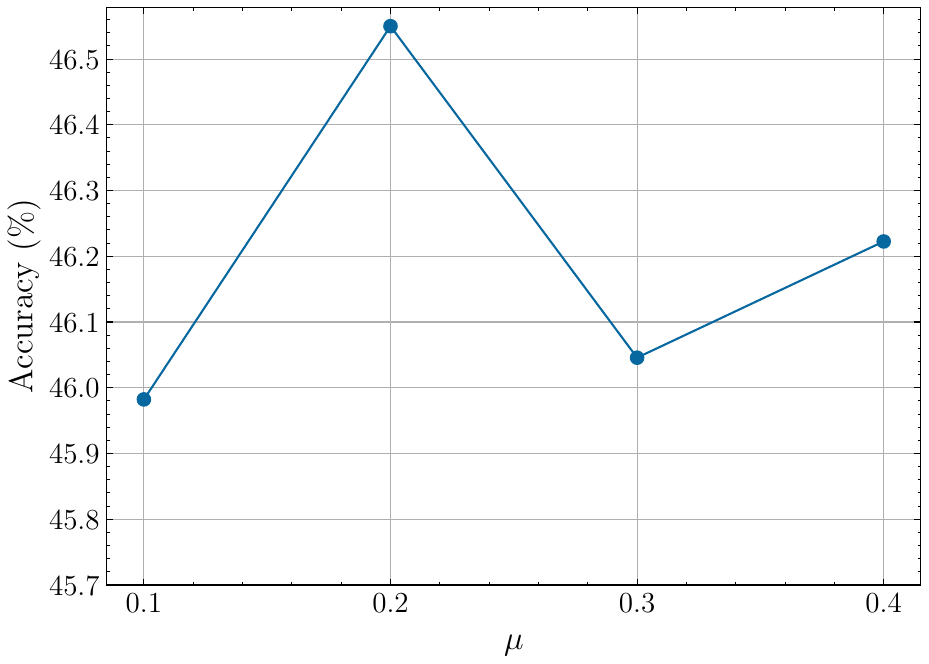} 
\end{subfigure}
\begin{subfigure}{0.30\linewidth}
    \centering
  \includegraphics[width=\linewidth]{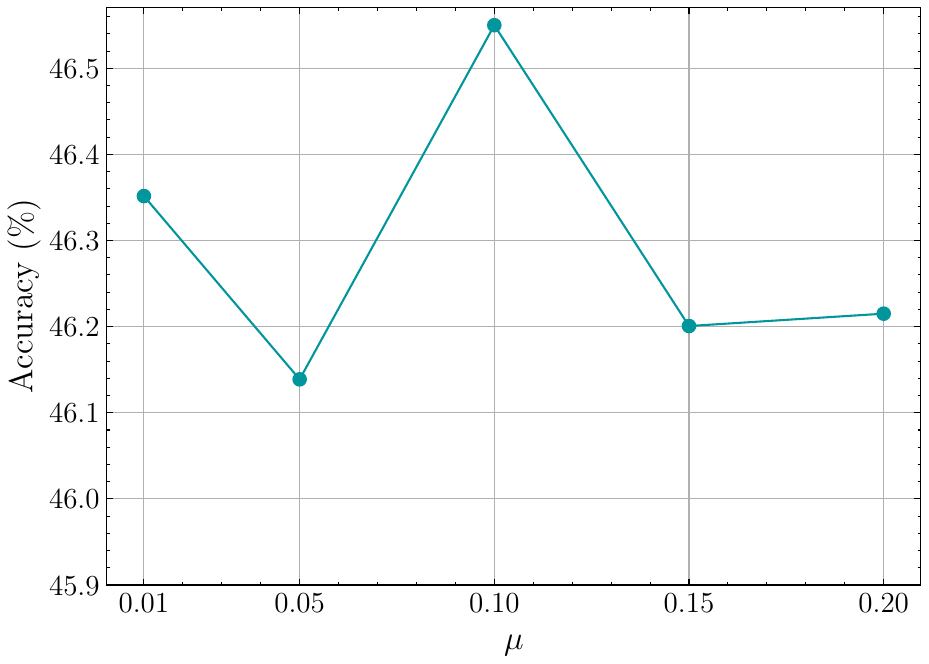} 
\end{subfigure} 
\begin{subfigure}{0.30\linewidth}
    \centering
  \includegraphics[width=\linewidth]{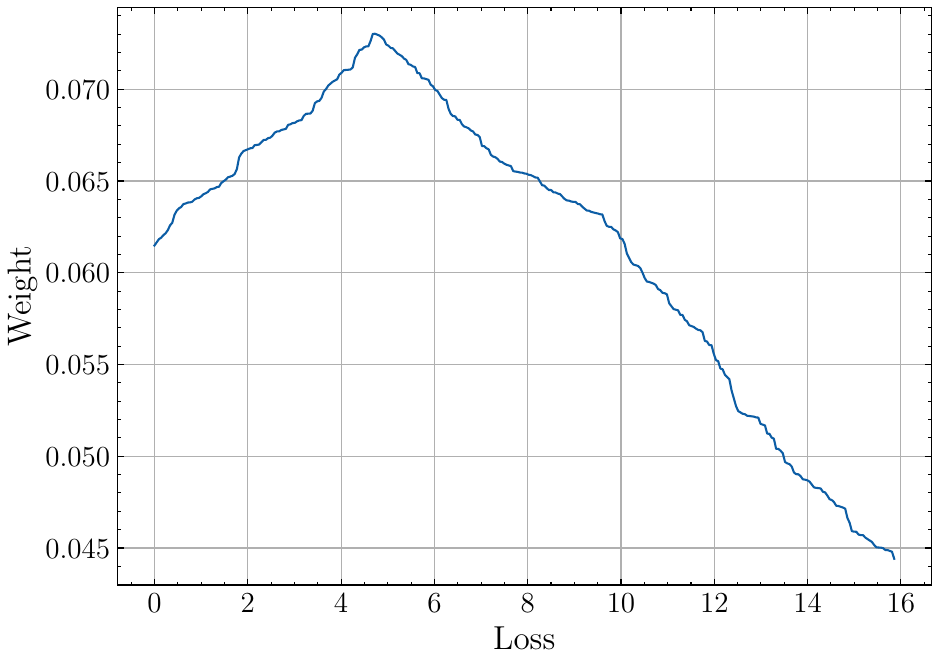} 
\end{subfigure} 
\end{figure}

\noindent\textbf{Varying the angular margin.} The angular margin, represented as $\mu$, is critical to control the relationship between semantically close video and language inputs. To better understand $\mu$, we experiment with manually varying the margin $\mu$ from 0.1 to 0.4 in Figure \ref{fig:analysis_mu_a0_weight} (left). We observe that the optimal performance is achieved when $\mu \sim 0.2$. Based upon this observation, we adopt $a_0 = 2$ and $a_{1} = 10$. Subsequently, we evaluate the impact of $a_2 \in \{0.01, 0.05, 0.1, 0.15, 0.2\}$ in Figure \ref{fig:analysis_mu_a0_weight} (right). We discover that $a_{2} = 0.1$ achieves the best performance and we adopt it for all experiments.

\noindent\textbf{Reweighting strategies.}  In addition to MLP, there exist various approaches to weigh training losses to control noisy pairings and data imbalance issues within the training data, \textit{\textit{e.g.}} focal loss \citep{lin2017focal}, self-paced learning (SPL) \citep{jiang2014self}, and L2RW \citep{ren2018learning}. We replace our MLP by these approaches and demonstrate the results in Table \ref{tab:exp_weighting_functions}. We find that our MLP-parameterized function significantly improves upon previous weighting functions that use manually designed formulation, \textit{i.e.} focal loss and SPL, and also the meta-learning approach without using MLP, \textit{i.e.} L2RW. The possible reason could be that MLP is a universal approximator, so that it can adaptively learn a reasonable weighting function based on data, thus outperforming methods that do not employ MLP.

\begin{table}[t]
\centering
\caption{Experiments on the LVLM choice for our prompting strategy.}    
\label{tab:exp_lvlm_choice}
\resizebox{\linewidth}{!}{
\begin{tabular}{l|cc|ccc|ccc}
\hline
\multirow{3}{*}{\textbf{LVLM choice}} & \multicolumn{2}{c|}{\textbf{VideoQA}}                              & \multicolumn{6}{c}{\textbf{Text$\rightarrow$video retrieval}}                           \\ \cline{2-9}
                                      & \multirow{2}{*}{\textbf{MSRVTT}} & \multirow{2}{*}{\textbf{MSVD}} & \multicolumn{3}{c|}{\textbf{MSRVTT}}         & \multicolumn{3}{c}{\textbf{DiDeMo}}         \\
                                      &                                  &                                & \textbf{R@1} & \textbf{R@5} & \textbf{R@10} & \textbf{R@1} & \textbf{R@5} & \textbf{R@10} \\ \hline
No augmentation                              &           45.6	                     &            54.7                    & 59.0		         &    80.3        &       87.6         &     58.0		        &  82.9            &    89.6          \\
BLIP-2                               &             45.7	                     &            54.9                    &     59.1		         &    80.4          &       87.7         &     58.1		        &  82.9            &    89.7           \\
InstructBLIP                     &              46.1	                    &             55.2                   &     59.2	         &      80.9	        &        88.1       &  58.5	             &        	83.5      &      89.8         \\
LLaVA-ChatGPT                                 &              46.1	               &           55.0             &      59.0		    &    80.3     &      87.7     &  58.7		    &  83.9     &  90.0   \\ \hline
LLaVA                                & \textbf{46.4}                             & \textbf{55.8}                           & \textbf{60.0}         & \textbf{82.2}         & \textbf{89.2}          & \textbf{62.7}         & \textbf{89.9}         & \textbf{96.0}         \\ \hline
\end{tabular}}
\end{table}

\begin{table}[t]
\centering
\caption{Experiments on weighting functions.}
\label{tab:exp_weighting_functions}
\resizebox{\linewidth}{!}{
\begin{tabular}{l|cc|ccc|ccc}
\hline
\multirow{3}{*}{\textbf{Weighting function}} & \multicolumn{2}{c|}{\textbf{VideoQA}}                              & \multicolumn{6}{c}{\textbf{Text$\rightarrow$video retrieval}}                                                                         \\  \cline{2-9}
                                             & \multirow{2}{*}{\textbf{MSRVTT}} & \multirow{2}{*}{\textbf{MSVD}} & \multicolumn{3}{c|}{\textbf{MSRVTT}}                                & \multicolumn{3}{c}{\textbf{DiDeMo}}                                \\ 
                                             &                                  &                                & \textbf{R@1}         & \textbf{R@5}         & \textbf{R@10}        & \textbf{R@1}         & \textbf{R@5}         & \textbf{R@10}        \\ \hline
Focal loss                                  &      45.0                            &         54.1                       &      58.3                &         80.0             &              86.9        &        59.7              &    85.6                  &        91.8              \\
SPL                                        &           45.3                       &    54.3                            &         58.7        &      80.8                &             88.1         &          61.4            &      87.5                &             93.8         \\
L2RW                                         &         45.9                         &        55.2                        &     59.5	            &        81.8              &      89.6                &             62.0         &       88.8               &      94.6                \\ \hline
MLP                                          &        \textbf{46.4}                                                        &          \textbf{55.8}                       &  \textbf{60.0}         & \textbf{82.2}         & \textbf{89.2}          &    \textbf{62.7}          &       \textbf{89.9}       &     \textbf{96.0} \\ \hline
\end{tabular}}
\end{table}

\noindent\textbf{Optimization strategies.} We explore different strategies to optimize the MLP and video-language model. Particularly, we experiment with the joint learning strategy, which simultaneously updates the parameters of the MLP and the video-language model to minimize the objective in Eq. \ref{eq:total_training_objective}. As shown in Table \ref{tab:exp_learning_strategy}, our meta-learning approach outperforms the joint learning one and achieves the best performance. This empirically validates the effectiveness of our meta-learning strategy which enables the training to follow gradient of the meta-data.

\begin{table}[t]
\centering
\caption{Experiments on the number of key frames $Q$ and the concatenated grid.}
\label{tab:exp_q_grid}
\resizebox{0.9\linewidth}{!}{
\begin{tabular}{c|c|cc|ccc|ccc}
\hline
\multirow{3}{*}{$Q$} & \multirow{3}{*}{\textbf{Grid}} & \multicolumn{2}{c|}{\textbf{VideoQA}}                                                                      & \multicolumn{6}{c}{\textbf{Text$\rightarrow$video retrieval}}                                                                                                                                                   \\ \cline{3-10}
                            &                                & \multirow{2}{*}{\textbf{MSRVTT}} & \multirow{2}{*}{\textbf{MSVD}} & \multicolumn{3}{c|}{\textbf{MSRVTT}}                                                                     & \multicolumn{3}{c}{\textbf{DiDeMo}}                                                                     \\
                            &                                &                                  &                                & \textbf{R@1} & \textbf{R@5} & \textbf{R@10} & \textbf{R@1} & \textbf{R@5} & \multicolumn{1}{c}{\textbf{R@10}} \\ \hline
1                           & $1\times1$                            &            45.0	                    &     55.1                           &    58.2		          &         79.3     &     86.2             &    60.2		   &     85.8       &   93.2          \\ \hline
\multirow{2}{*}{2}          & $1\times2$                            &             45.7	                     &        55.3                        &     	 58.8 	        &      80.0            &     87.0          &   60.5         &    85.8        &   93.3            \\
                            & $2\times1$                            &                      45.9	                                &       55.4                                             &          58.4                        &          79.2                     &               87.0                     &           60.9		                       &             86.1                     &                 94.2                  \\ \hline
\multirow{2}{*}{4}          & $1\times4$                            &           45.9	                       &     55.2                           &   59.1		 	         &   	    80.3      &       88.4          &   61.3		       &    86.3        &        94.6       \\
                            & $2\times2$                            &        45.6	                                              &                55.2                                    &       59.2                            &     80.4                             &          88.4                          &       61.3		                           &             86.7                     &           94.6                        \\ \hline
\multirow{2}{*}{8}          & $2\times4$                           &       45.9	                                               &       55.6                                             &             			59.8		             &     81.3                 &              88.7                    &        61.8		                          &                      89.0             &                     95.3            \\
                            & $4\times2$                            &                45.8	                                      &              55.5                                      &            59.3	                      &        80.7           &       88.5                       &     61.8		                             &           88.9                      &              95.3                       \\ \hline
\multirow{2}{*}{12}         & $3\times4$                            &      \textbf{46.4}                                                        &            \textbf{55.8}                       &    \textbf{60.0}         &   \textbf{82.2}         &   \textbf{89.2}          &      \textbf{62.7}          &         \textbf{89.9}       &       \textbf{96.0}                              \\
                            & $2\times6$                           &                   45.8                                   &           	55.6                                         &       59.3	                           &             81.7	                     &                 89.1                  &   62.7	                               &            	89.2                      &      95.6                             \\ \hline
\multirow{2}{*}{16}         & $2\times8$                            &                   45.8	                                   &             55.1                                       &                58.3		                  &              79.7                    &       87.6                            &              61.9		                    &               88.7                   &           94.8                        \\
                            & $4\times4$                            &                   45.7	                                   &              55.1                                      &         58.9		                         &             80.8                     &        88.7                           &         62.6	                        &              89.1                    &       95.9                             \\ \hline
\end{tabular}}
\end{table}

\noindent\textbf{Video-text data augmentation.} We assess the influence of different video-text data augmentation strategies. In particular, we ablate the LLaVA-generated samples and also explore different LVLM choices, \textit{i.e.} BLIP-2 \citep{li2023blip} and InstructBLIP \citep{dai2305instructblip}. In addition, we also use LLaVA model to generate image caption for each video frame, and ask ChatGPT to write a summary for the concatenated captions of all frames as the textual description. Table \ref{tab:exp_lvlm_choice} shows that neglecting LVLM-generated data results in slight performance degradation, which can be resolved by using other LVLMs, but still lack behind LLaVA model. We conjecture that LLaVA can better follow the language instruction to produce more accurate textual descriptions for videos \citep{liu2023visual}. Moreover, LLaVA-ChatGPT approach also leads to performance degradation, possibly because it is still challenging for ChatGPT to infer temporal relations among captions of separate video frames.

\begin{table}[t]
\centering
\caption{Experiments on the learning strategy.}
\label{tab:exp_learning_strategy}
\resizebox{\linewidth}{!}{
\begin{tabular}{l|cc|ccc|ccc}
\hline
\multirow{3}{*}{\textbf{Training strategy}} & \multicolumn{2}{c|}{\textbf{VideoQA}}                              & \multicolumn{6}{c}{\textbf{Text$\rightarrow$video retrieval}}                           \\ \cline{2-9}
                                            & \multirow{2}{*}{\textbf{MSRVTT}} & \multirow{2}{*}{\textbf{MSVD}} & \multicolumn{3}{c|}{\textbf{MSRVTT}}         & \multicolumn{3}{c}{\textbf{DiDeMo}}         \\
                                            &                                  &                                & \textbf{R@1} & \textbf{R@5} & \textbf{R@10} & \textbf{R@1} & \textbf{R@5} & \textbf{R@10} \\ \hline
Joint learning                              &               45.5                   &      54.9                          &    59.3          &  81.1            &        88.4       &       59.2       &       84.9       &      90.2         \\
Meta-learning                               &        \textbf{46.4}                                                        &          \textbf{55.8}                       &  \textbf{60.0}         & \textbf{82.2}         & \textbf{89.2}          &    \textbf{62.7}          &       \textbf{89.9}       &     \textbf{96.0}      \\ \hline  
\end{tabular}}
\end{table}

\begin{figure*}[t]
    \centering
    \caption{a) Relative R@1 improvement in the text-video retrieval task on MSRVTT for each topic; b) Relative accuracy improvement in the videoQA task on MSRVTT with respect to the proportion of questions for which each label is the answer.}
    \label{fig:improvement_figure}
    \begin{subfigure}[h!]{0.5\linewidth}
        \centering
        \includegraphics[width=\linewidth]{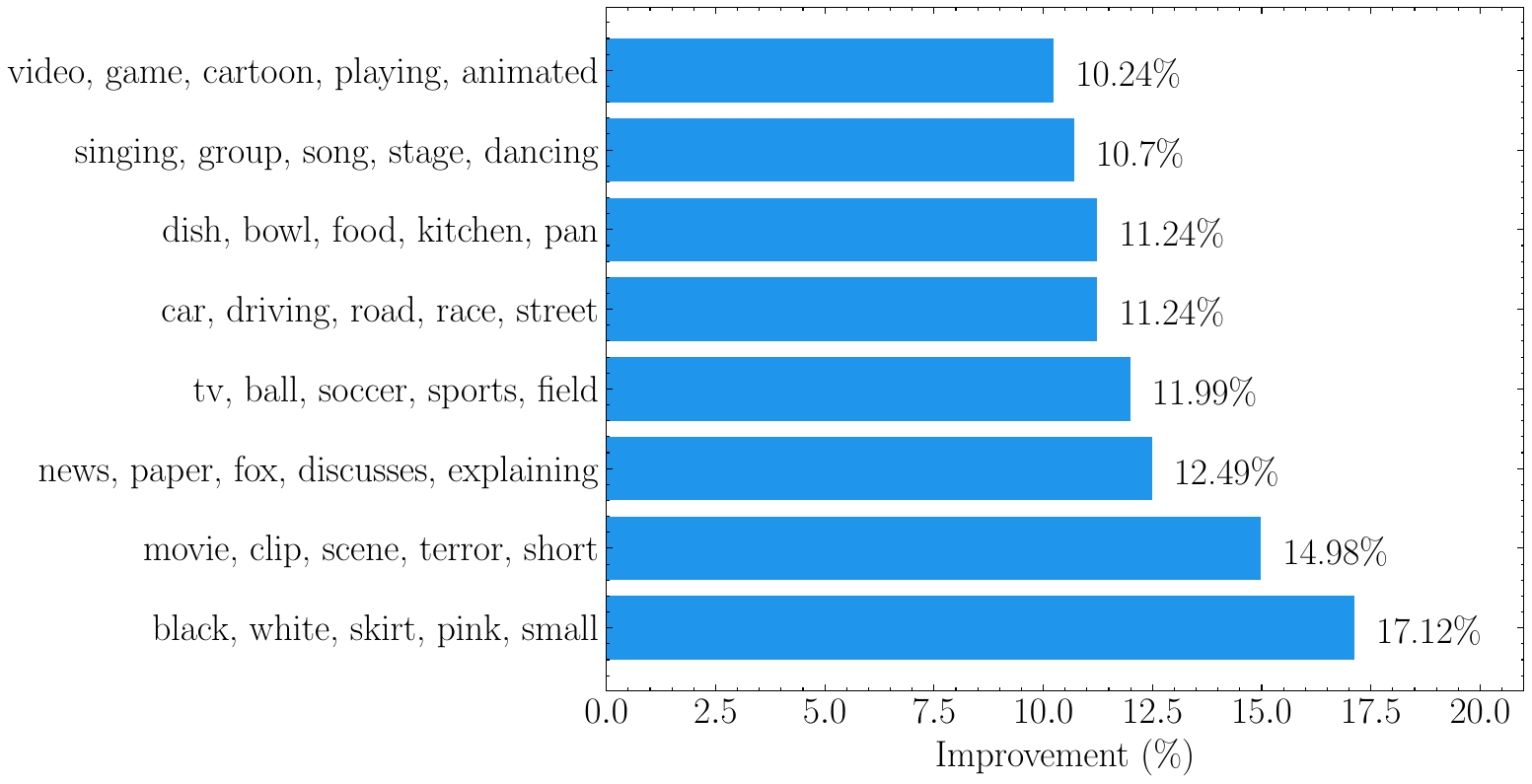}
    \end{subfigure}
    \hspace{3mm}
    \begin{subfigure}[h!]{0.46\linewidth}
        \centering
        \includegraphics[width=0.9\linewidth]{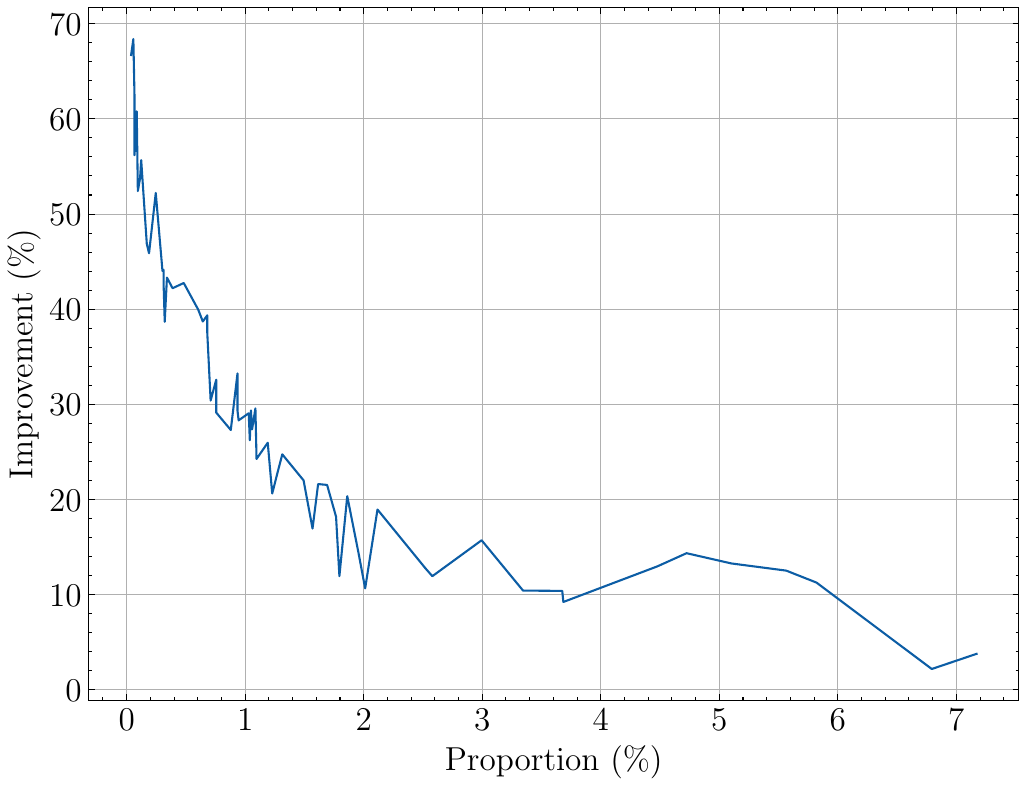}
    \end{subfigure}
\end{figure*}

\begin{table}[h!]
\centering
\caption{Case study of similarity scores for video-text pairs.}
\label{tab:similarity_score_video_text_pairs}
\resizebox{\linewidth}{!}{
\begin{tabular}{c|p{0.7\linewidth}|c|c}
\hline
\textbf{Video} & \multicolumn{1}{c|}{\textbf{Caption}} & \textbf{CLIP-ViP score} & \textbf{Our score} \\ \hline
\multirow{4}{*}{\includegraphics[width=0.1\linewidth]{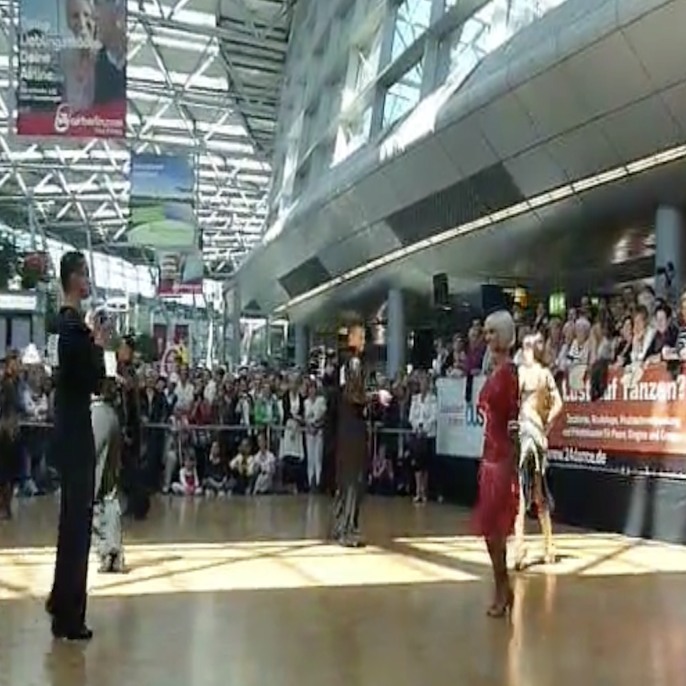} \includegraphics[width=0.1\linewidth]{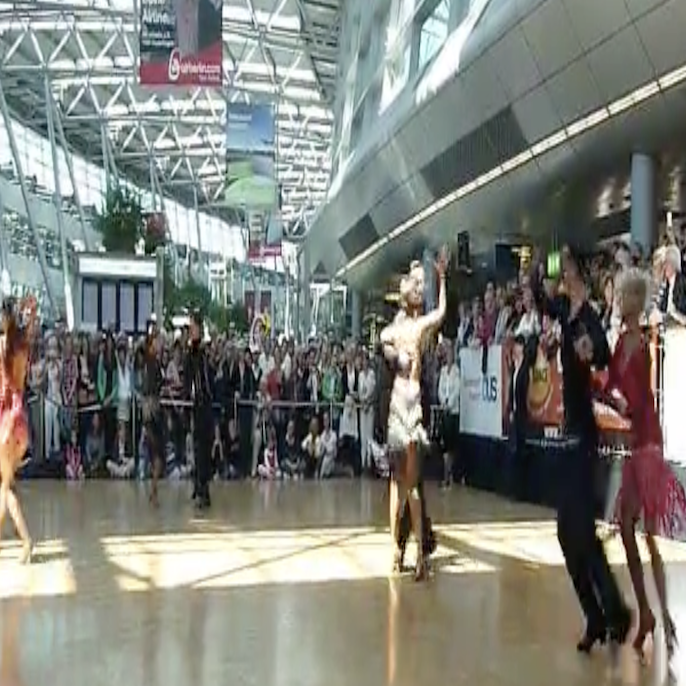}
\includegraphics[width=0.1\linewidth]{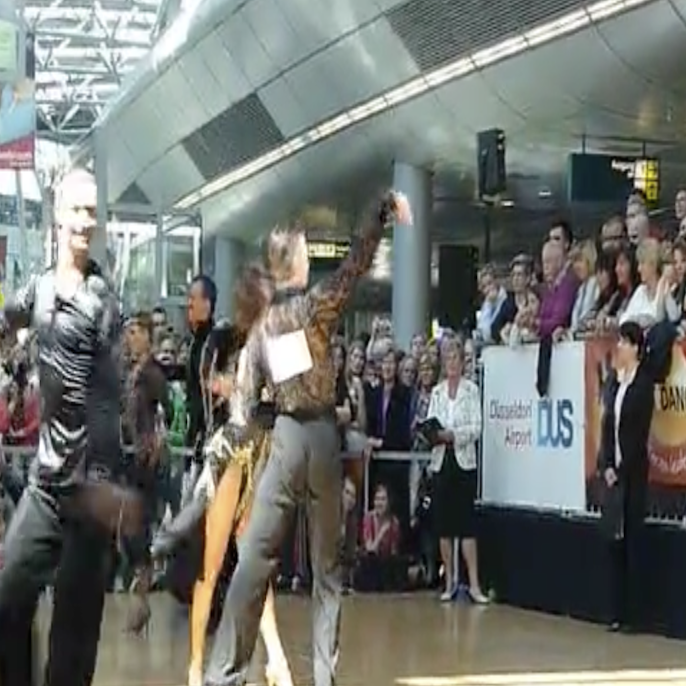} \includegraphics[width=0.1\linewidth]{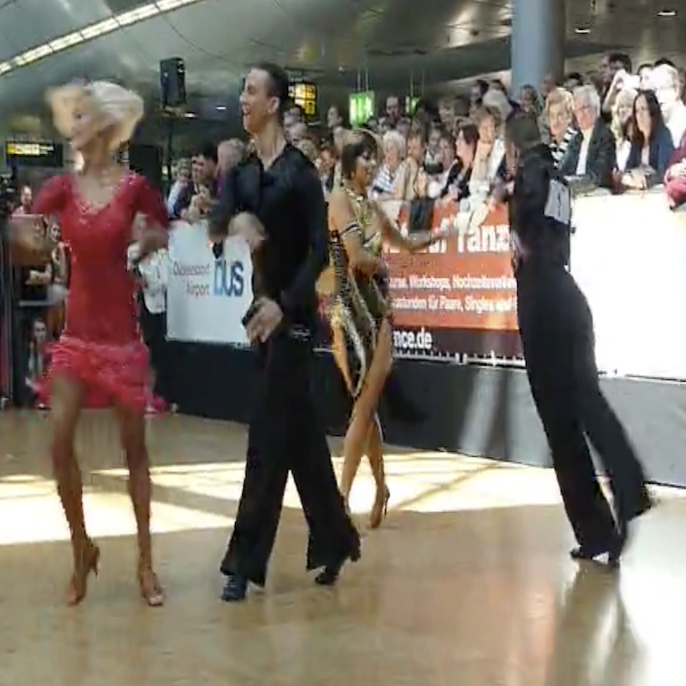}}                  &      a group of people dancing on a stage in front of a crowd.                                &    0.1727                                         &            0.1698                            \\ \cline{2-4}
                                   &      the man in black reaches his dance partner. man walks to the woman in red the camera zooms out on the dancers. we see a dancer in the back left spinning                                &   0.2175                                          &  0.3342                                      \\ \hline
\multirow{5}{*}{\includegraphics[width=0.1\linewidth]{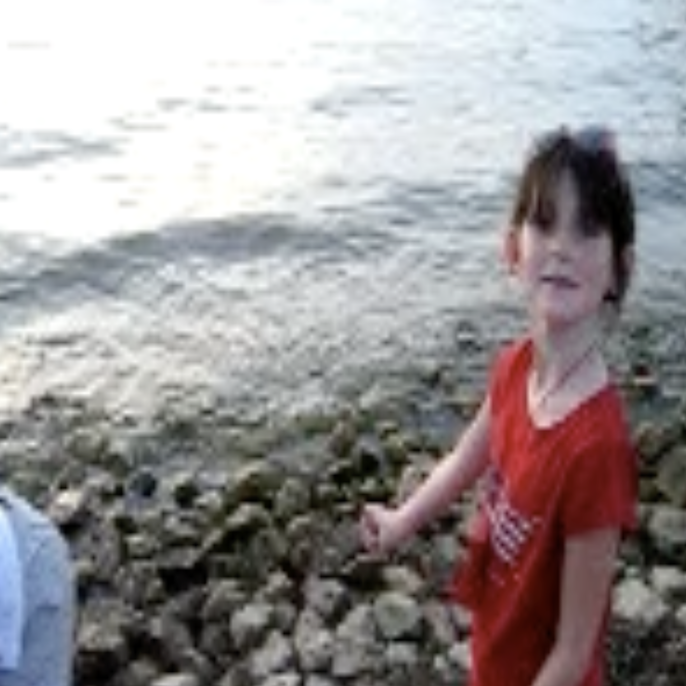} \includegraphics[width=0.1\linewidth]{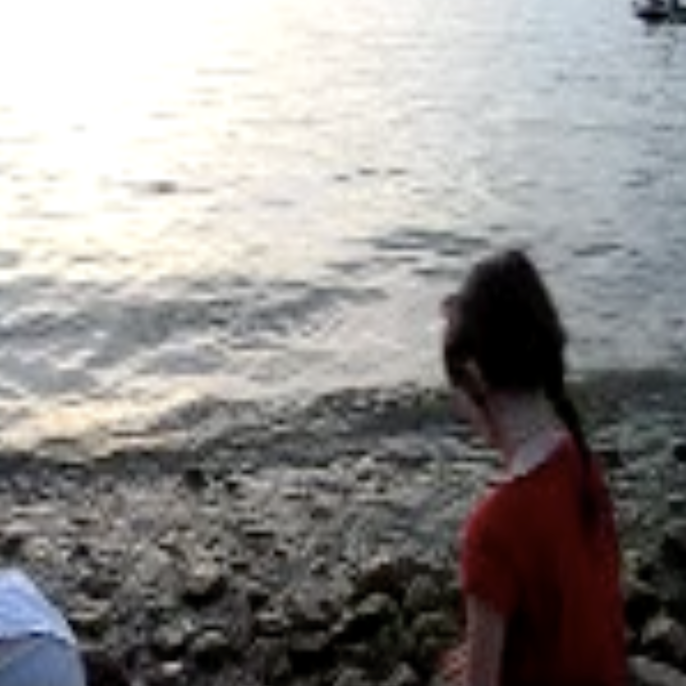}
\includegraphics[width=0.1\linewidth]{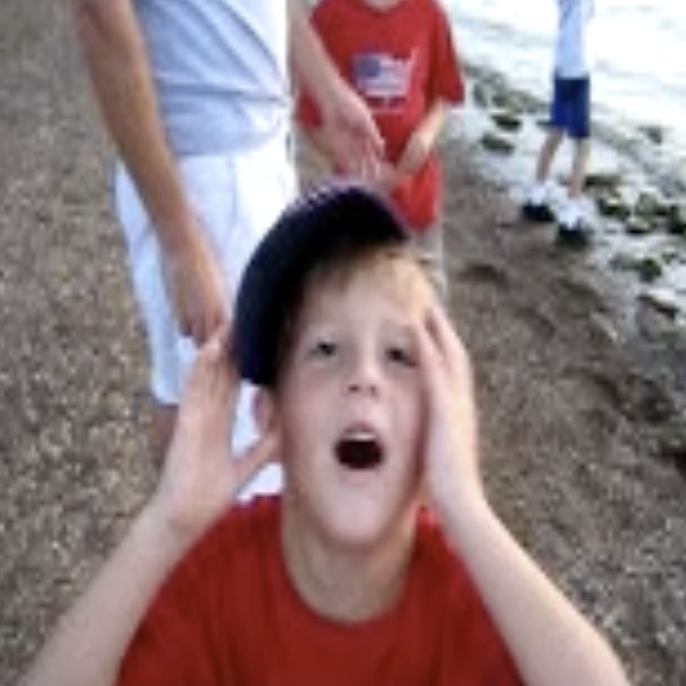} \includegraphics[width=0.1\linewidth]{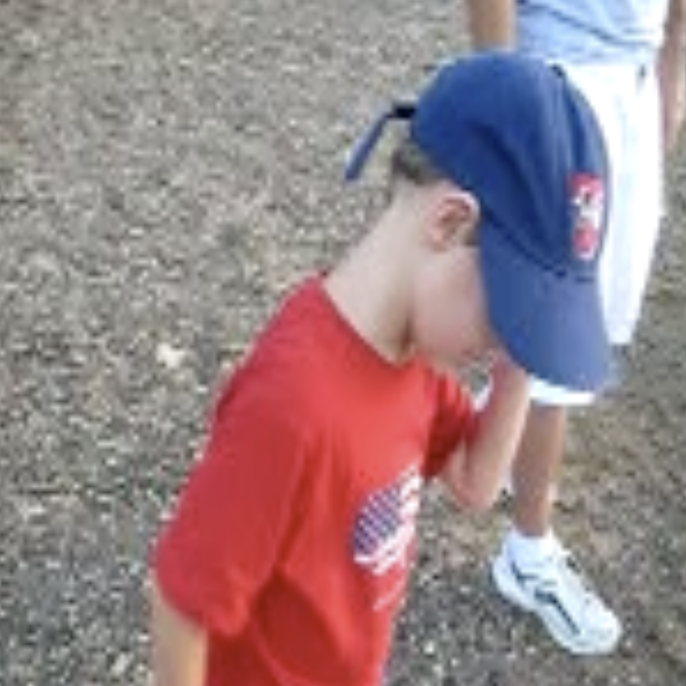}}                  &              a boy in a red shirt and a blue hat is playing.                        &       0.2594                                      &              0.2441                          \\ \cline{2-4}
                                   &     a girl throwing a rock. girl in the red shirt looks back over his shoulder the boy in a blue hat is seen for the first time. the boy in the baseball cap first looks at the camera boy wearing a blue ball cap                                 &   0.2869                                          &   0.3164  \\    \hline
\end{tabular}}
\end{table}

\noindent\textbf{Number of extracted key frames.} We explore the impact of varying number of extracted key frames and varying the approach to concatenate in Table \ref{tab:exp_q_grid}. While the performance is quite invariant to the concatenation strategy, we observe an increasing trend of the performance when the number of the extracted frames $Q$ increases, since there exists more information for LVLM to put into the descriptions. However, after surpassing $Q = 12$ frames, the performance dramatically decreases. The reason could be that when $Q$ is excessively large, visual frames become excessively small for LVLM to precisely detect details. Furthermore, balanced grids, \textit{i.e.} $3 \times 4$ and $4 \times 4$, tend to outperform skewed grids, \textit{i.e.} $2 \times 6$ and $2 \times 8$, respectively. The cause might be that video details are more difficult for an LVLM to capture when the video frames are presented on a long rectangle than when they are presented on a balanced one.

\subsection{Analysis}
\noindent\textbf{Effect of angular margin contrastive learning.} To better understand the effect of our angular margin contrastive method, we show in Table \ref{tab:similarity_score_video_text_pairs} examples of videos and language captions, along with the similarity scores which are generated by the baseline CLIP-ViP model and the model trained with our proposed subtractive contrastive learning strategy. We observe that when the caption consists of less details, \textit{i.e.} it aligns more weakly with the video content, our model assigns a lower similarity score compared to CLIP-ViP. In contrast, when the language caption becomes more detailed, we assign a higher score while CLIP-ViP only slightly elevates the score. These examples demonstrate that our angular margin contrastive strategy can control the similarity level for weakly aligned video-text pairs while being able to adapt to cases of strongly aligned pairs. 

\noindent\textbf{Effect of MLP-parameterized weighting function.} To closely study the impact of our MLP-based weighting function, we show the relative R@1 and the relative accuracy improvement of the text-video retrieval tasks in Figure \ref{fig:improvement_figure}a and \ref{fig:improvement_figure}b, respectively. Both figures demonstrate that we accomplish higher level of performance improvement on samples that belong to minority groups, \textit{i.e.} unpopular topic groups and answer labels. Such results intuitively substantiate the productivity of our MLP-based weighting function and showcase its capacity to learn the effective strategy to control the training effect of the training samples.

\section{Summary}
\noindent In this paper, we propose a meta-optimized contrastive framework to enhance video-language representation learning. In particular, we propose a contrastive learning framework with a subtractive margin between positive video and language to regularize their representations from reaching perfect similarity. Our framework also utilizes an MLP network that maps training losses to sample weights to enable the model to dynamically adjust the focus on data samples during training. Combined with a strategy to utilize large-vision language model to augment video-text data, our framework achieves superior results on commonly used video question answering and text-video retrieval tasks. Our framework is also applicable to a wide array of model architectures, which can promote its implementation in practical applications.
\chapter[READ: Recurrent Adapter with Partial Video-Language Alignment for Parameter-Efficient Transfer Learning in Low-Resource Video-Language Modeling]{READ: Recurrent Adapter with \\ Partial Video-Language Alignment for Parameter-Efficient Transfer \\ Learning in Low-Resource \\ Video-Language Modeling}
\label{ch2:read}

\section{Introduction}
\begin{figure*}
    \centering
    \includegraphics[width=\linewidth]{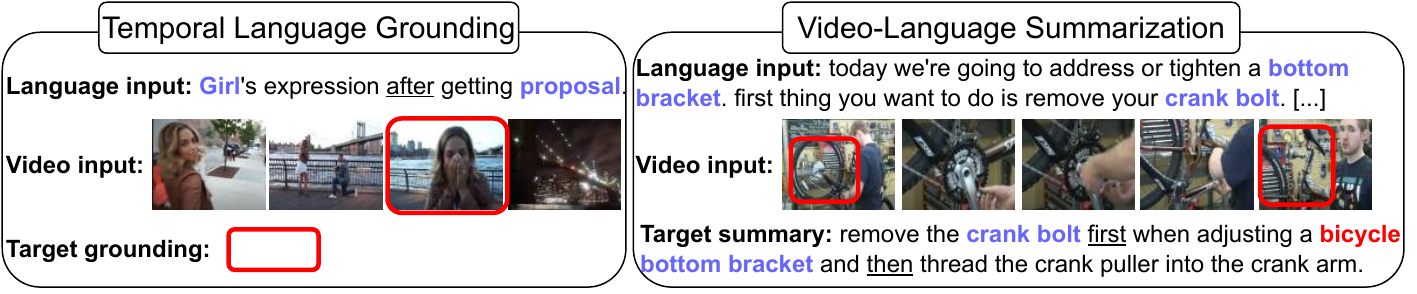}
    \caption{Examples of the TLG and VLS problems. TLG model needs to understand the meaning of language entities such as \textit{proposal} or \textit{girl}, and the existence of \textit{expression} in video frames. VLS model is expected to recognize salient information, \textit{e.g. crank bolt}, \textit{bottom bracket} from the language, and \textit{bicycle} from the video.}
    \label{fig:task_example}
\end{figure*}

\noindent Video-language modeling is a challenging problem since it involves understanding both video and language modalities. For example, temporal language grounding (TLG) model comprehends video detail and language query to localize semantically related video moments (Figure \ref{fig:task_example} (left)), or video-language summarization (VLS) model extracts information from both video content and language transcript to write the summary (Figure \ref{fig:task_example} (right)). 

Previous  video-language modeling methods \citep{liu2022umt, lei2021detecting, yu2021vision}  employ pretrained Transformer models such as Unified Multimodal Transformer (UMT) \citep{liu2022umt} and Vision-Guided BART (VG-BART) \citep{yu2021vision}, and fine-tune all the parameters of these models for every single task.
This results in substantial storage overhead since each task demands storing a separate model \citep{zhang2023multimodal}. Moreover, because of the difficulty of collecting video-language data \citep{pan2022st}, fully fine-tuning these over-parameterized models in low-resource scenarios, where limited training data is available, leads to instability and sub-optimal performance \citep{jiang2022cross, huang2023vop}.

To address these shortcomings, adapters are proposed as a parameter-efficient solution for finetuning video-language pretrained transformers \citep{jiang2022cross, zhang2023multimodal, yang2023aim, sung2022vl, chen2022adaptformer}. 
The strategy is to add additional adaptation module to each layer of the pre-trained network and only the adaptation modules are trained during fine-tuning to improve the 
 parameter-performance trade-off.
These modules rely on non-linear projections to downproject video-language inputs into low-dimensional space then up-project them back to the original high-dimensional space.
However, such projections consider video frames and textual words as separate tokens, thus ignoring the intrinsic temporal dependency among video frames or textual words. Without such dependency information, it is difficult to reason about temporal context in the video to properly ground the language (\emph{e.g.} in Figure \ref{fig:task_example}, determine the \emph{expression} of the \emph{girl} \emph{after}, not \emph{before}, the \emph{proposal}), or coherently link the entities in the summary (\emph{e.g.} in Figure \ref{fig:task_example}, recap the chronological order of \emph{bolt removing} and \emph{puller threading}). Moreover, because at fine-tuning time only adaptation modules are trained using limited video-language data, little attention is paid to the information flow that starts from the raw video-language inputs till the low-dimensional space of the adaptation modules. This may result in losing essential task-related information and carrying noise into these modules \citep{tsai2020multimodal, han2021improving}. 

\begin{figure}[t]
    \centering
    \includegraphics[width=0.6\linewidth]{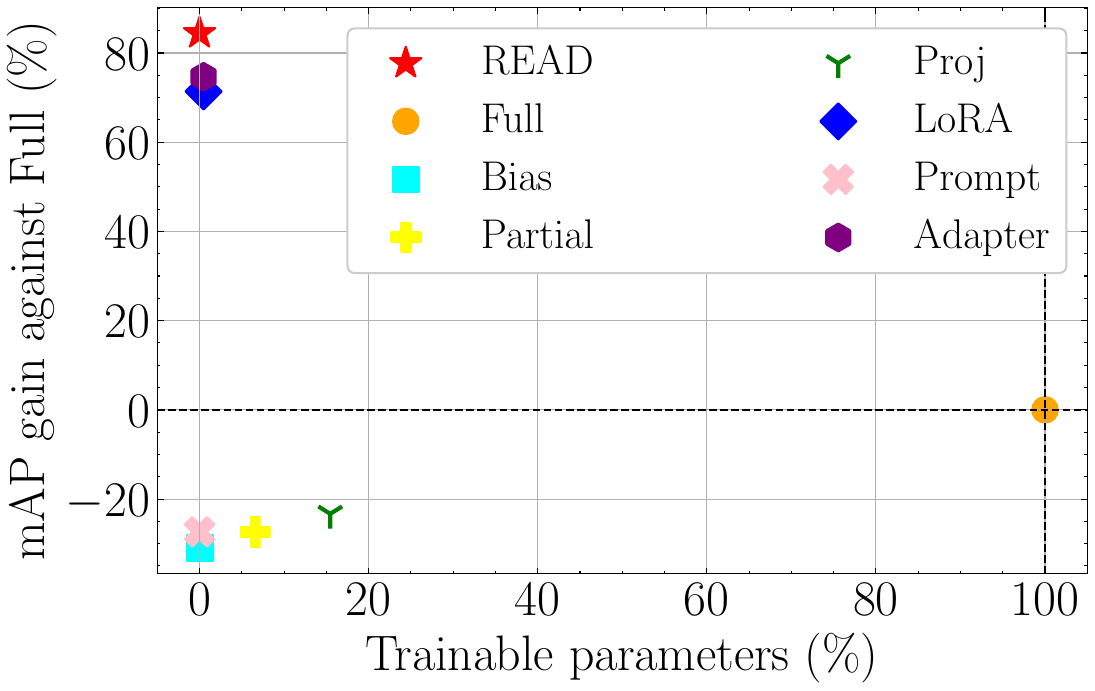}
    \caption{Comparison of our proposed READ method with the full fine-tuning and other parameter-efficient fine-tuning methods. For each method, we denote the mAP gain averaged over the domains of the YouTube Highlights dataset together with the number of trainable parameters.
    }
    \label{fig:map_gain_trainable_parameters}
\end{figure}
To resolve the first issue, we propose a novel adapter architecture, \textbf{RE}current \textbf{AD}apter (READ), for video-language modeling tasks. The key idea is to incorporate the recurrent modeling ability into the adaptation module to capture the temporal dependency of video frames and language entities \citep{goodfellow2016deep}. As such, we formulate READ as a parameter-efficient bottleneck with a sequence of operations including feature dimension reduction, recurrent modeling, and feature dimension recovery. Since the incorporated recurrent computation works in the low dimension (\textit{e.g.} 4-dimensional), our READ module stands as a lightweight design and can be cheaply integrated throughout the Transformer architecture for enhancing video-language modeling, using only up to 1.20\% trainable parameters.

As for the second issue, we propose \textbf{P}artial \textbf{V}ideo-\textbf{L}anguage \textbf{A}lignment (PVLA), a novel objective to explicitly encourage the alignment between video and language representations, thus capturing invariant aligned information across modalities that are critical for downstream tasks. The key concept is to minimize the Partial Optimal Transport (POT) distance between the distribution over video frame representations and the distribution over textual word representations. The rationale for our partial implementation of optimal transport lies in that video and language do not exhibit complete one-to-one correspondence. Typically, the language does not describe all aspects of the video, and only part of the language sequence is strongly related to part of the video frames, \textit{e.g.} in Figure \ref{fig:task_example} the language input about the \textit{girl’s expression} is only related to the target grounding. As such, utilizing POT for distribution matching is to focus on essential masses that are strongly related between modalities, hence optimizing towards better video-language alignment and gaining more control over video-language information passed into our READ modules.

Based on our novel proposals, we construct the READ framework that can be employed to finetune various pre-trained Transformer architectures such as multimodal transformer (UMT \citep{liu2022umt}, Moment-DETR \citep{lei2021detecting}), and generative vision-guided transformer models (VG-BART \citep{lewis2020bart} and VG-T5 \citep{raffel2020exploring}). %
Through freezing these pre-trained models and fine-tuning only our READ modules with PVLA objective, we outperform standard fine-tuning and other parameter-efficient methods with substantially fewer tunable parameters (Figure \ref{fig:map_gain_trainable_parameters}) for low-resource video-language tasks, including temporal language grounding and video-language summarization. To sum up, our contributions can be summarized as:
\begin{itemize}
    \item We propose \textbf{RE}current \textbf{AD}apter (READ), a novel adapter architecture,
    that better captures  temporal information for modeling video-language tasks.
    \item We propose \textbf{P}artial \textbf{V}ideo-\textbf{L}anguage \textbf{A}lignment (PVLA) objective to encourage the alignment between video and language modalities during the adaptation process.
    \item We validate our READ framework by extensive experiments using multiple low-resource temporal language grounding and video-language summarization datasets,
     where READ outperforms all existing fully or parameter-efficient fine-tuning strategies with only up to 1.20\% parameters tunable.
\end{itemize}

\begin{figure}[t]
    \centering
\includegraphics[width=0.7\linewidth]{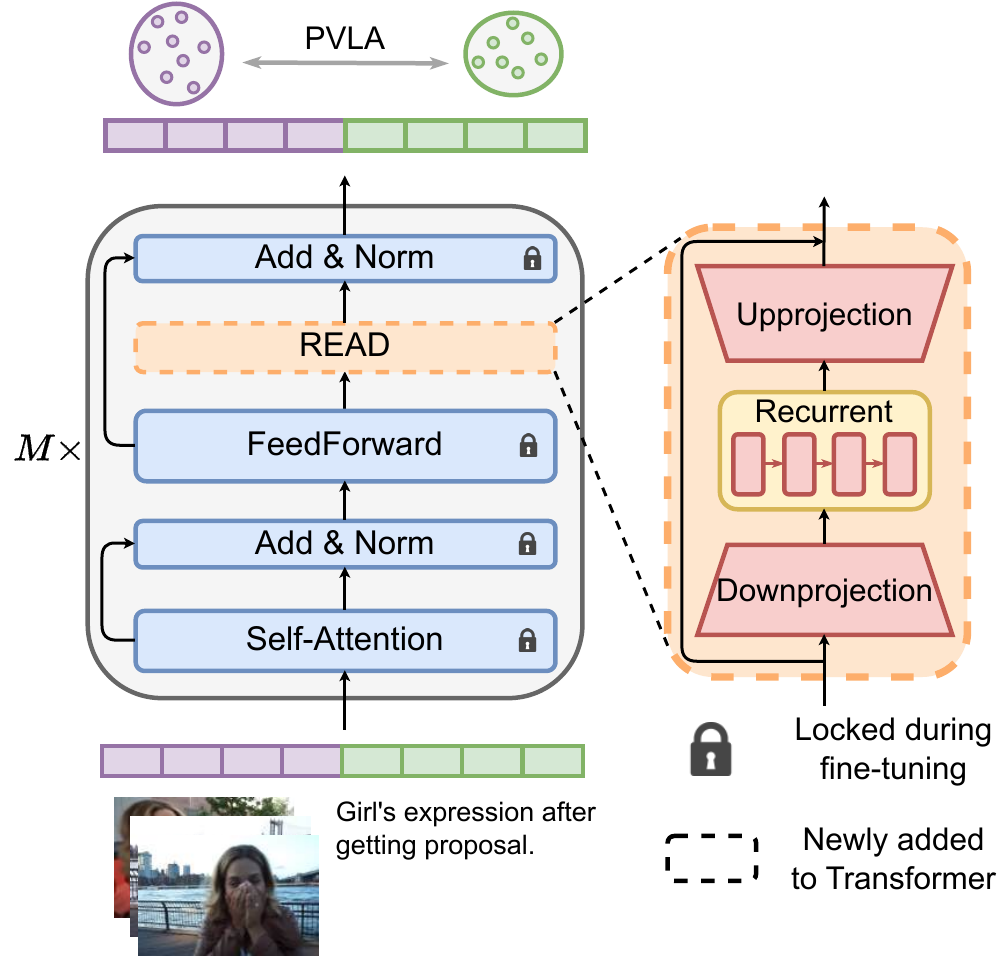}
    \caption{Overall illustration of the proposed recurrent adapter (READ) and partial video-language alignment (PVLA) framework.} 
    \label{fig:overall_illustration}
\end{figure}
\section{Methodology}
\noindent We present recurrent adapter (READ) to effectively develop the temporal modeling capability and efficiently transfer large pre-trained transformer models for video-language downstream tasks. We also introduce the partial video-language alignment (PVLA) task to optimize the alignment of in-distribution video-language inputs for better supporting video-language adaptation under low-resource settings. Our overall framework is illustrated in Figure \ref{fig:overall_illustration}.

\subsection{Preliminary – Transformer Architecture for Video-Language Modeling}
\label{sect:preliminary}
\noindent We concentrate our work upon the Transformer architecture \citep{vaswani2017attention}. The architecture consists of an embedding layer and $M$ consecutive Transformer blocks. As inputs to the Transformer model, we extract $N_{V}$ frames and $N_{L}$ words from the video and language input, respectively. The embedding layer would encode the extracted frames and words into sequences of initial video and language representations $H_{V}^{(0)} = \{\mathbf{h}_{v,i}^{(0)}\}_{i=1}^{N_V}$ and $H_{L}^{(0)} = \{\mathbf{h}_{l,j}^{(0)}\}_{j=1}^{N_L}$, respectively. The transformer then forwards these sequences into consecutive Transformer blocks, each of which is typically composed of a multi-head self-attention (MHSA) layer, a residual connection with normalization (Add \& Norm) layer, a feedforward layer, and another Add \& Norm layer. 

In MHSA for video-language modeling, the language representations are linearly projected into the query tensor $\mathbf{Q} \in \mathbb{R}^{N_L \times d}$, whilst the video representations into the key $\mathbf{K} \in \mathbb{R}^{N_V \times d}$ and value tensors $\mathbf{V} \in \mathbb{R}^{N_V \times d}$:
\begin{gather*}
\mathbf{Q}^{(m)} = \text{Linear}\left(\mathbf{H}_{L}^{(m)}\right), \; \mathbf{K}^{(m)} = \text{Linear}\left(\mathbf{H}_{V}^{(m)}\right), \\
\mathbf{V}^{(m)} = \text{Linear}\left(\mathbf{H}_{V}^{(m)}\right),
\end{gather*}
where $m$ denotes the index of the current Transformer block and $d$ the hidden dimension. Then, the self-attention computation is conducted upon these vectors as:
\begin{equation}
\begin{split}
\mathbf{X}^{(m)} = \text{Attention}\left(\mathbf{Q}^{(m)}, \mathbf{K}^{(m)}, \mathbf{V}^{(m)}\right) = \\ \text{Softmax}\left(\frac{\mathbf{Q}^{(m)}\cdot(\mathbf{K}^{(m)})^{\top}}{\sqrt{d}}\right) \cdot \mathbf{V}^{(m)}.
\end{split}
\end{equation}
The attention output $\mathbf{X}^{(m)}$ is subsequently sent to an Add \& Norm layer:
\begin{equation}
\mathbf{P}^{(m)} = \text{LN}\left(\mathbf{X}^{(m)} + \mathbf{H}^{(m)}_{L}\right),
\end{equation}
where LN denotes the layer normalization layer. Subsequently, $\mathbf{P}^{(m)}$ is forwarded to a FeedForward block to produce the output representation $\mathbf{O}^{(m)}$, which will be passed to another Add \& Norm layer to create the video-informed language representation for the next transformer block:
\begin{gather}
\mathbf{O}^{(m)} = \text{GeLU}\left(\text{Linear}\left(\mathbf{P}^{(m)}\right)\right), \\ 
\mathbf{H}_{L}^{(m+1)} = \text{LN}\left(\mathbf{P}^{(m)} + \mathbf{O}^{(m)} \right), \; \mathbf{H}_{V}^{(m+1)} = \mathbf{H}_{V}^{(m)}.
\end{gather}
The video-language representation of the last Transformer block $\mathbf{H}_{L}^{(M+1)}$ is finally adopted to perform a specific downstream task.

\subsection{Recurrent Adapter (READ)}
\noindent The objective of our READ is to incorporate the temporal modeling capability for the adaptation module. To this end, we construct a recurrent-based bottleneck layer which is composed of a downprojection layer, a recurrent neural network (RNN) layer, and an up-projection layer.

Formally, given the FeedForward output $\mathbf{O}$, our recurrent adapter can be expressed as:
\begin{gather}
\mathbf{\tilde{O}} = \mathbf{O} + \text{GELU}\left(\text{RNN}\left(\mathbf{O} \cdot W_{\text{down}}\right)\right)\cdot W_{\text{up}},
\end{gather}
where $W_{\text{down}} \in \mathbb{R}^{d \times k}, W_{\text{up}} \in \mathbb{R}^{k \times d}$, and $k \ll d$. Subsequently, we combine $\mathbf{P}$ and $\tilde{\mathbf{O}}$ via residual connection to generate the output $\mathbf{H}$:
\begin{equation}
\mathbf{H} = \text{LN}\left(\mathbf{\tilde{O}} + \mathbf{P}\right).
\end{equation}
In addition to RNN, we also experiment with other recurrent architectures in Table \ref{tab:recurrent_ablation_experiments} and observe that the performance is insensitive to the architectural choice. Therefore, for simplicity, we decide to implement the RNN architecture in our READ layer.

\noindent\textbf{Fine-tuning.} During the fine-tuning stage, we preserve the weights of the pre-trained Transformer model and only optimize our introduced READ layers. In detail, the original model components (blue blocks in Figure \ref{fig:overall_illustration}) are frozen, while the parameters of READ (the yellow block in Figure \ref{fig:overall_illustration}) are updated with respect to the task-specific and the partial video-language alignment losses, which will be delineated in the upcoming section.

\noindent\textbf{Testing.} During testing, we maintain the shared parameters of the pre-trained Transformer model and only load those of our extra READ modules that are fine-tuned in the previous phase. This would keep the storage cost from burgeoning because the number of added parameters is tiny.
\subsection{Partial Video-Language Alignment (PVLA)}
\noindent To encourage the control towards the information flow of video frames and language words, we propose to optimize the alignment between the in-distribution video and language representations $H_{V}$ and $H_{L}$ at all Transformer blocks. 

We consider video and language as two discrete distributions $\boldsymbol{\mu}$ and $\boldsymbol{\nu}$, whose $H_V$ and $H_L$ are their supports, respectively.  We formulate this setting as $\boldsymbol{\mu} = \sum\limits_{i=1}^{N_V} \mathbf{a}_{i} \delta_{\mathbf{h}_{v,i}}$ and $\boldsymbol{\nu} = \sum\limits_{j=1}^{N_L} \mathbf{b}_{j} \delta_{\mathbf{h}_{l,j}}$, with $\delta_{\mathbf{h}_{l,j}}$ and $\delta_{\mathbf{h}_{l,j}}$ being the Dirac functions respectively centered upon $\mathbf{h}_{v,i}$ and $\mathbf{h}_{l,j}$. The weight vector of the supports is $\mathbf{a} = \frac{\mathbf{1}_{N_V}}{N_V}$, and $\mathbf{b} = \frac{\mathbf{1}_{N_L}}{N_L}$.

Based upon the above setting, we propose the partial video-language alignment (PVLA) task, which is to minimize the following $\mathcal{L}_{\text{PVLA}}$ loss equal to the partial optimal transport (POT) distance $\mathcal{D}_{\text{POT}}$ between $\boldsymbol{\mu}$ and $\boldsymbol{\nu}$ as:
\begin{gather}
\mathcal{L}_{\text{PVLA}} = \mathcal{D}_{\text{POT}}(\boldsymbol{\mu}, \boldsymbol{\nu}) = \min_{\mathbf{T} \in \Pi(\mathbf{a}, \mathbf{b})} \sum\limits_{i=1}^{N_V} \sum\limits_{j=1}^{N_L} \mathbf{T}_{i,j} \cdot c\left(\mathbf{h}_{v,i}, \mathbf{h}_{l,j}\right), \\
\begin{split}
\text{s.t} \quad \Pi(\mathbf{a}, \mathbf{b}) = \{\mathbf{T} \in \mathbb{R}^{N_V \times N_L}_{+} \mid \mathbf{T}\mathbf{1}_{N_L} \leq \mathbf{a},  \mathbf{T}^{\top}\mathbf{1}_{N_V} \leq \mathbf{b}, \\\mathbf{1}_{N_V}^{\top} \cdot \mathbf{T} \cdot \mathbf{1}_{N_L} = s, \quad 0 \leq s \leq \min(N_L, N_V)\}.
\label{eq:pvla_formulation}
\end{split}
\end{gather}
Because the exact minimization over the transport plan $\mathbf{T}$ is intractable, we adopt the Sinkhorn-based algorithm to compute $\mathbf{T}$. We explicate our algorithm to calculate the partial video-language alignment loss in Algorithm \ref{alg:pvla_loss}.

Our PVLA formulation is flexible where it allows only $s$ samples from one distribution to be transported to the other, and enables the algorithm to decide the value of $s$, in case the input language only corresponds to certain video aspects (or vice versa).

\noindent\textbf{Training Strategy.} For training, we jointly optimize the video-language task-specific loss and our PVLA loss. It is worth noting that we only update our introduced READ layers while keeping the remaining components frozen.

\section{Experiments}
\setlength{\textfloatsep}{10pt}
\begin{algorithm}[t]
\caption{Computing the PVLA loss}
\label{alg:pvla_loss}
\begin{algorithmic}
\Require{$\mathbf{C} = \{\mathbf{C}_{i,j} = c\left(\mathbf{h}_{v,i}, \mathbf{h}_{l,j}\right) \mid 1 \leq i \leq N_V, 1 \leq j \leq N_L\} \in \mathbb{R}^{N_V \times N_L}$,\; temperature $\tau,\; \mathbf{a} \in \mathbb{R}^{N_V},\; \mathbf{b} \in \mathbb{R}^{N_L},\; s,\; N_{\text{iter}}$} \\
$\mathcal{L}_{\text{PVLA}} = \infty$ 
\For{$s=1$ to $\min(N_L, N_V)$}
    \State $\mathbf{T} = \text{exp}\left(-\frac{\mathbf{C}}{\tau}\right)$ 
    \State $\mathbf{T} = \frac{s}{\left(\mathbf{1}_{N_V}\right)^{\top} \cdot \mathbf{T} \cdot \mathbf{1}_{N_L}} \mathbf{T}$
    
    \For{$i=1$ to $N_{\text{iter}}$} 
        \State $\mathbf{p}_{a} = \min \left(\frac{\mathbf{a}}{\mathbf{T}\mathbf{1}_{N_L}}, \mathbf{1}_{N_V}\right)$, $\mathbf{T}_{a} = \text{diag}\left(\mathbf{p}_{a}\right) \cdot \mathbf{T}$
        \State $\mathbf{p}_{b} = \min \left(\frac{\mathbf{b}}{\mathbf{T}_{a}^{\top}\mathbf{1}_{N_V}}, \mathbf{1}_{N_L}\right)$, $\mathbf{T}_{b} = \text{diag}\left(\mathbf{p}_{b}\right) \cdot \mathbf{T}_{a }$
        \State $\mathbf{T} = \frac{s}{\left(\mathbf{1}_{N_V}\right)^{\top} \cdot \mathbf{T} \cdot \mathbf{1}_{N_L}} \mathbf{T}_{b}$
    \EndFor 
    \State $\mathcal{L}_{\text{PVLA}} = \min\left(\mathcal{L}_{\text{PVLA}}, \;\sum\limits_{i=1}^{N_V} \sum\limits_{j=1}^{N_L} \mathbf{T}_{i,j} \mathbf{C}_{i,j}\right)$ 
    \EndFor \\
\Return $\mathcal{L}_{\text{PVLA}}$
\end{algorithmic}
\end{algorithm}

{\renewcommand{\arraystretch}{1.2}
\begin{table*}[h!]
\centering
\fontsize{9pt}{9pt}\selectfont
\begin{tabular}{l|c|ccccccc}
\hline
\multicolumn{1}{c|}{\textbf{Method}} & \multicolumn{1}{c|}{\centering\textbf{\#params (M)}} & \multicolumn{1}{c}{\textbf{Dog}} & \multicolumn{1}{c}{\textbf{Gym}} & \multicolumn{1}{c}{\textbf{Par.}} & \multicolumn{1}{c}{\textbf{Ska.}} & \multicolumn{1}{c}{\textbf{Ski.}} & \multicolumn{1}{c}{\textbf{Sur.}} & \multicolumn{1}{c}{\textbf{Avg.}} \\ \hline
Full                       &       283.97 (100\%)                                   &      65.90\sig                   &      75.20\sig                   &        82.20\sig                  &         71.80\sig                 &         72.30\sig                 &       81.15\sig                   &              74.76\sig            \\
Bias                       &      0.51 (0.18\%)                                    &       46.23\sig                  &     61.19\sig                    &          56.73\sig                &         31.36\sig                 &          61.14\sig                &      49.77\sig                    &           51.07\sig               \\
Partial                    &       38.75 (13.65\%)                                   &       48.28\sig                  &       63.26\sig                  &           59.71\sig               &           32.66\sig               &           64.58\sig               &        56.22\sig                  &        54.12\sig                  \\
Proj                       &        5e-4 (1.76e-4\%)                                  &       57.05\sig                  &    65.70\sig                     &          63.03\sig                &        71.83\sig                  &            65.45\sig              &           79.71\sig               &         67.13\sig                 \\
LoRA                       &            13.12 (4.62\%)                             &               60.97\sig          &          67.68\sig               &          72.53\sig                &          66.62\sig                &              71.24\sig            &    79.15\sig                      &            69.70\sig              \\
Prompt                       &      0.02 (0.01\%)                                    &               48.28\sig          &               63.26\sig        &       59.71\sig                   &           35.67\sig               &       35.67\sig                   &        64.61\sig                  &           46.87\sig               \\
Adapter                    &       13.11 (4.62\%)                                   &         62.89\sig                &       67.09\sig                  &       74.56\sig                   &           62.56\sig               &        68.10\sig                  &                    78.73\sig      &          68.98\sig                \\ \hline
READ                     & 0.16 (0.06\%)                   & \textbf{67.65}                   & \textbf{78.05}                   & \textbf{83.25}                    & \textbf{72.40}                    & \textbf{72.98}                    & \textbf{82.36}                    & \textbf{76.12}  \\ \hline
    \end{tabular}
\caption{TLG results on the YouTube Highlights dataset. We report the mean average precision (mAP) and the number of trainable parameters (\#params). \sig means the gain of READ is statistically significant at the 0.05 level.}
\label{tab:youtube_highlights_results}
\end{table*}}
{\renewcommand{\arraystretch}{1.2}
\begin{table*}[h!]
\centering
\fontsize{9pt}{9pt}\selectfont
\begin{tabular}{p{0.11\linewidth}|>{\centering\arraybackslash}p{0.12\linewidth}|>{\centering\arraybackslash}p{0.04\linewidth}>{\centering\arraybackslash}p{0.04\linewidth}>{\centering\arraybackslash}p{0.04\linewidth}>{\centering\arraybackslash}p{0.04\linewidth}>{\centering\arraybackslash}p{0.04\linewidth}>{\centering\arraybackslash}p{0.04\linewidth}>{\centering\arraybackslash}p{0.04\linewidth}>{\centering\arraybackslash}p{0.04\linewidth}>{\centering\arraybackslash}p{0.04\linewidth}>{\centering\arraybackslash}p{0.04\linewidth}>{\centering\arraybackslash}p{0.04\linewidth}}
\hline
\textbf{Method} & \textbf{\#params (M)} & \textbf{VT} & \textbf{VU} & \textbf{GA} & \textbf{MS} & \textbf{PK} & \textbf{PR} & \textbf{FM} & \textbf{BK} & \textbf{BT} & \textbf{DS} & \textbf{Avg.} \\ \hline
Full            &         285.28 (100\%)                        &     84.17\sig                            &           81.50\sig                      &           88.20\sig                      &             71.54\sig                    &         81.40\sig                        &              84.31\sig                   &         72.30\sig                        &             76.53\sig                    &           78.86\sig                      &             77.70\sig  &   79.65\sig                      \\
Bias            &             0.25 (0.09\%)                    &                         38.08\sig        &                   69.62\sig              &         60.87\sig                        &             31.25\sig
&     68.84\sig                            &               51.71\sig                  &        50.72\sig                         &          65.38\sig                       &         54.42\sig                        &            59.05\sig & 54.99\sig                       \\
Partial         &             38.75 (13.58\%)                    &                   57.27\sig              &            62.57\sig                     &                      58.08\sig           &            52.35\sig                     &         61.58\sig                        &               63.94\sig                  &        50.82\sig                         &            62.36\sig                     &        58.05\sig                         &         47.79\sig & 57.48\sig                          \\
Proj            &            5e-4 (1.75e-4\%)                     &       57.65\sig                          &        65.80\sig                         &      64.40\sig                           &          55.57\sig                       &           64.67\sig                      &         67.07\sig                        &              59.08\sig                   &            74.70\sig                     &        63.29\sig                         &                  49.48\sig & 62.17\sig                 \\
LoRA            &      13.28 (4.66\%)                           &        77.87\sig                         &        77.01\sig                         &        77.82\sig                         &           66.38\sig                      &       80.21\sig                          &           82.23\sig                      &         66.89\sig                        &                  72.31\sig               &       69.58\sig                          &           72.09\sig & 74.24\sig                        \\
Prompt            &     0.02 (0.007\%)                            &                    61.67\sig             &         71.98\sig                        &          64.07\sig                       &              35.54\sig                   &                      72.74\sig           &              48.70\sig                   &      52.97\sig              & 67.59\sig             &         57.28\sig                        &           38.60\sig                      &          57.11\sig                         \\
Adapter         &            13.29 (4.66\%)                     &      78.46\sig                           &       76.38\sig                          &           77.36\sig                      &            67.12\sig                     &           80.33\sig                      &             82.51\sig                    &            67.77\sig                     &           71.71\sig                      &          69.58\sig                       &           71.24\sig & 74.25\sig                        \\ \hline 
READ     &  0.14 (0.05\%)      & \textbf{88.30}                           & \textbf{85.15}                           & \textbf{89.76}                           & \textbf{75.80}                           & \textbf{86.69}                           & \textbf{86.62}                           & \textbf{74.99}                           & \textbf{82.38}                           & \textbf{84.65}                           & \textbf{79.60}                           & \textbf{83.39}   \\                  \hline 
\end{tabular}
\caption{TLG results on the TVSum dataset. We report the mean average precision (mAP) and the number of trainable parameters (\#params). \sig means the gain of READ is statistically significant at the 0.05 level.}
\label{tab:tvsum_results}
\vspace{-5pt}
\end{table*}}

{\renewcommand{\arraystretch}{1.2}
\begin{table}[h!]
\centering
\fontsize{9pt}{9pt}\selectfont
\begin{tabular}{l|c|c}
\hline
\textbf{Method} & \textbf{\#params (M)} & \textbf{mAP} \\ \hline
Full            &           15.88 (100\%)                    &      36.14\sig        \\
Bias            &          0.06 (0.38\%)                     &       24.89\sig       \\
Partial         &         1.05 (6.61\%)                      &     26.37\sig         \\
Proj            &          7.31 (46.03\%)                     &   32.71\sig           \\
LoRA            &      0.19 (1.20\%)                       &     33.96\sig         \\
Prompt          &      0.04 (0.25\%)                         &    25.86\sig          \\
Adapter         &         0.20 (1.26\%)                      &       33.61\sig       \\ \hline
READ       &          0.19 (1.20\%)                     &     \textbf{36.74}        \\ \hline
\end{tabular}
\caption{TLG results on the QVHighlights dataset. We report the mean average precision (mAP) and the number of trainable parameters (\#params). \sig means the gain of READ is statistically significant at the 0.05 level.}
\label{tab:qvhighlights_resultss}
\end{table}}

\noindent We conduct extensive experiments to evaluate the effectiveness of our READ framework. We first describe the experimental settings, covering the downstream tasks, evaluation metrics, pre-trained backbones, baseline approaches, and implementation details. We then present the numerical results of our method with baseline models, then provide ablation study and thorough analysis to explore various configurations. Eventually, we perform qualitative assessments to further elucidate the behavior of our framework.
\subsection{Experimental Settings}
\label{subsect:experimental_settings}
\noindent\textbf{Downstream tasks.} We assess the effectiveness on the temporal language grounding and video-language summarization tasks. The corresponding datasets to each task are presented as follows:
\begin{itemize}
    \item \emph{Temporal Language Grounding (TLG):} The TLG’s task is to localize temporal boundaries of the video frames that semantically relate to the language query. The evaluation is performed upon three datasets, \emph{i.e.} YouTube Highlights \citep{sun2014ranking}, TVSum \citep{song2015tvsum}, and QVHighlights \citep{lei2021detecting}. YouTube Highlights consists of 40 video-language training inputs for each of the 6 domains. TVSum comprises 10 domains, each of which possesses 5 video-language training inputs. The QVHighlights benchmark includes 7,218 language-annotated video segments for training, 1,550 for development, and 1,542 for testing. 
    Following previous work on low-resource experiments \citep{boulanger2022generating}, we keep our training size at 700 samples, which is less than 10\% of the full data for the QVHighlights dataset, while preserving the original splits on the TVSum and YouTube Highlights datasets.
    \item \emph{Video-Language Summarization (VLG):} Given a video-language input, the VLS’s target is to generate a summary which takes into account both video and language content \citep{yu2021vision}. We consider the How2 dataset \citep{sanabria2018how2}, from which we randomly draw 2,000 out of 73,993 samples for training, \textit{i.e.} less than 3\% of the full data, to simulate the low-resource settings, while maintaining 2,520 samples for validation, and 2,127 samples for testing.
\end{itemize}
\textbf{Evaluation metrics.} For the TLG task, we follow previous works \citep{lei2021detecting, liu2022umt} to use the mean average precision (mAP) metric. Regarding VLS, we utilize the ROUGE score, which is a popular metric for summarization \citep{zhang2020pegasus, yu2021vision}.

{\renewcommand{\arraystretch}{1.2}
\begin{table}[h!]
\centering
\fontsize{9pt}{9pt}\selectfont
\begin{tabular}{l|>{\centering\arraybackslash}p{0.24\linewidth}|>{\centering\arraybackslash}p{0.1\linewidth}>{\centering\arraybackslash}p{0.1\linewidth}>{\centering\arraybackslash}p{0.1\linewidth}}
\hline
\textbf{Method}    & \multicolumn{1}{p{0.25\linewidth}|}{\centering\textbf{\#params (M)}}  & \textbf{R1} & \textbf{R2} & \textbf{RL} \\ \hline
Full      &          249.67 (100\%)                &  35.72\sig  &  11.88\sig   &   30.00\sig \\
Bias      &              0.20 (0.08\%)            &  30.51\sig   &  8.20\sig  &  23.00\sig  \\
Partial   &            16.54 (6.62\%)              &  31.55\sig  &  8.63\sig  &  13.65\sig  \\
Proj      &        38.60 (15.46\%)                  &  32.76\sig   &  9.11\sig   &  30.01\sig  \\
LoRA      &          1.19 (0.48\%)                &  40.04\sig  &  20.36\sig  & 35.98\sig   \\
Prompt       &       0.05 (0.02\%)                   &  31.55\sig  &  8.63\sig  &  14.90\sig  \\
Adapter   &           1.20 (0.48\%) &  41.52\sig   &  20.75\sig  &  36.88\sig  \\ \hline
READ &            1.17 (0.47\%)            &  \textbf{44.01} &  \textbf{21.91}  &  \textbf{37.91} \\ \hline
\end{tabular}
\caption{VLS results on the How2 dataset with the VG-BART model. We report the ROUGE-1, ROUGE-2, and ROUGE-L scores, with the number of trainable parameters (\#params). \sig means the gain of READ is statistically significant at the 0.05 level.}
\label{tab:how2_bart_results}
\end{table}}

{\renewcommand{\arraystretch}{1.2}
\begin{table}[h!]
\centering
\fontsize{9pt}{9pt}\selectfont
\begin{tabular}{l|c|ccc}
\hline
\textbf{Method}    & \multicolumn{1}{p{0.25\linewidth}|}{\centering\textbf{\#params (M)}}  & \textbf{R1} & \textbf{R2} & \textbf{RL} \\ \hline
Full      &          333.16 (100\%)                &  32.37\sig  & 8.07\sig   &  26.53\sig   \\
Bias      &           0.07 (0.02\%)               &  27.03\sig  &  4.53\sig  & 19.51\sig   \\
Partial   &           16.52 (4.96\%)               &  27.83\sig   &  4.92\sig  &  10.49\sig  \\
Proj      &           24.67 (7.40\%)               &  29.16\sig  &  5.50\sig  &  26.76\sig  \\
LoRA      &        1.93 (0.58\%)                  &  36.63\sig  &  16.72\sig  &  32.77\sig  \\
Prompt       &      0.18 (0.05\%)                &  28.33\sig   &  5.12\sig  &  11.75\sig  \\
Adapter   &           1.95 (0.59\%)               &  37.83\sig  &  17.45\sig  &  33.87\sig   \\ \hline
READ &           1.91 (0.57\%)              &  \textbf{40.12} &  \textbf{18.71}  &  \textbf{34.42} \\ \hline
\end{tabular}
\caption{VLS results on the How2 dataset with the VG-T5 model. We report the ROUGE-1, ROUGE-2, and ROUGE-L scores, with the number of trainable parameters (\#params). \sig means the gain of READ is statistically significant at the 0.05 level.}
\label{tab:how2_t5_results}
\end{table}}

{\renewcommand{\arraystretch}{1.2}
\begin{table}[h!]
\centering
\fontsize{9pt}{9pt}\selectfont
\begin{tabular}{l|cc}
\hline
\textbf{Method} & \textbf{mAP - YouTube Highlights} & \textbf{R2 - How2} \\ \hline
No VLA          &              73.80                     &      18.22              \\
VLA             &               74.41                    &         20.01           \\ 
\rowcolor{Gray}
PVLA            & 76.12                             & 21.91     \\ \hline     
\end{tabular}
\caption{Partial video-language alignment (PVLA) ablation experiments on YouTube Highlights and How2. We \colorbox{Gray}{color} the settings we implement for our READ method.}
\label{tab:pvla_ablation_experiments}
\end{table}}

{\renewcommand{\arraystretch}{1.2}
\begin{table*}[h!]
\centering
\fontsize{9pt}{9pt}\selectfont
\begin{tabular}{>{\centering\arraybackslash}p{0.3\linewidth}|p{0.3\linewidth}cc}
\hline
\textbf{Video}    & \centering\textbf{Language query}                                            & \textbf{POT distance} & \textbf{AP} \\ \hline
\multirow{2}{*}{\includegraphics[width=0.23\linewidth]{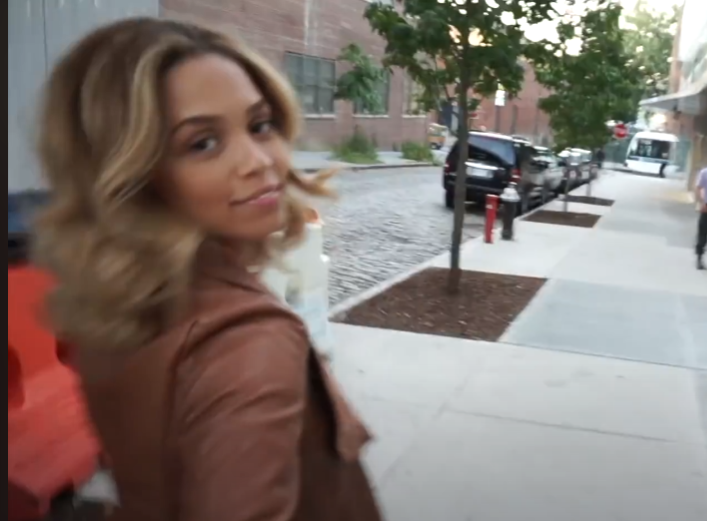} \includegraphics[width=0.23\linewidth]{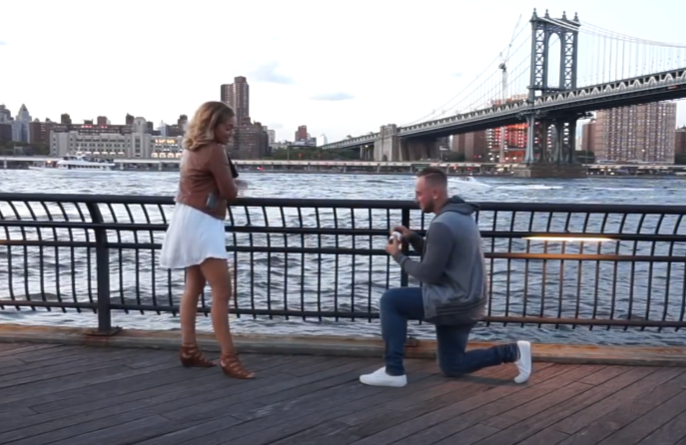}
\includegraphics[width=0.23\linewidth]{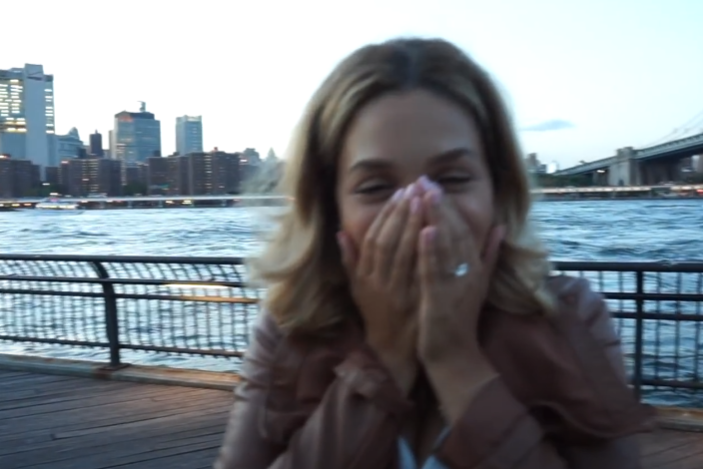} \includegraphics[width=0.23\linewidth]{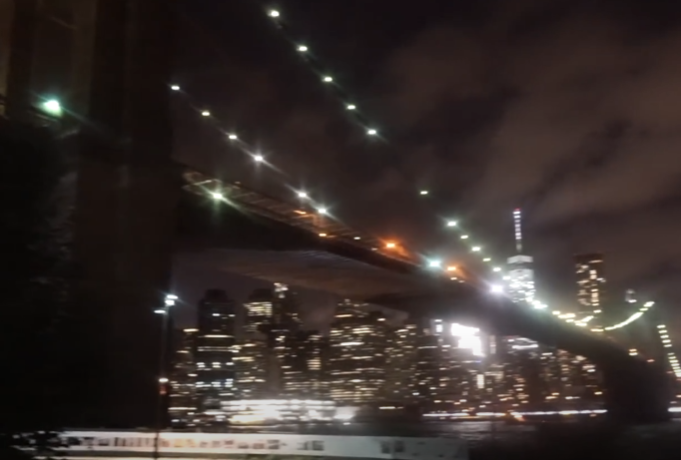}} &  Girl's expression after getting proposal & 27.17                 & 90.28       \\ \cline{2-4}
\multirow{2}{*}{}                  & A boy showing his arm after being stung at the beach                & 56.38                 & 30.00       \\ \hline
\end{tabular}
\caption{Case study on the temporal language grounding benchmark. We extract the POT distance between video and language of two inputs with different language queries and measure the respective AP performance change.}
\label{tab:case_study}
\end{table*}}

{\renewcommand{\arraystretch}{1.2}
\begin{table*}[h!]
\centering
\resizebox{\linewidth}{!}{
\begin{tabular}{p{0.2\linewidth}cccc}
\hline
\textbf{Recurrent architecture} & \textbf{\#params - UMT (M)} & \textbf{mAP - YouTube Highlights} & \textbf{\#params - VG-BART (M)} & \textbf{R2 - How2} \\ \hline
\centering GRU                      & 0.16                                      &                   76.07                                    & 1.17                                       &               21.85                         \\
\centering LSTM                     & 0.16                                     &                          76.08                             & 1.17                                       &         21.87                               \\
\rowcolor{Gray}
\centering RNN                      & 0.16                                      & 76.12                                                 & 1.17                                       & 21.91              \\ \hline
\end{tabular}
}
\caption{Recurrent architecture ablation experiment on YouTube Highlights and How2. We \colorbox{Gray}{color} the settings we implement for our READ method.}
\label{tab:recurrent_ablation_experiments}
\end{table*}}

{\renewcommand{\arraystretch}{1.2}
\begin{table}[h!]
\centering
\fontsize{9pt}{9pt}\selectfont
\begin{tabular}{c|>{\centering\arraybackslash}p{0.3\linewidth}>{\centering\arraybackslash}p{0.2\linewidth}}
\hline
\textbf{Distance method} & \textbf{mAP - YouTube Highlights} & \textbf{R2 - How2} \\ \hline
AvgPool - Cosine         &           71.65                        &           20.32        \\
MaxPool - Cosine         &       74.37                             &      21.36              \\
AvgPool - L2             &             72.11                      &           20.73         \\
MaxPool - L2             &              74.39                     &      21.02              \\
\rowcolor{Gray}
Partial OT                      & 76.12                             & 21.91   \\ \hline          
\end{tabular}
\caption{Distance method ablation experiments on YouTube Highlights and How2. We \colorbox{Gray}{color} the settings we implement for our READ method.}
\label{tab:distance_method_ablation_experiments}
\end{table}}

\noindent\textbf{Pre-trained backbones.} We adopt the Transformer encoder-decoder architecture \citep{vaswani2017attention} pre-trained with both supervised and self-supervised objectives. Specifically, for TLG, we use the unified multimodal transformer (UMT) \citep{liu2022umt} and Moment-DETR \citep{lei2021detecting} models pre-trained upon the automatic speech recognition task. For VLS, we carry out the parameter-efficient adaptation on the generative vision-guided BART (VG-BART) and T5 (VG-T5) \citep{yu2021vision} pre-trained upon reconstruction and masked language modeling tasks \citep{raffel2020exploring}. 

\noindent\textbf{Baseline methods.} We compare our method with a comprehensive list of baseline approaches for efficient video-language transfer learning:
\begin{itemize}
    \item \emph{Full}: update all parameters of the pre-trained backbone.
    \item \emph{Partial}: only update the last layers of the encoder and decoder in the Transformer model.
    \item \emph{Bias} \citep{zaken2022bitfit}: only fine-tune the bias terms in the Transformer backbone.
    \item \emph{Proj}: fine-tune only the last linear projection layer in the Transformer. 
    \item \emph{LoRA} \citep{hu2021lora}: solely fine-tune the decomposition matrices introduced to the linear weights of the Transformer model.
    \item \emph{Prompt} \citep{jia2022visual}: Append a sequence of learnable prompt tokens to both video and language inputs and only fine-tune the appended sequence.
    \item \emph{Adapter} \citep{houlsby2019parameter}: update only the adaptation modules consisting of downprojection and up-projection layers inserted into the Transformer model.
\end{itemize}

\noindent\textbf{Implementation details.} For the TLG task, we use the SlowFast \citep{feichtenhofer2019slowfast} and video encoder of CLIP \citep{radford2021learning} to extract features every 2 seconds. For the VLS task, we use a 3D ResNeXt-101 model to extract a 2048-dimensional embedding for every 16 non-overlapping frames. Similar to previous works \citep{houlsby2019parameter, chen2022adaptformer}, to support training stability, we initialize the weights of the down-projection layer $W_{\text{down}}$ with the Kaiming normal \citep{he2015delving} method, whereas those of the up-projection $W_{\text{up}}$, recurrent layer RNN, and biases of our READ layers are configured with zero initialization. In our PVLA framework, we implement the cost distance $c\left(\mathbf{h}_{v,i}, \mathbf{h}_{l,j}\right)$ as the cosine distance $c\left(\mathbf{h}_{v,i}, \mathbf{h}_{l,j}\right) = 1 - \frac{\mathbf{h}_{v,i} \cdot \mathbf{h}_{l,j}}{||\mathbf{h}_{v,i}||_{2} \cdot ||\mathbf{h}_{l,j}||_{2}}$, and set the maximum number of iterations $N_{\text{iter}}$ to 1,000 and the temperature $\tau$ to 0.05. We fine-tune all models leveraging the AdamW optimizer on 4 NVIDIA Tesla V100 GPUs and report average results of 5 runs. Specific details about the epoch, batch size, learning rate, and the number of Transformer blocks for each task can be found in the Appendix.
\subsection{Main Results}

\noindent For main comparison of our READ with baseline methods, we denote the results of YouTube Highlights in Table \ref{tab:youtube_highlights_results}, TVSum in Table \ref{tab:tvsum_results}, QVHighlights in Table \ref{tab:qvhighlights_resultss}, and How2 in Tables \ref{tab:how2_bart_results} and \ref{tab:how2_t5_results}.

\noindent\textbf{Temporal language grounding (TLG).} For the YouTube Highlights dataset, our READ framework substantially outperforms the Full fine-tuning approach (\emph{e.g.} 1.36\% on average, 1.75\% in the Dog domain, and 1.21\% in the Surfing domain), while updating far less parameters (0.16M vs. 283.97M). We significantly surpass all other efficient fine-tuning methods as well, \textit{e.g.} with an improvement of 10.96\% over the Adapter in the Gym category, or 5.78\% over LoRA in the Skating category.

For the TVSum dataset, we observe that our method enhances the Full fine-tuning direction with only 0.14M versus 285.28M tunable parameters. For instance, we obtain an increase of 4.26\% in the MS subset and 3.74\% on average. Compared with the best parameter-efficient approach, \emph{i.e.} the Adapter, we achieve a gain of 15.07\% in BT, 10.67\% in BK, and 8.77\% in the VU domain. 

Our improvement also generalizes across different pre-trained backbone. On the QVHighlights dataset, in which we work with the Moment-DETR architecture, we accomplish a gain of 0.6\% over the standard fine-tuning method, while our tunable parameters are only 0.19M versus its 15.88M. We also surpass the efficient approach LoRA with an enhancement of 2.78\% in mAP.

These results demonstrate that our READ framework can efficiently model video-language inputs to polish the low-resource temporal language grounding performance of various pre-trained Transformer models.

\noindent\textbf{Video-language summarization (VLS).} Analogous to the TLG experiment, on the VG-BART backbone, we improve upon the full fine-tuning approach with 8.29 points of ROUGE-1, 10.03 points of ROUGE-2, and 7.91 points of ROUGE-L. Importantly, we only update 1.17M parameters, which account for 0.47\% total parameters of the overall model. On the VG-T5 backbone, we exceed the full approach by 7.75 points in ROUGE-1, 10.64 points in ROUGE-2, and 7.89 points in ROUGE-L, whilst keeping 99.43\% parameters frozen. 

In addition, our framework substantially outperforms other fine-tuning strategies, \emph{e.g.} LoRA with 3.97 points in ROUGE-1, 1.55 points in ROUGE-2, and 1.93 points in ROUGE-L on the VG-BART architecture, along with 3.49 points in ROUGE-1, 1.99 points in ROUGE-2, and 1.65 points in ROUGE-L on the VG-T5 one. 

These results substantiate that our method is applicable to diverse benchmarks and model architectures, particularly not only multimodal Transformers for temporal language grounding but also generative Transformers for video-language summarization. We hypothesize that our advantages are due to the recurrent adapter’s ability to model temporal information and the PVLA task to align video and language signals to maintain more essential information during the efficient fine-tuning stage.
\subsection{Ablation Studies}
\noindent We ablate our READ framework to discover what factors result in the demonstrated efficiency and observe several intriguing properties. Our ablation studies are all conducted on the YouTube Highlights and How2 test set.

\noindent\textbf{Effects of video-language alignment.} We evaluate our framework without the assistance of the PVLA task and with the one of the VLA variant that requires all masses of one distribution to be transferred (we set $s = \min\left(N_V, N_L\right)$ in formulation (\ref{eq:pvla_formulation})). As shown in Table \ref{tab:pvla_ablation_experiments}, the performance drops dramatically when we remove the PVLA task from the fine-tuning procedure. We conjecture that the model has become deficient in managing the information injected into the low-dimensional space of the READ layers, thus passing detrimental noise to the downstream task. Moreover, the VLA variant brings slight performance decrease, which could be due to the VLA’s restrictive nature of transporting all masses from the language distribution to the video one or vice versa.

\noindent\textbf{Effects of the recurrent architecture.} In addition to RNN, there exist various recurrent architectures in the literature, particularly the gated recurrent unit (GRU) \citep{cho2014learning} and long short-term memory (LSTM) \citep{hochreiter1997long}. We experiment with different recurrent choices and explicate the results in Table \ref{tab:recurrent_ablation_experiments}. As can be observed, the performance is insensitive to the choice of recurrent design. Therefore, we select the simplest option, \emph{i.e.} recurrent neural network (RNN) for our READ layers.

\noindent\textbf{Distance methods.} We further ablate on the distance metrics to estimate the distance between video and language distributions. Technically, we perform the average- and max-pooling of the video and language representations. Then, we consider the cosine distance or the L2 distance of the two pooled vectors as the video-language distance. Results in Table \ref{tab:distance_method_ablation_experiments} substantiate the superiority of our POT distance for the PVLA objective. Such success illustrates the POT-based PVLA’s advantage of modeling the relationship nature between video and language representations
\subsection{Qualitative Assessment}
\noindent\textbf{Case study.} We display a TLG example on the YouTube Highlights dataset, along with the POT distance estimated by our PVLA framework and the AP score in Table \ref{tab:case_study}. We observe that when the language query semantically corresponds to a moment in the video, \textit{i.e. a girl expression after she gets the proposal}, the POT distance is small and correlates with the high value of AP. In contrast, when we replace the original query with an out-of-distribution one, the POT distance burgeons significantly, causing the AP to decrease from 90.28\% to 30.00\%. Therefore, we conclude that our READ framework is capable of intelligently adjusting the information flowing through the READ layers in order to produce the final output consistent with the video-language input and downstream tasks.

\section{Conclusion}
\noindent We propose a novel READ-PVLA framework for parameter-efficient transfer learning to video-language modeling tasks. Our READ-PVLA utilizes recurrent computation component to enable temporal modeling capability and partial video-language alignment objective to preserve critical information for bottleneck adaptation modules. Experiments demonstrate that READ-PVLA consistently outperforms both the full fine-tuning and competitive strategies, whilst bringing the benefit of parameter-efficiency (at most 1.20\% trainable parameters). Our method is also applicable to diverse pre-trained models, which has the potential to employ more powerful video-language models in the future.

\chapter[Encoding and Controlling Global Semantics for Long-form Video Question Answering]{Encoding and Controlling Global Semantics for Long-form Video \\ Question Answering}
\label{ch3:global}

\section{Introduction}
\noindent VideoQA has been extensively studied to develop systems to assist humans in daily activities \citep{grauman2022ego4d, lei2021assistsr}, \textit{e.g.}, remind users of their past actions, help users locate their belongings, and provide assistance with complex tasks. To implement these functions, we expect videoQA systems  to understand and extract relevant information from long-form videos with diverse objects and complex spatial-temporal interactions.

\begin{figure*}[t]
    \centering
    \includegraphics[width=\linewidth]{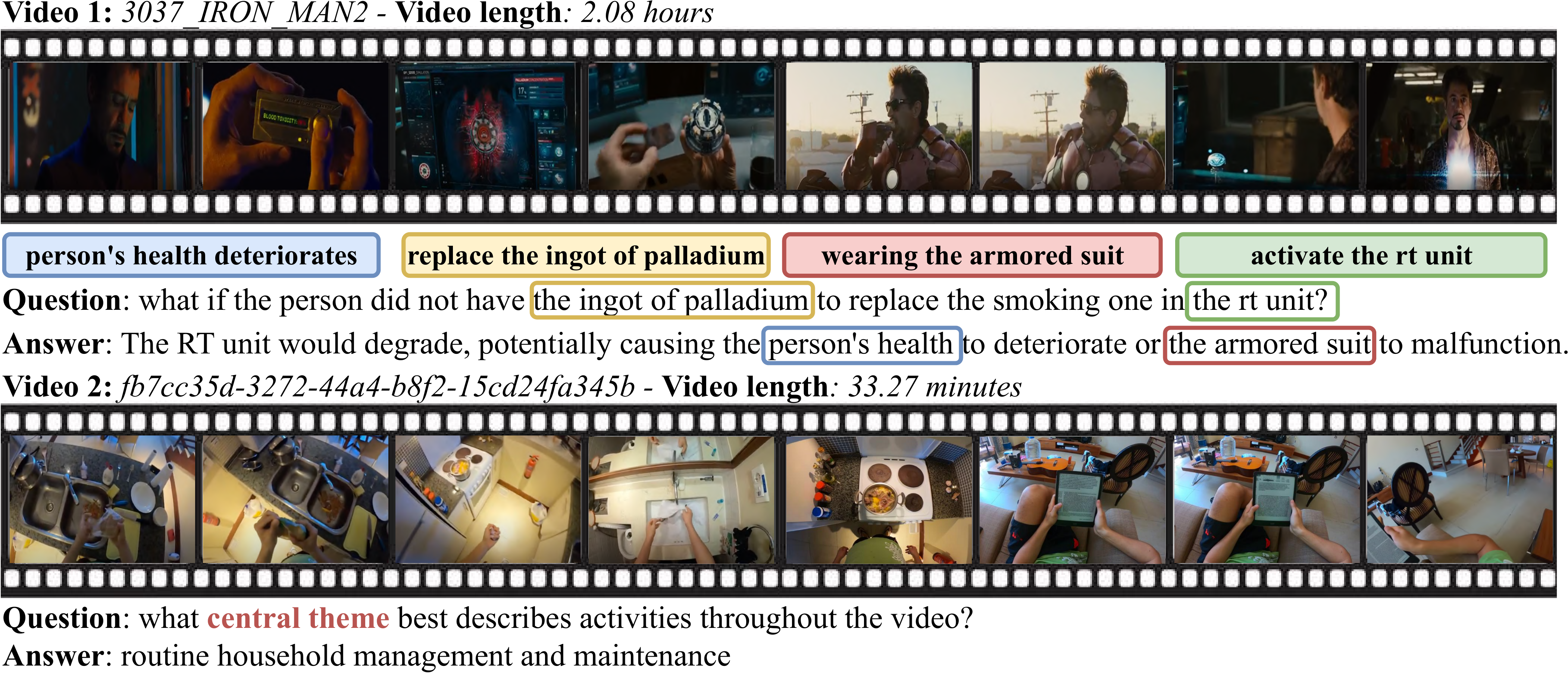}
    \caption{Long-form videoQA examples, with videos taken from MAD \citep{soldan2022mad} and Ego4D \citep{grauman2022ego4d} datasets, respectively. Question in video 1 requires the model to reason about the relation chain of replacing \textit{ingot of palladium} to activate the \textit{rt unit} that powers the \textit{armored suit} and protects \textit{person’s health}. Question in video 2 necessitates an understanding of the overall theme in video 2.}
    \label{fig:longform_videoqa_example}
\end{figure*}

Compared with short clips, long-form videos pose more challenges for videoQA.
They consist of a higher number of objects and events. As such, comprehensively encoding information from them requires expensive computations. Moreover, a high amount of information may be unrelated to the posed question. To address these problems, recent studies \citep{bain2021frozen, wang2023all, gao2023mist} adaptively select a subset of video frames and visual regions associated with the question. Nevertheless, if a question necessitates a reasoning of the entire sequence of events (\textit{e.g.} video 1’s Figure \ref{fig:longform_videoqa_example}), or an understanding of the overall video narration (\textit{e.g.} video 2’s Figure \ref{fig:longform_videoqa_example}), a mere handful of selected frames or regions might not sufficiently encapsulate necessary details.

To tackle these problems, we introduce a state space layer (SSL). Before forwarding video frames to selection modules, SSL fixes long-term dependency patterns for integrating global information into visual representations. Such global information offers the selected frames the global context within the video, so that they can relate to other frames even though those frames are not selected for attention computation. However, a considerable amount of unrelated global information may flow into visual representations. Therefore, we first equip SSL with a gating mechanism to provide more controllability over the flow of global semantics into visual representations, resulting in our \textbf{G}ated \textbf{S}tate space \textbf{M}ulti-modal \textbf{T}ransformer (GSMT) architecture. 
Furthermore, we promote global semantics that is more aligned with the question. In particular, we introduce \textbf{C}ross-modal \textbf{C}ompositional \textbf{C}ongruence (C$^3$) objective that compares visual attention with its version transitioned to the language basis via cross-modal attention, effectively measuring cross-modal congruence between intra-modal relations. Our rationale behind focusing on intra-modal relations is because videoQA models often need to understand spatial and temporal relationships between entities and events posed by the question \citep{gandhi2022measuring}, thus we encourage globally informed visual representations to maintain compositional consistency between visual patches and question entities. 

Remarkably, we observe that recent long-form videoQA works \citep{gao2023mist, islam2024video} still mostly evaluate on videos lasting at most one minute or two, and use short-natured questions which necessitate watching only a short period of video, \textit{i.e.}, about 100 seconds  \citep{mangalam2023egoschema}, to determine the answer. To more rigorously evaluate long-form videoQA capacity, we introduce a construction procedure which utilizes large language model (LLM) to generate questions and associated answers for egocentric and movie videos whose average lengths are 17.5 minutes and 1.9 hours, respectively. Additionally, we also conduct automatic and manual filtering to obtain high-quality questions which require watching a video up to 1200 seconds to answer, longer than any existing long-form videoQA benchmarks \citep{xiao2021next, wu2021star, mangalam2023egoschema}.

To sum up, our contributions are as follows:
\begin{itemize}[leftmargin=*]
    \item We propose a Gated State space Multi-modal Transformer (GSMT) with state space layer (SSL) to integrate global information into visual representations for long-form videoQA.
    
    \item We equip SSL with a gating mechanism to provide controllability over the flow of global video semantics and a Cross-modal Compositional Congruence (C$^3$) objective to encourage question-aligned visual representations. 
    
    \item We curate two new datasets with excessively long video lengths and long-natured questions for long-form videoQA. Comprehensive experiments on our curated and five standard datasets substantiate the superiority of our framework over various competitive baselines.
\end{itemize}

\section{Methodology}
\begin{figure*}[t]
    \centering
    \includegraphics[width=\linewidth]{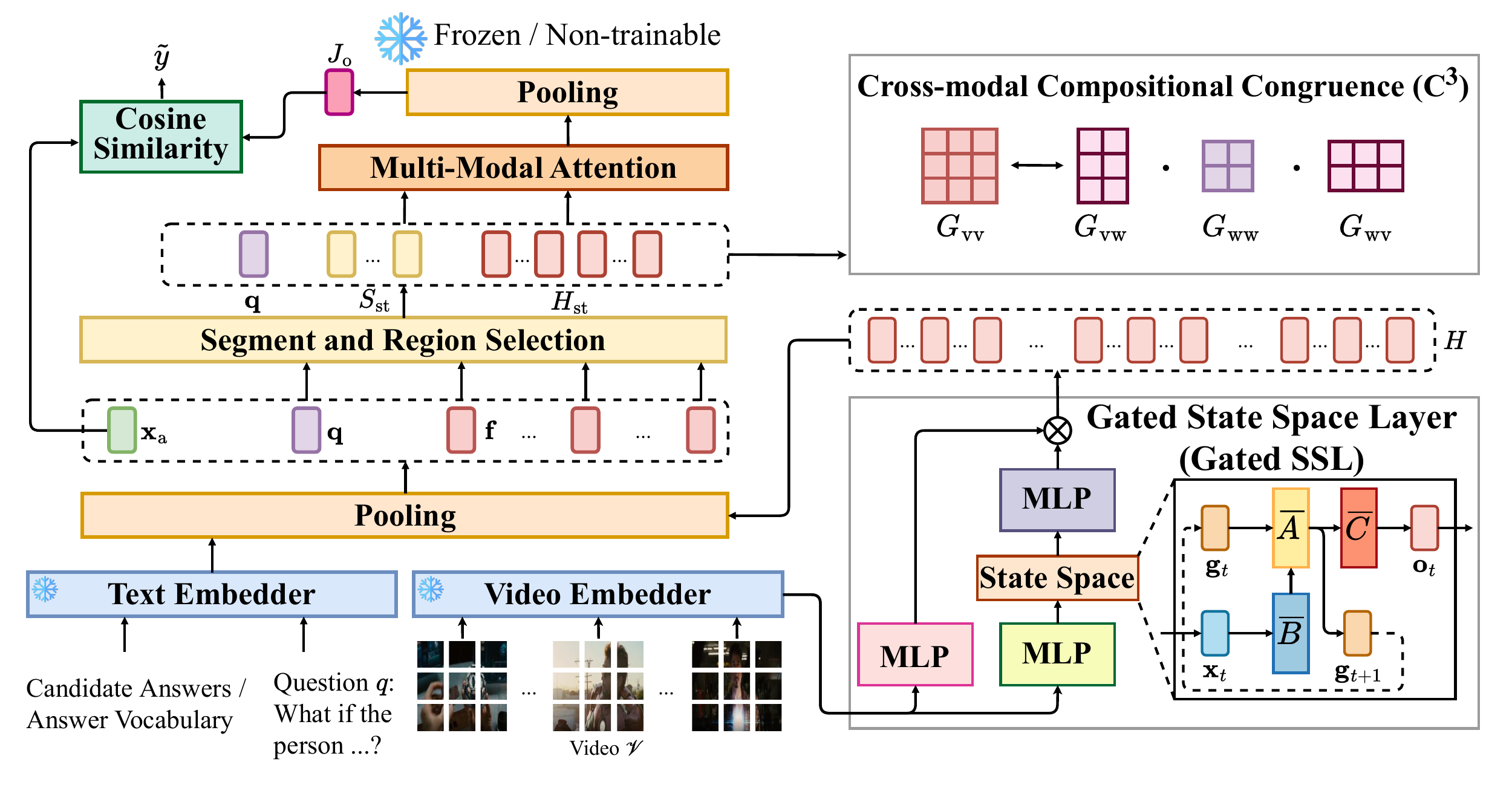}
    \caption{Illustration of the GSMT architecture empowered by gated SSL and C$^3$ training objective.}
    \label{fig:overall_architecture}
\end{figure*}

\noindent The formulation of video question answering (videoQA) is to predict the answer $y$ for a question $q$ about a video $\mathcal{V}$ as follows:
\begin{equation}
\tilde{y} = \argmax_{y \in \mathcal{A}} \mathcal{F}_{\theta} (y|q, \mathcal{V}, \mathcal{A}),
\end{equation}
where $\tilde{y}$ denotes the predicted answer chosen from the set of candidate options $\mathcal{A}$, and $\theta$ denotes the trainable parameters of a videoQA model. With this videoQA task formulation, we explain our proposed GSMT architecture as the model $\mathcal{F}_{\theta}$ and C$^3$ objective to support GSMT. The overall framework is illustrated in Figure \ref{fig:overall_architecture}.

\subsection{Gated State Space Multi-Modal Transformer (GSMT)}
\noindent Our GSMT takes videos and questions as input, divides each video frame into visual patches and a question into textual words, and then forwards visual patches and textual words into video and text embedder to extract initial representations.

\noindent\textbf{Input Embedder.} For video embedder, we utilize frozen pre-trained vision-language Transformer to extract patch-level features $X = \{\mathbf{x}_{0}, \mathbf{x}_{1}, ..., \mathbf{x}_{L-1}\}$, of all $T$ frames, where $t$-th frame consists of $N$ patch-level features $\{\mathbf{x}_{t,j}\}_{j=0}^{N-1}$, hence $L = NT$, and $\mathbf{x} \in \mathbb{R}^{d}$. For text embedder, a similar frozen pre-trained vision-language Transformer is used to obtain word-level features $W = \{\mathbf{w}_{0}, \mathbf{w}_{1}, ..., \mathbf{w}_{M-1}\}$, where $\mathbf{w}_{0}$ corresponds to the \texttt{[CLS]} token and $\mathbf{w}_{1}, ..., \mathbf{w}_{M-1}$ correspond to words in the question.

\noindent\textbf{Gated State Space Layer (Gated SSL).} Inspired by \citep{gu2021efficiently}, we define a sequence-to-sequence map $SSL(X)$ from a sequence of patch-level features to $d_{S}$-dim hidden states, parameterized by learnable state matrices $A \in \mathbb{R}^{d_{S} \times d}, B \in \mathbb{R}^{d_{S} \times d}, C \in \mathbb{R}^{d \times d_{S}}$, and step size $\Delta$ as:
\begin{gather}
\mathbf{g}_{t+1} = \overline{A} \cdot \mathbf{g}_{t} + \overline{B} \cdot \mathbf{x}_{t+1}, \quad \mathbf{o}_{t+1} = \overline{C} \cdot \mathbf{g}_{t+1}, \\
\overline{A} = e^{A\Delta}, \quad \overline{B} = (\overline{A} - I) A^{-1} B, \quad \overline{C} = C.
\end{gather}
We unroll the mapping to obtain:
\begin{equation}
\mathbf{o}_{t} = \sum\limits_{j=0}^{t} \overline{CA}^{j}\overline{B} \cdot \mathbf{x}_{t-j}.
\end{equation}
This can be written as a convolutional representation $O = \overline{\Gamma} * X$, where $\overline{\Gamma} = \left(\overline{CB}, \overline{CAB}, ..., \overline{CA}^{L-1}\overline{B}\right)$ denotes the convolutional kernel, $*$ the discrete convolution operator, $X$ the input sequence, $O$ the corresponding output sequence. This convolution denotes the fixed global dependency pattern that facilitates the computation of global information among visual patches. We use Fast Fourier Transformer (FFT) \citep{cooley1965algorithm} to compute the convolution in parallel provided that $\bar{\Gamma}$ has been obtained. 

Computing the kernel $\bar{\Gamma}$ is non-trivial since it requires $L$ distinct matrix powers. Instead, inspired by \citep{gupta2022diagonal}, we initialize $A$ to be a diagonal matrix $\text{diag}\left(\lambda_{1}, \lambda_{2}, ..., \lambda_{d_{S}}\right)$ and $B$ to all-one matrix $\mathbbm{1}_{d_{S} \times d}$. Due to this initialization, the kernel can be computed as:
\begin{equation}
\overline{\Gamma} = \left(C \odot E\right) \cdot P,
\end{equation}
where $E = \left(\frac{e^{\lambda_{1}\Delta-1}}{\lambda_{1}}, \frac{e^{\lambda_{2}\Delta-1}}{\lambda_{2}}, ..., \frac{e^{\lambda_{d_{S}}\Delta-1}}{\lambda_{d_{S}}}\right) \in \mathbb{R}^{d_{S}}$, in which $P_{i,j} = e^{\lambda_{i} \cdot j \cdot \Delta} \in \mathbb{R}^{d_{S} \times L}$, and $\odot$ denotes the element-wise multiplication.

To equip SSL with the control over which global semantics to integrate into visual representations, we construct a gating unit as: 
\begin{gather}
U = \phi\left(\text{Linear}(X)\right), \;\; V = \phi\left(\text{Linear}(X)\right), \\
O = \text{Linear}\left(\text{SSM}\left(U\right)\right), \\
H = \text{Linear}\left(O \odot V\right),
\end{gather}
where $U, V, O \in \mathbb{R}^{L \times d_{\text{gating}}}$ denote the intermediate gating representations, $H \in \mathbb{R}^{L \times d_{h}}$ the visual representations output by gated SSL, $\phi$ the non-linear activation, and $d_{\text{gating}} < d, d_{h}, d_{s}$.

\noindent\textbf{Visual Segment and Region Selection.} After obtaining visual patch-level hidden representations $H$, we proceed to obtain frame features by pooling each frame $t$’s visual patches:
\begin{equation}
\mathbf{f}_{t} = \text{pool}\left(\mathbf{h}_{t,0}, \mathbf{h}_{t,1}, ..., \mathbf{h}_{t,N-1} \right).
\end{equation}
Subsequently, we group non-overlapping consecutive frames into a segment to obtain $I$ segments, each of which contains $N_{p} = \lceil\frac{L}{I}\rceil$ patches and $N_{f} = \lceil \frac{T}{I}\rceil$ frames. We proceed to compute segment features through pooling frame features corresponding to the segment:
\begin{equation}
\mathbf{s}_{i} = \text{pool}\left(\mathbf{f}_{i,0}, \mathbf{f}_{i,1}, ..., \mathbf{f}_{i,N_{f}-1}\right).
\end{equation} 
Similarly, we also pool word features to obtain the question representation:
\begin{equation}
\mathbf{q} = \text{pool}\left(\mathbf{w}_{0}, \mathbf{w}_{1}, ..., \mathbf{w}_{M-1}\right).
\end{equation}
Given the segment features $S = \{\mathbf{s}_{i}\}_{i=0}^{I-1}$ and question feature $\mathbf{q}$, we conduct top-$k$ segment selection:
\begin{gather}
Q = \text{Linear}\left(\mathbf{q}\right),\; K = \text{Linear}\left(S\right), \\
\mathcal{B} = \mathop{selector}_{Top_{k}} \left(\text{softmax}\left(\frac{QK^{\top}}{\sqrt{d_{k}}}\right)\right).
\end{gather}
We implement $selector$ as a differentible Gumbel-softmax selection function. The output of the $selector$ is a sequence of segment index $\mathcal{B}$. Thereby, we extract respective segment features with respect to the selected segment indices, \textit{i.e.} $S_{\text{st}} = \{\mathbf{s}_{b} \;|\; b \in \mathcal{B}\}$.

For every frame $\tau$ in the selected segments, we then perform cross-modal attention between its patch-level hidden representations $H_{\tau} = \{\mathbf{h}_{\tau,j}\}_{j=0}^{N-1}$, $\tau \in \{\lfloor\frac{b}{N_{f}}\rfloor \; | \; b \in \mathcal{B}\}$ and the question to select top-$j$ question-related patches:
\begin{gather}
Q = \text{Linear}\left(\mathbf{q}\right), \; K = \text{Linear}\left(H_{\tau}\right), \\
\mathcal{T} = \mathop{selector}_{Top_{j}} \left(\text{softmax}\left(\frac{QK^{\top}}{\sqrt{d_{k}}}\right)\right).
\end{gather}
Lastly, we stack the selected patches of all selected frames to obtain $H_{\text{st}} = \{\mathbf{h}_{\tau,j}|\tau \in \mathcal{T}\}_{j=0}^{N-1}$.

\noindent\textbf{Multi-Modal Attention.} At present, we employ self-attention to produce multi-modal hidden representations that fuse the information of question and video. In particular, we concatenate the question word-level features $Q = \{\mathbf{w}_{i}\}_{i=0}^{M-1}$, selected segment features $S_{\text{st}} = \{\mathbf{s}_{b} \; | \; b \in \mathcal{B} \}$, and selected patch features $H_{\text{st}} = \{\mathbf{h}_{\tau,j}|\tau \in \mathcal{T}\}_{j=0}^{N-1}$:
\begin{equation}
J = \left[\text{Linear}(S_{\text{st}}), \text{Linear}(H_{\text{st}}), \text{Linear}(\mathbf{q})\right],
\end{equation}
where $[;]$ denotes the concatenation operator. Thereafter, we iteratively conduct self-attention over the concatenated features for $N_{L}$ layers to accomplish multi-modal contextual representations:
\begin{equation}
J^{(i+1)} = \text{SelfAttention} \left(J^{(i)}\right), \; 0 \leq i \leq N_{L} - 1.
\label{eq:multimodal_self_attention}
\end{equation}

\noindent\textbf{Answer Prediction.} Afterwards, we pool the features of all multi-modal self-attention layers:
\begin{equation}
J_{\text{o}} = \text{pool}\left(J^{(0)}, J^{(1)}, ..., J^{(N_{L}-1)}\right).
\end{equation}
Then, we calculate the cosine similarity between $J_{o}$ and the feature of all candidate answers $X_{\text{ans}} = \{\mathbf{x}_{a} \; | \; a \in \mathcal{A}\}$ obtained by utilizing the pre-trained model. We choose the candidate answer of the highest similarity as the final prediction $\tilde{y}$:
\begin{equation}
\tilde{y} = \argmax_{y \in \mathcal{A}} \left(J_{\text{o}} \cdot \left(X_{\text{ans}}\right)^{\top}\right).
\end{equation}

\subsection{Cross-modal Compositional Congruence (C$^3$) Objective} 
\noindent Based on Eq. (\ref{eq:multimodal_self_attention}), we denote the output representations of visual patches as $J_{\text{v}}$ and those of question words as $J_{\text{w}}$. We calculate cross-modal attention:
\begin{gather}
G_{\text{vw}} = J_{\text{v}} \cdot \left(J_{\text{w}}\right)^{\top}, \;\; G_{\text{wv}} = J_{\text{w}} \cdot \left(J_{\text{v}}\right)^{\top},
\end{gather}
and intra-modal visual and textual attention as:
\begin{gather}
G_{\text{vv}} = J_{\text{v}} \left(J_{\text{v}}\right)^{\top}, \;\; G_{\text{ww}} =  J_{\text{w}} \left(J_{\text{w}}\right)^{\top}.
\end{gather}

Given the intra-modal and cross-modal attention, we perform the change of basis to compute the intra-modal visual attention in the language space:
\begin{gather}
R_{\text{vv}} = G_{\text{vw}} G_{\text{ww}}  \left(G_{\text{vw}}\right)^{\top}.
\end{gather}
As such, we define the loss objective to align the original $G_{\text{vv}}$ with the change-of-basis version $R_{\text{vv}}$:
\begin{equation}
\mathcal{L}_{\text{C}^{3}} = \text{m-KL}\left(\text{softmax}\left(R_{\text{vv}}\right), \text{softmax}\left(G_{\text{vv}}\right)\right), 
\end{equation}
where m-KL denotes the symmetric Kullback-Leilber Divergence (KL) between $R$ and $G$:
\begin{equation}
\text{m-KL}(R,G) = \text{KL}(R||G) + \text{KL}(G||R).
\end{equation}

\subsection{Overall Objective}
\noindent We jointly optimize the softmax cross-entropy loss $\mathcal{L}_{\text{CE}}$ between the log-likelihood of the model prediction $\tilde{y}$ and the groundtruth answer $y$, with our proposed cross-modal alignment objective $\mathcal{L}_{\text{C}^{3}}$:
\begin{equation}
\mathcal{L} = \mathcal{L}_{\text{CE}} + \gamma \cdot \mathcal{L}_{\text{C}^{3}},
\end{equation}
where $\gamma$ denotes the hyperparameter to control the effect of the cross-modal alignment objective.

\section{Ego-QA and MAD-QA Benchmarks}
\noindent To more rigorously evaluate long-form videoQA models, we construct two new datasets, \textit{\textit{i.e.}} Ego-QA and MAD-QA.

\subsection{Ego-QA} 
\noindent We inherit 3k hours of 8640 egocentric videos from the Ego4D dataset \citep{grauman2022ego4d}. Each video is associated with about 280 dense captions of consecutive moments. Based on these captions, we create our dataset in 2 stages, \textit{\textit{i.e.}} question-answer generation and data filtering. 

\noindent\textbf{Question-answer generation.} In this first stage, we concatenate a video’s dense captions following the time order to construct its language description. We utilize GPT-4 \citep{achiam2023gpt} to generate 20 questions per video. In our prompt, we encourage GPT-4 to avoid questions that are visually biased and can be answered by a short video moment. Then, we present the generated questions to GPT-4 to generate the correct answer along with 4 wrong answer choices. 

\noindent\textbf{Data filtering.} In the second stage, we filter out questions that include clue words, \textit{e.g.} ``\textit{passage}'', ``\textit{text}'', and ``\textit{description}''. Moreover, we also remove questions that GPT-4 can answer without looking at the concatenated narration or the question. Then, we adopt manual filtering by asking ten graduate students who are native English speakers to ensure the veracity and temporal certificate length for every question-answer sample. Particularly, annotators are instructed to verify that 1) questions are valid and the correct answer is indeed correct, 2) all distractor answers are incorrect, and 3) the video length to watch to determine the correct answer is at least 2 minutes. 

The filtering stage reduces the number of admissible questions by a factor of $4\times$ to $5\times$. We accomplish 18.8K questions for 992 videos, which we split into 80\% train, 10\% val, and 10\% test. 

\subsection{MAD-QA} 
\noindent We follow the same process for Ego-QA to obtain MAD-QA by utilizing 1.2K hours of 650 videos from the MAD dataset \citep{soldan2022mad}. Since video lengths and the number of dense captions are larger than Ego-QA, we ask GPT-4 to generate 60 instead of 20 questions per video. Because GPT-4 might store external knowledge about the movie, we replace the name of characters in the caption with $person\_1$, $person\_2$, etc. Afterwards, we obtain 15.7K questions for 650 videos, and we split them into 80\%  train, 10\% val, and 10\% test.

The average video lengths in Ego-QA and MAD-QA are 17.5 minutes and 1.9 hours, respectively. Moreover, the average necessary video lengths humans need to watch to determine the answer for the two datasets are respectively 1204.4 and 396.07 seconds, longer than the average 100-second length of the recent very long-form videoQA dataset EgoSchema \citep{mangalam2023egoschema}. 

\section{Experiments}
\begin{table*}[t]
\centering
\resizebox{\linewidth}{!}{
\begin{tabular}{l|c|c|c|c|c|c|c|c|c}
\hline
\multicolumn{1}{c|}{\textbf{Question Types}} & \textbf{Obj-Rel} & \textbf{Rel-Act} & \textbf{Obj-Act} & \textbf{Super} & \textbf{Seq} & \textbf{Exists} & \textbf{Dur-Comp} & \textbf{Act-Rec} & \textbf{All} \\ \hline
AIO                                         & 48.34                    & 48.99                    & 49.66                  & 37.53                & 49.61               & 50.81           & 45.36                        & 18.97                         & 48.59        \\
ATP                                         & 50.15                    & 49.76                    & 46.25                  & 39.78                & 48.25               & 51.79           & 49.59                        & 18.96                         & 49.79        \\
MIST-AIO                                    & 51.43                    & 54.67                    & 55.37                  & 41.34                & 53.14               & 53.49           & 47.48                        & 20.18                         & 50.96        \\
MIST-CLIP                                   & 51.68                    & 67.18                    & 68.99                  & 42.05                & 67.24               & 60.33           & 54.62                        & 19.69                         & 54.39        \\ \hline
GSMT-AIO                            &       53.67                   &           56.10               &                 56.61       &          43.44            &          53.84           &         56.26        &                            49.83   &                21.73               &       52.61       \\
GSMT-CLIP                &    \textbf{53.94}       &           \textbf{69.84}               &     \textbf{72.53}                        &            \textbf{44.19}           &           \textbf{69.12}             &      \textbf{61.45}              &          \textbf{57.32}                          &               \textbf{21.20}                &     \textbf{56.16}        \\ \hline
\end{tabular}}
\caption{Results of videoQA on AGQA-v2.}
\label{tab:exp_agqav2}
\end{table*}

\begin{table}[t]
\centering
\resizebox{0.8\linewidth}{!}{
\begin{tabular}{l|ccccc|c}
\hline
\multicolumn{1}{c|}{\textbf{Method}} & \textbf{Attribute} & \textbf{State} & \textbf{Event} & \textbf{Order} & \textbf{Number} & \textbf{All} \\ \hline
STAGE                               & 39.49              & 49.93          & 34.52          & 55.32          & 38.54           & 41.97        \\
AIO                                 & 41.78              & 52.98          & 37.57          & 55.16          & 38.50           & 44.86        \\
ATP                                 & 42.87              & 53.49          & 38.35          & 55.25          & 38.65           & 45.43        \\
MIST-AIO                            & 43.63              & 55.17          & 40.99          & 55.44          & 39.54           & 47.19        \\
MIST-CLIP                           & 44.05              & 58.13          & 42.54          & 56.83          & 40.32           & 48.97        \\ \hline
GSMT-AIO                          &      48.76              &       57.99         &             44.96    &           57.39      &    43.18            &       50.81       \\
GSMT-CLIP                          &        \textbf{49.23}            &     \textbf{61.10}           & \textbf{46.66}               &     \textbf{58.83}           &       \textbf{44.03}          &     \textbf{52.73}       \\ \hline 
\end{tabular}}
\caption{Results of videoQA on Env-QA.}
\label{tab:exp_envqa}
\end{table}

\begin{table}[t]
\centering
\resizebox{0.8\linewidth}{!}{
\begin{tabular}{l|cccc|c}
\hline
\multicolumn{1}{c|}{\textbf{Method}} & \textbf{Interaction} & \textbf{Sequence} & \textbf{Prediction} & \textbf{Feasibility} & \textbf{Mean} \\ \hline
CLIP                                & 39.80                & 40.50             & 35.50               & 36.00                & 38.00         \\
RESERVE-B                           & 44.80                & 42.40             & 38.80               & 36.20                & 40.50         \\
Flamingo-9B                         & -                    & -                 & -                   & -                    & 43.40         \\
AIO                                 & 47.53                & 50.81             & 47.75               & 44.08                & 47.54         \\
ATP                                 & 50.63                & 52.87             & 49.36               & 40.61                & 48.37         \\
CoVGT                           & -               & -             & -               & -                & 46.23         \\
MIST-AIO                            & 53.00                & 52.37             & 49.52               & 43.87                & 49.69         \\
MIST-CLIP                           & 55.59                & 54.23             & 54.24               & 44.48                & 51.13         \\ 
GLOBAL                   &             \textbf{59.36}         &          \textbf{57.52}         &            \textbf{57.21}         &          \textbf{46.68}            &     \textbf{52.85}         \\ \hline
\end{tabular}}
\caption{Results of videoQA on STAR.}
\label{tab:exp_star}
\end{table}

\begin{table}[t]
\centering
\resizebox{0.7\linewidth}{!}{
\begin{tabular}{l|ccc|c}
\hline
\multicolumn{1}{c|}{\textbf{Method}} & \textbf{Causal} & \textbf{Temporal} & \textbf{Descriptive} & \textbf{All} \\ \hline
HQGA                                & 48.48           & 51.24             & 61.65                & 51.42        \\
CLIP                                & 46.30           & 39.00             & 53.10                & 43.70        \\
VQA-T                               & 49.60           & 51.49             & 63.19                & 52.32        \\
AIO                                 & 48.04           & 48.63             & 63.24                & 50.60        \\
ATP                                 & 53.10           & 50.20             & 66.80                & 54.30        \\
CoVGT                               & 59.69           & 58.00             & 69.88                & 60.73        \\
MIST-AIO                            & 51.54           & 51.63             & 64.16                & 53.54        \\
MIST-CLIP                           & 54.62           & 56.64             & 66.92                & 57.18        \\ \hline
GSMT-AIO                    &           59.72      &        59.04           &         69.91             &         60.76     \\
GSMT-CLIP                   &         \textbf{60.87}        &         \textbf{61.16}          &           \textbf{70.26}           &         \textbf{62.49}     \\ \hline
\end{tabular}}
\caption{Results of videoQA on NExT-QA.}
\label{tab:exp_nextqa}
\end{table}

\begin{table}[t]
\centering
\resizebox{0.3\linewidth}{!}{
\begin{tabular}{l|c}
\hline
\multicolumn{1}{c|}{\textbf{Method}} & \textbf{Accuracy} \\ \hline
EgoVLP                              & 34.86             \\
VideoReCap                 & 50.23             \\
MIST-AIO                            & 56.27             \\
MIST-CLIP                           & 56.42             \\ \hline
GSMT-AIO                    & 58.28             \\
GSMT-CLIP                   & \textbf{58.55}            \\ \hline
\end{tabular}}
\caption{Results of videoQA on EgoSchema.}
\label{tab:exp_egoschema}
\end{table}

\begin{table}[t]
\centering
\resizebox{0.6\linewidth}{!}{
\begin{tabular}{l|c|c|c}
\hline
\multicolumn{1}{c|}{\textbf{Method}} & \textbf{Params} & \textbf{Ego-QA} & \textbf{MAD-QA} \\ \hline
AIO [CVPR 2023]       &          110M                & 24.19           & 16.14           \\
CoVGT [ECCV 2022]      &      159M                   & 26.72           & 15.71           \\
MIST-AIO [CVPR 2023]    &        382M                & 27.71           &  14.19           \\
MIST-CLIP [CVPR 2023]   &     382M                  & 29.73           &   17.15         \\ 
SeViLa [NeurIPS 2023]        &   4.2B    & 21.96           & OOM   \\
GLOBAL         &    384M              &    \textbf{32.40}      &   \textbf{19.11}       \\ \hline
\end{tabular}}
\caption{Results on constructed Ego-QA and MAD-QA.}
\label{tab:exp_egoqa_madqa}
\end{table}

\begin{table}[t]
\centering
\resizebox{0.6\linewidth}{!}{
\begin{tabular}{l|c|c|c|c}
\hline
\multicolumn{1}{c|}{\textbf{Method}} & \textbf{NExT-QA} & \textbf{STAR} & \textbf{Ego-QA} &  \textbf{MAD-QA} \\ \hline
SSL                            &  61.91     & 52.40            & 31.30                & 18.95           \\
Non-diag SSL                     &  60.44           & 51.63            & 30.51                  & 18.86           \\
Attention                 &   59.74             & 51.07            & 30.06                  & 18.41  \\ 
Convolution                 &   59.34             & 50.85            & 30.17                  & 18.33  \\ \hline
Gated SSL                 &   \textbf{62.49}            & \textbf{52.85}           & \textbf{32.40}                 & \textbf{19.11} \\ \hline
\end{tabular}}
\caption{Ablation results of gated SSL.}
\label{tab:ablation_gss}
\end{table}

\begin{table}[t]
\centering
\resizebox{0.8\linewidth}{!}{
\begin{tabular}{l|c|c|c|c}
\hline
\multicolumn{1}{c|}{\textbf{Method}} & \textbf{NExt-QA} & \textbf{STAR} & \textbf{Ego-QA} & \textbf{MAD-QA} \\ \hline
GSMT           &      60.78      & 49.67            & 29.73         &      18.00    \\
GSMT w/ OT        &   60.89      & 50.30            & 30.14         & 18.22        \\  
GSMT w/ POT       &     60.93    & 50.59            & 30.55         & 18.26        \\  \hline 
GSMT w/ C$^3$      &    \textbf{62.49}      & \textbf{52.85}           & \textbf{32.40}       & \textbf{19.11}        \\  
\hline
\end{tabular}}
\caption{Ablation results of cross-modal alignment.}
\label{tab:ablation_cross_modal_alignment}
\end{table}

\begin{table}[t]
\centering
\resizebox{0.8\linewidth}{!}{
\begin{tabular}{l|c|c|c|c|c|c}
\hline
\multicolumn{1}{c|}{\textbf{Dataset / $d_{\text{gating}}$}} & \textbf{32} & \textbf{64} & \textbf{128} & \textbf{256} & \textbf{512} & \textbf{No gating} \\ \hline
NExT-QA                                                    & 60.74       & 61.09       & \textbf{62.49}        & 61.90        & 61.33        & 60.61              \\
STAR                                                       & 51.82       & 52.63       & \textbf{52.85}        & 52.70        & 51.86        & 51.27              \\
Ego-QA                                                     & 29.96       & 31.37       & \textbf{32.40}        & 31.55        & 31.25        & 29.88   \\ \hline
\end{tabular}}
\caption{Effect of various $d_{\text{gating}}$ dimensions and no gating on NExT-QA, STAR, and Ego-QA datasets.}
\label{tab:ablation_study_gating_unit}
\end{table}

\begin{table}[h!]
\centering
\resizebox{0.8\linewidth}{!}{
\begin{tabular}{l|c|c|c|c}
\hline
\multicolumn{1}{c|}{\textbf{Method}} & \textbf{NExT-QA} & \textbf{STAR} & \textbf{Ego-QA} & \textbf{MAD-QA} \\ \hline
Multi-modal SSL                     & 60.69            & 50.87         & 29.74           & 18.61           \\
GSMT                        & \textbf{62.49}            & \textbf{52.85}        & \textbf{32.40}           & \textbf{19.11}           \\ \hline
\end{tabular}}
\caption{Effect of the position of SSL.}
\label{tab:ablation_study_position_ssl}
\end{table}

\subsection{Standard Benchmarks}
\noindent In addition to our constructed Ego-QA and MAD-QA datasets, we follow previous works \citep{gao2023mist, xiao2023contrastive, wang2023all} to evaluate GSMT on four additional publicly available datasets for long-form videoQA: AGQA \citep{grunde2021agqa}, NExT-QA \citep{xiao2021next}, STAR \citep{wu2021star}, Env-QA \citep{gao2021env}, and EgoSchema \citep{mangalam2023egoschema}.

\noindent\textbf{AGQA} \citep{grunde2021agqa} is a videoQA dataset for compositional spatio-temporal reasoning. As recommended by the dataset creator, we employ its v2 version, which possesses more balanced distributions. AGQA consists of 2.27M QA pairs for 9.7K videos.

\noindent\textbf{NExT-QA} \citep{xiao2021next} focuses on causal and temporal reasoning. The dataset comprises 5,440 videos associated with 52K questions.

\noindent\textbf{STAR} \citep{wu2021star} concentrates on situated reasoning questions. The dataset provides 60K questions related to 22K videos clips.

\noindent\textbf{Env-QA} \citep{gao2021env} is curated for dynamic environment understanding. Env-QA contains 23K egocentric videos collected on virtual environment AI2THOR \citep{kolve2017ai2}, which are used to generate 85K questions.

\noindent\textbf{EgoSchema} \citep{mangalam2023egoschema} consists of egocentric videos of 3-minute length. Questions in EgoSchema require humans on average 100 seconds watching the video to answer. 

\begin{figure}[t]
    \centering
    \includegraphics[width=\linewidth]{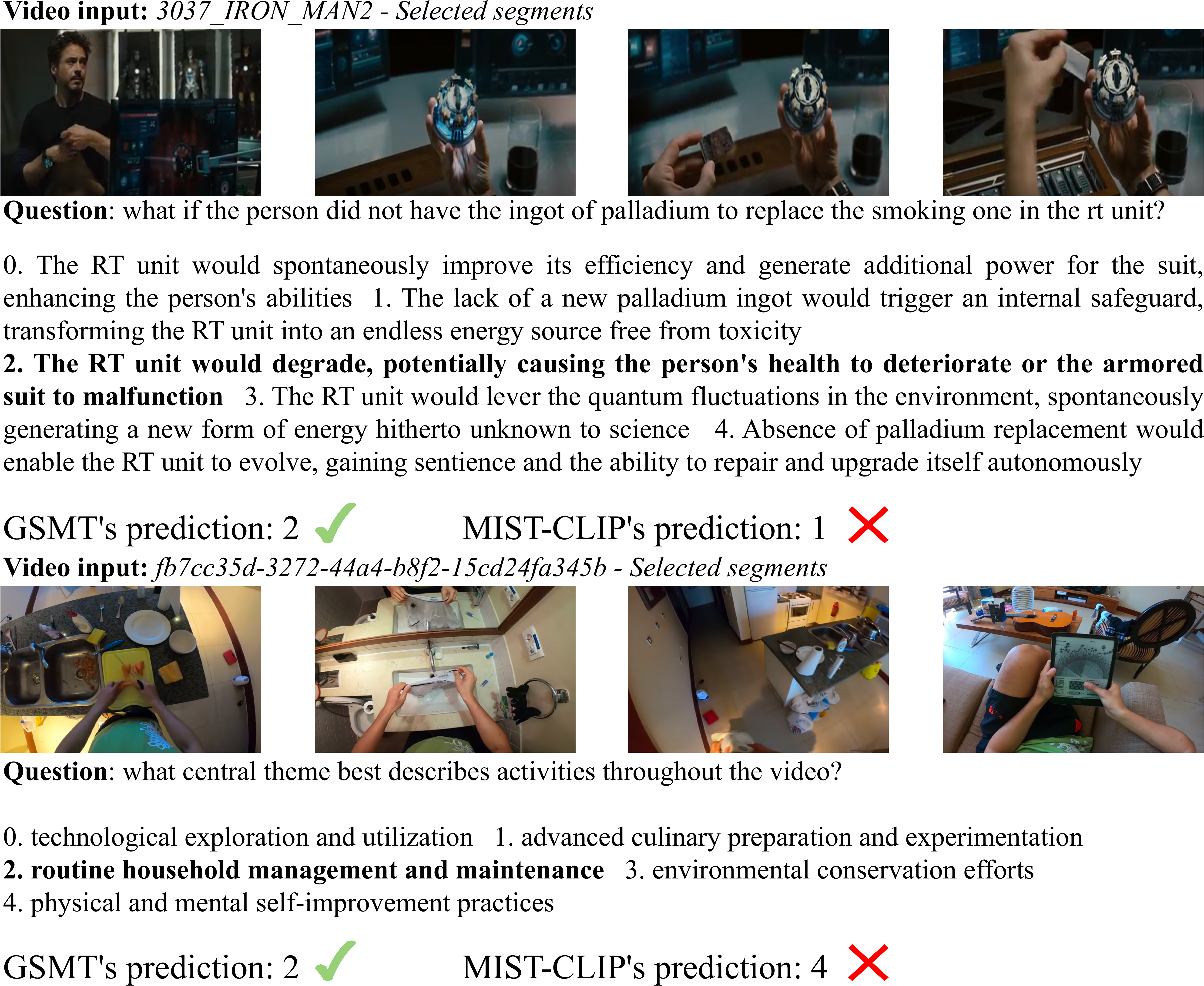}
    \caption{Qualitative results on the constructed MAD-QA and Ego-QA datasets.}
    \label{fig:example_qualitative_analysis}
\end{figure}

\begin{figure}[t]
    \centering
    \includegraphics[width=0.6\linewidth]{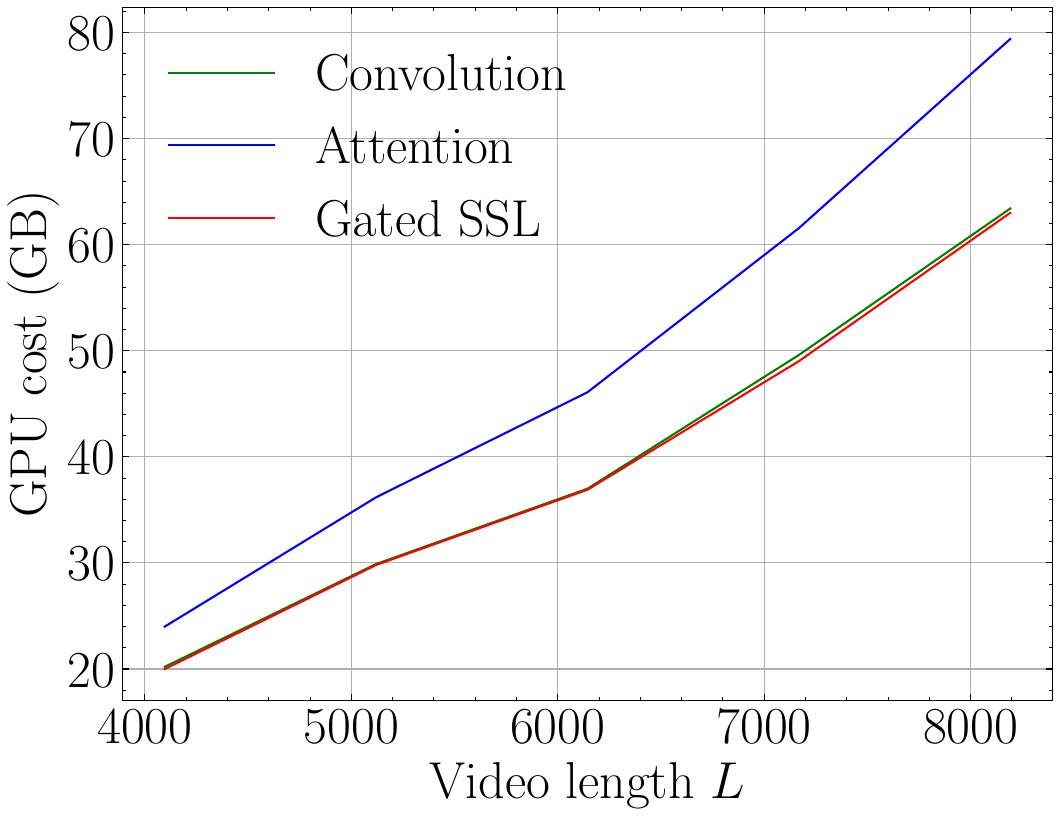}
    \caption{GPU memory cost with respect to visual sequence length $L$ of attention, convolution, and gated SSL mechanism.}
    \label{fig:ablation_study_gpu_cost}
\end{figure}

\subsection{Implementation Details}
\noindent Our framework can be implemented on most multi-modal Transformers. To fairly compare with previous works \citep{wang2023all, gao2023mist}, we evaluate upon two popular types of pre-trained models, \textit{\textit{i.e.}} CLIP (ViT/B-32) \citep{radford2021learning} and All-in-One-Base \citep{wang2023all}. We divide a video frame into $4\times4$ patches and send them to video embedder. For our gated SSL, we use $d_S = d = d_{h} = 512, d_{\text{gating}} = 128$. For selection modules, we use top-$k$ = 4, top-$j$ = 12. For the multimodal attention module, we use $N_L = 2$. Based on validation, we employ max-pooling for all pooling operations, ReLU activation for $\phi$, and $N_L = 2$. We apply $\mathcal{L}_{C^{3}}$ upon $J^{(2)}$, and observe no difference between applying on $J^{(1)}$ and $J^{(2)}$.  For fair comparison on AGQA, NExT-QA, STAR, Env-QA, and EgoSchema datasets, we sample 32 frames per video, and split them into $K = 8$ segments. Since video lengths are longer in our EgoQA and MAD-QA datasets, we sample 128 and 8192 frames per video, respectively, and split into $K = 8$ segments. For language modality, we embed the question with the same pre-trained model as the video embedder, and embed the answer with the pre-trained BERT-base model. We apply $\lambda = 0.005$ to balance the scale of $\mathcal{L}_{C^{3}}$ and $\mathcal{L}_{\text{CE}}$.

\subsection{Baselines}
We compare our proposed framework against the following baseline:
\begin{itemize}
    \item \textbf{HQGA} \citep{xiao2022hqga}: models video as a conditional graph hierarchy which aggregates visual facts in a level-wise manner with the guidance of textual modality,
    \item \textbf{CoVGT} \citep{xiao2023contrastive}: a model which exploits a graph transformer to encode video by capturing visual objects and their relations for spatio-temporal reasoning, along with contrastive learning to align its graph transformer with the textual encoder,
    \item \textbf{ATP} \citep{buch2022revisiting}: a model which learns to select one salient video frame to perform the videoQA task,
    \item \textbf{AIO} \citep{wang2023all}: an end-to-end videoQA model that releases the need of unimodal encoders and leverages non-parametric token rolling operation,
    \item \textbf{STAGE} \citep{lei2020tvqa+}: a model that jointly carries out videoQA with moment localization and object grounding,
    \item \textbf{VQA-T} \citep{yang2021just}: a cross-modal Transformer which is pre-trained with supervised contrastive learning on a large-scale videoQA dataset,
    \item \textbf{RESERVE-B} \citep{zellers2022merlot}: jointly represents video frames, texts, and audio, and learns to predict the masked textual and audio tokens given video frames,
    \item \textbf{Flamingo-9B} \citep{alayrac2022flamingo}: a model which is able to process multi-modal prompt and cast videoQA as text prediction,
    \item \textbf{EgoVLP} \citep{lin2022egocentric}: pre-trains multi-modal Transformer on egocentric videos for egocentric videoQA.
    \item \textbf{VideoReCap} \citep{islam2024video}: a recursive model that synergizes different video hierarchies to process hour-long videos,
    \item \textbf{MIST} \citep{gao2023mist}: a multi-modal Transformer which decomposes video into frames, patches, and segments to efficiently process with its self-attention mechanism.
\end{itemize}

\subsection{Quantitative Results}
\noindent We show our experiments on AGQA-v2, Env-QA, STAR, NExT-QA, and EgoSchema in Table \ref{tab:exp_agqav2}, \ref{tab:exp_envqa}, \ref{tab:exp_star}, \ref{tab:exp_nextqa}, and \ref{tab:exp_egoschema}, respectively, while revealing results on our constructed Ego-QA and MAD-QA in Table \ref{tab:exp_egoqa_madqa}. We can observe that our method achieves superior performance over the latest methods on all datasets. In terms of the overall accuracy, we outperform the second-best method on AGQA-v2, Env-QA, STAR, and EgoSchema, \textit{\textit{i.e.}} MIST-CLIP, by 1.77\%, 3.76\%, 1.72\%, and 2.13\%, respectively. Equivalently, we outperform CoVGT, which is the second-best method on NExT-QA, by 1.20\%.

Inspecting more closely, we note that our framework obtains more significant performance increase on questions that require the capacity of reasoning among visual concepts, \textit{\textit{i.e.}} improving 2.66\% and 3.54\% respectively for \textit{relation-action} and \textit{object-action} on AGQA-v2, 6.25\% and 4.52\% respectively for causal and temporal on NExT-QA, than those that require the ability to extract information within one frame, \textit{\textit{i.e.}} improving 2.26\% for \textit{object-relation} on AGQA-v2 and 3.34\% for \textit{descriptive} on NExT-QA. These results demonstrate our global semantics signal can address the challenging long-range temporal reasoning problems of long-form videoQA.

Remarkably, existing methods demonstrate significantly low performance on our curated datasets. For example, MIST-CLIP only achieves 29.73\% on Ego-QA, and 17.15\% accuracy on MAD-QA, which is less than random chance. In contrast, humans obtain 80.29\% and 73.21\% accuracy on Ego-QA and MAD-QA, respectively. These results suggest that previous methods might not encompass sufficient information in their selected segments and visual regions. Conversely, with the integrated global information, our framework can enhance videoQA performance on these challenging datasets. However, the accuracy remains substantially below human performance. Future research should focus more on genuine long-form videoQA, where videos can extend to several hours.

\subsection{Ablation Study}

\noindent\textbf{Gated SSL implementation.} We explore the effect of our gated SSL in Table \ref{tab:ablation_gss} on NExT-QA, STAR, Ego-QA, and MAD-QA datasets. As can be observed, removing the gating unit, \textit{i.e.} the SSL approach, results in performance drops, since redundant and noisy information might be passed to the visual representations. Additionally, not initializing state space parameters as diagonal matrices, \textit{i.e.} non-diag SSL, does not remarkably impact the performance. However, the time and memory complexity would become $O(L^2)$, which is significantly more costly than our initialization approach.

\noindent\textbf{Choices for Global Semantics.} We compare gated SSL with other choices to extract global semantics among visual elements, \textit{i.e.} self-attention and convolution, in terms of videoQA performance in Table \ref{tab:ablation_gss} and GPU memory cost in Figure \ref{fig:ablation_study_gpu_cost}. As can be observed, our gated SSL not only brings less computational cost than self-attention but also higher accuracy, validating the effectiveness of its global information signal. Moreover, whereas convolution pays equivalent computational cost to our gated SSL, its local pattern does not provide productive contextual information among visual elements, resulting in lower accuracy than gated SSL. 

\noindent\textbf{Effect of Gating Unit.} We ablate the gating unit and vary the gating dimension $d_{\text{gating}}$ in our gated SSL. As shown in Table \ref{tab:ablation_study_gating_unit}, increasing the gating dimension leads to higher videoQA accuracy, as model has more controllability towards global information into visual representations. 
However, when the gating dimension becomes larger, the performance saturates and deteriorates. We posit that the model might become more constrained to allow the encoding of global semantics, degenerating to the architecture with limited global semantics. Apparently, removing gating unit results in performance drops, because irrelevant global information for the question could flow into visual hidden states without model controllability.

\noindent\textbf{Effect of C$^3$ objective.} We compare our $C^{3}$ alignment objective with alternative approaches, \textit{i.e.} optimal transport (OT) \citep{pramanick2022multimodal} and its partial variant (POT) \citep{chapel2020partial}. As shown in Table \ref{tab:ablation_cross_modal_alignment}, both OT and POT can polish videoQA performance, while C$^3$ yields the highest performance. This shows that compositional consistency is important for long-form videoQA, since the model needs to grasp the relations among entities, specifically those specified by the question.

\noindent\textbf{Position of State Space Layer.} We replace the penultimate multi-modal attention with SSL. As shown in Table \ref{tab:ablation_study_position_ssl}, such design choice deteriorates the videoQA performance. The reason might be that since the frames and regions have already been selected, limited global information can be extracted from the video. Moreover, SSL does not explicitly calculate dependency between tokens, thus producing little refined representations to compute the final answer, which has been observed by previous work \citep{zuo2022efficient}. This substanstiates our decision to adopt SSL to integrate global semantics of video in an earlier stage. 

\subsection{Qualitative Results}
\noindent We visualize videoQA cases in Figure \ref{fig:longform_videoqa_example}. As can be observed, our model can choose the correct answer for questions that require information over the video with a limited number of video segments. We posit that due to our integrated global semantics, visual representations not only encode information of the selected segments but also the video context, thus furnish the model with sufficient cues to ascertain the correct answer. In contrast, constrained to the selected segments, previous state-of-the-art, \textit{i.e.} MIST-CLIP \citep{gao2023mist}, struggles in such questions and produces the incorrect output.
\section{Summary}
\noindent We introduce a Gated State space Multi-modal Transformer (GSMT) with a state space layer (SSL) to integrate global semantics of video into visual representations to tackle long-form videoQA. We further incorporate a gating unit to provide more controllability over the integrated global semantics and a cross-modal compositional congruence (C$^3$) objective to encourage the semantics aligned with the question. To comprehensively evaluate long-form videoQA, we curate two long-form videoQA datasets with excessively long video lengths and long-natured questions. Extensive experiments on these and standard datasets validate the superiority of our framework.
\chapter{Motion-aware Contrastive Learning for Temporal Panoptic Scene Graph Generation}
\label{ch4:motionpvsg}

\section{Introduction}
\noindent The advent of autonomous agents, intelligent systems, and robots warrants a comprehensive understanding of real-world environments \citep{ma2022sqa3d, driess2023palm, raychaudhuri2023reduce, cheng2022masked, li2023tube, li2023transformer}. Such understanding encompasses beyond merely recognizing individual entities, but also a sophisticated understanding of their relationships. To construct a detailed understanding, scene graph generation (SGG) research \citep{li2022dynamic,bin2019mr, sudhakaran2023vision, nag2023unbiased, wang2024oed} has sought to provide relational perspective on scene understanding. In SGG frameworks, scene graphs utilize nodes to represent entities and edges to represent relationships, constructing a comprehensive and structured understanding of visual scenes. 

However, due to being primarily based on bounding boxes to denote entities, scene graphs fall short of replicating human visual perception with a lack of granularity \citep{yang2023panoptic}. To overcome this limitation, panoptic scene graph generation \citep{yang2022panoptic, zhao2023textpsg} has been presented to expand the scope of SGG to incorporate pixel-level precise entity localization and thorough scene understanding including background components. Subsequently, because the temporal dimension undoubtedly provides richer information than the static spatial dimension, recent works \citep{yang2023panoptic, yang20244d} have shifted attention to the domain of videos and 4D scenes, resulting in the tasks of panoptic video and 4D scene graph generation.

\begin{figure}[t]
\centering
\includegraphics[width=0.7\linewidth]{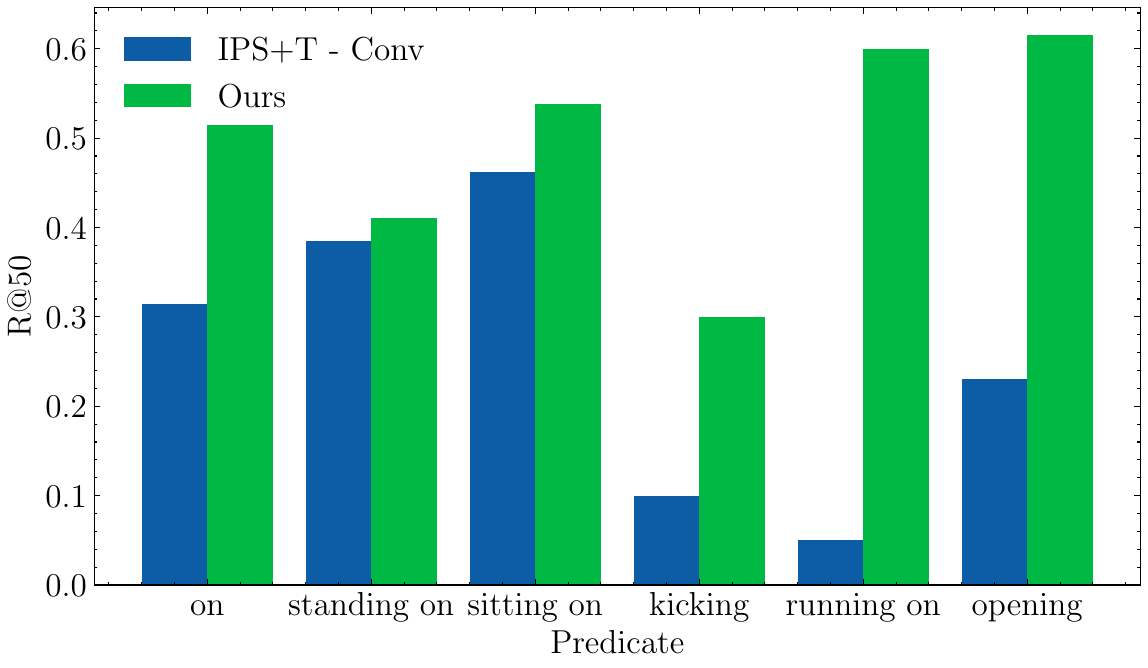} 
\caption{State-of-the-art IPS+T - Convolution \citep{yang2023panoptic} exhibits high R@50 scores for static relations, \textit{e.g.} \textit{on}, \textit{sitting on}, and \textit{standing on}, than dynamic relations, \textit{e.g.} \textit{kicking}, \textit{running on}, and \textit{opening}. In contrast, our method can perform effectively on both static and dynamic relations.}
\label{fig:r@50}
\end{figure}

\begin{figure*}[t]
\centering
\includegraphics[width=\linewidth]{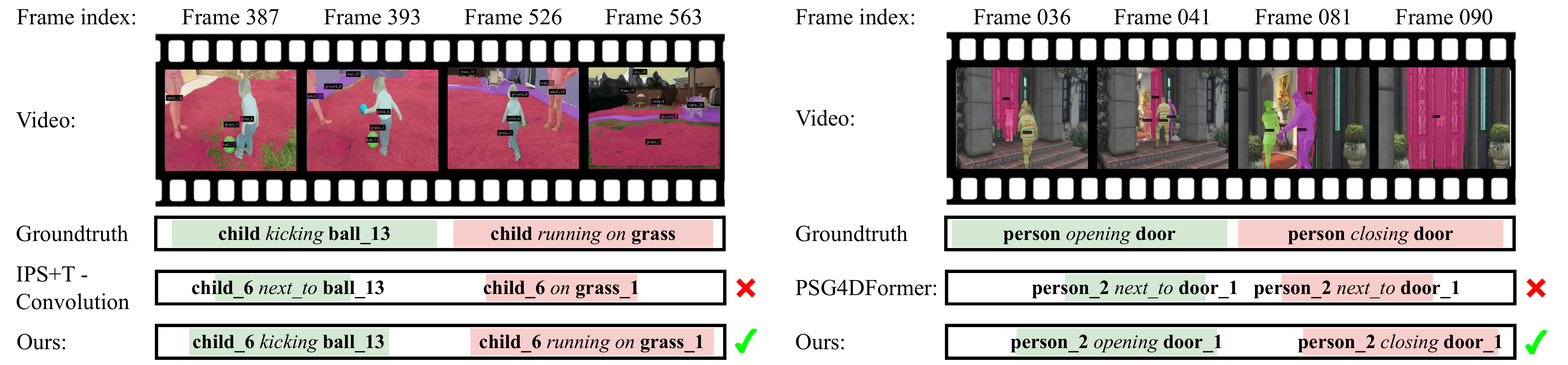} 
\caption{Examples of temporal panoptic scene graph generation of state-of-the-art  \citep{yang2023panoptic, yang20244d} and our method.}
\label{fig:qualitative_example}
\end{figure*}

Popular methods \citep{yang2023panoptic, yang20244d, yang2022panoptic} for temporal panoptic scene graph generation produce entity masks tracked across the temporal dimension, \textit{i.e.} mask tubes, then predict temporal relations among them. To conduct relation prediction, these methods encode the segmentation mask tubes, apply global pooling, then forward to a multi-layer perceptron for classifying their relations. However, such global pooling operation is well-known to be ineffective in representing temporal or motion patterns, which are useful for determining the relation among the entities. Consequently, this would result in higher misclassification rates of more dynamic relations \citep{wang2023taking, nag2023unbiased, zhou2022context}, as illustrated in Figure \ref{fig:r@50}. 

To encourage temporal representation learning, current research \citep{nguyen2023demaformer, liu2022spatial, zhou2023learning} uses contrastive learning for videos. However, they mainly seek to force two clips from the same video to be close together. As such, they mostly capture the semantics of visual scenes and disregard motions \citep{chen2020graph}. Moreover, instead of paying attention to precise entity localization, they work on frame-level representations. This would inadvertently inject motions from non-target entities into visual representations, which might not benefit relation classification in panoptic scene graph generation.

In this paper, to encourage representation learning to capture motion patterns for temporal panoptic scene graph generation, we propose a novel contrastive learning framework that focuses on mask tubes of the segmented entities. First, we force a mask tube and the one of similar subject-relation-object but of a different video to obtain close representations. Since positive mask tubes originate from distinct video clips, the model cannot rely upon visual semantics to optimize the contrastive objective, but instead depends on the motion trajectory evolution, which is our target component for representation learning. Second, we propel negative mask tubes generated by temporally shuffling the original tubes. Moreover, we also push apart representations of mask tubes from the same video but belonging to different triplets. Because mask tubes of different triplets from a common video with close visual features are separated from each other, we once again motivate the model to generate representations that are less reliant upon visual semantics but motion-sensitive features. In addition, the visually similar negative mask tubes can play a role as hard negative samples, thus accelerating the contrastive learning process \citep{chen2024curriculum}.

Moreover, in order to implement our motion-aware contrastive learning framework, there is a need to quantify the relationship between mask tubes. This quantification marks a challenging problem as mask tubes are a sequence of segmentation masks that span over the sequence of video frames. Furthermore, mask tubes of two triplets might exhibit different lengths since two events often occur at different speed. Unfortunately, the popular pipeline of temporal pooling and then similarity estimation flattens the temporal dimension of the mask tubes and neglects their motion features. To resolve this problem, we consider mask tubes of two triplets as two distributions and seek the optimal transportation map between them, then utilize the transport distance as the distance between two triplets' tubes. Such scheme of transporting can play a role of synchronizing the motion states of two triplets and takes advantage of the mask tubes' evolutionary trajectory. 

To sum up, our contributions are as follows:
\begin{itemize}
    \item We propose a novel contrastive learning framework for temporal panoptic scene graph generation which pulls together entity mask tubes with similar motion patterns and pushes away those of distinct motion patterns.
    \item We utilize optimal transport distance to estimate the relationship between two events' mask tubes for the proposed contrastive framework.
    \item Comprehensive experiments demonstrate that our framework outperforms state-of-the-art methods on both natural and 4D video datasets, especially on recognizing dynamic subject-object relations.
\end{itemize}
\section{Problem Formulation}
\noindent Temporal panoptic scene graph generation (TPSGG) is a task to generate a dynamic scene graph given an input video. In the generated scene graph, each node corresponds to an entity and each edge corresponds to a spatial-temporal relation between two entities. Formally, the input of a TPSGG model is a video clip $V$, particularly $V \in \mathbb{R}^{T \times H \times W \times 3}$ for a natural video, $V \in \mathbb{R}^{T \times H \times W \times 4}$ for a 4D RGB-D video, and $V \in \mathbb{R}^{T \times M \times 6}$ for a 4D point cloud video, $T$ denotes the number of frames, $M$ the number of point clouds of interest, and the frame size $H \times W$ should remain consistent across the video. The output of the model is a dynamic scene graph $G$. The TPSGG task can be formulated as follows:
\begin{equation}
P(G|V) = P(M, O, R|V).
\end{equation}
In particular, G consists of binary mask tubes $M = \{\mathbf{m}_{1}, \mathbf{m}_{2}, ..., \mathbf{m}_{N}\}$ and entity labels $O = \{o_{1}, o_{2}, ..., o_{N}\}$ which are associated with $N$ entities in the video, and their relations are denoted as $R = \{r_{1}, r_{2}, ..., r_{L}\}$. With respect to entity $o_{i}$, the mask tube $\mathbf{m}_{i} \in \{0, 1\}^{T \times H \times W}$ composes all tracked masks in all video frames, and its category $o_{i} \in \mathbb{C}^{O}$. For all entities in frame $t$, their masks must not overlap, \textit{i.e.} $\sum\limits_{i=1}^{N} \mathbf{m}_{i}^{t} \leq \textbf{1}^{H \times W}$. The relation $r_{i} \in \mathbb{C}^{R}$ associates two entities, one of which is the subject and the other is an object, with a relation class and a time period. $\mathbb{C}^{O}$ and $\mathbb{C}^{R}$ denote the entity and relation set class, respectively.

\begin{figure*}[t]
\centering
\includegraphics[width=\linewidth]{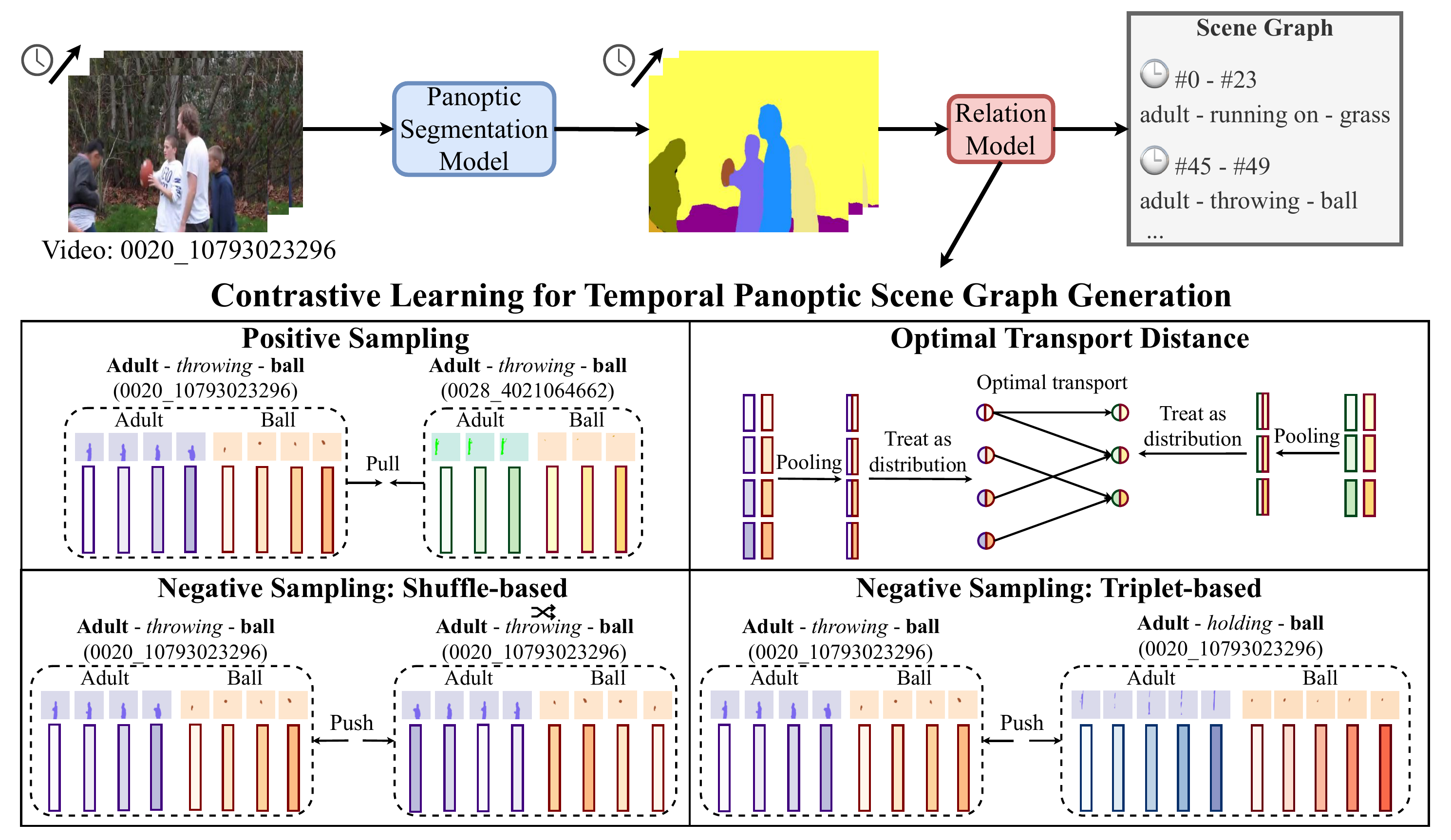} 
\caption{Framework overview of contrastive learning for temporal scene graph generation.}
\label{fig:framework}
\end{figure*}
\section{Methodology}
\begin{figure}[t]
\centering
\includegraphics[width=0.7\linewidth]{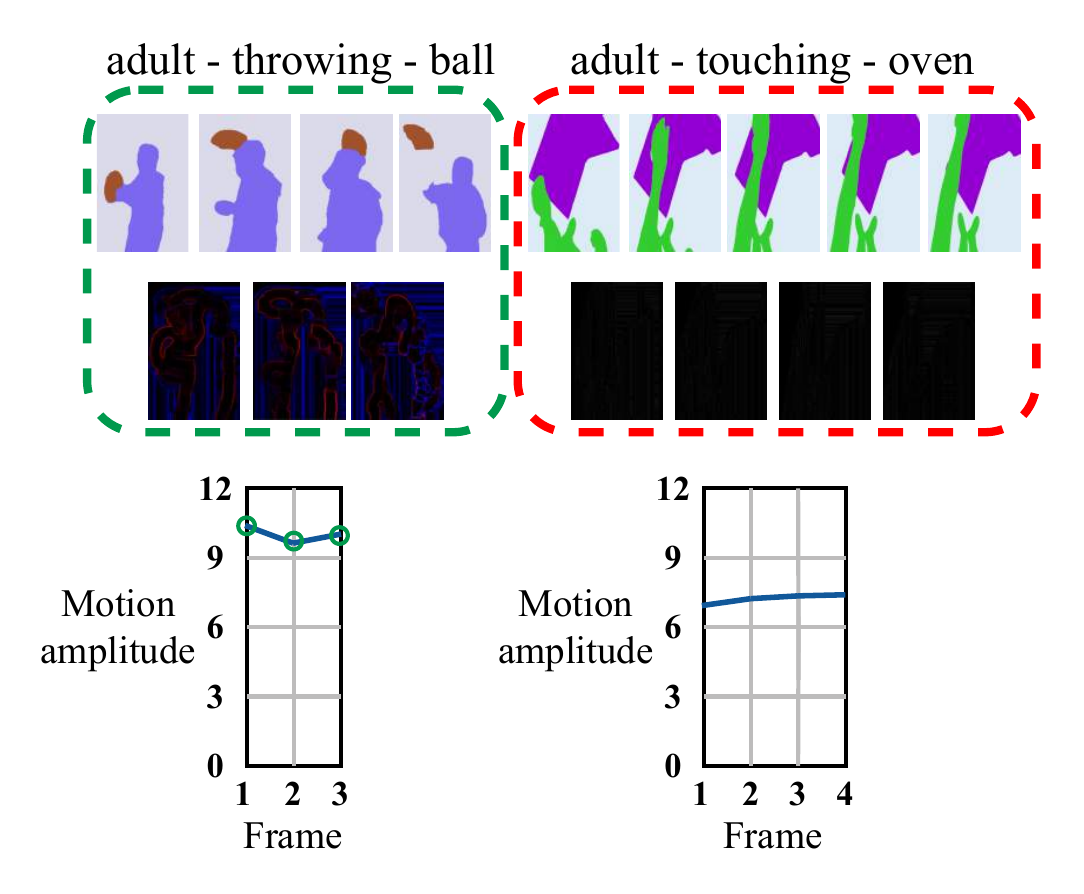} 
\caption{Proposed strategy to select strong-motion tubes.}
\label{fig:selecting_strong_motion}
\end{figure}

\noindent We firstly present the backbone pipeline to conduct temporal panoptic scene graph generation. Then, we explain our proposed contrastive learning framework to facilitate motion-aware mask tube representation learning. We also present the detail of our optimal transport approach to estimate the relation between two mask tubes for the contrastive objective. Our overall framework is illustrated in Figure \ref{fig:framework}.

\subsection{Temporal Panoptic Segmentation}
\noindent Given a video clip $V \in \mathbb{R}^{T \times H \times W \times 3}$, $V \in \mathbb{R}^{T \times H \times W \times 4}$, or $V \in \mathbb{R}^{T \times M \times 6}$, the initial step is to segment and track each pixel in a non-overlapping manner. Formally, the model produces a set of entity masks $\{y_{i}\}_{i=1}^{N} = \{(\mathbf{m}_{i}, p_{i}(c))\}_{i=1}^{N}$, where $\mathbf{m}_{i} \in \{0,1\}^{T \times H \times W}$ denotes the tracked video mask, \textit{i.e.} the mask tube, and $p_{i}(c)$ denotes the probability of assigning class $c$ to the tube $\mathbf{m}_{i}$. $N$ denotes the number of entities, which consist of both foreground (thing) and background (stuff) classes.

\noindent\textbf{Segmentation module.} Inspired by \citep{yang2023panoptic, yang20244d}, we adopt the Transformer-based encoder-decoder segmentation model. There are two types of segmentation procedure: 1) image panoptic segmentation combined with a tracker (IPS+T) and 2) video panoptic segmentation (VPS). IPS+T procedure will process each video frame separately and uses the tracker to connect the mask tubes across the video frames, while VPS processes each video frame with its reference frame from a nearby timestamp. 

Both procedures are initiated by producing a set of object queries which interacts with encoded visual patches via masked cross-attention. Receiving a video $V$, the model produces a set of queries $\{\mathbf{q}_{i}\}_{i=1}^{N}$, where each query $\mathbf{q}_{i}$ corresponds to one entity. Subsequently, every query is forwarded to two multi-layer perceptrons (MLPs) to project the queries into mask classification and mask regression outputs. 

\noindent\textbf{Training and inference.} During training, each query is matched to a groundtruth mask through mask-based bipartite matching to calculate the segmentation loss. During inference, IPS+T generates panoptic segmentation masks for each frame, and uses the tracker to achieve $N$ tracked mask tubes. In contrast, VPS employs two query embeddings of the target and reference frame, and performs query-wised similarity tracking to obtain $N$ tracked mask tubes. 

\subsection{Relation Classification} 
\noindent After the segmentation step, if the relation module is to be trained, we match query tubes with the annotated groundtruth masks based on the tube IoU values with the groundtruth. Otherwise, we directly forward mask tubes to self-attention or convolutional layers for encoding them into hidden representations $\{H_{i}\}_{i=1}^{N}$, $H_{i} \in \mathbb{R}^{T \times D}$, where $D$ denotes the hidden dimension. Then, we construct query pairs from every two query tubes' representations $H_{i}$ and $H_{j}$, $i, j \in \{1, 2, ..., N\}, \; i \neq j$. Inspired by \citep{yang2023panoptic, yang20244d}, in every pair, we perform global pooling over the temporal dimension for each mask tube representation:
\begin{equation}
\mathbf{h}_{i} = \text{Pooling}\left(H_{i}\right),
\end{equation}
where $\mathbf{h}_{i} \in \mathbb{R}^{D}$. 	Afterwards, we concatenate $\mathbf{h}_{i}$ and $\mathbf{h}_{j}$, and forward to a MLP to generate the relation category:
\begin{equation}
\log p(r_{i,j}) = \text{MLP}\left(\left[\mathbf{h}_{i}, \mathbf{h}_{j}\right]\right).
\end{equation}
To train the relation classification module, we use the cross-entropy loss calculated based on the predicted relation log-likelihood and the groundtruth. For inference, we extract the relation of the highest log-likelihood.

\subsection{Contrastive Learning for Temporal Panoptic Scene Graph Generation}

\noindent Our goal is to encourage mask tube representations $\{H_{i}\}_{i=1}^{N}$ to become motion-aware. In the beginning, we concatenate the representations of two mask tubes $H_{i}^{\text{sub}}$ and $H_{j}^{\text{obj}}$, which have been matched to a groundtruth subject-relation-object triplet, to form an anchor representation $H_{i,j}$ (anchor):
\begin{equation}
H^{a}_{i,j} = [H_{i}^{\text{sub}}, H_{j}^{\text{obj}}],
\end{equation}
where $H^{a}_{i,j} \in \mathbb{R}^{T \times 2D}$. Then, we propose a contrastive learning framework in which we motivate the model to associate mask tubes based upon the motion information. The objective of contrastive learning is to produce a representation space through attracting positive pairs, \textit{i.e.} $H^{a}$ and $H^{p}$ (positive), while pushing apart negative pairs, \textit{i.e.} $H^{a}$ and $H^{n}$ (negative). We accomplish this by optimizing the contrastive objective, which is formulated as follows:
\begin{equation}
\mathcal{L}_{\text{cont}} = -\text{log}\frac{e^{\text{sim}\left(H^{a}, H^{p}\right)}}{e^{\text{sim}\left(H^{a}, H^{p}\right)} + \sum\limits_{z=1}^{N_{n}} e^{\text{sim}\left(H^{a}, H^{n}_{z}\right)}},
\end{equation}
where sim denotes the similarity function defined upon a pair of mask tube representations. The formulation shows that what the model learn is largely dependent upon how positive and negative samples are generated. 

\noindent\textbf{Positive sampling.} To satisfy our motion-aware requirement for contrastive learning, we extract mask tube representations from the entities of the same subject and object category that exhibit a similar groundtruth relation from another video. Since two videos possess distinct visual features, the model must rely on the shared motion pattern of similar subject-relation-object triplets to associate the anchor and the positive sample. 

\noindent\textbf{Negative sampling.} For negative sampling, we design two strategies, which result in two contrastive approaches, \textit{i.e.} shuffle-based and triplet-based contrastive learning.

\subsection{Shuffle-based contrastive learning} 
\noindent In our shuffle-based approach, we create negative samples by utilizing a series of temporal permutations $\pi$ to the anchor tube, \textit{i.e.} shuffling:
\begin{equation}
H^{n} = \pi\left(H^{a}\right).
\end{equation}
As such, the contrastive objective will force the model to propel representations of the anchor tube, which is in the normal order, from the shuffled tube, which exhibits a distorted motion due to the shuffled order. This would make the learned representation sensitive to frame ordering, \textit{i.e.} motion-aware, as the anchor $H^{a}$ and the negative tube $H^{n}$ share visual semantics and can only be distinguished using motion information. 

\noindent\textbf{Selecting strong-motion mask tubes.} However, there exists a potential risk: for static relations such as \textit{on}, \textit{next to}, and \textit{in}, mask tubes might involve almost no motion. As a result, the shuffled tube would become identical to the anchor one and the model would not be able to differentiate them and learn reasonably. To address this problem, we propose a strategy to select strong-motion tubes for shuffling, which we illustrated in Figure \ref{fig:selecting_strong_motion}.

Given a video, our aim is to select mask tubes that carry strong motion for shuffling. To measure the motion of the mask tube, we utilize optical flow edges \citep{xiao2021modist}. We estimate flow edges via employing a Sobel filter \citep{sobel2022sobel} onto the flow magnitude map and take the median over the flow edge pixels of the entity masks. Then, we select mask tubes whose the maximum value across the optical flow surpasses a threshold $\gamma$.

\subsection{Triplet-based contrastive learning}
\noindent To take advantage of motion-aware signals from triplets of similar subject-relation-object category, we design a triplet-based approach to create negative samples. A naive approach would be to sample mask tubes of any distinct subject-relation-object triplet from the anchor sample. However, if we run into triplets with all distincgt subject, relation, and object categories, the negative pair would be trivial for the model to distinguish, resulting in less effective learning.

In order to create harder negative samples, we choose negative mask tubes from the same video with the anchor. We create a multi-nomial distribution, where triplets that share more subject, relation, or object categories with the anchor will be more likely to be drawn. Hence, our negative samples can hold close visual semantics with the anchor sample, and increase the likelihood that the model depends on motion semantics to push them apart. From contrastive learning perspective, these samples form hard negative samples to accelerate the learning process \citep{chen2024curriculum}.

\subsection{Optimal Transport for Mask Tube Relation Quantification}
\noindent There is one remaining problem, \textit{i.e.} how to define the similarity function $\text{sim}$ for two mask tubes' representations $H_{i}$ and $H_{j}$. In this work, we consider two mask tubes as two discrete distributions $\boldsymbol{\mu}$ and $\boldsymbol{\nu}$, whose $H_{i}$ and $H_{j}$ are their supports, respectively. Formally, $\boldsymbol{\mu} = \sum\limits_{k=1}^{T_{i}} \mathbf{a}_{k} \delta_{\mathbf{h}_{i,k}}$ and $\boldsymbol{\nu} = \sum\limits_{l=1}^{T_{j}} \mathbf{b}_{l} \delta_{\mathbf{h}_{j,l}}$, where $\delta_{\mathbf{h}_{i,k}}$ and $\delta_{\mathbf{h}_{j,l}}$ denote the Dirac functions centered upon $\mathbf{h}_{i,k}$ and $\mathbf{h}_{j,l}$, respectively. The weights of the supports are $\mathbf{a} = \frac{\mathbf{1}_{T_{i}}}{T_{i}}$ and $\mathbf{b} = \frac{1_{T_{j}}}{T_{j}}$.

After defining the distribution scheme, we propose the tube alignment optimization problem, which is to find the transport plan that achieves the minimum distance between $\boldsymbol{\mu}$ and $\boldsymbol{\nu}$ as follows:
\begin{gather}
d_{\text{OT}} = \mathcal{D}_{\text{OT}}(\boldsymbol{\mu}, \boldsymbol{\nu}) = \min_{\mathbf{T} \in \Pi(\mathbf{a}, \mathbf{b})} \sum\limits_{k=1}^{T_{i}} \sum\limits_{l=1}^{T_{j}} \mathbf{T}_{i,j} \cdot c\left(\mathbf{h}_{i,k}, \mathbf{h}_{j,l}\right), \\
\begin{split}
\text{s.t} \quad \Pi(\mathbf{a}, \mathbf{b}) = \{\mathbf{T} \in \mathbb{R}^{T_i \times T_j}_{+} \mid \mathbf{T}\mathbf{1}_{T_i} \leq \mathbf{a},  \mathbf{T}^{\top}\mathbf{1}_{T_j} \leq \mathbf{b}, \\\mathbf{1}_{T_i}^{\top} \cdot \mathbf{T} \cdot \mathbf{1}_{T_j} = s, \quad 0 \leq s \leq \min(T_i, T_j)\},
\label{eq:pvla_formulation}
\end{split}
\end{gather}
where $c$ denotes a pre-defined distance between two vectors. We implement the cost distance $c\left(\mathbf{h}_{i,k}, \mathbf{h}_{j,l}\right) = 1 - \frac{\mathbf{h}_{i,k} \cdot \mathbf{h}_{j,l}}{||\mathbf{h}_{i,k}||_{2} ||\mathbf{h}_{i,k}||_{2}}$ as the cosine distance. As the exact optimization over the transport plan $\mathbf{T}$ is intractable, we adopt the Sinkhorn-based algorithm to estimate $\mathbf{T}$. We delineate the algorithm to calculate the distance in Algorithm \ref{alg:ot_distance}. To turn the distance into similarity value, we take its negative value and add to a pre-defined margin $\alpha$:
\begin{equation}
    \text{sim}\left(\mathbf{h}^{a}, \mathbf{h}^{p}\right) = \alpha - d_{\text{OT}}.
\vspace{-10pt}
\end{equation}

\setlength{\textfloatsep}{5pt}
\begin{algorithm}[t]
{\footnotesize \caption{Computing the optimal transport distance}
\label{alg:ot_distance}
}
{\footnotesize
\begin{algorithmic}
\Require{$\mathbf{C} = \{\mathbf{C}_{l,k} = c\left(\mathbf{h}_{i,l}, \mathbf{h}_{j,k}\right) \mid 1 \leq i \leq N_V, 1 \leq j \leq N_L\} \in \mathbb{R}^{T_i \times T_j}$,\;  $\mathbf{a} \in \mathbb{R}^{T_i},\; \mathbf{b} \in \mathbb{R}^{T_j},\; s,\; N_{\text{iter}}$} \\
$d_{\text{OT}} = \infty$ 
\For{$s=1$ to $\min(T_i, T_j)$}
    \State $\mathbf{T} = \text{exp}\left(-\frac{\mathbf{C}}{\tau}\right)$ 
    \State $\mathbf{T} = \frac{s}{\left(\mathbf{1}_{T_i}\right)^{\top} \cdot \mathbf{T} \cdot \mathbf{1}_{T_j}} \mathbf{T}$
    
    \For{$i=1$ to $N_{\text{iter}}$} 
        \State $\mathbf{p}_{a} = \min \left(\frac{\mathbf{a}}{\mathbf{T}\mathbf{1}_{T_j}}, \mathbf{1}_{T_i}\right)$, $\mathbf{T}_{a} = \text{diag}\left(\mathbf{p}_{a}\right) \cdot \mathbf{T}$
        \State $\mathbf{p}_{b} = \min \left(\frac{\mathbf{b}}{\mathbf{T}_{a}^{\top}\mathbf{1}_{T_i}}, \mathbf{1}_{T_j}\right)$, $\mathbf{T}_{b} = \text{diag}\left(\mathbf{p}_{b}\right) \cdot \mathbf{T}_{a }$
        \State $\mathbf{T} = \frac{s}{\left(\mathbf{1}_{T_i}\right)^{\top} \cdot \mathbf{T} \cdot \mathbf{1}_{N_L}} \mathbf{T}_{b}$
    \EndFor 
    \State $d_{\text{OT}} = \min\left(d_{\text{OT}}, \;\sum\limits_{k=1}^{T_i} \sum\limits_{l=1}^{T_j} \mathbf{T}_{k,l} \mathbf{C}_{k,l}\right)$ 
    \EndFor \\
\Return $d_{\text{OT}}$
\end{algorithmic}}
\end{algorithm}
\section{Experiments}
{\renewcommand{\arraystretch}{1.1}
\begin{table*}[t]
\centering
\resizebox{\linewidth}{!}{
\begin{tabular}{l|ccc|ccc}
\hline
\multicolumn{1}{c|}{\multirow{2}{*}{\textbf{Method}}} & \multicolumn{3}{c|}{\textbf{vIoU threshold = 0.5}}       & \multicolumn{3}{c}{\textbf{vIoU threshold = 0.1}}       \\
\multicolumn{1}{c|}{}                                 & \textbf{R/mR@20} & \textbf{R/mR@50} & \textbf{R/mR@100} & \textbf{R/mR@20} & \textbf{R/mR@50} & \textbf{R/mR@100} \\ \hline
IPS+T - Vanilla                                      & 3.04 / 1.35      & 4.61 / 2.94      & 5.56 / 3.33       & 8.28 / 5.68      & 14.47 / 9.92     & 18.24 / 11.84     \\
IPS+T - Handcrafted filter                           & 2.52 / 1.72      & 3.77 / 2.36      & 4.72 / 2.79       & 8.07 / 5.61      & 13.42 / 8.27     & 16.46 / 10.11     \\
IPS+T - Transformer                                  & 3.88 / 2.81      & 5.66 / 4.12      & 6.18 / 4.44       & 9.01 / 6.69      & 14.88 / 11.28    & 17.51 / 13.20     \\
IPS+T - Convolution                                  & 3.88 / 2.55      & 5.24 / 3.29      & 6.71 / 5.36       & 10.06 / 8.98     & 14.99 / 12.21    & 18.13 / 15.47     \\
Ours - Transformer                                               &    \underline{3.98} / \underline{2.98}					              &      \underline{5.97} / \underline{4.20}            &      \underline{7.44} / \underline{5.15}             &          \underline{10.59} / \underline{9.56}        &     \underline{16.98} / \underline{12.39}             &      \underline{22.33} / \underline{17.47}             \\ 
Ours - Convolution                                                &   \textbf{4.51} / \textbf{3.56}				              &             \textbf{6.08} / \textbf{4.38}     &      \textbf{7.76} / \textbf{5.86}	             &     \textbf{11.43} / \textbf{9.57}             &    \textbf{17.30} / \textbf{13.13}              &       \textbf{22.85} / \textbf{17.48}            \\ \hline
VPS - Vanilla                                        & 0.21 / 0.10      & 0.21 / 0.10      & 0.31 / 0.18       & 6.29 / 3.04      & 9.64 / 6.74      & 12.89 / 9.60      \\
VPS - Handcrafted filter                             & 0.42 / 0.13      & 0.52 / 0.50      & 0.94 / 0.92       & 5.24 / 2.84      & 7.65 / 7.14      & 9.64 / 8.22       \\
VPS - Transformer                                    & 0.42 / 0.61      & 0.73 / 0.76      & 1.05 / 0.92       & 6.50 / 5.75      & 9.64 / 8.25      & 12.26 / 9.51      \\
VPS - Convolution                                    & 0.42 / 0.25      & 0.63 / 0.67      & 0.63 / 0.67       & 8.07 / 7.84      & 11.01 / 9.78     & 12.89 / 10.77     \\
Ours - Transformer                                                &    \underline{0.63} / \underline{0.83}              &    \underline{1.05} / \underline{0.76}              &      \underline{1.05} / \underline{0.76}             &     \underline{6.71} / \underline{6.94}             &   \underline{10.27} / \underline{8.68}               &     \underline{13.42} / \underline{12.09}              \\ 
Ours - Convolution                                               &   \textbf{0.84} / \textbf{0.98}					               &         \textbf{1.26} / \textbf{1.22}         &        \textbf{1.26} / \textbf{1.22}           &       \textbf{8.18} / \textbf{8.00}           &     \textbf{12.90} / \textbf{11.47}             &      \textbf{14.22} / \textbf{13.59}             \\ \hline
\end{tabular}}
\caption{Experimental results on the OpenPVSG dataset.}
\label{tab:exp_openpvsg}
\end{table*}}

{\renewcommand{\arraystretch}{1.1}
\begin{table*}[t]
\centering
\resizebox{\linewidth}{!}{
\begin{tabular}{l|l|ccc|ccc}
\hline
\multirow{2}{*}{\textbf{Input type}} & \multicolumn{1}{c|}{\multirow{2}{*}{\textbf{Method}}} & \multicolumn{3}{c|}{\textbf{PSG4D-GTA}}                  & \multicolumn{3}{c}{\textbf{PSG4D-HOI}}                  \\ 
                                     &                                  & \textbf{R/mR@20} & \textbf{R/mR@50} & \textbf{R/mR@100} & \textbf{R/mR@20} & \textbf{R/mR@50} & \textbf{R/mR@100} \\ \hline
\multirow{3}{*}{Point cloud videos}  & 3DSGG                            & 1.48 / 0.73      & 2.16 / 0.79      & 2.92 / 0.85       & 3.46 / 2.19      & 3.15 / 2.47      & 4.96 / 2.84       \\
                                     & PSG4DFormer                      & 4.33 / 2.10      & 4.83 / 2.93      & 5.22 / 3.13       & 5.36 / 3.10      & 5.61 / 3.95      & 6.76 / 4.17       \\
                                     & Ours                             & \textbf{5.88} / \textbf{3.45}      & \textbf{6.31} / \textbf{3.70}      & \textbf{7.31} / \textbf{4.70}       & \textbf{7.28} / \textbf{5.09}      & \textbf{7.62} / \textbf{6.49}      & \textbf{9.18} / \textbf{6.85}       \\ \hline
\multirow{3}{*}{RGB-D videos}        & 3DSGG                            & 2.29 / 0.92      & 2.46 / 1.01      & 3.81 / 1.45       & 4.23 / 2.19      & 4.47 / 2.31      & 4.86 / 2.41       \\
                                     & PSG4DFormer                      & 6.68 / 3.31      & 7.17 / 3.85      & 7.22 / 4.02       & 5.62 / 3.65      & 6.16 / 4.16      & 6.28 / 4.97       \\
                                     & Ours                             & \textbf{9.07} / \textbf{5.52}      & \textbf{9.73} / \textbf{6.32}      & \textbf{9.73} / \textbf{6.32}       & \textbf{7.63} / \textbf{6.09}      & \textbf{8.36} / \textbf{6.94}      & \textbf{8.53} / \textbf{8.29} \\ \hline      
\end{tabular}}
\caption{Experimental results on both PSG4D-GTA and PSG4D-HOI groups of PSG4D dataset.}
\label{tab:exp_psg4d}
\end{table*}}

{\renewcommand{\arraystretch}{1.1}
\begin{table}[t]
\centering
\resizebox{0.7\linewidth}{!}{
\begin{tabular}{l|ccc}
\hline
\multicolumn{1}{c|}{\textbf{Method}} & \textbf{R/mR@20} & \textbf{R/mR@50} & \textbf{R/mR@100} \\ \hline
w/o shuffle-based                   & 4.41 / 3.43      & 5.90 / 4.24      & 7.30 / 5.79      \\
w/o triplet-based                   & 4.44 / 3.50      & 6.02 / 4.28      & 7.36 / 5.83       \\
Ours                                &   \textbf{4.51} / \textbf{3.56}               &   \textbf{6.08} / \textbf{4.38}               &    \textbf{7.44} / \textbf{5.86}               \\ \hline
\end{tabular}}
\caption{Ablation results for contrastive learning approaches on OpenPVSG dataset. We adopt the vIoU threshold of 0.5.}
\label{tab:exp_ablation_contrastive_openpvsg}
\end{table}}

{\renewcommand{\arraystretch}{1.1}
\begin{table}[t]
\centering
\resizebox{0.8\linewidth}{!}{
\begin{tabular}{l|ccc}
\hline
\multicolumn{1}{c|}{\textbf{Tube relation quantification}} & \textbf{R/mR@20} & \textbf{R/mR@50} & \textbf{R/mR@100} \\ \hline
Pooling - Cosine similarity                   & 4.44 / 3.40      & 6.04 / 4.36      & 7.36 / 5.84       \\
Pooling - L2                   & 4.36 / 3.37      & 3.77 / 5.95      & 7.29 / 5.80       \\
Optimal transport                                &    \textbf{4.51} / \textbf{3.56}              &    \textbf{6.08} / \textbf{4.38}              &  \textbf{7.44} / \textbf{5.86}                 \\ \hline
\end{tabular}}
\caption{Ablation results for mask tube relation quantification method between mask tubes on OpenPVSG dataset.}
\label{tab:exp_ablation_ot_distance_openpvsg}
\end{table}}

{\renewcommand{\arraystretch}{1.1}
\begin{table*}[t]
\centering
\resizebox{\linewidth}{!}{
\begin{tabular}{c|l|ccc|ccc}
\hline
\multirow{2}{*}{\textbf{Input type}} & \multicolumn{1}{c|}{\multirow{2}{*}{\textbf{Method}}} & \multicolumn{3}{c|}{\textbf{PSG4D-GTA}}                  & \multicolumn{3}{c}{\textbf{PSG4D-HOI}}                  \\
                                     &                                 & \textbf{R/mR@20} & \textbf{R/mR@50} & \textbf{R/mR@100} & \textbf{R/mR@20} & \textbf{R/mR@50} & \textbf{R/mR@100} \\ \hline
\multirow{3}{*}{Point cloud videos}  & w/o shuffle-based                                    & 5.56 / 2.92      & 5.57 / 2.98      & 6.51 / 4.36      & 6.56 / 4.29      & 6.98 / 6.25      & 8.76 / 6.43       \\
& w/o triplet-based                                    & 5.77 / 2.93      & 5.59 / 3.26      & 6.53 / 4.39       & 6.67 / 4.85     & 7.52 / 6.31      & 8.84 / 6.43      \\
                                     & Ours                                                 & \textbf{5.88} / \textbf{3.45}      & \textbf{6.31} / \textbf{3.70}      & \textbf{7.31} / \textbf{4.70}       & \textbf{7.28} / \textbf{5.09}      & \textbf{7.62} / \textbf{6.49}      & \textbf{9.18} / \textbf{6.85}        \\ \hline
\multirow{3.2}{*}{RGB-D videos}     & w/o shuffle-based                                    & 8.35 / 5.34      & 8.76 / 5.68      & 8.88 / 5.53       & 7.00 / 5.53      & 7.51 / 6.02      & 7.56 / 7.42       \\
                              & w/o triplet-based                                    & 9.00 / 5.46      & 9.71 / 5.95      & 9.63 / 5.82       & 7.12 / 6.03      & 8.31 / 6.51      & 8.24 / 7.95       \\
                                     & Ours                                                 & \textbf{9.07} / \textbf{5.52}      & \textbf{9.73} / \textbf{6.32}      & \textbf{9.73} / \textbf{6.32}       & \textbf{7.63} / \textbf{6.09}      & \textbf{8.36} / \textbf{6.94}      & \textbf{8.53} / \textbf{8.29}      \\ \hline
\end{tabular}}
\caption{Ablation results for contrastive learning approaches on PSG4D dataset.}
\label{tab:exp_ablation_contrastive_psg4d}
\end{table*}}

{\renewcommand{\arraystretch}{1.1}
\begin{table*}[t]
\centering
\resizebox{\linewidth}{!}{
\begin{tabular}{c|l|ccc|ccc}
\hline
\multirow{2}{*}{\textbf{Input type}} & \multicolumn{1}{c|}{\multirow{2}{*}{\textbf{Tube relation quantification}}} & \multicolumn{3}{c|}{\textbf{PSG4D-GTA}}                  & \multicolumn{3}{c}{\textbf{PSG4D-HOI}}                  \\
                                     &                                 & \textbf{R/mR@20} & \textbf{R/mR@50} & \textbf{R/mR@100} & \textbf{R/mR@20} & \textbf{R/mR@50} & \textbf{R/mR@100} \\ \hline
\multirow{3}{*}{Point cloud videos}  & Pooling - Cosine similarity                                   & 5.76 / 2.87     & 6.02 / 3.62      & 6.84 / 4.11       & 7.24 / 4.45      & 7.44 / 6.27      & 8.20 / 6.64       \\
                                     & Pooling - L2                                  & 5.46 / 2.78      & 5.38 / 3.39      & 6.51 / 3.86       & 6.72 / 4.11      & 6.74 / 6.05      & 7.96 / 6.11       \\
                                     & Optimal transport                                                 & \textbf{5.88} / \textbf{3.45}      & \textbf{6.31} / \textbf{3.70}      & \textbf{7.31} / \textbf{4.70}       & \textbf{7.28} / \textbf{5.09}      & \textbf{7.62} / \textbf{6.49}      & \textbf{9.18} / \textbf{6.85}       \\ \hline
\multirow{3.2}{*}{RGB-D videos}        & Pooling - Cosine similarity                                    & 9.03 / 5.37     & 9.47 / 5.86      & 9.70 / 6.02       & 7.36 / 5.43      & 7.93 / 6.70     & 8.06 / 7.42      \\
                                     & Pooling - L2                                  & 8.89 / 4.70      & 8.90 / 5.41      & 9.08 / 5.78       & 6.65 / 5.26      & 7.74 / 6.29      & 7.95 / 7.39       \\
                                     & Optimal transport                                                 & \textbf{9.07} / \textbf{5.52}      & \textbf{9.73} / \textbf{6.32}      & \textbf{9.73} / \textbf{6.32}       & \textbf{7.63} / \textbf{6.09}      & \textbf{8.36} / \textbf{6.94}      & \textbf{8.53} / \textbf{8.29}     \\ \hline
\end{tabular}}
\caption{Ablation results for mask tube relation quantification method between mask tubes on PSG4D dataset.}
\label{tab:exp_ablation_ot_distance_psg4d}
\end{table*}}

We conduct comprehensive experiments to evaluate the effectiveness of our motion-aware contrastive framework. We first describe the experiment settings, covering the evaluation datasets, evaluation metrics, baseline methods, and implementation details. Next, we present quantitative results of our method, then provide ablation study and careful analysis to explore properties of our motion-aware contrastive framework. Eventually, we conduct qualitative analysis to concretely examine its behavior.

\subsection{Experiment Settings}
\noindent\textbf{Datasets.} We assess the effectiveness of our method on natural and 4D video inputs. The corresponding dataset to each input type is as follows:
\begin{itemize}
    \item \textbf{Open-domain Panoptic video scene graph generation (OpenPVSG)} \citep{yang2023panoptic}: OpenPVSG consists of scene graphs and associated segmentation masks with respect to subject and object nodes in the scene graph. The dataset comprises 400 videos, including 289 third-person videos from ViDOR \citep{shang2019annotating}, 111 egocentric videos from Epic-Kitchens \citep{damen2022epic} and Ego4D \citep{grauman2022ego4d}. 

    \item \textbf{Panoptic scene graph generation for 4D (PSG4D)} \citep{yang20244d}: The PSG4D dataset is divided into two groups, \textit{i.e.} PSG4D-GTA and PSG4D-HOI. PSG4D-GTA comprises 67 third-view videos with an average length of 84 seconds, 35 object categories, and 43 relationship categories. On the contrary, PSG4D-HOI contains 2,973 videos from an egocentric perspective, whose average duration is 20 seconds. The PSG4D-HOI's videos are mostly related to indoor scenes, covering 46 object categories and 15 relationship categories. 
\end{itemize}

\noindent\textbf{Evaluation metrics.} We use the recall at $K$ (R@$K$) and mean recall at $K$ (mR@$K$) metrics, which are standard metrics used in scene graph generation tasks. Both R@$K$ and mR@$K$ consider the top-$K$ triplets predicted by the panoptic scene graph generation model. A successful recall of a predicted triplet must satisfy the following criteria: 1) correct category labels for the subject, object, and predicate; 2) a volume Intersection over Union (vIoU) greater than or equal to 0.5 between the predicted mask tubes and the groundtruth tubes. For extensive comparison, we also report results with the vIoU threshold of 0.1. 

\noindent\textbf{Baseline methods.} We compare our method with a comprehensive list of baseline approaches for temporal panoptic scene graph generation: (i) \textbf{IPS+T - Vanilla} \citep{yang2023panoptic} uses image panoptic segmentation (IPS) model with a tracker for segmentation, and fully-connected layers to separately encode temporal states of entity mask tubes; (ii) \textbf{IPS+T - Handcrafted filter} \citep{yang2023panoptic} uses image panoptic segmentation (IPS) model with a tracker for segmentation, and a manually-designed kernel to encode entity mask tubes; (iii) \textbf{IPS+T - Convolution} \citep{yang2023panoptic} uses image panoptic segmentation (IPS) model with a tracker for segmentation, and learnable convolutional layers to encode entity mask tubes; (iv) \textbf{IPS+T - Transformer} \citep{yang2023panoptic} uses image panoptic segmentation model (IPS) with a tracker for segmentation, and Transformer-based encoder with self-attention layers to encode entity mask tubes; (v) \textbf{VPS - Vanilla} \citep{yang2023panoptic} is similar to IPS+T - Vanilla, but uses video panoptic segmentation (VPS) model for panoptic segmentation; (vi) \textbf{VPS - Handcrafted filter} \citep{yang2023panoptic} is similar to IPS+T - Handcrafter filter, but uses video panoptic segmentation (VPS) model for  segmentation; (vii) \textbf{VPS - Convolution} \citep{yang2023panoptic} is similar to IPS+T - Convolution, but uses video panoptic segmentation (VPS) model for segmentation; (viii) \textbf{VPS - Transformer} \citep{yang2023panoptic} is similar to IPS+T - Transformer, but uses video panoptic segmentation (VPS) model for segmentation; (ix) \textbf{3D-SGG} \citep{wald2020learning} is based on PointNet \citep{qi2017pointnet} and graph convolutional network \citep{kipf2016semi} but neglects the depth dimension and generates panoptic scene graphs for 4D video inputs; (x) \textbf{PSG4DFormer} \citep{yang20244d} is a specialized model for 4D inputs, using Mask2Former \citep{cheng2022masked} for segmentation and a spatial-temporal Transformer to encode object mask tubes for relation classification.

\noindent\textbf{Implementation details.} For fair comparison, we experiment our contrastive framework with both IPS+T and VPS as segmentation module for panoptic video scene graph generation. In the former case, we leverage the UniTrack tracker \citep{wang2021different} combined with Mask2Former model \citep{cheng2022masked}, which is initialized from the best-performing COCO-pretrained weights and fine-tuned for 8 epochs using AdamW optimizer with a batch size of 32, learning rate of 0.0001, weight decay of 0.05, and gradient clipping with a max L2 norm of 0.01. In the latter case, we utilize Video K-Net \citep{li2022video}, also initialized from COCO-pretrained weights and fine-tuned with the same strategy as IPS+T. In the relation classification step, we conduct fine-tuning with a batch size of 32, employing the Adam optimizer with a learning rate of 0.001. For 4D panoptic scene graph generation, we adopt the PSG4DFormer baseline. To work with RGB-D and point cloud videos, we use an ImageNet pretrained on ResNet-101 \citep{russakovsky2015imagenet} and the DKNet \citep{wu20223d} as the visual encoder, respectively. We fine-tune the segmentation module for RGB-D and point cloud videos for 12 and 200 epochs, respectively. We use additional 100 epochs to train the relation classification module. Based on validation, we adopt a threshold $\gamma = 9.0$ and a margin $\alpha = 10.0$. We set the maximum number of iterations $N_{\text{iter}}$ to 1,000. 

\begin{figure}[t]
\centering
\includegraphics[width=0.7\linewidth]{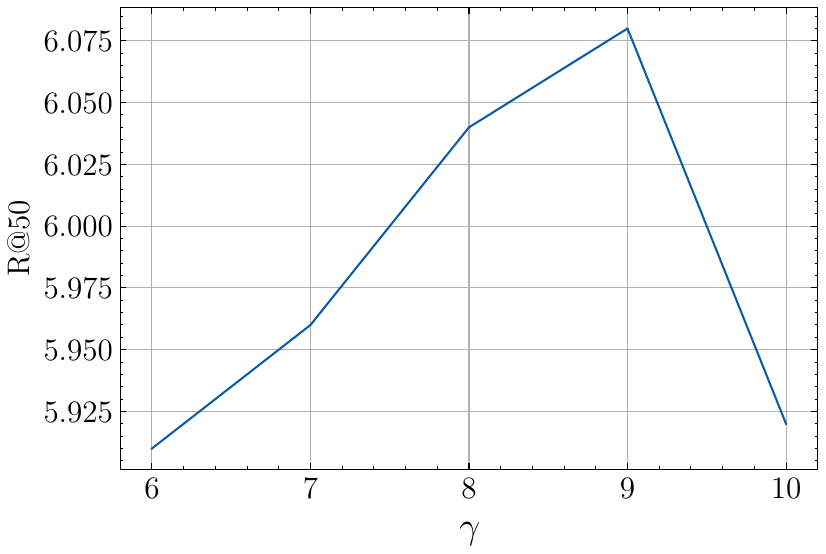}
\caption{Ablation results on threshold $\gamma$.}
\label{fig:ablation_gamma}
\end{figure}

\subsection{Main Results}
\noindent\textbf{Results on OpenPVSG.} As shown in Table \ref{tab:exp_openpvsg}, we substantially outperform both IPS+T - Convolution and IPS+T - Transformer when we use IPS+T for segmentation. In particular, using a higher vIoU threshold to filter out inaccurate segmentation, we surpass IPS+T - Transformer by 1.3/0.7 points of R/mR@100, while surpassing IPS+T - Convolution by 0.8/1.1 points of R/mR@50. In addition, for a less strict vIoU threshold, we outperform IPS+T - Transformer by 1.6/2.9 points of R/mR@20, and IPS+T - Convolution by 2.3/0.9 points of R/mR@50. These results demonstrate that our method makes a propitious contribution to temporal panoptic scene graph generation, not only to popular but also to unpopular relation classes. 

\noindent\textbf{Results on PSG4D.} Table \ref{tab:exp_psg4d} shows that our method also achieves significantly higher performance than the PSG4DFormer model. Specifically, when working with point cloud videos, on PSG4D-GTA, we outperform the baseline method by 1.6/1.4 points. Analogously, on PSG4D-HOI, we outperform PSG4DFormer by 2.0/2.5 points of R/mR@50. These results indicate that our framework bears a valuable impact to both egocentric and third-view videos. We hypothesize that both video types consist of dynamic actions among objects whose mask tube representations should be polished. In addition, when working with RGB-D videos, on PSG4D-GTA, we enhance the baseline method by 2.4/2.2 points of R/mR@20,. Furthermore, on PSG4D-HOI, our motion-aware contrastive learning also considerably refines PSG4DFormer by 2.0/2.4 points of R/mR@20. 
Such results have verified the generalizability of our motion-aware contrastive framework over natural, point cloud, and RGB-D videos.

\subsection{Ablation Study}
\noindent\textbf{Effect of the contrastive components.} We evaluate our framework without the assistance of either the shuffle-based or the triplet-based contrastive objective. As shown in Table \ref{tab:exp_ablation_contrastive_openpvsg} and \ref{tab:exp_ablation_contrastive_psg4d}, the performance degrades when we both remove shuffle-based and triplet-based contrastive approaches. In addition, triplet-based contrastive learning plays a more fundamental role than the shuffle-based one. We hypothesize that shuffle-based contrastive learning is better at focusing on motion semantics than triplet-based one.

\noindent\textbf{Effect of selecting strong-motion tubes.} We evaluate the impact of our strategy to filter out weak motion tubes. In Figure \ref{fig:ablation_gamma}, we observe a performance boost when we increase the threshold to select mask tubes with strong motion. However, further elevating the threshold results in performance degradation, since there are more mask tubes eliminated, thus limiting the effect of our motion-aware contrastive framework. 

\noindent\textbf{Effect of optimal transport distance.} In this ablation, we compare various strategies to calculate the similarity between two mask tubes. Results in Table \ref{tab:exp_ablation_ot_distance_openpvsg} and \ref{tab:exp_ablation_ot_distance_psg4d} show that the proposed optimal transport achieves much higher performance for both natural and 4D video inputs. We conjecture that other method such as pooling then cosine similarity or L2 neglects the temporal or flattens the motion nature of the entity mask tubes, thus reducing the effectiveness.

\subsection{Qualitative Analysis}
\noindent We visualize examples processed by the state-of-the-art models and ours in Figure \ref{fig:qualitative_example}. As can be observed, our model successfully produces mask tubes overlapping with the groundtruth, and importantly predicts the correct relations of the subject-object pairs. On the other hand, baseline models tend to prefer more static relations, since during training they do not explicitly focus on motion-sensitive features. Statistics in Figure \ref{fig:r@50} also substantiate our proposition, in which we achieve considerably higher recalls for dynamic relations than baseline approaches.
\section{Summary}
\noindent In this paper, we propose a motion-aware contrastive learning framework for temporal panoptic scene graph generation. In our framework, we learn close representations for temporal masks of similar entities that exhibit common relations. Moreover, we separate temporal masks from their shuffled version, and also separate temporal masks of different subject-relation-object triplets. To quantify the relationship among temporal masks in the proposed contrastive framework, we utilize optimal transport to preserve the temporal nature among temporal entity masks. Extensive experiments substantiate the effectiveness of our framework for both natural and 4D videos.
\chapter{Multi-Scale Contrastive Learning for Video Temporal Grounding}
\label{ch5:mstg}

\section{Introduction}
\noindent Temporal video grounding aims to localize moments of interest in an untrimmed video given a free-form textual description. It is a challenging multimodal task since it involves understanding temporal information in videos and reasoning about their connections to semantic information in texts. Recently, temporal grounding has drawn increasing attention \citep{mu2024snag, jung2023overcoming, xu2023boundary, pan2023scanning}, due to its wide range of applications such as surveillance \citep{zhang2016context}, robotics \citep{burgner2015continuum}, and autonomous driving \citep{claussmann2019review}.

Previous methods \citep{zhang2020learning, soldan2021vlg, zhang2020span} for temporal grounding concentrate on grounding merely a few queries in short video snippets. However, recently the growing availability of long videos, \textit{e.g.} on streaming platforms, and demands to query their rich content have necessitated productive grounding of large volumes of queries in long videos. Because of such short-to-long video paradigm shift, latest methods \citep{zhang2022actionformer, mu2024snag} have utilized local self-attention to restrict attention within a local window, following the intuition that temporal context beyond a certain range is less helpful for moment localization.

\begin{figure*}[t]
\centering
\includegraphics[width=0.45\linewidth]{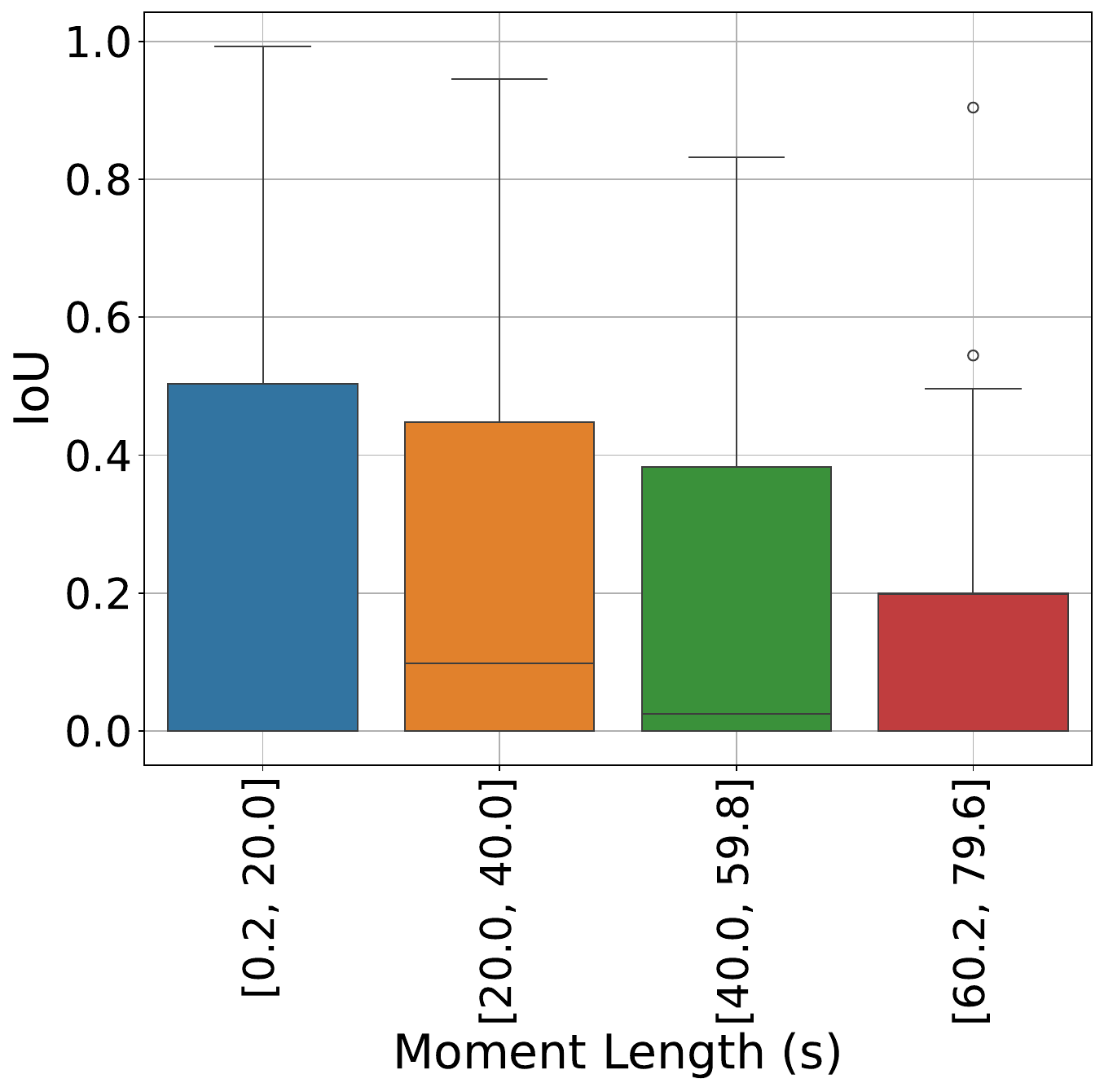}\,
\includegraphics[width=0.45\linewidth]{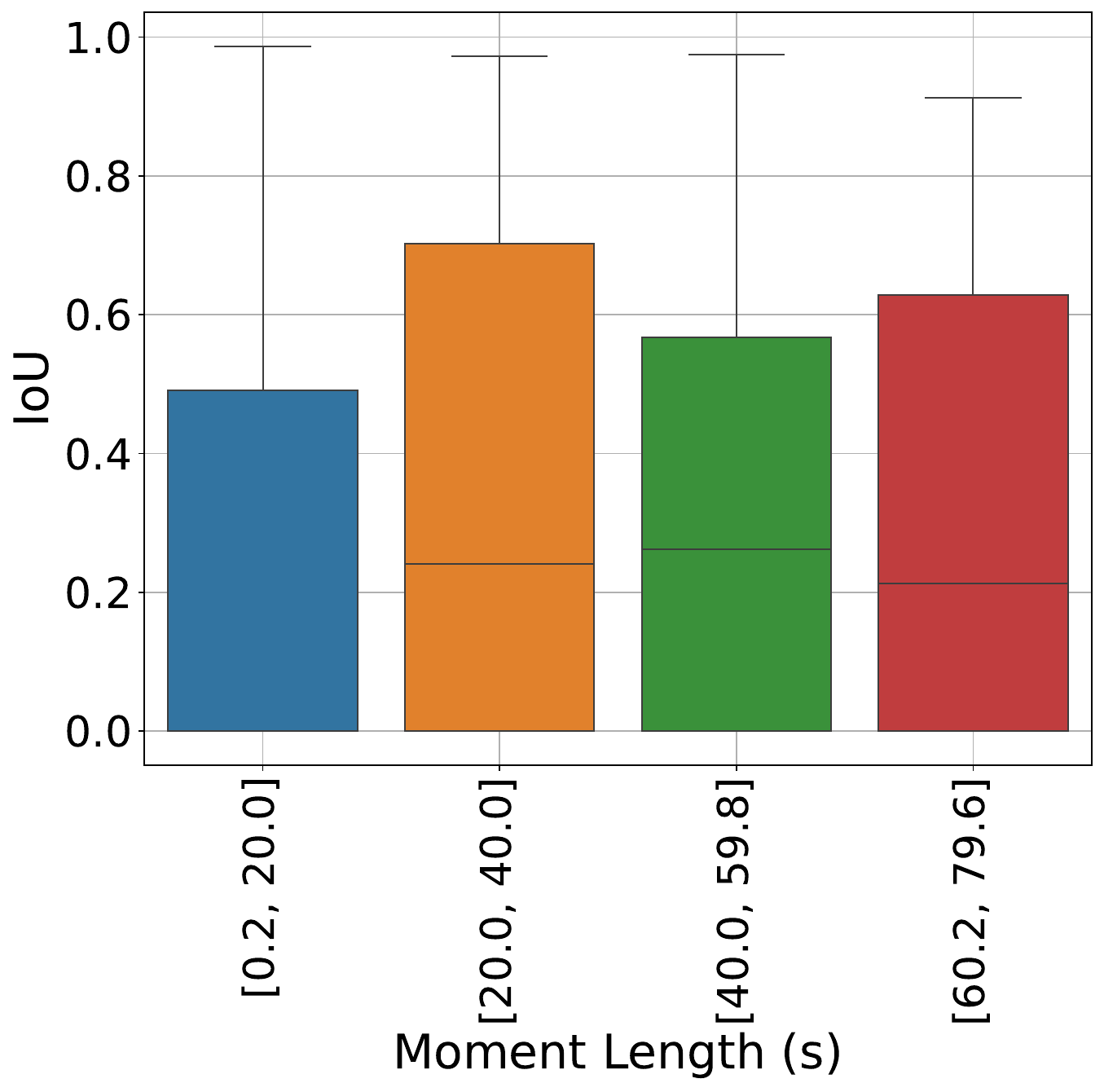} \\
\includegraphics[width=0.45\linewidth]{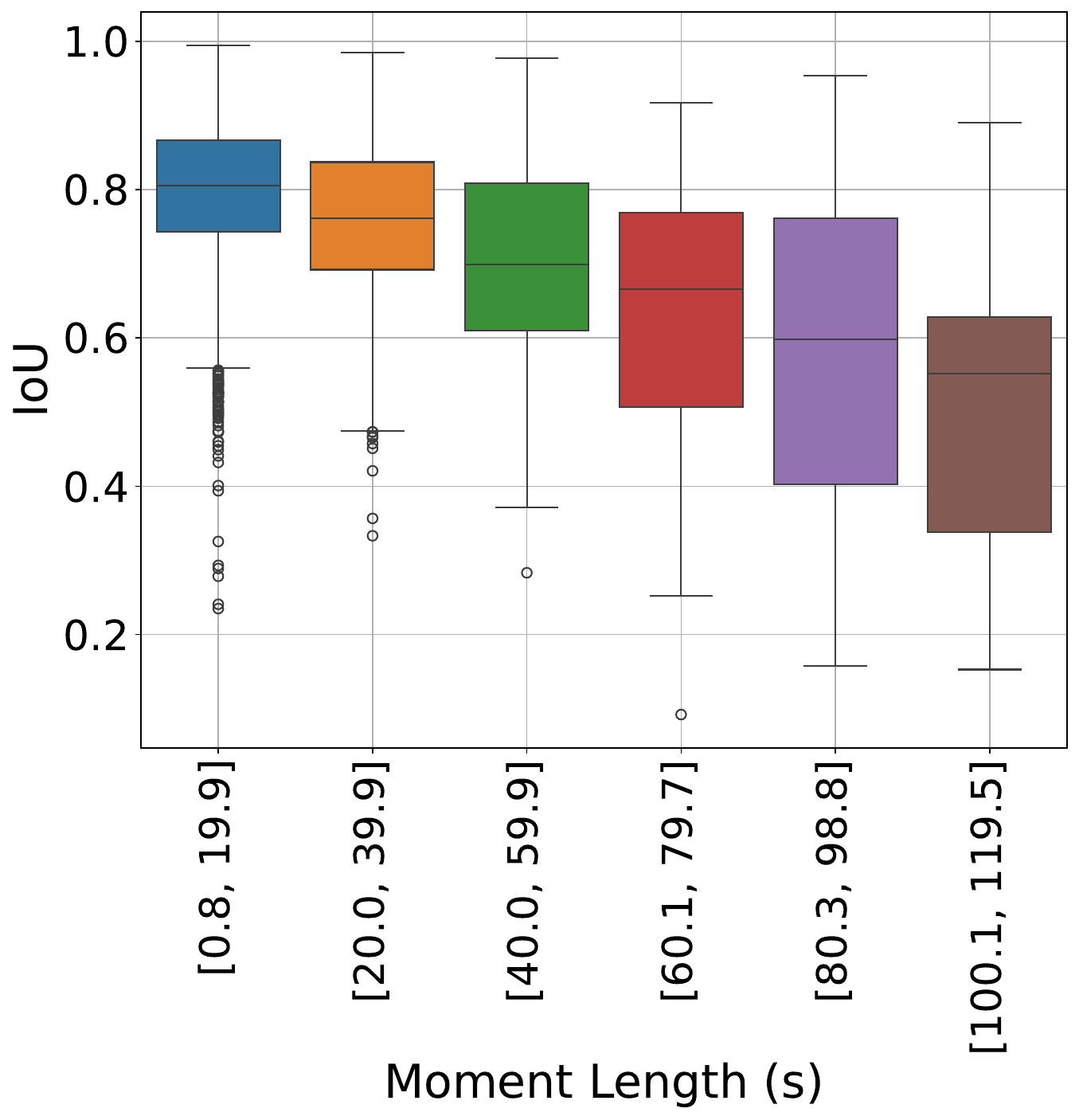}
\includegraphics[width=0.45\linewidth]{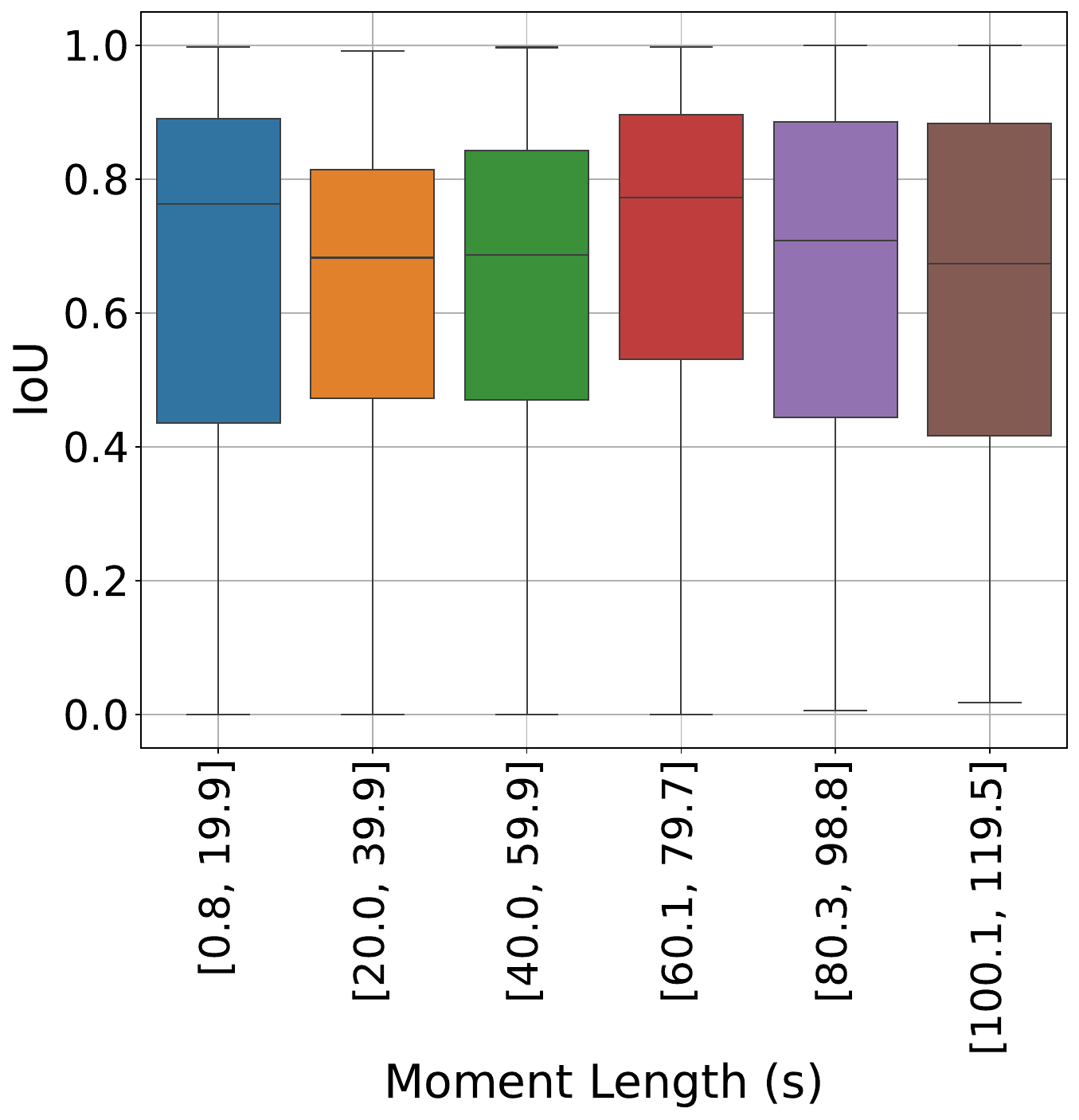}
\caption{First and Second: IoU results with respect to target video moment length on Ego4D-NLQ \citep{grauman2022ego4d} of baseline SnAG \citep{mu2024snag} and our model. Third and Fourth: IoU results with respect to target video moment length on TACoS \citep{regneri2013grounding} datasets of baseline SnAG \citep{mu2024snag} and our model.}
\label{fig:ego4d_tacos_moment_length_iou}
\end{figure*}

To capture moments at different temporal scales without enlarging the window size of the local self-attention, recent methods \citep{zhang2022actionformer, mu2024snag} need to combine several Transformer blocks with downsampling between every two blocks, resulting in a feature pyramid of moment representations, as illustrated in Figure \ref{fig:temporal_grounding_example} (left). Unfortunately, due to such downsampling operation, when moment representations are propagated from lower levels of short-range (local) moments to higher levels of long-range (global) moments, information contained in representations of longer moments will gradually degrade \cite{guo2020augfpn, yang2023afpn}. This could explain why performance of these methods tends to degrade as the duration of target moments increase, as shown in Figure \ref{fig:temporal_grounding_example} (right) and statistically shown with Intersection-over-Union (IoU) results in Figure \ref{fig:ego4d_tacos_moment_length_iou}, respectively.

\begin{figure}[t]
    \centering
    \includegraphics[width=0.4\linewidth]{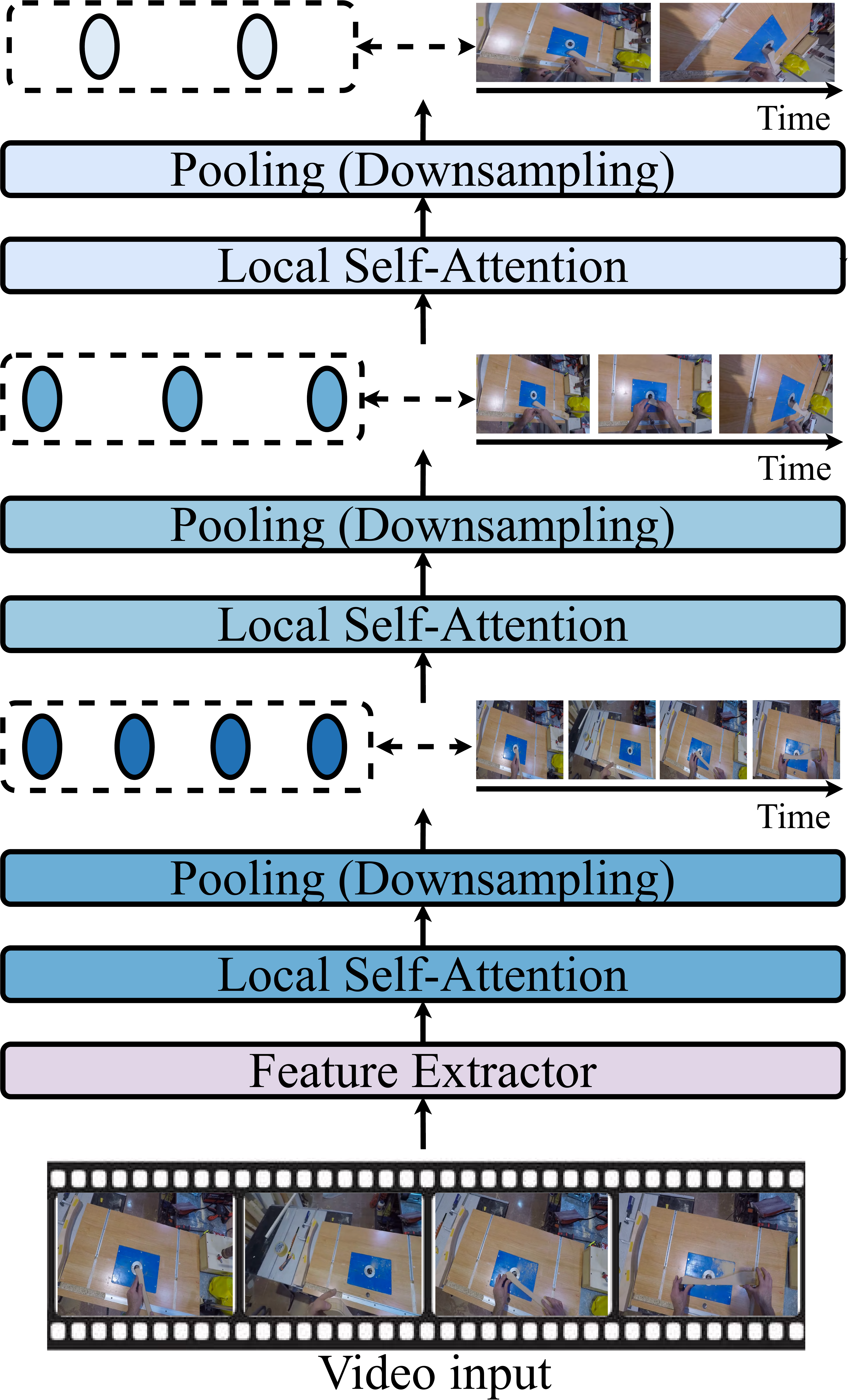}
    \caption{Illustration of feature pyramid to encode video moments of different lengths. }
    \label{fig:temporal_grounding_example}
\end{figure}

\begin{figure*}[t]
    \centering
    \includegraphics[width=0.96\linewidth]{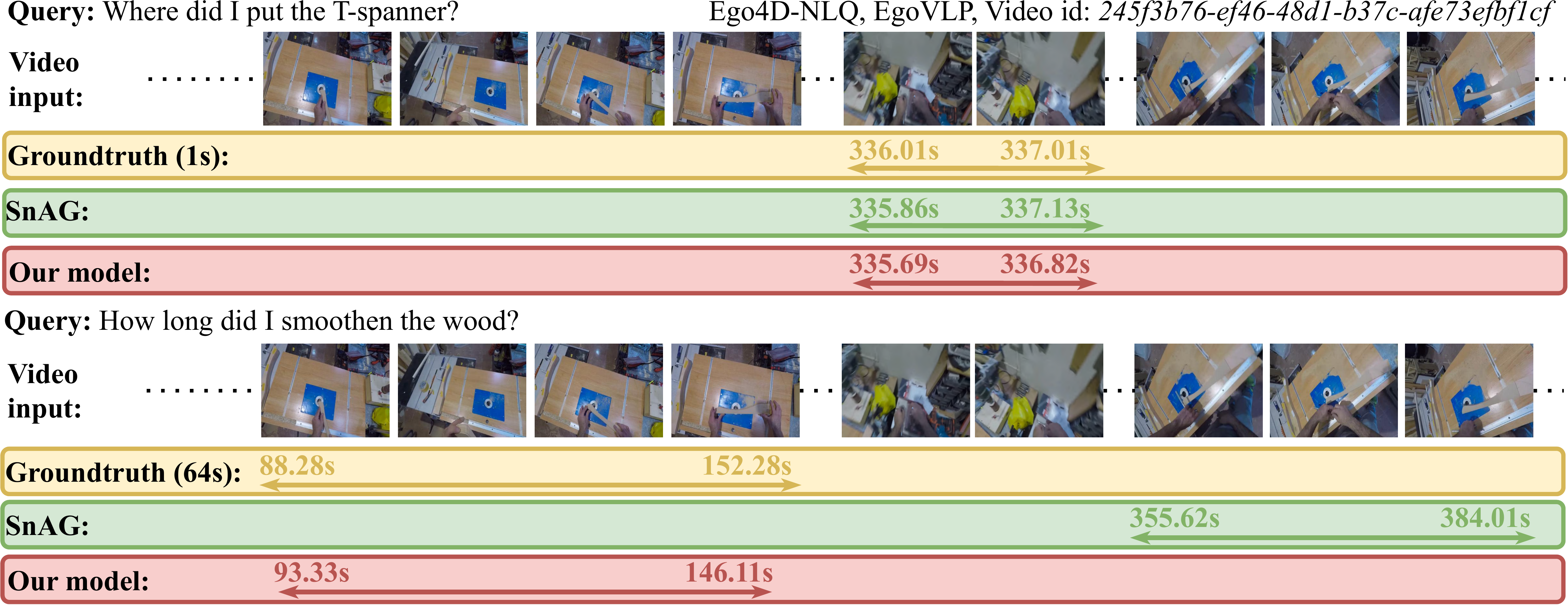} 
    \caption{An example where recent method SnAG \citep{mu2024snag} accurately localizes short video moment but fails on long moment. }
    \label{fig:temporal_grounding_example}
\end{figure*}

To enrich information in video moment representations, recent works \citep{panta2024cross, xiao2024bridging, ji2024weakly, liu2024towards} have employed contrastive learning for temporal grounding. The intuition is to capture mutual information between video moments and textual query to preserve salient semantics in moment representations. These works mainly involve query-moment pairs in which queries relate to video moments of distinct videos, hence the learned semantics among moment representations would be independent from each other. However, such approach might not be suitable for the latest scalable video-centric approach \citep{zhang2022actionformer, mu2024snag}, in which multiple textual queries are related to one video. Therefore, if the grounding of two textual queries results in temporal overlapping, there might be a conflict in compact moment representations \citep{an2023unicom}. Furthermore, focusing upon moment-query relations limits these works to the feature space of the final encoder layer, which could not effectively utilize all hidden representations across encoding layers. For multi-scale temporal grounding, such cross-scale representations should be fully used since they express semantics in video moments of various lengths. 

To resolve the above issues, in this paper, we propose a multi-scale contrastive learning framework for multi-scale temporal grounding. In our framework, instead of leveraging moment-query relationships, we utilize the association among video moments. Particularly, to avoid representation conflict among video moments, we introduce a query-centric contrastive approach that draws temporally separate video moments corresponding to a common textual query. A central component of our framework is the creation of positive and negative video moment samples, which previous works primarily apply data augmentation \citep{kim2022exploring, xing2023svformer}. However, because most long-form videos consist of a high volume of video moments, choosing an appropriate augmentation strategy that suits every moment is a non-trivial and lengthy tuning step.  Another common approach is to introduce a memory bank to store positive or negative samples' representations, which are created by aggregating input representations iteratively during training \citep{panta2024cross, han2023momentum}. Nevertheless, a memory bank would present additional hyperparameters such as the bank size and update frequency, which demand laborious tuning effort \citep{wang2021exploring}. 

To prevent these problems, we directly draw samples from the feature space of video moment encoder. Specifically, we take advantage of internal, intermediate representations of video moments from the encoder that are readily available through the feed-forward step of the network without the need to rely upon external steps such as data augmentation or online storing of samples in memory banks. Accordingly, we introduce a within-scale and cross-scale approach to create positive and negative moment samples for contrastive learning. Regarding the within-scale approach, we seek to pull together representations of such semantically close video moments on the same scale of similar temporal range. Moreover, we also push apart representations of video moments which are unrelated to the textual query. Regarding the cross-scale approach, we compel the model to relate global long-range video moments to local short-range moments, while simultaneously repelling semantically distant cross-scale representations in an analogous cross-scale manner. This cross-scale approach would enable long-range moment representations to capture nuanced details of short-range moments, thereby mitigating informational degradation within long-range representations.

To sum up, our contributions are the following:
\begin{itemize}
    \item We propose a multi-scale contrastive framework that focuses on moment-moment relations to mitigate informational degradation in video moment representations.
    \item We propose a within- and cross-scale strategy that supports semantic consistency not only between similar-range but also cross-range video moment representations emanating across layers of the video encoder.
    \item Our framework achieves superior results across major benchmark datasets concerning both short-form and long-form video grounding.
\end{itemize}

\section{Methodology}
\noindent In this section, we delineate our proposed contrastive framework for multi-scale temporal grounding, particularly focusing on a sampling procedure to draw video moment representations across temporal scales.
\begin{figure}[t]
\centering
\includegraphics[width=\linewidth]{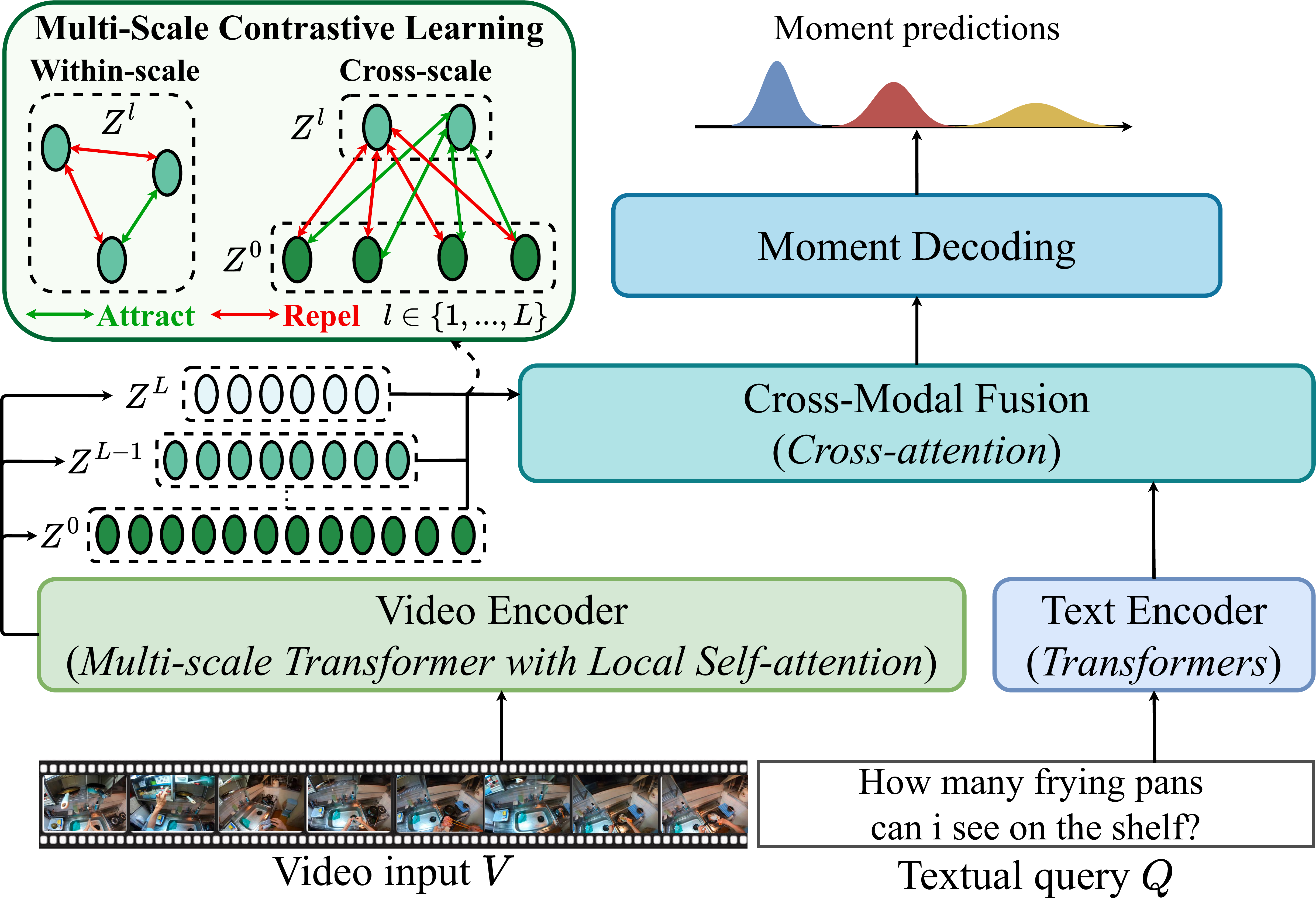}
\caption{Overall illustration of the proposed framework.}
\label{fig:overall_illustration}
\end{figure}

\subsection{Preliminary - Video Temporal Grounding}
\noindent We denote an input video $V$ as a sequence of video clips $\{v_{t}\}_{t=1}^{T} = \{v_{1}, v_{2}, ..., v_{T}\}$, where $v_{t}$ denotes a video moment (clip) centered at time $t$. We use a pre-trained feature extractor to embed each  $v_{t}$ into a moment embedding $\mathbf{v}_{t}$. Given the video $V$, our task is to localize a moment $y = \left(s, e\right)$ based on a sentence query $Q = \{q_{1}, q_{2}, ..., q_{K}\}$. Similar to the input video, we also embed the query $Q$ into a sequence of word embeddings $\{\mathbf{q}_{1}, \mathbf{q}_{2}, ..., \mathbf{q}_{K}\}$.

\paragraph{Video encoder.} After embedding video clips, we use a convolution-based projection function to encode local context of video clips:
\begin{equation}
Z^{0} = \{\mathbf{z}_{t}^{0}\}_{t=1}^{T} = \text{Conv}\left(\mathbf{v}_{1}, \mathbf{v}_{2}, ..., \mathbf{v}_{T}\right).
\end{equation}
Subsequently, we designate $L$ Transformer layers to encode temporal context among video clips. In detail, each Transformer layer consists of a local multi-head self-attention (LocalMSA) with a window size of $W$ and an MLP block, in which we restrict the attention to be within a local window:
\begin{gather}
\hspace{-5pt}
\bar{Z}^{l} = \alpha^{l}\cdot\text{LocalMSA} \left(\text{LN}\left(Z^{l-1}\right)\right) + Z^{l-1}, \\
\hat{Z}^{l} = \bar{\alpha}^{l} \cdot\text{MLP}\left(\text{LN}\left(\bar{Z}^{l}\right)\right) + \bar{Z}^{l}, \\
Z^{l} = \downarrow\left(\hat{Z}^{l}\right), \quad l \in \{1, 2, ..., L\},
\end{gather}
where $Z^{l-1}, \bar{Z}^{l}, \hat{Z}^{l} \in \mathbb{R}^{T^{l-1} \times D}$, $Z^{l} \in \mathbb{R}^{T^{l} \times D}$. $T^{l-1} / T^{l}$ is the downsampling ratio, $\alpha^{l}$ and $\bar{\alpha}^{l}$ are learnable per-channel scaling factors \citep{touvron2021going}, $D$ is the hidden dimension, and LN is the layer normalization. 

Inspired by \citep{mu2024snag}, we implement the downsampling operator $\downarrow$ as a strided depthwise 1D convolution. The downsampling operation engenders the multi-scale property of the encoder, generating representations for longer video moments.

\paragraph{Text encoder.} We use Transformer layers, where each layer includes a vanilla self-attention followed by an MLP. Thus, the textual encoder produces textual representations $E = \{\mathbf{e}_{1}, \mathbf{e}_{2}, ..., \mathbf{e}_{K}\}$ for query embeddings $\{\mathbf{q}_{1}, \mathbf{q}_{2}, ..., \mathbf{q}_{K}\}$.

\paragraph{Cross-modal fusion.} Our architecture uses cross-attention to fuse video clip and query word representations. Technically, we modulate video clip representations $\{Z^{l}\}_{l=1}^{L}$ with word representations $E$ as follows:
\begin{gather}
\label{eq:cross_modal_fusion_ln}
\tilde{Z}^{l} = \text{LN}\left(Z^{l}\right), \;\; \tilde{E} = \text{LN}\left(E\right), \\
O^{l} = \sigma\left(\frac{\left(\tilde{Z}^{l}\right)^{\top} \cdot \tilde{E}}{\sqrt{D}}\right) \cdot \tilde{Z}^{l}, \\
X^{l} = \beta^{l} \cdot \text{MLP}\left(\text{LN}\left(O^{l}\right)\right) + O^{l}, 
\end{gather}
where $\beta^{l}$ denotes a learnable per-channel scale and $\sigma$ the Softmax activation function. 

\paragraph{Moment decoding.} After cross-modal fusion, our model converts each time step $t$ to a moment candidate. Specifically, given $\mathbf{x}_{t}^{l}$, we use a convolutional network comprising 1D convolutional layers as the classification head to predict a score $p_{t}^{l}$. In a similar vein, we use a similar 1D convolutional network attached with a ReLU activation function to regress the normalized distances from $t$ to the moment boundaries $(d_{t}^{s}, d_{t}^{e})$ if $\mathbf{x}_{t}^{l}$ is classified as positive. Formally, the decoded moment is computed as:
\begin{gather}
(t, l) = \argmax_{t,l} p_{t}^{l}, \\
\hat{s} = 2^{l-1} \left(t - d_{t}^{s}\right), \quad \hat{e} = 2^{l-1} \left(t + d_{t}^{e}\right).
\end{gather}
During testing, we employ Soft-NMS \citep{bodla2017soft} to merge overlapping moment predictions.

\subsection{Cross-scale Contrastive Learning}
\noindent\textbf{Query-centric sampling.} As randomly sampling moment-query pairs for contrastive learning might lead the model to representation conflict if the groundings of two queries overlap with each other, we instead introduce a sampling approach that draws a text query $Q$ and its temporally separate video moments associated with a common video $V$:
\begin{equation}
Q_{j'}, \{y_{j'}^{l}\}_{l=1}^{L} \sim \mathcal{U} \left(\{Q_{j}, \{y_{j'}^{l}\}_{l=1}^{L}\}_{j=1}^{N_{Q}}\right),
\end{equation}
where $\mathcal{U}$ denotes a discrete uniform distribution, $\{y_{j'}^{l}\}_{l=1}^{L}$ the set of target video moments in each layer $l$, and $N_Q$  the number of textual queries related to video $V$. We generate the target set $\mathcal{P}(l)$ via center sampling \citep{zhang2022actionformer, mu2024snag}, \textit{i.e.} given any moment centered at $t$, any time step $c \in \left[t - \alpha \frac{T}{T^{l}}, t + \alpha \frac{T}{T^{l}}\alpha\right]$  in layer $l$ is considered as a target. After sampling the query and target moments, we directly utilize the representations $\{\mathbf{z}_{j'}^{l}\}_{l=1}^{L}$ of the target moments $\{y_{j'}^{l}\}_{l=1}^{L}$ extracted by the aforementioned multi-scale video encoder.

\paragraph{Within-scale contrastive learning.} Having obtained the representations of target moment samples, we directly utilize moments within each scale as positive and negative samples. Particularly, we iterate over every layer $l$ of the video encoder, and for each anchor video moment $y_{j'}^{l}$, we consider all video moments of layer $l$ corresponding to query $Q_{j'}$ to become positive moment set $P(l)$, and randomly draw those not corresponding to query $Q_{j'}$ to be negative set $\mathcal{N}(l)$. 

Then, we formulate multi-scale contrastive objective over all layers $l \in \{1, 2, ..., L\}$, which pushes positive moments closer while negative moments further:
\begin{equation}
\begin{split}
\mathcal{L}_{\text{within}} &= -\sum\limits_{l = 1}^{L}\sum\limits_{i \in \mathcal{P}(l)}\sum\limits_{j \in \mathcal{P}(l), i \neq j} \\ 
&\log \frac{e^{ \left(\mathbf{z}_{i}^{l} \cdot \mathbf{z}_{j}^{l}\right)}}{e^{ \left(\mathbf{z}_{i}^{l} \cdot \mathbf{z}_{j}^{l}\right)} + \sum\limits_{n \in \mathcal{N}(l)} e^{\left(\dot{\mathbf{z}}_{i}^{l} \cdot \mathbf{z}_{n}^{l}\right)}}.
\end{split}
\end{equation}

\paragraph{Cross-scale contrastive learning.} We further associate semantically close moment representations from across different scales. Specifically, we push short-range moment representations closer to semantically close long-range moment representations. This would enable short-range moments to relate to longer video context while long-range features to capture nuanced details of short-range moments. 

As video moment features of layer 0 $\{\mathbf{z}_{j'}^{0}\}$ are the most likely to preserve salient video information compared to other levels, we employ features of the target moments from the lowest level as the anchor set for cross-scale contrastive learning. To construct positive and negative moment set, we utilize features of higher levels $l \in \{1, 2, ..., L\}$ in the feature pyramid corresponding to video moments that involve and do not involve the textual query, respectively. Denoting the set of moment indices in level $l$ that are related to the query as $\mathcal{P}(l)$ and the set of those that are unrelated as $\mathcal{N}(l)$, we define the cross-scale contrastive learning objective as:
\begin{equation}
\begin{split}
\mathcal{L}_{\text{cross}} &= -\sum\limits_{i \in \mathcal{P}(0)}\sum\limits_{l = 1}^{L} \sum\limits_{j \in \mathcal{P}(l)} \\ 
&\log \frac{e^{ \left(\mathbf{z}_{i}^{0} \cdot \mathbf{z}_{j}^{l}\right)}}{e^{ \left(\mathbf{z}_{i}^{0} \cdot \mathbf{z}_{j}^{l}\right)} + \sum\limits_{n \in \mathcal{N}(l)} e^{\left(\mathbf{z}_{i}^{0} \cdot \mathbf{z}_{n}^{l}\right)}}.
\end{split}
\end{equation}

\subsection{Training Objective}
\noindent For temporal grounding training, we adopt a focal loss $\mathcal{L}_{\text{cls}}$ for target moment classification and a Distance-IoU loss $\mathcal{L}_{\text{reg}}$ for distance regression from a positive time step $t$ to the target moment. Then, we  combine these losses with our within- and cross-scale contrastive loss:
\begin{equation}
\mathcal{L} = \mathcal{L}_{\text{cls}} + \rho_{\text{reg}} \cdot \mathcal{L}_{\text{reg}} + \rho_{\text{within}} \cdot \mathcal{L}_{\text{within}} + \rho_{\text{cross}} \cdot \mathcal{L}_{\text{cross}},
\end{equation}
where $\rho_{\text{reg}}$, $\rho_{\text{within}}$, and $\rho_{\text{cross}}$ denote hyperparameters to balance the regression, within-scale, and cross-scale contrastive losses, respectively. 
\section{Experiments}
\noindent To validate the effectiveness, we conduct extensive experiments against recent methods for temporal grounding. We  also perform ablation study to investigate each component. 

\subsection{Datasets}
\noindent Following previous works, we work on five challenging datasets of temporal grounding, which belong to two main categories, \textit{i.e.} 1) Long videos, many queries (Ego4D-NLQ \citep{grauman2022ego4d}, MAD \citep{soldan2022mad}, and TACoS \citep{regneri2013grounding}) and 2) Short videos, few queries (ActivityNet-Captions \citep{krishna2017dense} and Charades-STA \citep{sigurdsson2016hollywood}). 

\noindent\textbf{Ego4D-NLQ} \citep{grauman2022ego4d} consists of egocentric videos recording daily human activities. Each video possesses length from 3.5 to 20 minutes and is associated with 11.6 queries on average. 

\noindent\textbf{MAD} \citep{soldan2022mad} comprises 1.2K hours of movies with 384K queries transcribed from audio description. Since each video is a movie, each exhibits 47 to 202 minutes long.

\noindent\textbf{TACoS} \citep{regneri2013grounding} focuses on cooking topics. The total video length is 10.1 hours and each video is tasked with 143.5 queries for the temporal grounding operation.

\noindent\textbf{ActivityNet-Captions} \citep{krishna2017dense} targets dense video captioning and is subsequently adapted to temporal grounding. Its video length is two minutes on average and the average number of queries per video is approximately 3.65 queries. 

\noindent\textbf{Charades-STA} \citep{sigurdsson2016hollywood} is an action recognition dataset transformed into a temporal grounding one. Each video lasts approximately 30 seconds and possesses 2.4 queries. 

\subsection{Evaluation Metrics}
\noindent We report Recall@K at different temporal intersection-over-union $\theta$ (R@K, tIoU = $\theta$) for all datasets. The metric measures the percentage of textual queries whose at least one of the top-$K$ moment predictions temporally overlap with the groundtruth moment more than $\theta$. 

\subsection{Implementation Details}
\noindent To fairly compare with previous works and satisfy the scalability of temporal grounding operation for long videos, we adopt video-centric sampling approach \citep{mu2024snag}. For Ego4D-NLQ, we use pre-trained 1) SlowFast video features \citep{feichtenhofer2019slowfast} with BERT textual features \citep{devlin2018bert}, and 2) EgoVLP video and textual features \citep{lin2022egocentric}. For testing, we report R@$\{1,5\}$, tIoU = $\{0.3, 0.5\}$. For MAD dataset, we use CLIP features \citep{radford2021learning} for both videos and texts, and report R@$\{1,5,10,50\}$, tIoU = $\{0.1, 0.3, 0.5\}$. For the TACoS dataset, we use C3D video features \citep{tran2015learning} and GloVe textual features \citep{pennington2014glove}. We report results in terms of R@$\{1,5\}$, tIoU = $\{0.5, 0.7\}$. In addition, we utilize I3D features \citep{carreira2017quo} pre-trained on Kinetics \citep{kay2017kinetics} for Charades-STA and C3D features \citep{tran2015learning} for ActivityNet-Captions experiments. For both datasets, similar to TACoS, we take advantage of GloVe textual features \citep{pennington2014glove}. We report R@$\{1,5\}$, tIoU = $\{0.5, 0.7\}$ for testing on Charades-STA, and R@$\{1,5\}$, tIoU = $\{0.3, 0.5\}$ for testing on ActivityNet-Captions. For more details regarding model architecture, we direct interested readers to the appendix.
For both within-scale and cross-scale contrastive learning implementation, we keep the size of the negative sample set $\mathcal{N}(l)$ in every level $l$ to be equal to the size of the positive video clips $\mathcal{P}(l)$ that correspond to the target video moments. Based upon validation and fair comparison with previous methods, we use $\rho_{\text{ref}} = \rho_{\text{within}} = \rho_{\text{cross}}$ = 1.0.

\subsection{Baselines}
\noindent We consider the following temporal grounding models as baselines: 

\begin{itemize}
    \item \textbf{VSL-Net} \citep{zhang2020span}: utilizes textual query to highlight regions potential to comprise the target moment.
    \item \textbf{VLG-Net} \citep{soldan2021vlg}: models temporal grounding as a graph matching problem.
    \item \textbf{Moment-DETR} \citep{lei2021detecting}: a Transformer encoder-decoder architecture that views temporal grounding as a set prediction problem.
    \item \textbf{CONE} \citep{hou2022cone}: subsequently slices a video input into windows, selects relevant windows, and ranks the selected windows to obtain target moments.
    \item \textbf{MMN} \citep{wang2022negative}: a Siamese-like network architecture that is trained with video-query and query-video contrastive learning.
    \item \textbf{SSRN} \citep{zhu2023rethinking}: enriches anchor frames with additional consecutive frames.
    \item \textbf{G2L} \citep{li2023g2l}: measures moment-query similarities using geodesic distance and quantifies cross-modal interactions with game-theoretic interactions. \item \textbf{SOONet} \citep{pan2023scanning}: an anchor-based framework that conducts grounding by pre-ranking, re-ranking, and regression.
    \item \textbf{MESM} \citep{liu2024towards}: a fine-grained moment-query contrastive approach modeled for query word and video moment representations,
    \item \textbf{Contrastive-MSAT} \citep{panta2024cross}: applies moment-query contrastive loss supported by a momentum-based memory bank.
    \item \textbf{UVCOM} \citep{xiao2024bridging}: a moment-query contrastive approach for a unified video comprehension framework,
    \item \textbf{SnAG} \citep{mu2024snag}: achieves scalable grounding  with cross-modal late fusion.
\end{itemize}

\section{Experimental Results}
\subsection{Main Results}

{\renewcommand{\arraystretch}{1.2}
\begin{table}[t]
\centering
\resizebox{0.7\linewidth}{!}{
\begin{tabular}{l|l|cccccc}
\hline
\multicolumn{1}{c|}{\multirow{2}{*}{\textbf{Features}}} & \multicolumn{1}{c|}{\multirow{2}{*}{\textbf{Model}}} & \multicolumn{3}{c}{\textbf{R@1}}           & \multicolumn{3}{c}{\textbf{R@5}}           \\ 
\multicolumn{1}{c|}{}                                   & \multicolumn{1}{c|}{}                                & \textbf{0.3} & \textbf{0.5} & \textbf{Avg} & \textbf{0.3} & \textbf{0.5} & \textbf{Avg} \\ \hline
\multirow{6}{*}{SF+BERT}                                & VSL-Net                                             & 5.45         & 3.12         & 4.29         & 10.74        & 6.63         & 8.69         \\
                                                       & CONE                                                & 10.40        & 5.03         & 7.72         & 22.74        & 11.87        & 17.31        \\
                                                       & SOONet                                              & 8.00         & 3.76         & 5.88         & 22.40        & 11.09        & 16.75        \\
                                                       & SnAG                                                & 9.83         & 6.83         & 8.33         & 27.93        & 19.27        & 23.60        \\
                                                       & MSTG & \textbf{10.80} & \textbf{7.22} & \textbf{9.49} & \textbf{28.54} & \textbf{20.38} & \textbf{25.06} \\ \hline
\multirow{4}{*}{EgoVLP}                                & VSL-Net                                             & 10.84        & 6.81         & 8.83         & 18.84        & 13.45        & 16.15        \\
                                                       & CONE                                                & 14.15        & 8.18         & 11.17        & 30.33        & 18.02        & 24.18        \\
                                                       & SnAG                                                & 15.72        & 10.78        & 13.25        & 38.39        & 27.44        & 32.92        \\
                                                       & MSTG & \textbf{16.37} & \textbf{11.27} & \textbf{13.96}  &  \textbf{39.97} & \textbf{28.70} &  \textbf{34.43} \\ \hline 
\end{tabular}}
\caption{Results on Ego4D-NLQ.}
\label{tab:ego4d_nlq_results}
\end{table}}

{\renewcommand{\arraystretch}{1.2}
\begin{table*}[t]
\centering
\resizebox{\linewidth}{!}{
\begin{tabular}{l|ccc|ccc|ccc|ccc}
\hline
\multicolumn{1}{c|}{\multirow{2}{*}{\textbf{Model}}} & \multicolumn{3}{c|}{\textbf{R@1}}           & \multicolumn{3}{c|}{\textbf{R@5}}           & \multicolumn{3}{c|}{\textbf{R@10}}          & \multicolumn{3}{c}{\textbf{R@50}}          \\
         & \textbf{0.1} & \textbf{0.3} & \textbf{0.5} & \textbf{0.1} & \textbf{0.3} & \textbf{0.5} & \textbf{0.1} & \textbf{0.3} & \textbf{0.5} & \textbf{0.1} & \textbf{0.3} & \textbf{0.5} \\ \hline
VLG-Net                                             & 3.64         & 2.76         & 1.65         & 11.66        & 9.31         & 5.99         & 17.39        & 14.56        & 9.77         & 39.78        & 34.27        & 24.93        \\
Moment-DETR                                         & 0.31         & 0.24         & 0.16         & 1.52         & 1.14         & 0.28         & 2.79         & 2.06         & 1.20         & 11.08        & 7.97         & 4.71         \\
CONE                                                & 8.90         & 6.87         & 4.10         & 20.51        & 16.11        & 9.59         & 27.20        & 21.53        & 12.82        & 43.36        & 34.73        & 20.56        \\
SOONet                                              & 11.26        & 9.00         & 5.32         & 23.21        & 19.64        & 13.14        & 30.36        & 26.00        & 17.84        & 50.32        & 44.78        & 32.59        \\
SnAG                                                & 10.28        & 8.46         & 5.55         & 24.42        & 20.60        & 13.75        & 32.23        & 27.50        & 19.00        & 52.28        & 46.68        & 35.24        \\
MSTG                                           &       \textbf{12.76}       &    \textbf{10.94}          &   \textbf{6.92 }          &     \textbf{26.43}         &     \textbf{22.60}         &     \textbf{15.43}         &     \textbf{34.08}         &     \textbf{29.41}         &   \textbf{20.70}           &       \textbf{54.84}       &        \textbf{48.26}      &      \textbf{37.77}        \\ \hline
\end{tabular}}
\caption{Results on MAD.}
\label{tab:mad_results}
\end{table*}}

{\renewcommand{\arraystretch}{1.2}
\begin{table*}[t]
\centering
\resizebox{\linewidth}{!}{
\begin{tabular}{l|cccc|cccc|cccc}
\hline
\multicolumn{1}{c|}{\multirow{3}{*}{\textbf{Model}}} & \multicolumn{4}{c|}{\textbf{TACoS}}                                  & \multicolumn{4}{c|}{\textbf{ActivityNet-Captions}}                   & \multicolumn{4}{c}{\textbf{Charades-STA}}                           \\  
\multicolumn{1}{c|}{}                                & \multicolumn{2}{c}{\textbf{R@1}} & \multicolumn{2}{c|}{\textbf{R@5}} & \multicolumn{2}{c}{\textbf{R@1}} & \multicolumn{2}{c|}{\textbf{R@5}} & \multicolumn{2}{c}{\textbf{R@1}} & \multicolumn{2}{c}{\textbf{R@5}} \\
                                & \textbf{0.3}    & \textbf{0.5}   & \textbf{0.3}    & \textbf{0.5}   & \textbf{0.5}    & \textbf{0.7}   & \textbf{0.5}    & \textbf{0.7}   & \textbf{0.5}    & \textbf{0.7}   & \textbf{0.5}    & \textbf{0.7}   \\ \hline
VLG-Net                                             & 45.46           & 34.19          & 70.38           & 56.56          & 46.32           & 29.82          & 77.15           & 63.33          & -               & -              & -               & -              \\
MGSL-Net                                            & 42.54           & 32.27          & 63.39           & 50.13          & 51.87           & 31.42          & 82.60           & 66.71          & 63.98           & 41.03          & 93.21           & 63.85          \\
SSRN                                                & 45.10           & 34.33          & 65.26           & 51.85          & 54.49           & 33.15          & 84.72           & 68.48          & 65.59           & 42.65          & 94.76           & 65.48          \\
MMN                                                 & 39.24           & 26.17          & 62.03           & 47.39          & 48.59           & 29.26          & 79.50           & 64.76          & -               & -              & -               & -              \\
G2L                                                 & 42.74           & 30.95          & 65.83           & 49.86          & 51.68           & 33.35          & 81.32           & 67.60          & -               & -              & -               & -              \\
MESM                                                &  52.69            & 39.52         & -           & -          & -           & -         & -           & -          & 61.24            & 38.04          & -          & -          \\
Contrastive-MSAT                                                & 49.77             & 37.99          & 68.31           & 58.31          & 47.73              & 31.21          & 78.06           & 63.63          &   -         &      -     &    -       &    -       \\
UVCOM                                                & 36.39             & 23.32          & -           & -          & -              & -          & -           & -          &   59.25         &      36.64     &    -       &    -       \\
SnAG                                                & 56.44           & 44.86          & 81.15           & 70.66          & 48.55           & 30.56          & 81.71           & 63.41          & 64.62           & 46.26          & 92.55           & 71.94          \\
MSTG                                          &  \textbf{58.17}               &      \textbf{47.04}          &         \textbf{84.84}        &         \textbf{73.55}       &       \textbf{54.83}          &    \textbf{33.56}            &       \textbf{84.78}          &         \textbf{68.91}       &         \textbf{66.64}        &      \textbf{47.03}          &      \textbf{93.66}           &    \textbf{72.53}            \\ \hline
\end{tabular}}
\caption{Results on TACoS, ActivityNet-Captions, and Charades-STA.}
\label{tab:tacos_activitynet_charades_results}
\end{table*}}

\noindent\textbf{Results on Ego4D-NLQ} (Table \ref{tab:ego4d_nlq_results}). Our framework significantly outperforms recent temporal grounding methods. For example, using SlowFast+BERT features, we outperform previous best method, \textit{i.e.} SnAG, by mean improvements of 1.16\%  and 1.46\% in terms of R@1 and R@5 metrics, respectively. In addition, we accomplish more significant performance gains on the more stringent tIoU threshold of 0.5, denoting more precise moment localization.

\noindent\textbf{Results on MAD} (Table \ref{tab:mad_results}). Similar to results on Ego4D-NLQ, our framework obtains an outstanding improvement over previous temporal grounding methods. Specifically, we enhance SOONet with 1.68 and 2.82 points of R@1 and R@5 on average. Moreover, our model outperforms CONE and SnAG in terms of mean R@1 / R@5 by 3.58 / 6.08 and 2.11 / 1.90 points, respectively, especially for the more stringent tIoU threshold. 

\noindent\textbf{Results on TACoS} (Table \ref{tab:tacos_activitynet_charades_results} (left)). Our model achieves R@1 / R@5 of 47.04\% / 73.55\% at tIoU = 0.5, outperforming the strongest baseline, \textit{i.e.} SnAG, by a substantial margin, \textit{i.e.} +2.18\% R@1 and +2.89\% R@5. Combined with the results on Ego4D-NLQ and MAD, these results demonstrate that our contrastive framework provides beneficial signals to counter informational degradation in the feature pyramid for long-form video grounding. 

\noindent\textbf{Results on ActivityNet-Captions} (Table \ref{tab:tacos_activitynet_charades_results} (middle)). We achieve R@1 / R@5 scores of 33.56\% / 68.91\% at tIoU = 0.7. These results indicate that we outperform SSRN by 0.41\% and 0.43\% with regards to R@1 and R@5, respectively, even though we use the backbone SnAG which is significantly weaker than SSRN.

\noindent\textbf{Results on Charades-STA} (Table \ref{tab:tacos_activitynet_charades_results} (right)). Our model outperforms previous methods by a wide margin. Particularly, we accomplish 47.03\% R@1 and 72.53\% R@5 at tIoU = 0.7, exceeding SSRN by 4.38\% R@1 and 7.04\% R@5. These outcomes on Charades-STA and ActivityNet-Captions show that mutual information signals among video moments contributed by our contrastive framework can polish video moment representations to help temporal grounding on short-form videos.

\subsection{Ablation Study}
\label{sect:ablation}
\noindent We conduct extensive experiments on TACoS to study the influence of the design choices. 

{\renewcommand{\arraystretch}{1.2}
\begin{table}[t]
\centering
\resizebox{0.8\linewidth}{!}{
\begin{tabular}{l|cc|cc}
\hline
\multicolumn{1}{c|}{\multirow{2}{*}{\textbf{Positive-negative sampling approach}}} & \multicolumn{2}{c|}{\textbf{R@1}} & \multicolumn{2}{c}{\textbf{R@5}} \\
                                              & \textbf{0.3}    & \textbf{0.5}   & \textbf{0.3}    & \textbf{0.5}   \\ \hline
Data augmentation                                                       &    57.00		             &    45.46          &        83.13         &    72.06            \\
Memory bank                                                        &        57.69		       &    46.62            & 84.13                 &         72.94       \\
MSTG                                                        & \textbf{58.17}           & \textbf{47.04}          & \textbf{84.84}           & \textbf{73.55}        \\ \hline  
\end{tabular}}
\caption{Ablation results on TACoS with various positive and negative sampling approaches.}
\label{tab:ablation_positive_negative}
\end{table}}

{\renewcommand{\arraystretch}{1.2}
\begin{table}[t]
\centering
\resizebox{0.7\linewidth}{!}{
\begin{tabular}{l|cc|cc}
\hline
\multicolumn{1}{c|}{\multirow{2}{*}{\textbf{Contrastive component}}} & \multicolumn{2}{c|}{\textbf{R@1}} & \multicolumn{2}{c}{\textbf{R@5}} \\
                               & \textbf{0.3}    & \textbf{0.5}   & \textbf{0.3}    & \textbf{0.5}   \\ \hline
w/o within-scale                                     & 57.40           & 46.00          & 83.46           & 72.39          \\
w/o cross-scale                                      & 57.00           & 45.85          & 82.34           & 71.58          \\
MSTG                                                 & \textbf{58.17}           & \textbf{47.04}          & \textbf{84.84}           & \textbf{73.55}     \\ \hline    
\end{tabular}}
\caption{Ablation results on TACoS with multi-scale contrative components.}
\label{tab:ablation_contrastive_components}
\end{table}}

{\renewcommand{\arraystretch}{1.2}
\begin{table}[t]
\centering
\resizebox{0.6\linewidth}{!}{
\begin{tabular}{l|cc|cc}
\hline
\multicolumn{1}{c|}{\multirow{2}{*}{\textbf{Association approach}}} & \multicolumn{2}{c|}{\textbf{R@1}}                      & \multicolumn{2}{c}{\textbf{R@5}}                      \\
& \textbf{0.3}              & \textbf{0.5}              & \textbf{0.3}              & \textbf{0.5}              \\ \hline
Query-query                                                        & 55.61 & 45.06 & 81.25 & 71.75 \\
Moment-query                                                       & 57.00                     & 46.24                     & 82.44                     & 72.37                     \\
CLIP-based moment-moment                                           & 57.13                     & 46.94                     & 83.28                     & 72.96                     \\
MSTG                                                               & \textbf{58.17}                     & \textbf{47.04}                     & \textbf{84.84}                     & \textbf{73.55}              \\ \hline      
\end{tabular}}
\caption{Ablation results on TACoS with various association approaches.}
\label{tab:ablation_association}
\end{table}}

\paragraph{Effect of contrastive components.} We explore what extent each component of our contrastive framework, \textit{i.e.} within- or cross-scale objective, contributes to the overall performance improvement. In Table \ref{tab:ablation_contrastive_components}, cross-scale objective plays a more fundamental role in polishing video moment representations than the within-scale counterpart. Since cross-scale contrastive objective concentrates more upon long-range moment representations by relating them with the short-range moment ones, these results validate our hypothesis that informational degradation is a fundamental problem to resolve in multi-scale temporal grounding.

\paragraph{Effect of moment-moment association.} In addition to our proposed moment-moment association, we experiment with various approaches, \textit{i.e.} moment-query association, query-query association, and one approach to associate video moments but based on the semantic closeness of their corresponding textual queries. For the last approach, we consider two textual queries to be semantically similar if their CLIP-based cosine similarity score is greater than or equal to 0.8 (for positive sampling) and semantically distant if the similarity score is smaller than or equal to 0.2 (for negative sampling). As can be observed in Table \ref{tab:ablation_association}, query-query association performs the worst, as the approach does not polish moment representations. The moment-moment approach outperforms moment-query contrastive learning, but underperforms our method. We hypothesize that there might exist representation conflict between two video moments temporally overlap with each other.

\paragraph{Effect of direct utilization of moment representations.} We study the impact of our direct utilization of moment representations for positive and negative sample generation, and compare with Tube TokenMix \citep{xing2023svformer} as the data augmentation and the momentum-based memory bank approach \citep{panta2024cross}. Table \ref{tab:ablation_positive_negative} shows that we significantly surpass other methods, on average by 1.38 / 1.60 points of R@1 / R@5 over the augmentation approach, and 0.45 / 0.66 points of R@1 / R@5 over the memory bank approach. We hypothesize that while memory bank may maintain a high number of samples for contrastive learning, expensive hyperparameter tuning is essential to achieve an effective performance. 

\subsection{Qualitative Analysis}
\noindent In Figure \ref{fig:ego4d_tacos_moment_length_iou}, we observe that our model does not encounter degraded performance when the lengths of the target moments increase. Moreover, we visualize moment predictions of the recent method, \textit{i.e.} SnAG \citep{mu2024snag}, and our model in Figure \ref{fig:temporal_grounding_example}. Even though SnAG could precisely detect the shorter-length moment, it misses the moment of longer length, due to the degraded information issue. In contrast, our framework is able to localize both the short and long moments. We hypothesize that our contrastive framework can hold salient semantics for video moment representations to resolve the degraded signals in the grounding model, thus enhancing the grounding operation towards long video moments.
\section{Summary}
\noindent In this paper, we propose a multi-scale contrastive framework for multi-scale temporal grounding. Essentially, our framework utilizes a query-centric approach to associate temporally separate video moments which correspond to a common textual query to avoid representation conflict. Accordingly, we define a within-scale contrastive objective to model relations among similar-range video moments, and a cross-scsale objective to model relations among cross-range moments. Comprehensive experiments validate the effectiveness of our framework for both short-term and long-term temporal grounding.
\chapter{Temporal-Oriented Recipe for Transferring Large Vision-Language Model to Video Understanding}
\label{ch6:recipe}

\section{Introduction}
\noindent Empowered by the elevating popularity of video-text data \citep{nguyen2024encoding, nguyen2024video} and outstanding advances in large language model (LLM)-based designs, recent years have encountered remarkable progress in video understanding with large vision-language models (LVLMs). From the advent of models such as BLIP \citep{li2022blip}, BLIP-2 \citep{li2023blip}, and LLaVA \citep{liu2023visual}, video question answering (VideoQA) has improved from 33.8, 16.7, and 12.4 on MSVD \citep{wu2017deep}, MSRVTT \citep{xu2016msr}, and ActivityNet \citep{krishna2017dense} to more than 60.0 in terms of GPT-3.5 evaluation. Not only VideoQA but also long-term action recognition \citep{kuehne2014language, tang2019coin, wu2021towards} and video captioning \citep{zhou2018towards, islam2024video} have achieved significant breakthroughs.

In recent years, model architectures and training protocols have witnessed significant advancements. However, as these systems grow in diversity and scale, their computational demands pose substantial challenges for comparison, analysis, and reproducibility. Despite these advancements, many approaches have overlooked the core nature of video understanding. Rather than explicitly modeling temporal relationships, they often rely on spatial inductive biases, assuming that spatial knowledge can seamlessly extend to temporal comprehension. For instance, several methods focus on creating a unified representation space for visual and textual modalities \citep{lin2023video, chen2023videollm, zhang2023video}. Others emphasize aggregating or selecting salient visual tokens aligned with prompts \citep{shang2024traveler, xu2024pllava} or leverage large-scale pretraining with instruction-following datasets \citep{maaz2023video, luo2023valley, wang2024gpt4video}. Therefore, existing models fall short of realizing the full potential of video understanding. For example, while VideoQA systems can accurately answer questions about object detection or describe isolated actions, they struggle with queries involving causal and temporal relationships \citep{xiao2021next}. As shown in Table \ref{tab:examples}, they often generate inaccurate responses when faced with questions about temporal order or causality.

\begin{figure*}[t]
    \centering
    \caption{Our temporal-oriented recipe for large vision-language model.}
    \label{fig:recipe}
    \includegraphics[width=0.7\linewidth]{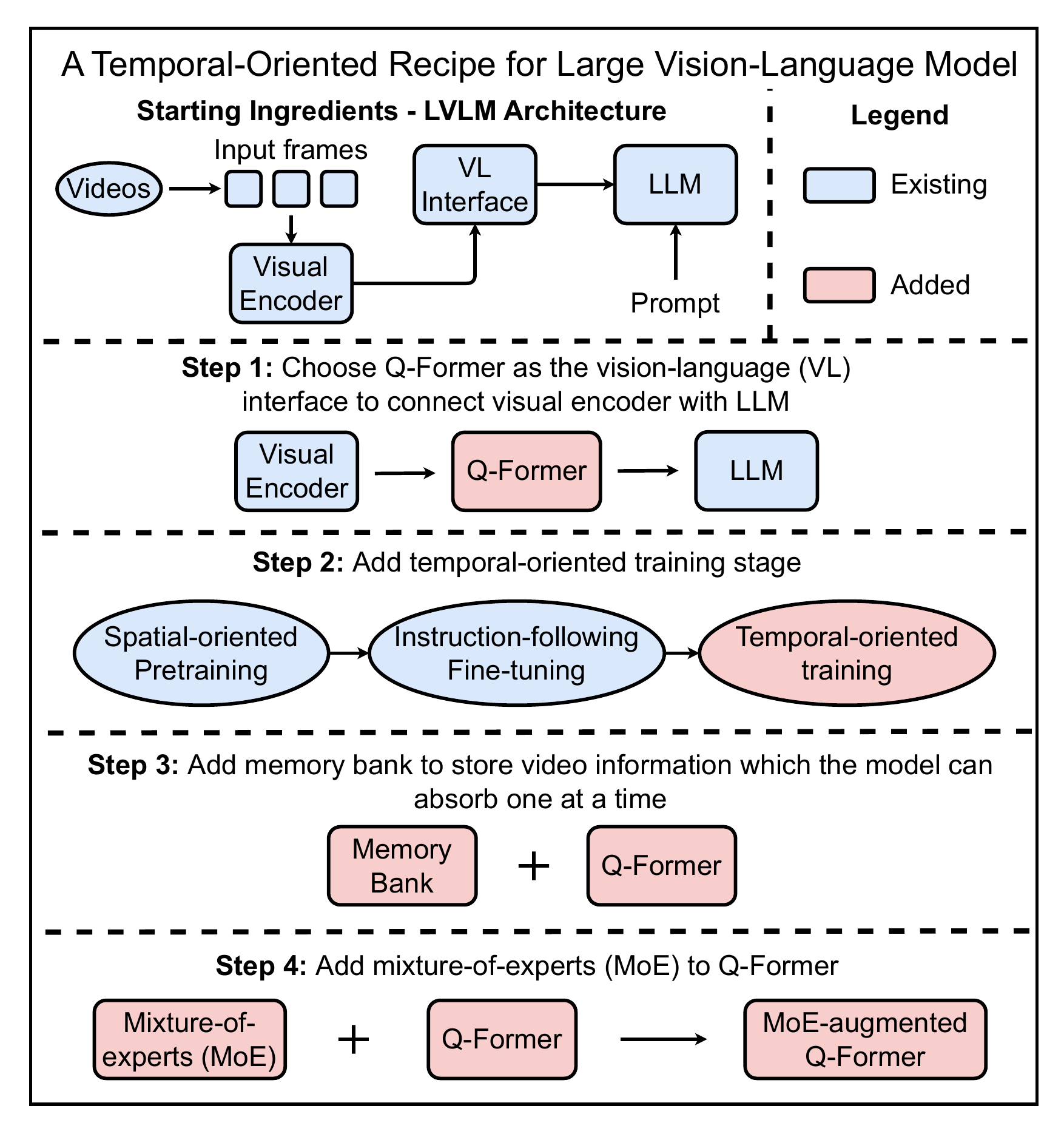}
\end{figure*}

To overcome this limitation, we aim to enhance temporal understanding capabilities of large vision-language models (LVLMs) by advancing temporal-critical components within their architectures. As illustrated in Figure \ref{fig:recipe}, an LVLM is fundamentally composed of three main components: a visual encoder, a vision-language interface, and a large language model (LLM). However, due to the large-scale nature of LLMs and the multimodal complexity of video data, identifying the primary factors driving model effectiveness is challenging \citep{he2024ma, qian2024streaming, chandrasegaran2024hourvideo}, hindering further progress in the field. Our focus is to bridge this gap by ensuring that temporal understanding is treated as a core aspect of video comprehension, rather than an implicit outcome of spatial knowledge.

\begin{table}[t]
    \caption{On the first row, a correct answer should comprise details related to cutting ginger and garlic on a chopping board, whereas other models wrongly mention ``\textit{rub salt}'', ``\textit{cut chicken}'' and ``\textit{add to the pot}'', and ``\textit{pour milk}''. On the second row, we need to respond with ``\textit{getting to the bus}'', but the models mistakenly note ``\textit{late for exam}'', ``\textit{to the hospital}'', and ``\textit{feeling sad}''.}
  \label{tab:examples}
  \vspace{5pt}
  \centering
  \resizebox{0.9\linewidth}{!}{
  \begin{tabular}{@{} l | p{0.18\linewidth}  |p{0.18\linewidth} | p{0.18\linewidth}  |p{0.18\linewidth} | p{0.18\linewidth} @{} }
    \toprule
    \rowcolor{gray!30}
    \multicolumn{1}{c|}{\textbf{Video}}        & \multicolumn{1}{c|}{\textbf{Question}}    & \multicolumn{1}{c|}{\textbf{Sample Answer}}  & \multicolumn{1}{c|}{\textbf{Video-LLaMA}} & \multicolumn{1}{c|}{\textbf{Video-LLaVA}} & \multicolumn{1}{c|}{\textbf{Qwen2.5-VL}} \\
    \midrule
    \multirow{3}{*}{%
      \includegraphics[width=0.09\linewidth]{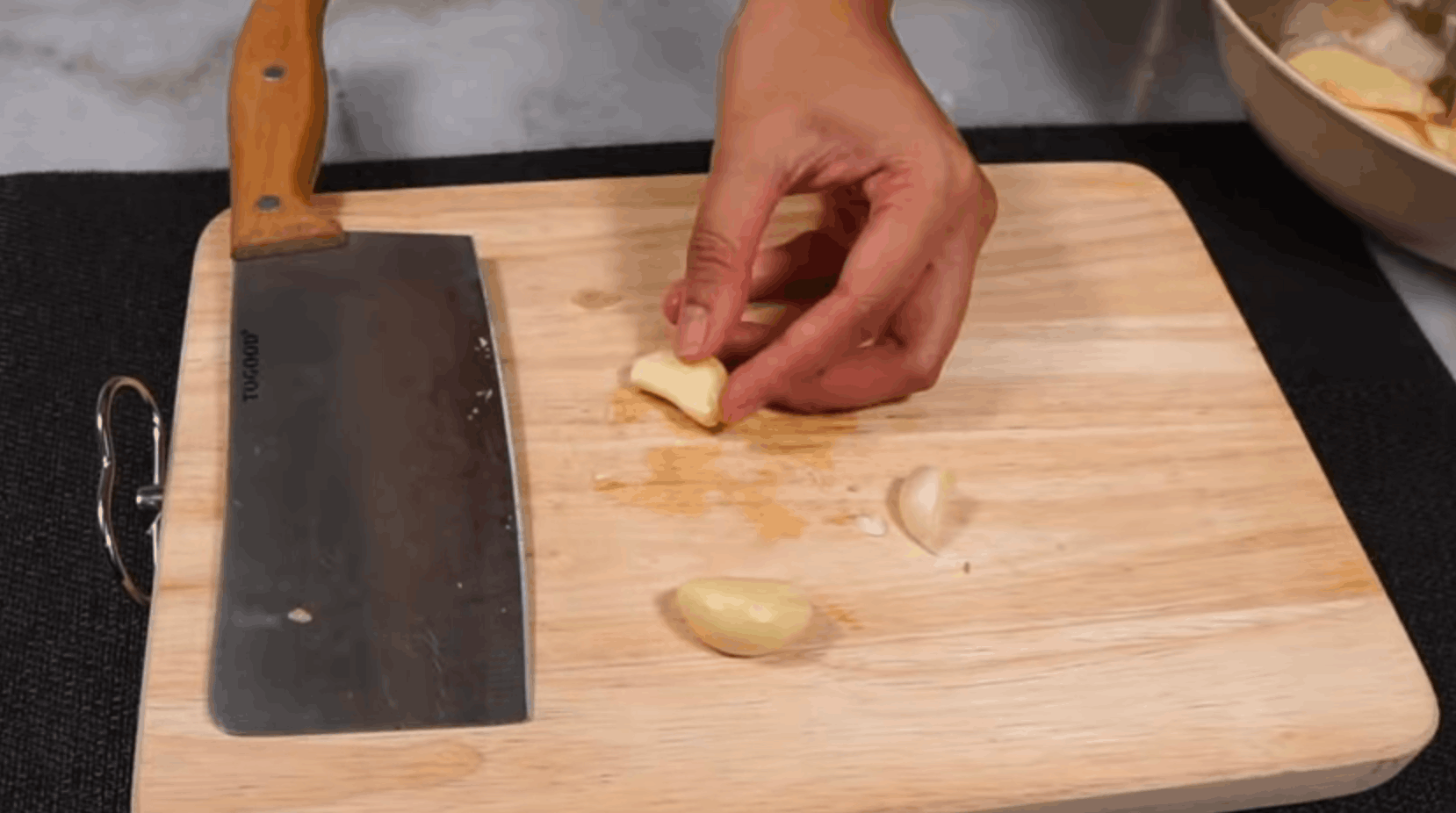}\quad
      \includegraphics[width=0.09\linewidth]{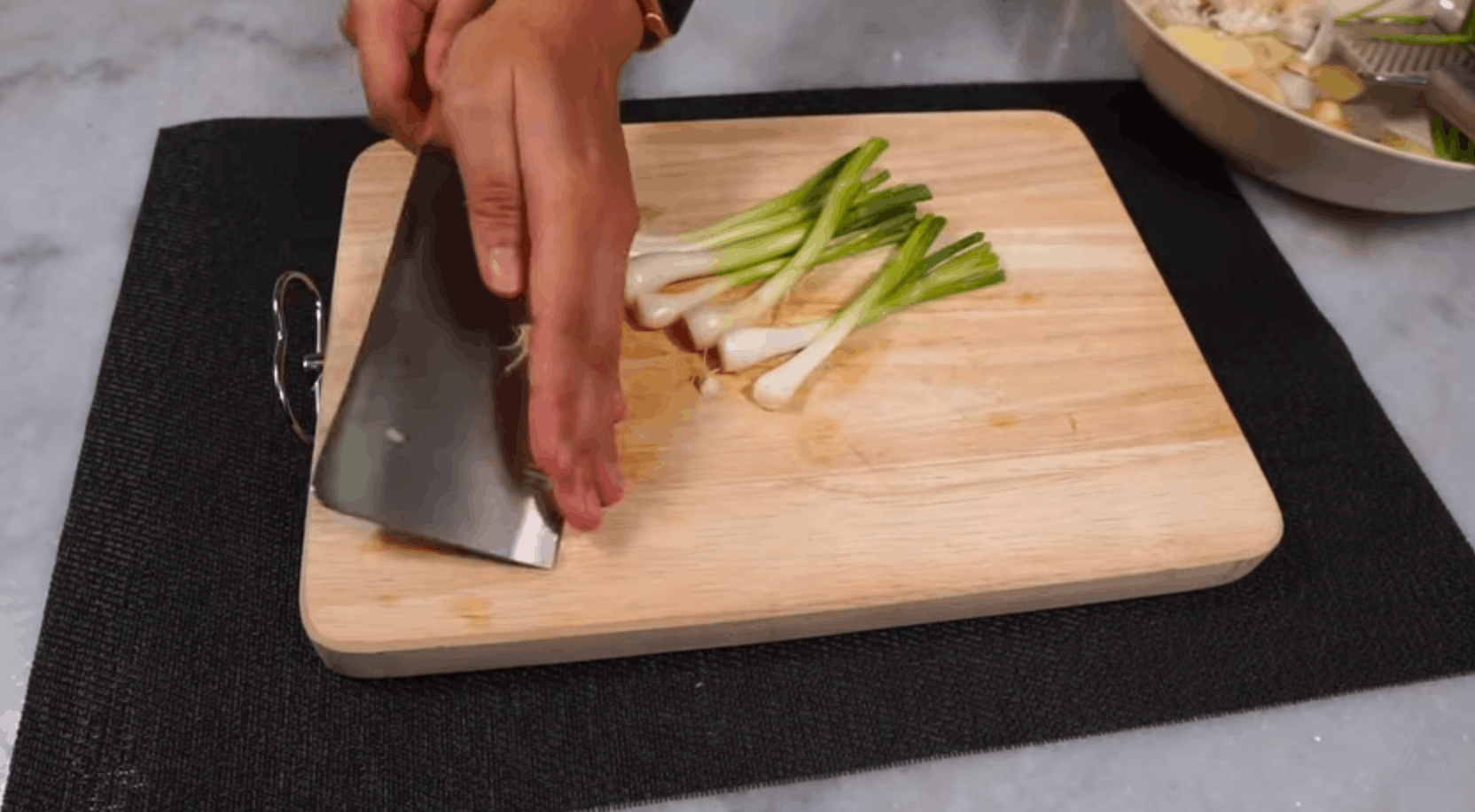}\quad
      \includegraphics[width=0.09\linewidth]{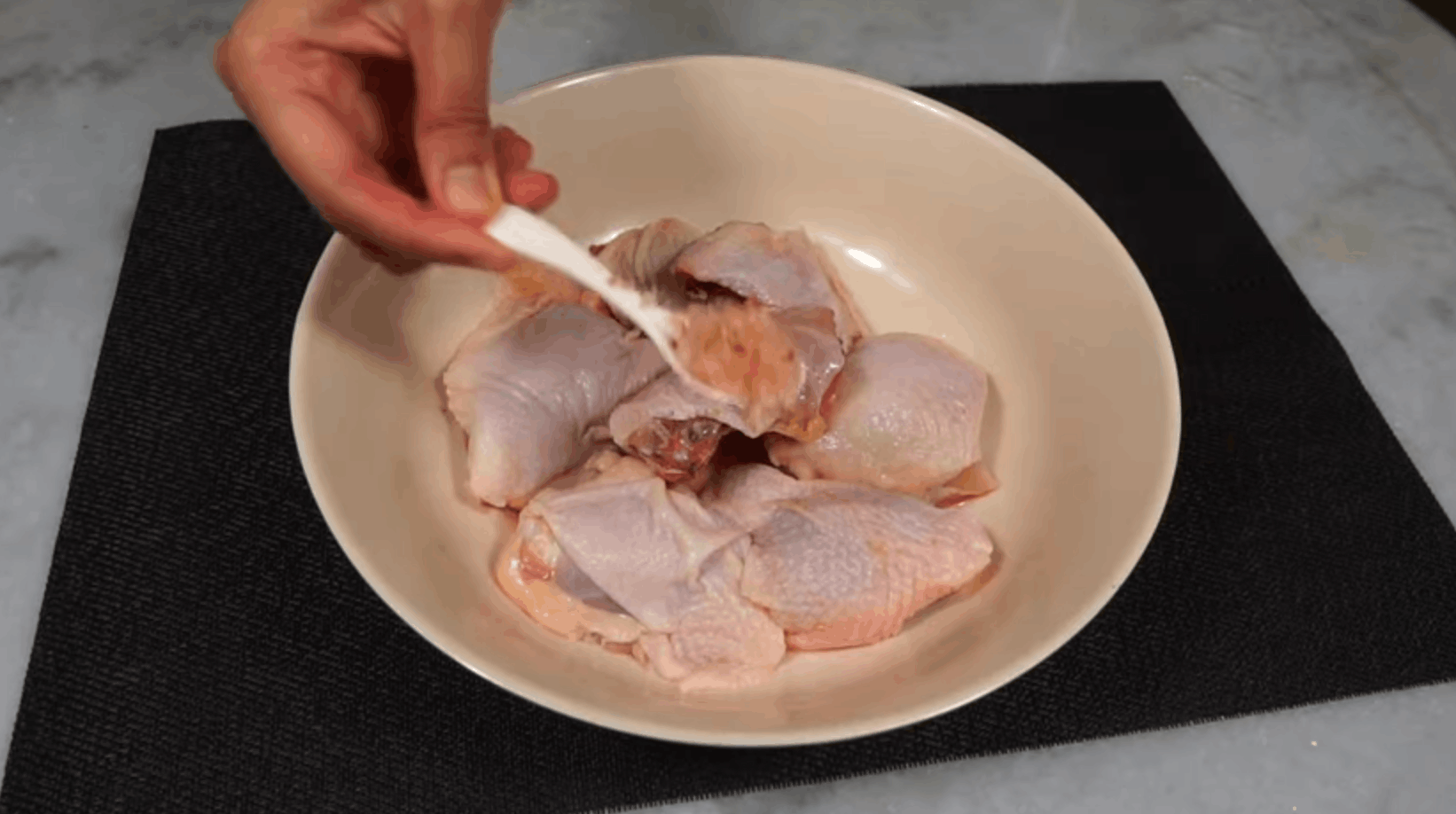}\quad
      \includegraphics[width=0.09\linewidth]{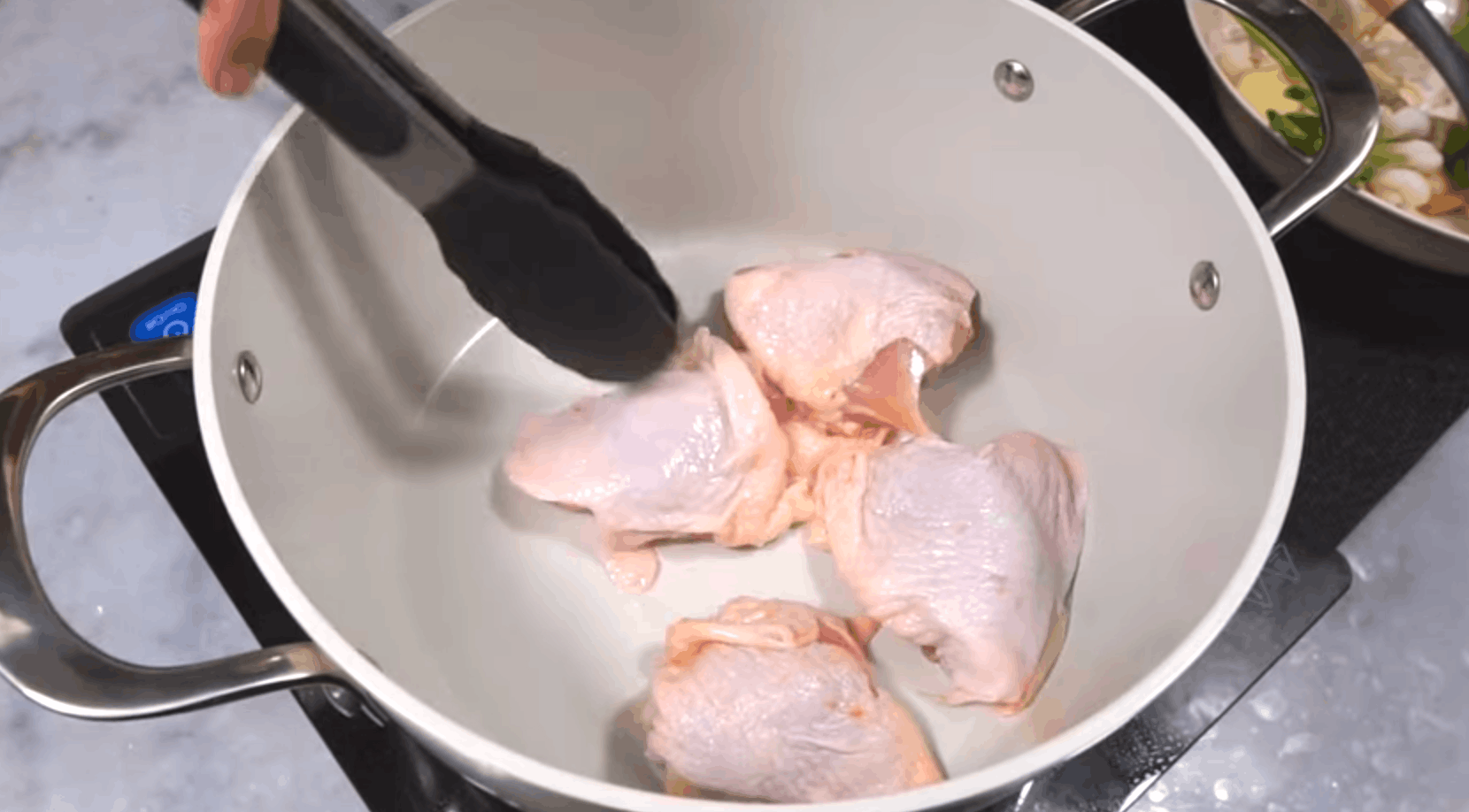}}  
      & What does the person do before seasoning chicken?  
      & The person cuts ginger and garlic into small pieces on a chopping board  
      & The person rubs salt on the chicken.  
      & Before seasoning chicken, the person cuts it into small pieces and adds it to a pot.  
      & Before seasoning chicken, the person pours milk into a bowl of rice.  \\ \hline
    \addlinespace[3pt]
    \multirow{3}{*}{%
      \includegraphics[width=0.09\linewidth]{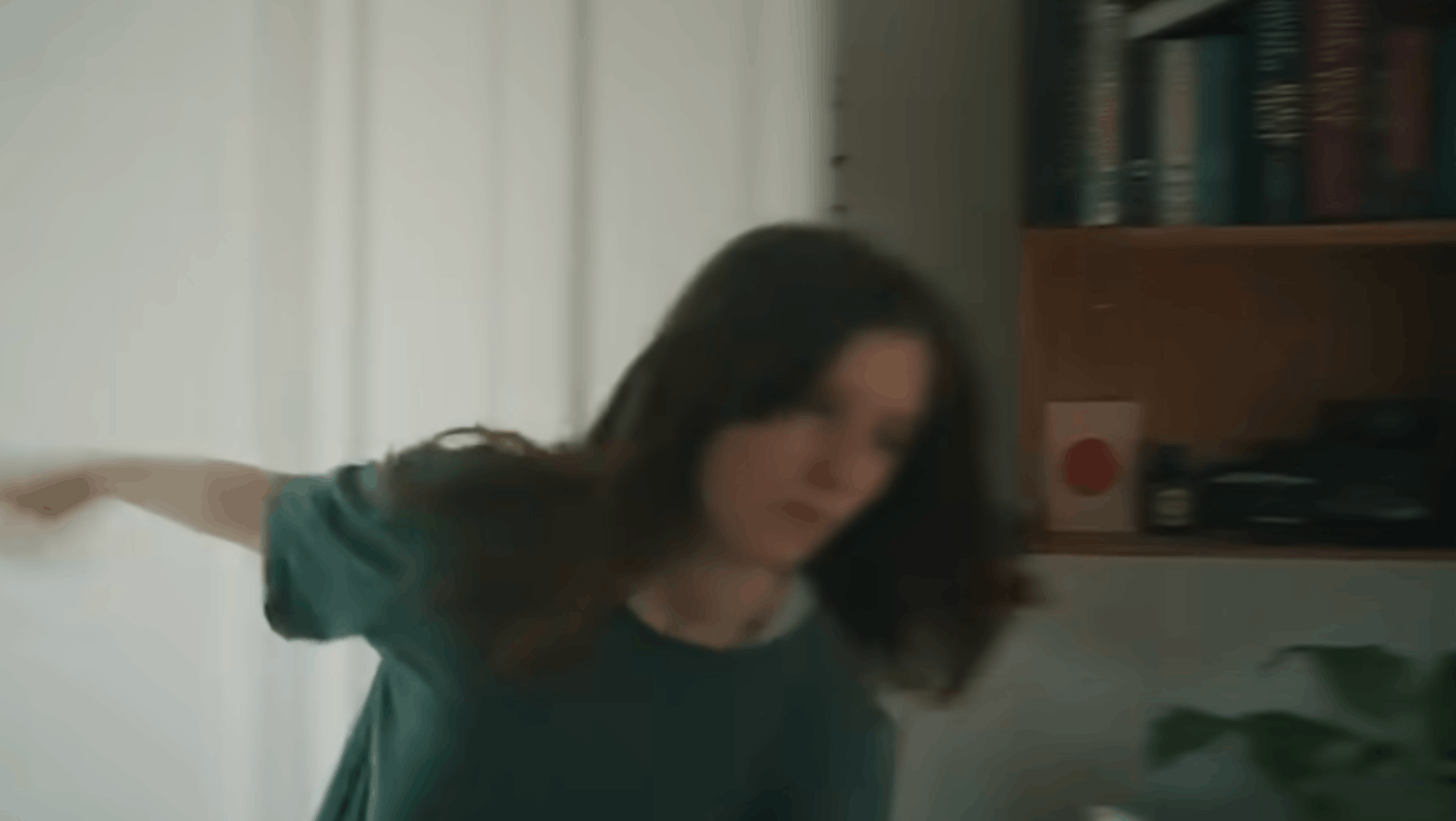}\quad
      \includegraphics[width=0.09\linewidth]{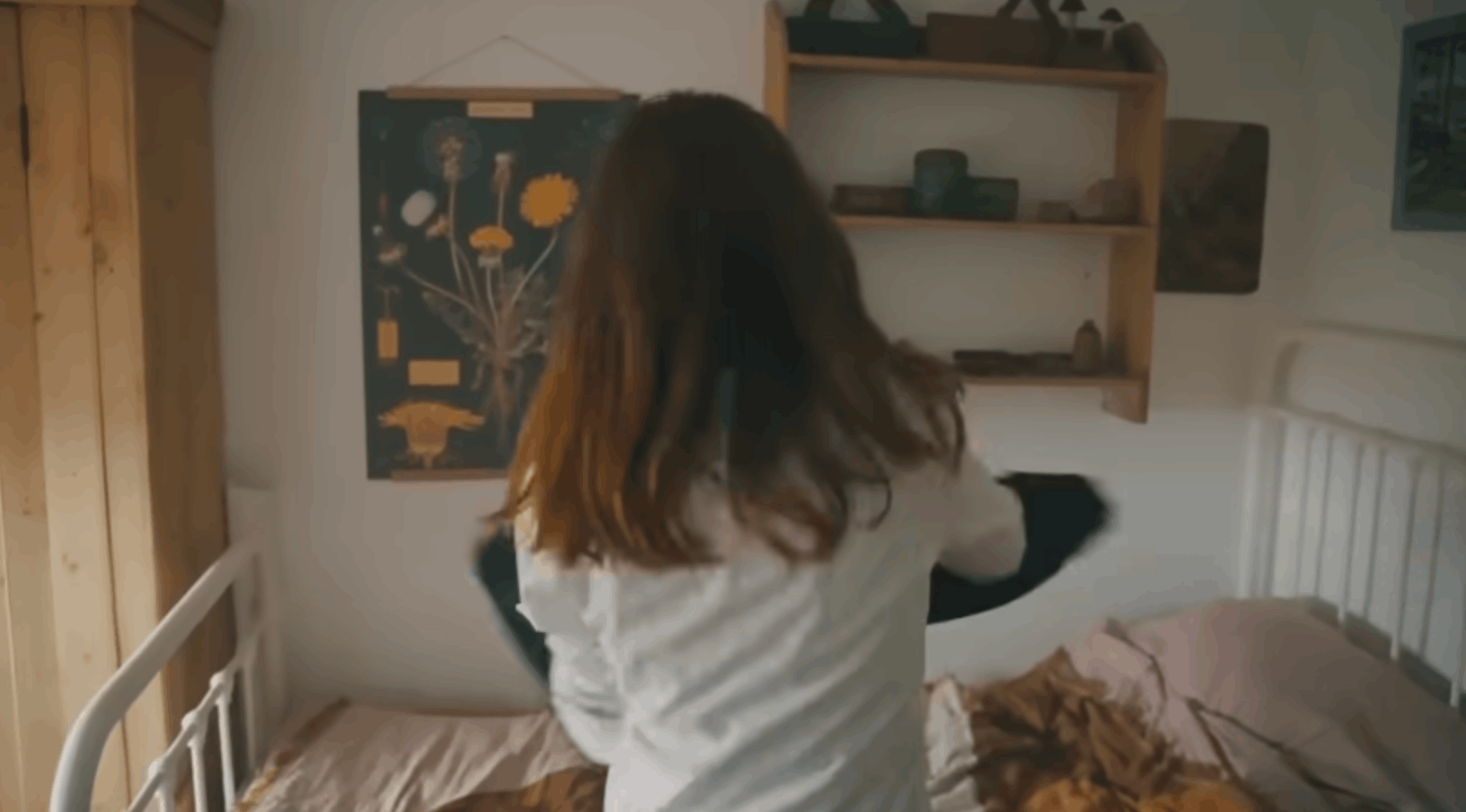}\quad
      \includegraphics[width=0.09\linewidth]{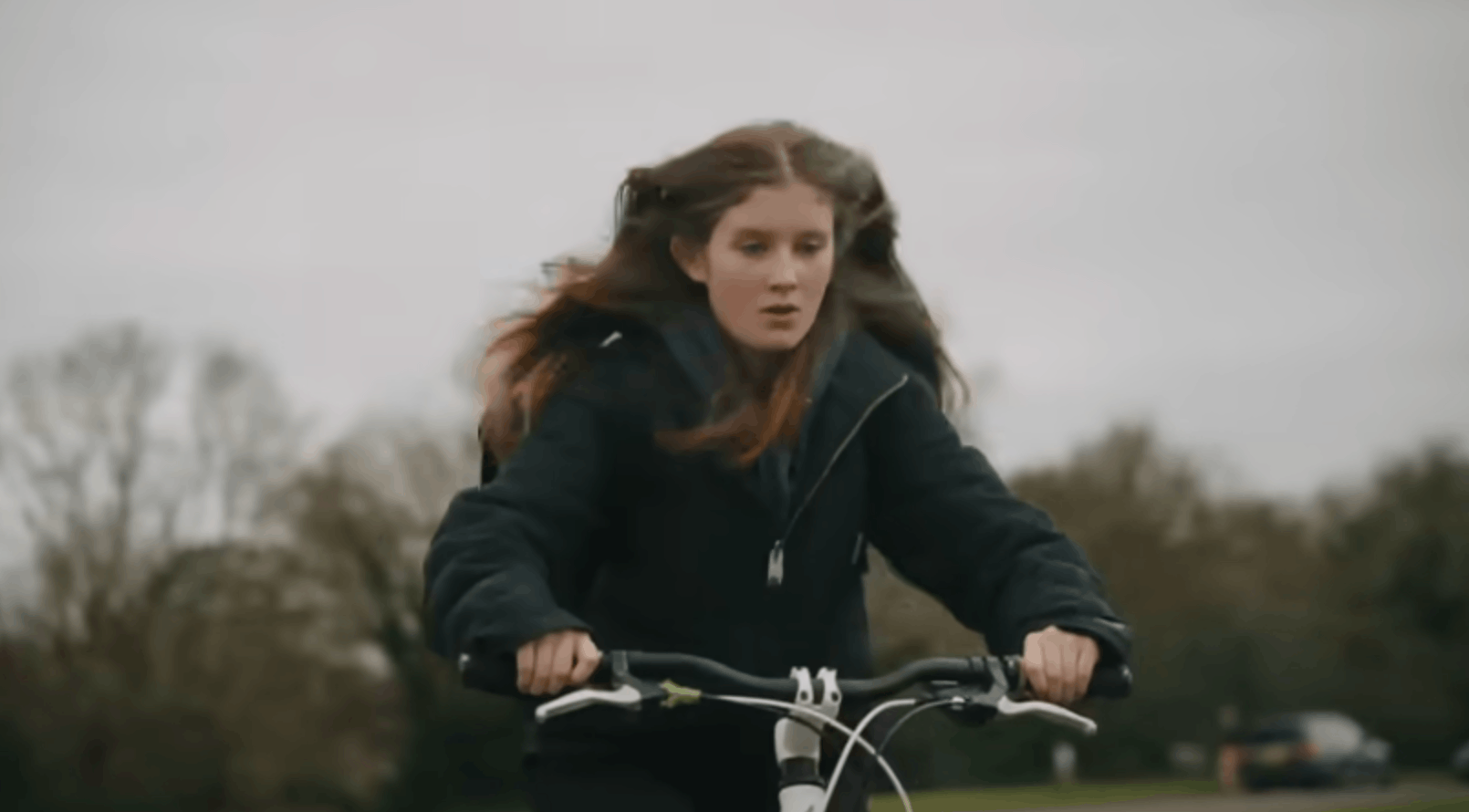}\quad
      \includegraphics[width=0.09\linewidth]{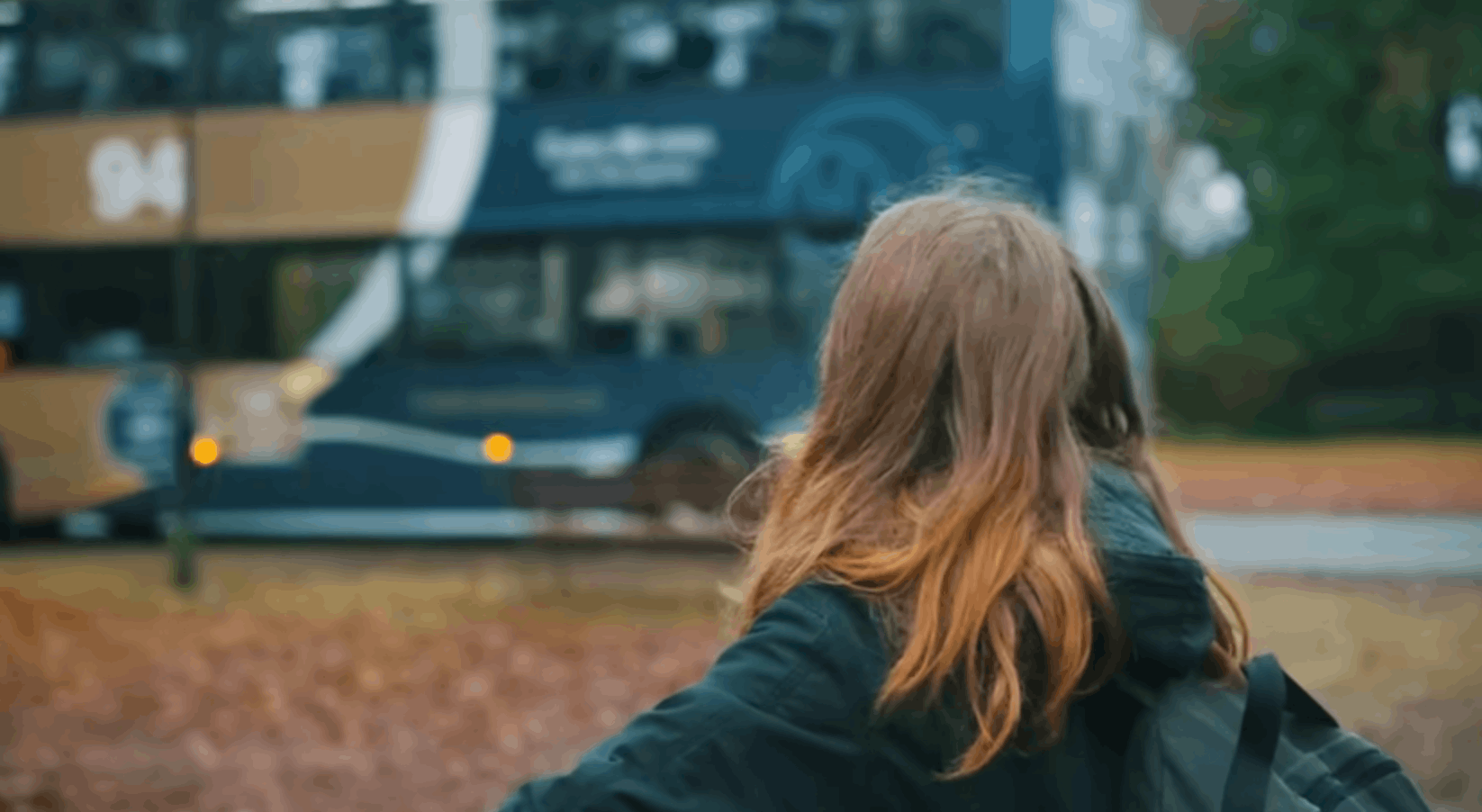}}  
      & Why did the girl rush?   
      & The girl rushed to get to the bus on time, as she was running to catch the bus  
      & The girl rushed because she was late for her exam.  
      & The girl rushed to the hospital because she was feeling unwell.  
      & The girl in the picture appears to be looking down and might be feeling sad or contemplative.  \\
    \bottomrule
  \end{tabular}}
\end{table}

To illustrate our points, in Table \ref{tab:lvlm_models}, we explicate the diversity of modern LVLMs for video understanding by examining them along various dimensions, including visual encoder, vision-language interface, LLM, and training data. Based on this examination, we observe that there exist stark differences among these models. Nevertheless, it is not straightforward to dissect which factors make an important contribution to the overall video understanding performance and which do not. 

{\renewcommand{\arraystretch}{1.2}
\begin{table}[t]
\caption{Existing LVLM models exhibit stark distinctions among themselves, making it challenging to reproduce, analyze, and compare these methods. Therefore, we aim to answer the question: ``\textit{Is there a straightforward recipe to build temporal understanding capacity for LVLMs?}''}
\vspace{5pt}
\label{tab:lvlm_models}
\centering
\resizebox{\linewidth}{!}{
\begin{tabular}{l|c|l|c|c|c}
\toprule
\rowcolor{HeaderBlue}
\multicolumn{1}{c|}{\textbf{Method}} & \textbf{Visual Encoder} & \multicolumn{1}{c|}{\textbf{Vision-Language Interface}} & \textbf{LLM Size} & \textbf{Training Data Size (Pretraining)} & \textbf{Training Data Size (IFT)} \\ 
\midrule
VALLEY \citep{luo2023valley}         & ViT-L        & Transformer + Mean-pooling                       & 13B      & 702K                 & 73K  \\
\rowcolor{GrayRow}
Video-LLaMA \citep{zhang2023video}   & CLIP-G       & Q-Former                                         & 7B       & 3M                   & 18K  \\
LLaMA-VID \citep{li2024llama}        & CLIP-G       & Linear Projection                                & 13B      & 790K                 & 763K \\
\rowcolor{GrayRow}
VideoChat \citep{li2023videochat}    & CLIP-G       & Q-Former                                         & 7B       & 25M                  & 18K  \\
VideoChat2 \citep{li2024mvbench}     & UMT-L        & Q-Former                                         & 7B       & 25M                  & 2M   \\
\rowcolor{GrayRow}
Video-ChatGPT \citep{maaz2023video}  & ViT-L        & Mean-pooling + Linear Projection                  & 7B       & 595K                 & 100K \\
Video-LLaVA \citep{lin2023video}     & ViT-L        & Linear Projection                                & 7B       & 1.3M                 & 765K \\
\rowcolor{GrayRow}
GPT4Video \citep{wang2024gpt4video}  & ViT-L        & Q-Former                                         & 7B       & 11K                  & 50K  \\
PLLaVA \citep{xu2024pllava}          & ViT-L        & Linear Projection + Adaptive Pooling             & 7B/13B/34B & 25M              & 783K \\
\rowcolor{GrayRow}
ST-LLM \citep{liu2024st}             & BLIP-2       & Linear Projection                                & 7B       & 25M                  & 2M   \\
Chat-UniVi \citep{jin2024chat}       & ViT-L        & Clustering-based Merging + Linear Projection      & 7B/13B   & 1.6M                 & 649K \\
\bottomrule
\end{tabular}}
\end{table}}

As one of the works that initiates empirical analysis research line, METER \citep{dou2022empirical} studies a wide variety of components in the context of image-language modeling. Unfortunately, its analysis mostly works on images and neglects many aspects related to video modeling, such as spatio-temporal architectural design, video pretraining data, and video pretraining objectives. To fill in such gap, recent VindLU work \citep{cheng2023vindlu} conducts an analysis towards important factors for video-language understanding models. Unfortunately, their analysis is limited to small-scale frameworks with millions of parameters. Similarly, \citet{fu2023empirical} performs an empirical study of video-language transformers, but narrowly concentrates on masked visual modeling objective. 

Our primary objective in this work is to answer the question ``\textit{Is there a straightforward recipe to build temporal understanding capacity for LVLMs?}'' Our answer is \textbf{yes}. To arrive at the answer, we conduct a thorough empirical study that demystifies the importance of various design choices and ultimately leads to a temporal-oriented recipe that significantly enhances video understanding results of previous LVLMs. Our recipe starts from a standard paradigm of a large vision-language model then proceeds with a progressive expansion scheme, where at each stage, we investigate a specific aspect of LVLM framework design (\textit{e.g.}, architecture, training objective, training data, etc.) and choose the most effective option. Particularly, we study the following LVLM design components: (i) the vision-language interface, (ii) the video training protocols, and (iii) temporal memory bank, and (iv) scaling of the essential component. We present our recipe in Figure \ref{fig:recipe}. 

The key lessons of our study include:
\begin{itemize}
    \item Among components in an LVLM architecture, we discover that enhancing vision-language interface significantly advances the temporal modeling strength of the LVLM. 
    \item A query transformer that incorporates query tokens to interact with video representations combined with a temporal memory bank to compress salient video information is crucial for satisfactory video understanding performance.
    \item We can further obtain gains of temporal understanding level with techniques to scale up the interface, including mixture-of-experts and number of query tokens to store video information.
    \item An additional training stage for LVLMs with temporal-oriented data is sufficient to remarkably enhance temporal understanding capability and achieve impressive results on video understanding.
\end{itemize}

\section{Temporal-Oriented Recipe for Large Vision-Language Model}
\label{sect:recipe}
\noindent In this section, we delineate our temporal-oriented recipe for large vision-language model. We start with a standard large vision-language model (LVLM), which consists of a visual encoder such as ViT and a large language model (LLM). Then, we progressively expand it to a model that achieves impressive temporal understanding results on various video understanding datasets and tasks. At each step of our recipe, we investigate how design choices have an impact upon temporal capacity of the LVLM. Throughout our procedure, we will discover answers to the following questions about the temporal-oriented recipe design:
\begin{itemize}
    \item Can explicitly constructing temporal understanding capacity help LVLM, particularly provided that various video understanding benchmarks are spatially biased \citep{buch2022revisiting, lei2022revealing}? If so, what is the best mechanism for LVLM to conduct temporal modeling?
    \item Given that video lengths vary with a wide range, what is the most productive mechanism for LVLM to read/absorb video information? Several approaches use visual encoder combined with linear projection \citep{maaz2023video, xu2024pllava} or query transformer (Q-Former) \citep{li2023videochat, li2024mvbench}, then proceed with a pooling mechanism, whereas others adopt a memory bank \citep{he2024ma}. Which of these is the most effective?
    \item Which temporal-oriented training schemes are most useful for temporal representation learning? There exist a wide variety of schemes, including video captioning \citep{abdar2024review}, moment captioning \citep{qasim2025dense, yang2023vid2seq}, moment grounding \citep{nguyen2023demaformer, lei2021detecting}, and video summarization \citep{apostolidis2021video, nguyen2023read}. How significant is each of these schemes? Are they complementary to each other?
    \item How can we optimize temporal capacity of LVLM? Can we inherit the mixture-of-experts (MoE) approach from LLM works, or increase the number of query tokens?
\end{itemize}

\subsection*{Step 0: Starting Ingredients}
\noindent\textbf{Large Vision-Language Model for Video Understanding.} We start with a standard ViT-G/14 \citep{dosovitskiy2020image} from EVA-CLIP \citep{fang2023eva}. For LLM, we use either Vicuna-7B or Vicuna-13B \citep{chiang2023vicuna}, forming either a 7B-LVLM or a 13B-LVLM, respectively. 

Formally, given a paired video and text prompt $(V, T)$, we first sample a sequence of frames from the video. Each frame is uniformly divided into equal-sized patches, which are then encoded by a visual encoder to obtain patch-level visual features. A positional embedding layer (PE) is applied to incorporate temporal ordering across frames, producing frame-level spatial-temporal representations. These visual features are then passed through a vision-language interface module, which projects them into the language representation space of the LLM. In parallel, the LLM embeds the text prompt $T$ into word embeddings. The projected visual embeddings and text embeddings are concatenated to form a joint input sequence, which the LLM processes to generate a response for the video-text pair. 

\noindent\textbf{Experimental Setup.} As our initialization, we directly inherit the pretrained and instruction-tuned model on image-based data \citep{liu2023visual, li2022blip, li2023blip}. Afterwards, we either conduct an additional temporal-oriented training step or go straight to finetuning and evaluating the model on the seven popular video understanding datasets: MSRVTT \citep{xu2016msr}, MSVD \citep{chen2011collecting}, ActivityNet-QA \citep{caba2015activitynet}, Breakfast \citep{kuehne2014language}, COIN \citep{tang2019coin}, and LVU \citep{wu2021towards}. For our empirical investigation, we choose the video question answering (VideoQA) task and report the accuracy across these datasets.

In the following subsections, we progressively expand this baseline by adding more components of elevating complexity. Specifically, we start by incorporating vision-language interface (step 1), integrate a temporal-oriented training stage (step 2), insert a temporal memory bank (step 3), and upscaling the interface (step 4). Note that due to the large computational cost, we cannot ablate the order of the steps in our recipe. Therefore, the order of the steps is primarily determined by the computational cost (\textit{i.e.} the steps that can be implemented most efficiently are investigated before other steps, subsequently moving to more computationally costly steps).

\subsection*{Step 1: Vision-Language Interface for LVLM}

{\renewcommand{\arraystretch}{1.2}
\begin{table}[t]
\caption{Effect of different types of vision-language interface on 7B-LVLM}
\label{tab:step1_vision_language_interface_7b}
\centering
\resizebox{0.9\linewidth}{!}{
\begin{tabular}{l|c|c|c|c|c|c}
\toprule
\rowcolor{LightBlue}
\multicolumn{1}{c|}{\textbf{Vision-Language Interface}} & \textbf{MSRVTT} & \textbf{MSVD} & \textbf{ActivityNet-QA} & \textbf{Breakfast} & \textbf{COIN} & \textbf{LVU}  \\
\midrule
Linear Projection                                      & 45.3            & 56.9          & 46.3                    & 85.9               & 83.9          & 57.1         \\
\rowcolor{Gray}
Q-Former w/o SA + Mean-pooling - S = 3                 & 45.8            & 57.4          & 46.4                    & 86.3               & 84.8          & 58.0         \\
Q-Former w/o SA + Adaptive-pooling - S = 3             & 46.2            & 57.5          & 46.8                    & 87.0               & 85.6          & 58.2         \\
\rowcolor{Gray}
Q-Former w/o SA + ESA - S = 3                          & 46.3            & 57.7          & 47.2                    & 87.6               & 86.0          & 58.7         \\
Q-Former w/o SA + Mean-pooling - S = 6                 & 46.3            & 58.1          & 47.6                    & 87.9               & 86.5          & 58.8         \\
\rowcolor{Gray}
Q-Former w/o SA + Adaptive-pooling - S = 6             & 46.6            & 58.1          & 47.6                    & 88.3               & 87.1          & 59.2         \\
Q-Former w/o SA + ESA - S = 6                          & 47.0            & 58.2          & 48.1                    & 88.4               & 87.6          & 59.8         \\
\rowcolor{Gray}
Q-Former w/o SA + Mean-pooling - S = 9                 & 47.2            & 58.2          & 48.3                    & 88.5               & 87.9          & 60.3         \\
Q-Former w/o SA + Adaptive-pooling - S = 9             & 47.4            & 58.7          & 48.6                    & 89.1               & 88.5          & 60.8         \\
\rowcolor{Gray}
Q-Former w/o SA + ESA - S = 9                          & 47.5            & 59.1          & 48.8                    & 89.9               & 88.6          & 61.2         \\
Q-Former w/o SA + Mean-pooling - S = 12                & 47.6            & 59.2          & 49.3                    & 90.1               & 89.1          & 61.8         \\
\rowcolor{Gray}
Q-Former w/o SA + Adaptive-pooling - S = 12            & 47.9            & 59.8          & 49.5                    & 90.8               & 89.4          & 62.2         \\
Q-Former w/o SA + ESA - S = 12                         & 48.2            & 59.9          & 49.5                    & 90.9               & 90.0          & 62.4         \\
\rowcolor{Gray}
Q-Former w/ SA - S = 3                                 & 48.7            & 60.2          & 49.8                    & 91.7               & 90.8          & 62.7         \\
Q-Former w/ SA - S = 6                                 & 48.8            & 60.4          & 49.8                    & 91.9               & 91.6          & 52.0         \\
\rowcolor{Gray}
Q-Former w/ SA - S = 9                                 & 49.0            & 60.4          & 50.3                    & 92.1               & 92.5          & 63.5         \\
Q-Former w/ SA - S = 12                                & 49.3            & 60.4          & 50.3                    & 92.4               & 92.7          & 63.7         \\
\rowcolor{Best}
Pre-trained Q-Former w/ SA - S = 12                    & \textbf{49.4}            & \textbf{60.8}          & \textbf{50.6}                    & \textbf{93.1}               & \textbf{93.4}          & \textbf{63.8}    \\
\bottomrule
\end{tabular}}
\end{table}}

{\renewcommand{\arraystretch}{1.2}
\begin{table}[t]
\caption{Effect of different types of vision-language interface on 13B-LVLM}
\label{tab:step1_vision_language_interface_13b}
\centering
\resizebox{0.9\linewidth}{!}{
\begin{tabular}{l|c|c|c|c|c|c}
\toprule
\rowcolor{LightBlue}
\multicolumn{1}{c|}{\textbf{Vision-Language Interface}} & \textbf{MSRVTT} & \textbf{MSVD} & \textbf{ActivityNet-QA} & \textbf{Breakfast} & \textbf{COIN} & \textbf{LVU}  \\
\midrule
Linear Projection                                      & 56.2            & 67.9          & 48.7                    & 86.5               & 84.5          & 65.2         \\
\rowcolor{Gray}
Pre-trained Q-Former w/ SA - S = 12                    & 56.8            & 68.2          & 48.7                    & 86.6               & 85.1          & 67.2         \\
Q-Former w/o SA + Mean-pooling - S = 3                 & 56.9            & 68.3          & 49.0                    & 87.2               & 85.2          & 67.3         \\
\rowcolor{Gray}
Q-Former w/o SA + Adaptive-pooling - S = 3             & 57.1            & 68.7          & 49.3                    & 87.7               & 86.0          & 67.6         \\
Q-Former w/o SA + ESA - S = 3                          & 57.5            & 69.3          & 49.5                    & 88.0               & 86.0          & 68.1         \\
\rowcolor{Gray}
Q-Former w/o SA + Mean-pooling - S = 6                 & 57.8            & 69.7          & 49.5                    & 88.4               & 86.4          & 68.1         \\
Q-Former w/o SA + Adaptive-pooling - S = 6             & 58.0            & 69.8          & 49.8                    & 88.7               & 87.3          & 68.5         \\
\rowcolor{Gray}
Q-Former w/o SA + ESA - S = 6                          & 58.1            & 70.4          & 50.0                    & 88.8               & 87.8          & 69.0         \\
Q-Former w/o SA + Mean-pooling - S = 9                 & 58.2            & 71.0          & 50.2                    & 89.3               & 88.2          & 69.1         \\
\rowcolor{Gray}
Q-Former w/o SA + Adaptive-pooling - S = 9             & 58.3            & 71.2          & 50.3                    & 89.4               & 88.9          & 69.4         \\
Q-Former w/o SA + ESA - S = 9                          & 58.7            & 72.8          & 50.4                    & 90.2               & 89.5          & 69.6         \\
\rowcolor{Gray}
Q-Former w/o SA + Mean-pooling - S = 12                & 59.1            & 72.2          & 50.5                    & 90.6               & 90.2          & 69.8         \\
Q-Former w/o SA + Adaptive-pooling - S = 12            & 59.2            & 72.3          & 50.6                    & 91.3               & 90.7          & 70.1         \\
\rowcolor{Gray}
Q-Former w/o SA + ESA - S = 12                         & 59.5            & 72.5          & 51.1                    & 92.1               & 90.9          & 70.3         \\
Q-Former w/ SA - S = 3                                 & 59.8            & 72.5          & 51.5                    & 92.4               & 91.1          & 71.0         \\
\rowcolor{Gray}
Q-Former w/ SA - S = 6                                 & 59.9            & 73.1          & 51.7                    & 92.4               & 91.5          & 71.2         \\
Q-Former w/ SA - S = 9                                 & 60.3            & 73.8          & 51.8                    & 92.9               & 92.0          & 71.5         \\
\rowcolor{Gray}
Q-Former w/ SA - S = 12                                & 60.5            & 74.3          & 52.0                    & 93.7               & 92.4          & 71.5         \\
\rowcolor{Best}
Pre-trained Q-Former w/o SA - S = 12                   & \textbf{60.6}            & \textbf{74.3}          & \textbf{52.5}                    & \textbf{93.7}               & \textbf{93.2}          & \textbf{71.8}     \\
\bottomrule
\end{tabular}}
\end{table}}

\noindent In the first stage of our temporal-oriented recipe, we investigate the interface between the vision and language domain for our LVLM. Such interface will enable the LLM to have access to visual information from the video input. For compactness, we study three interface schemes:

\begin{itemize}
    \item \textbf{Linear projection:} In this interface, the linear projection maps visual embeddings into appropriate dimensional space for the LLM. Due to its simplicity, this approach has been widely adopted by previous LLaVA-based LVLMs \citep{xu2024pllava, lin2023video}.

    \item \textbf{Query Transformer with Self-Attention (Q-Former w/ SA):} Following \citep{zhang2023video, he2024ma}, we use a number of transformer submodules which consist of cross-attention and self-attention layers. Cross-attention layers will enable a set of learnable query embeddings to interact with video representations to extract video information. In this variant, our Q-Former also contains self-attention layers, which can perform temporal modeling since they relate video frames together. We vary the number of submodules $S \in \{3,6,9,12\}$. Parameters of Q-Former can be either randomly initialized or initialized from a pre-trained model. In our work, if we initialize Q-Former from a pre-trained model, we follow MA-LMM \citep{he2024ma} to use the \textit{bert-base-uncased} with $S = 12$ submodules.

    \item \textbf{Query Transformer without Self-Attention (Q-Former w/o SA):} This version is similar to the previous one, except the fact that Q-Former does not comprise self-attention layers. Therefore, we need to incorporate an additional component after Q-Former for temporal modeling. We experiment with possible choices, including mean-pooling, adaptive pooling, and external self-attention (ESA) layers. 
\end{itemize}

As Table \ref{tab:step1_vision_language_interface_7b} and \ref{tab:step1_vision_language_interface_13b} show, Q-Former demonstrates critical performance improvement over the linear projection approach. The improvement is indicated by average +6.0\% and +6.2\% accuracy boost of our 12-layer Q-Former variant over the 7B and 13B linear-projection baseline, respectively. We also observe that initializing Q-Former self-attention layers with pretrained BERT encoder makes a significant contribution to the performance boost. This suggests that temporal semantics among words can be related to temporal relations among video frames. 

Interestingly, our findings contradict the conclusions of several prior studies \citep{liu2023visual, koh2023grounding}, which suggest that a simple linear projection is sufficient---and even more effective---than the Q-Former approach. In contrast, we observe that Q-Former plays a crucial role due to its ability to model diverse temporal relations across a broad range of video scenarios. We hypothesize that, particularly for temporally-intensive datasets, the integration of stacked cross- and self-attention layers provides the necessary capacity to capture and reason about complex temporal dependencies across video frames.

\textit{Takeaway 1: For all subsequent experiments, we use 12-layer pretrained Q-Former w/ SA as our vision-language interface for video understanding with large vision-language model (LVLM).}

\subsection*{Step 2: Temporal-Oriented Training Schemes}

\noindent Existing methods \citep{zhang2023video, li2024llama, lin2023video} typically follow a pipeline of pretraining and instruction-tuning, followed by downstream finetuning. In our work, we investigate whether introducing an additional training stage specifically aimed at enhancing temporal understanding can further improve the video comprehension capabilities of LVLMs. To this end, we explore several temporal-oriented training strategies, which are illustrated in Table \ref{tab:step2_temporal_oriented_training_scheme_examples}.
\begin{itemize}
    \item \textbf{Video Captioning (VC):} VC scheme aims to generate compact content of the video by leveraging the encoded information from the video. This objective resembles the next token prediction scheme to pretrain text-only LLM. To implement this objective, we provide the LVLM with a video input and the prompt ``\textit{what does the video describe?}’’, then train it to generate the groundtruth caption. For training data, we utilize 661K video-text pairs from 10M samples of the VIDAL-10M dataset \citep{zhu2023languagebind}.
    \item \textbf{Moment Captioning (MC):} Slightly different from VC, MC aims to caption only a specified part of the video. To implement this objective, we leverage the 745K samples from the InternVid dataset \citep{wang2023internvid}, each of which consists of a query and the specific starting and ending timestamps of the related moment in the video. Based on these timestamps, we convert them to discrete frame indices, then provide the model with the prompt ``\textit{Explain what happened from frame <start> to frame <end> in the video.}''
    \item \textbf{Moment Grounding (MG):} The MG task is the reverse variant of MC. Instead of training the model to write a caption, we let it generate the indices of the start and end frame index of the moment caption. Analogous to MC, we also employ the 745K samples from the InternVid dataset \citep{wang2023internvid}.
    \item \textbf{Dense Captioning (DC):} This task is the more complete and fine-grained version of MC and VC, respectively. In particular, we ask the LVLM ``\textit{Can you give me a breakdown of the occurrences at different timestamps in the video?}''. As Table \ref{tab:step2_temporal_oriented_training_scheme_examples} shows, the model is expected to describe a list of moments with the respective frame indices related to the moment. 
\end{itemize}

{\renewcommand{\arraystretch}{1.2}
\begin{table}[t]
\centering
\caption{Effects of Temporal-Oriented Training Schemes on 7B-LVLM}
\label{tab:step2_temporal_oriented_training_7b}
\resizebox{0.75\linewidth}{!}{
\begin{tabular}{l|c|c|c|c|c|c}
\toprule
\rowcolor{HeaderBlue}
\multicolumn{1}{c|}{\textbf{Training Scheme}} & \textbf{MSRVTT} & \textbf{MSVD} & \textbf{ActivityNet-QA} & \textbf{Breakfast} & \textbf{COIN} & \textbf{LVU} \\ 
\midrule
No temporal training        & 49.4 & 60.8 & 50.6 & 93.1 & 93.4 & 63.8 \\
VC                          & 50.3 & 61.5 & 51.0 & 93.2 & 93.7 & 64.4 \\
MC                          & 51.9 & 62.9 & 52.0 & 93.9 & 93.9 & 64.5 \\
MG                          & 50.9 & 62.3 & 51.4 & 93.3 & 93.9 & 64.6 \\
DC                          & 53.1 & 64.3 & 51.2 & 93.5 & 93.6 & 64.1 \\
\rowcolor{BestRow}
VC + MC + MG + DC           & \textbf{54.5} & \textbf{66.4} & \textbf{52.4} & \textbf{93.7} & \textbf{94.1} & \textbf{65.5} \\ 
\bottomrule
\end{tabular}}
\end{table}}

{\renewcommand{\arraystretch}{1.2}
\begin{table}[t]
\centering
\caption{Effects of Temporal-Oriented Training Schemes on 13B-LVLM}
\label{tab:step2_temporal_oriented_training_13b}
\resizebox{0.75\linewidth}{!}{
\begin{tabular}{l|c|c|c|c|c|c}
\toprule
\rowcolor{HeaderBlue}
\multicolumn{1}{c|}{\textbf{Training Scheme}} & \textbf{MSRVTT} & \textbf{MSVD} & \textbf{ActivityNet-QA} & \textbf{Breakfast} & \textbf{COIN} & \textbf{LVU} \\ 
\midrule
No temporal training        & 60.6 & 74.3 & 52.5 & 93.7 & 93.2 & 71.8 \\
VC                          & 62.0 & 74.8 & 53.9 & 94.0 & 93.6 & 72.5 \\
MC                          & 61.2 & 74.6 & 53.0 & 93.5 & 93.9 & 71.9 \\
MG                          & 61.8 & 74.6 & 53.4 & 93.6 & 93.5 & 72.1 \\
DC                          & 62.2 & 75.3 & 54.3 & 94.7 & 94.0 & 72.7 \\
\rowcolor{BestRow}
VC + MC + MG + DC           & \textbf{62.7} & \textbf{75.8} & \textbf{54.5} & \textbf{95.1} & \textbf{94.7} & \textbf{72.8} \\ 
\bottomrule
\end{tabular}}
\end{table}}

{\renewcommand{\arraystretch}{1.1}
\begin{table}[t]
\caption{Examples of temporal-oriented training schemes.}
\label{tab:step2_temporal_oriented_training_scheme_examples}
\centering
\resizebox{0.95\linewidth}{!}{
\begin{tabular}{c|c|p{0.4\linewidth}|p{0.4\linewidth}}
\hline
\rowcolor{HeaderBlue}
\textbf{Video} & \multicolumn{1}{c|}{\textbf{Training Scheme}} & \multicolumn{1}{c|}{\textbf{Example prompt}}                                                         & \multicolumn{1}{c}{\textbf{Sample Expected Output}}   \\ \hline
\multirow{2}{*}
   {\includegraphics[width=0.07\linewidth]{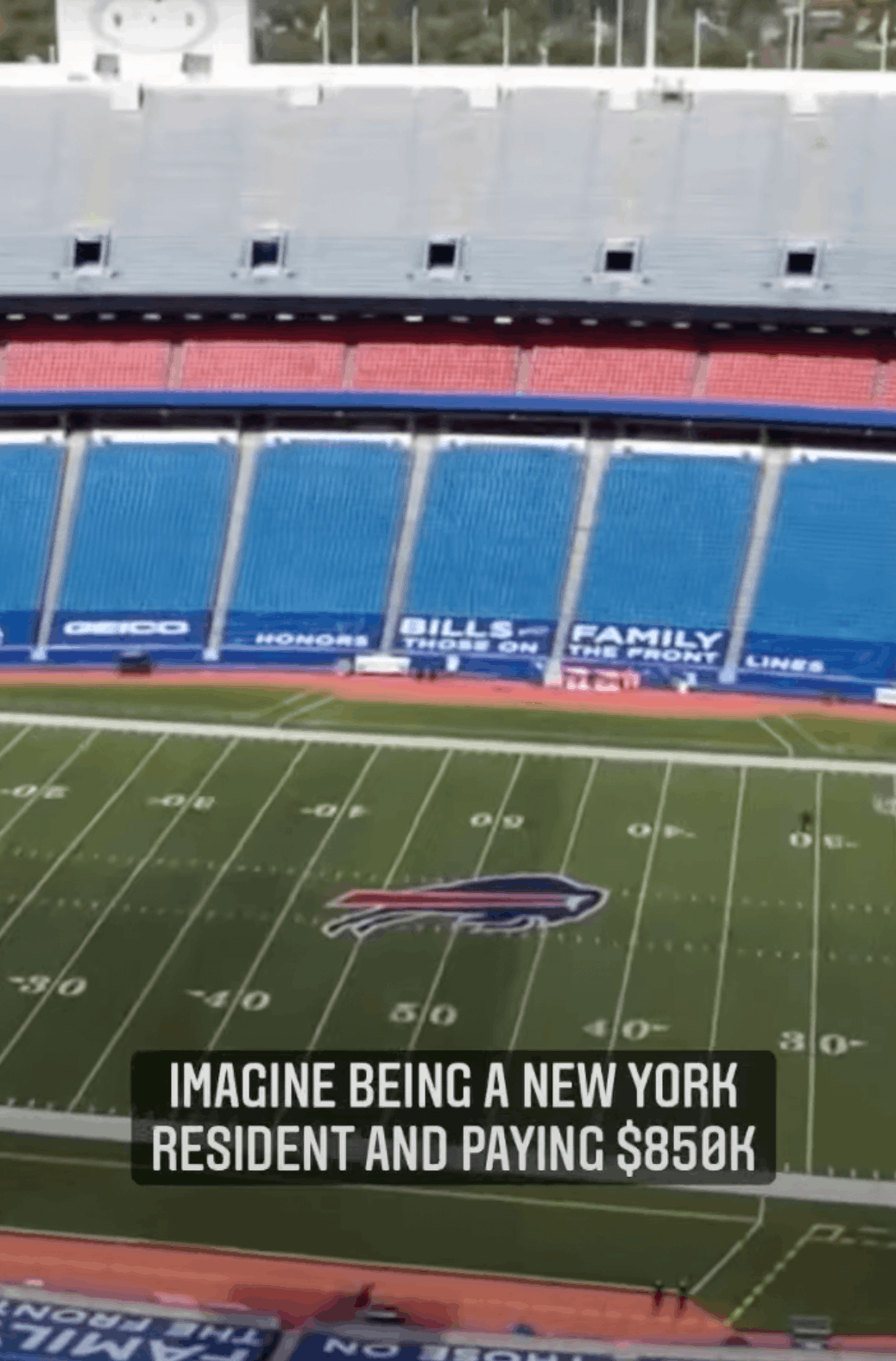} \includegraphics[width=0.07\linewidth]{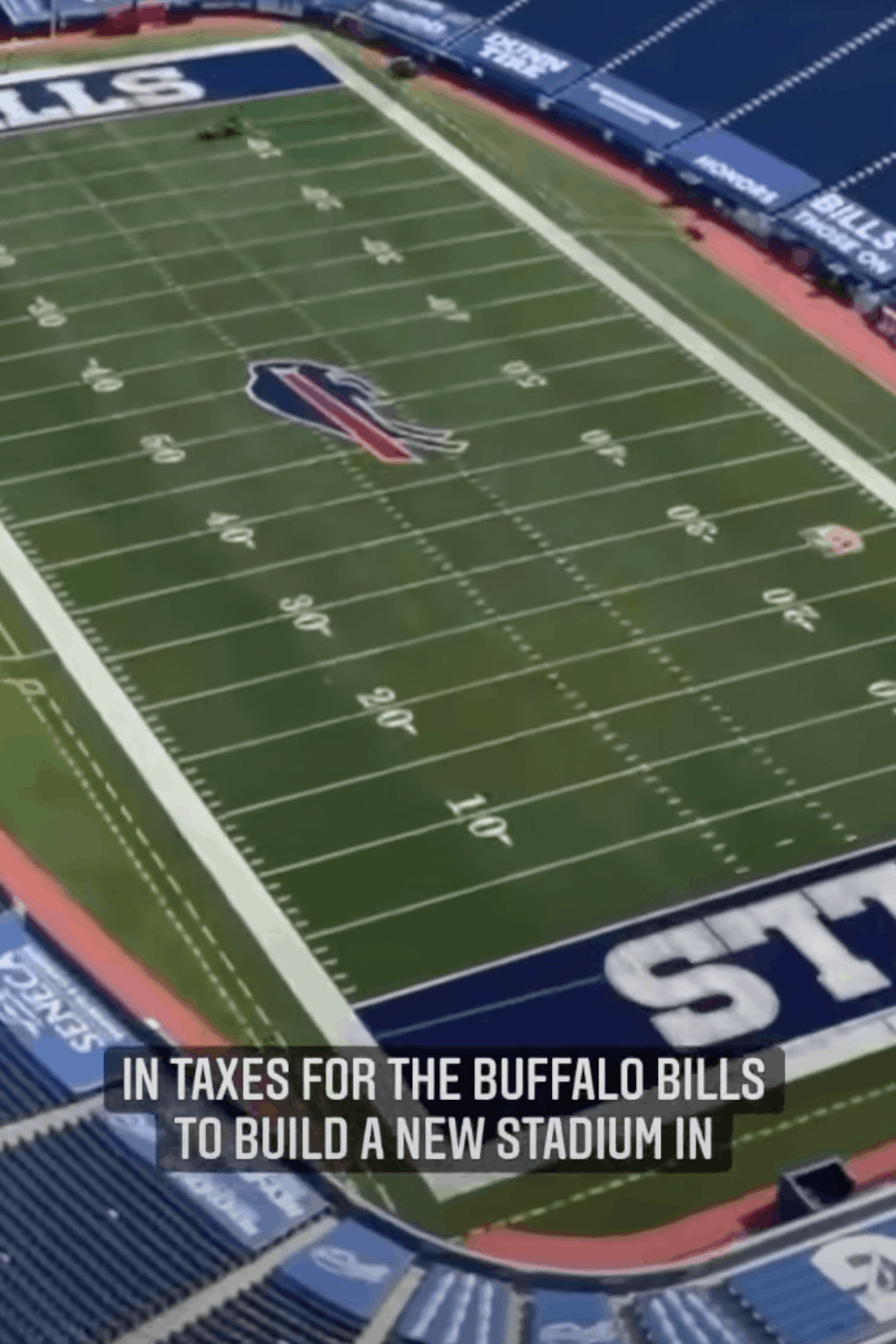}
\includegraphics[width=0.07\linewidth]{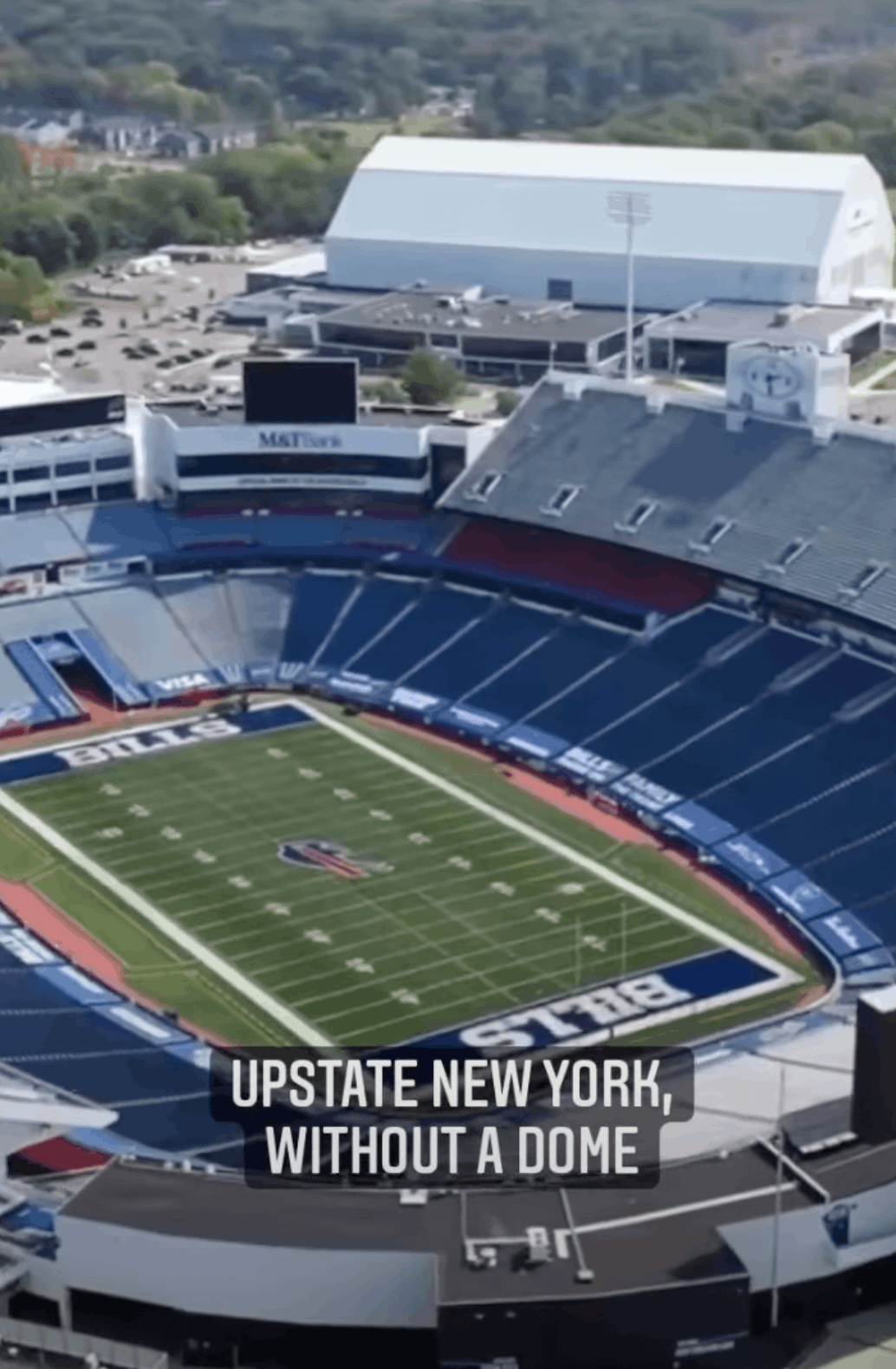} \includegraphics[width=0.07\linewidth]{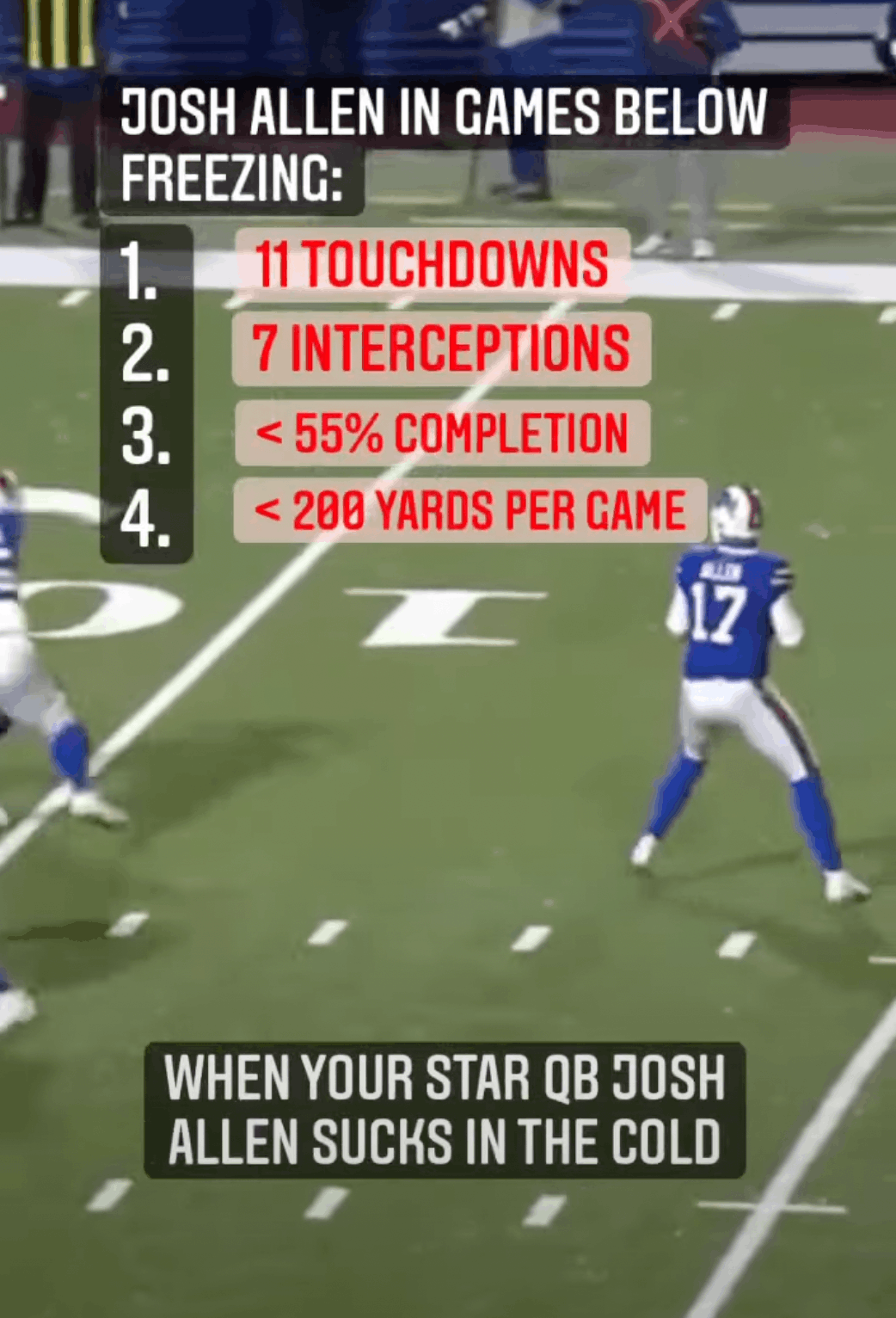}}              & VC                                           & What does the video describe?                                                                             & the buffalo bills new stadium is discussed as being deemed ineffective and not worth the investments made by the city and state.     \\ \hline
   \multirow{5}{*}
   {\includegraphics[width=0.07\linewidth]{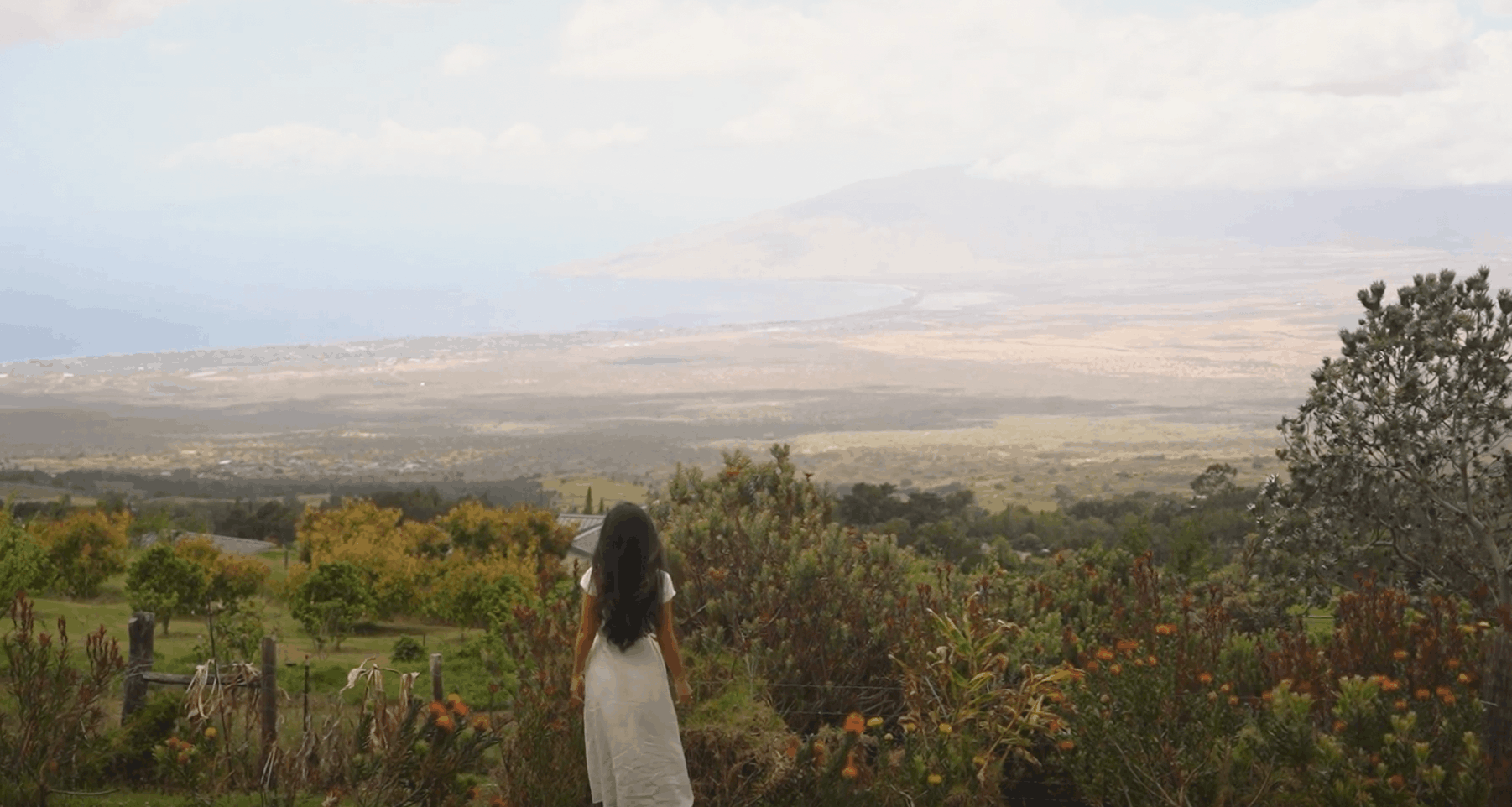} \includegraphics[width=0.07\linewidth]{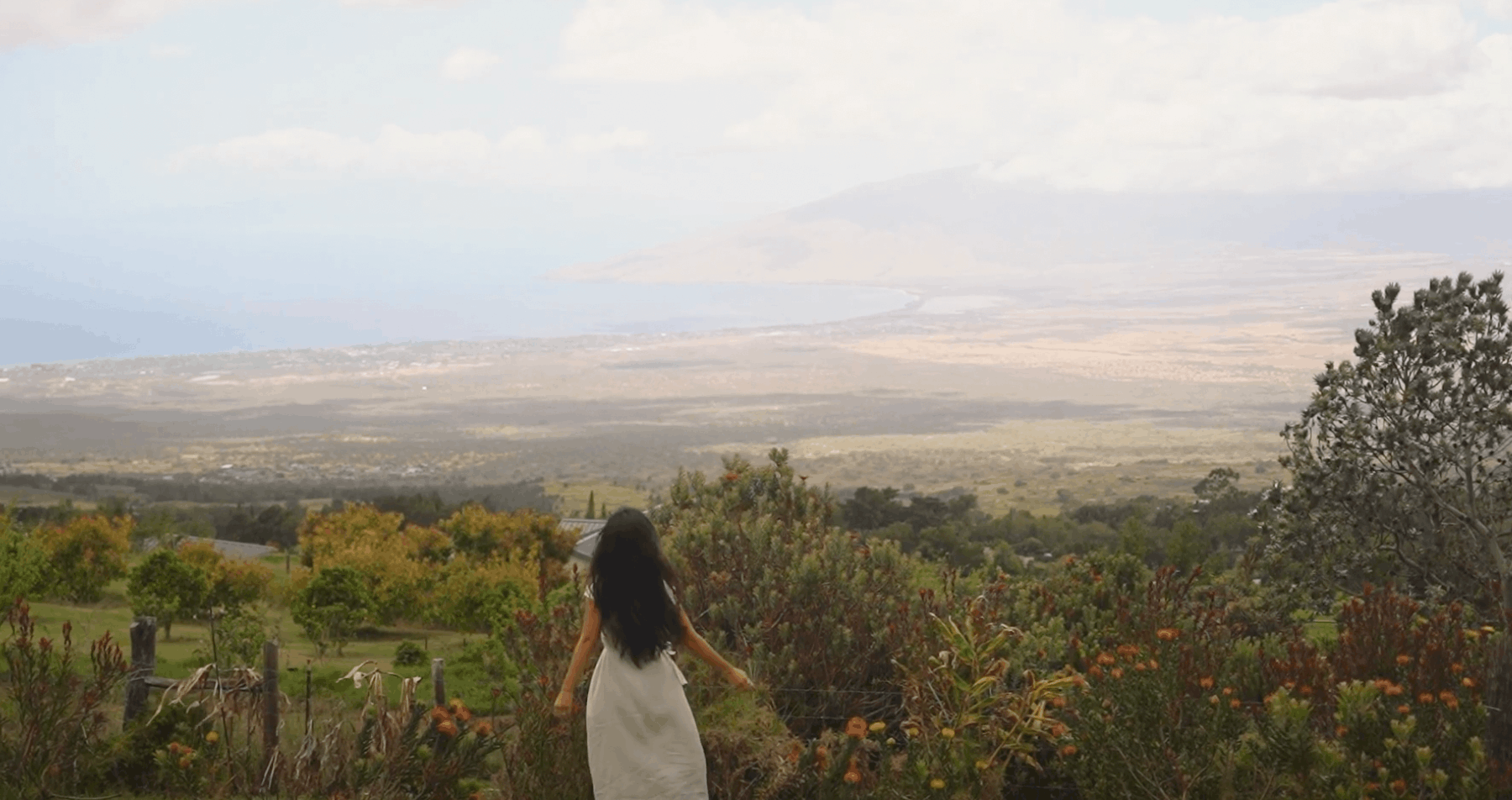}
\includegraphics[width=0.07\linewidth]{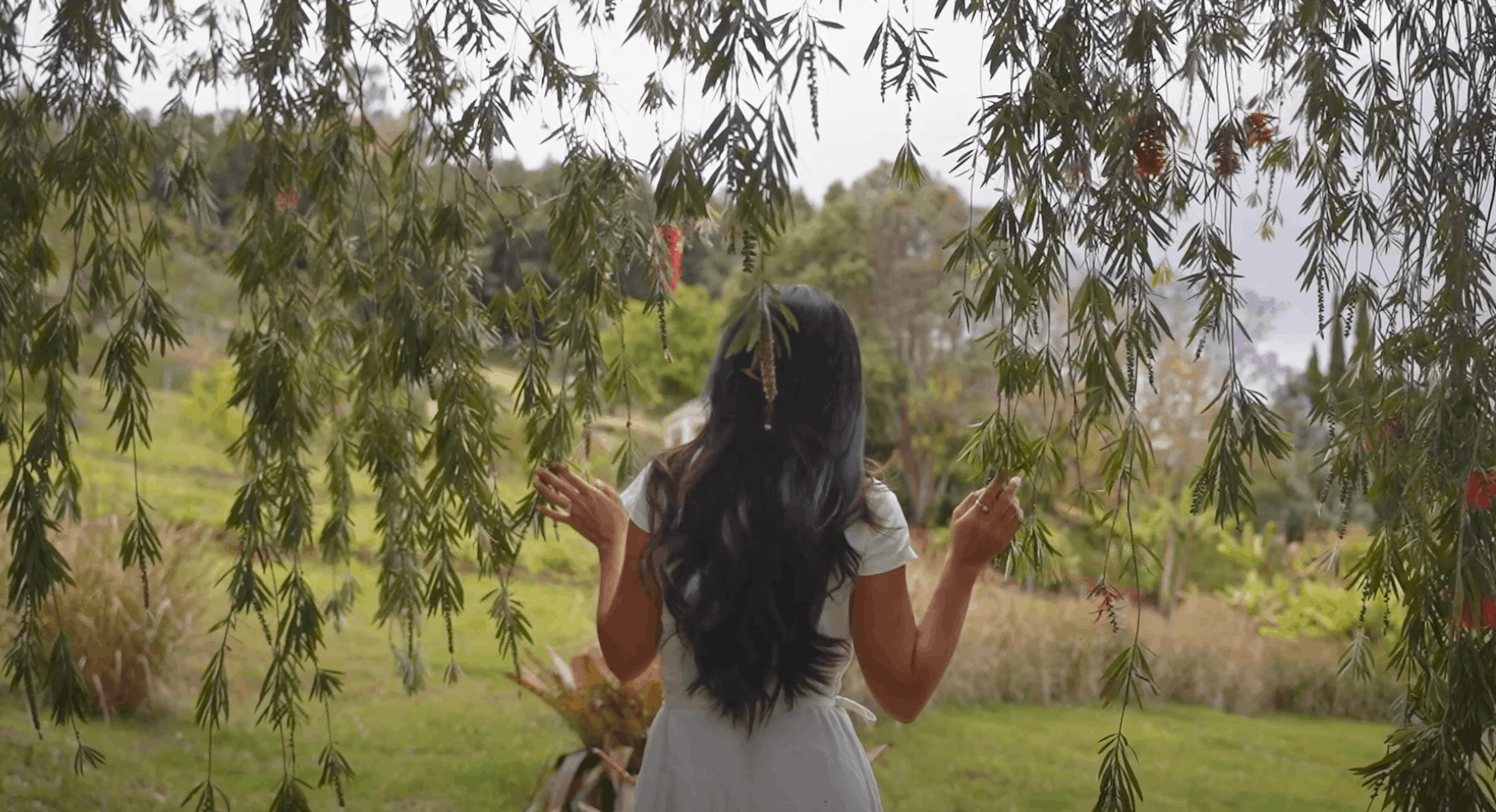} \includegraphics[width=0.07\linewidth]{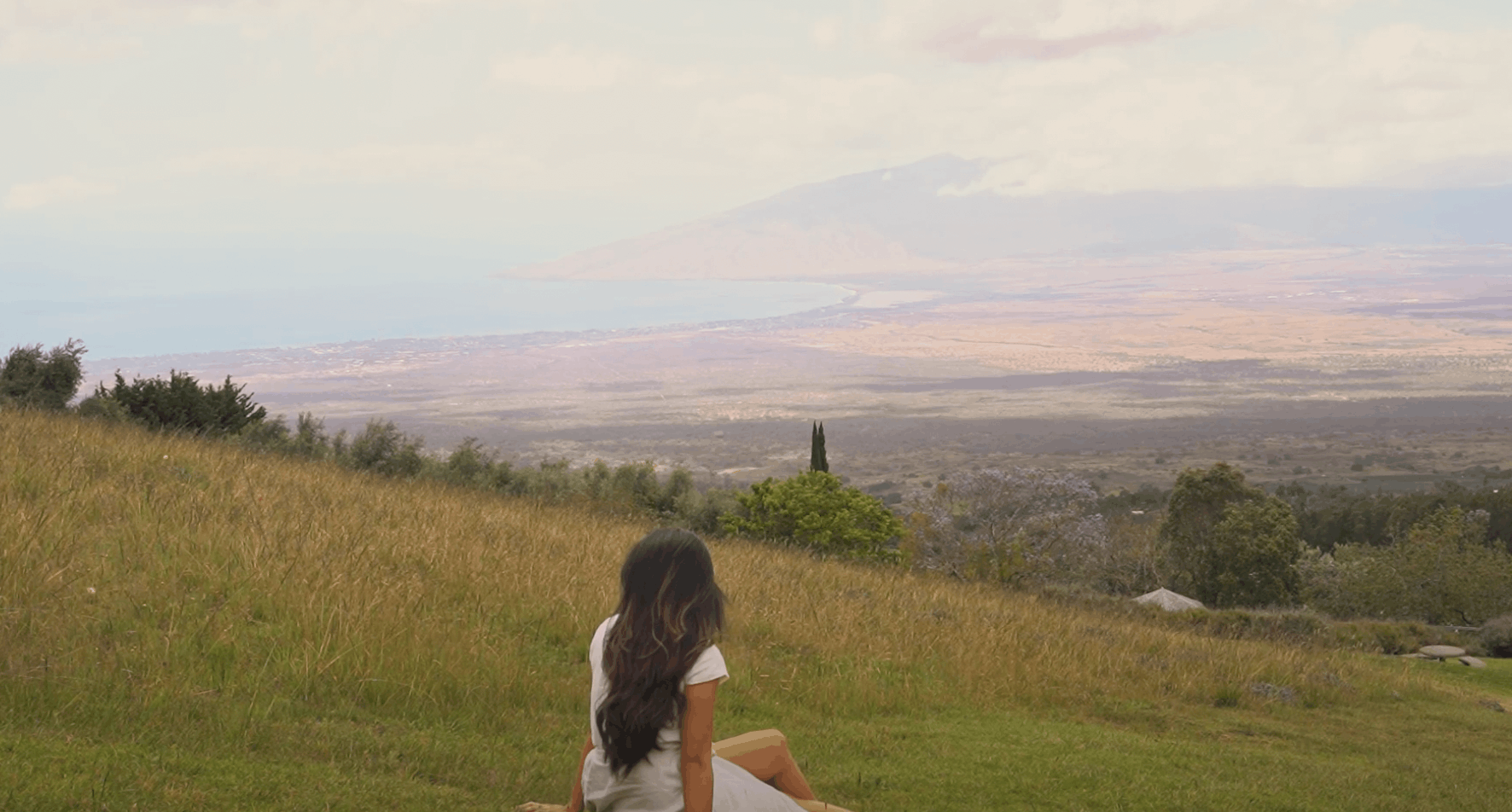}}
  & MC                                           & Explain what happened from frame 1 to frame 4 in the video.                                               & woman in long white dress walking up a hillside path.    \\ \cline{2-4}
 \multirow{5}{*} {\includegraphics[width=0.07\linewidth]{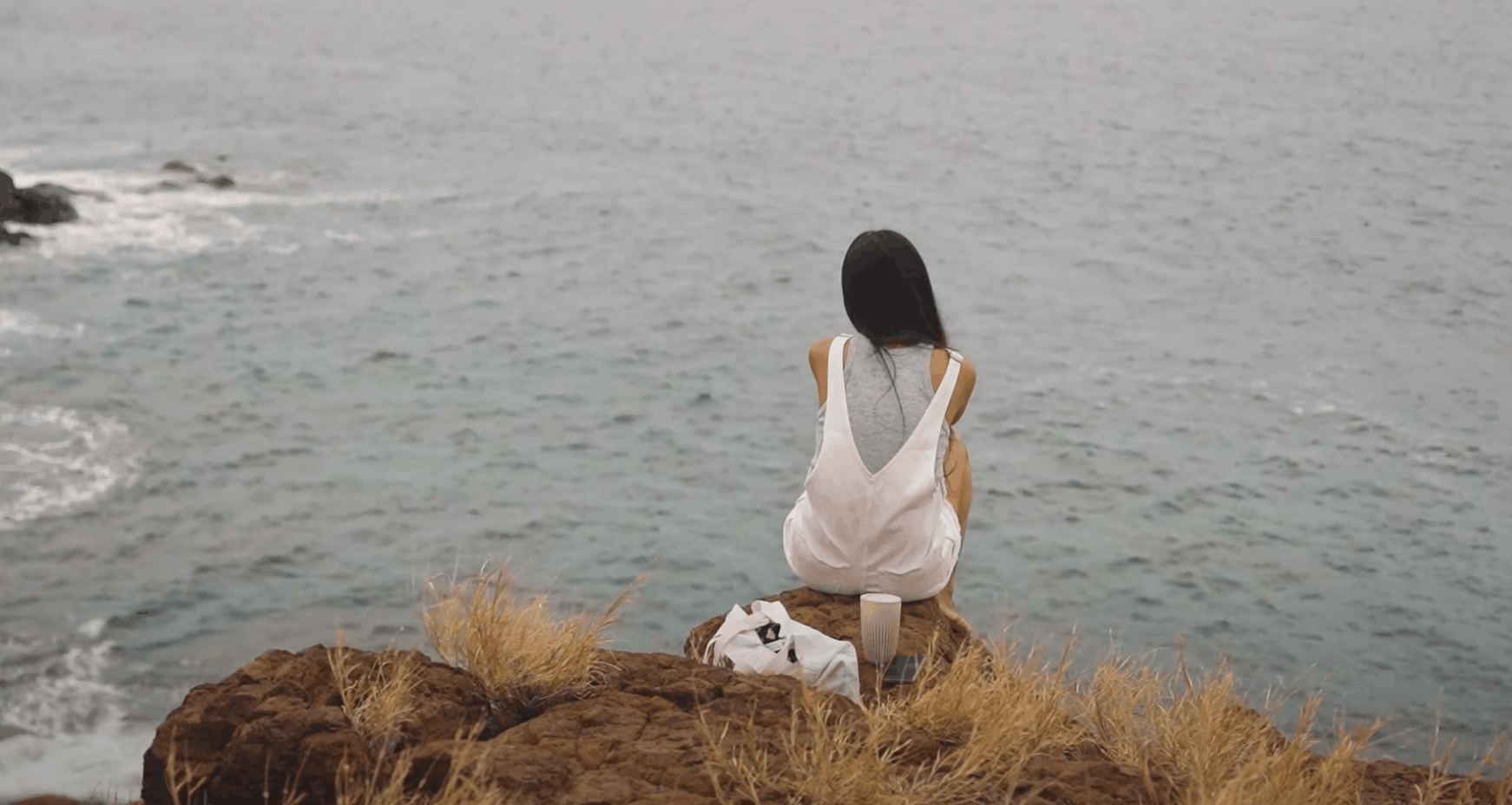} \includegraphics[width=0.07\linewidth]{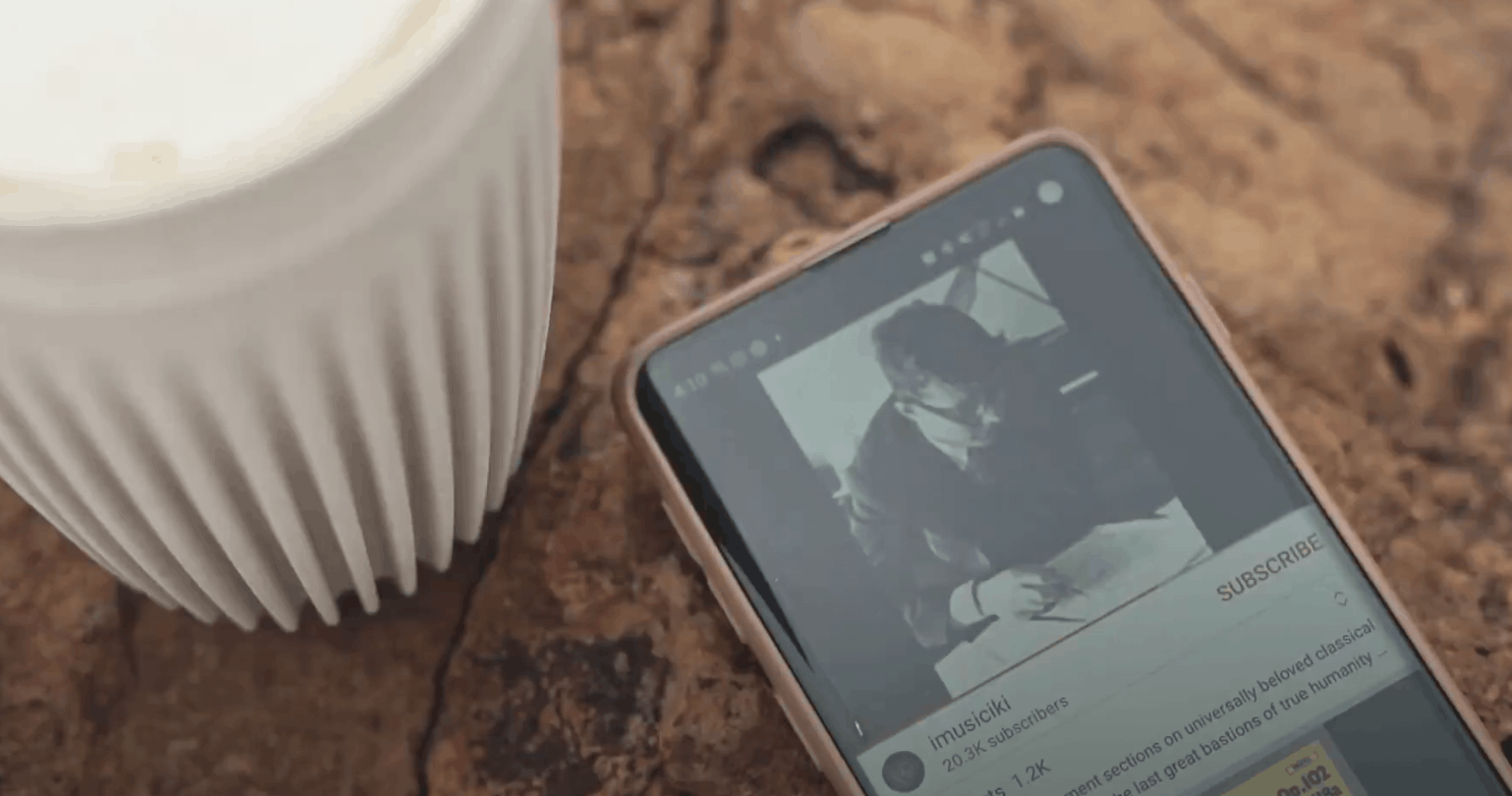}
\includegraphics[width=0.07\linewidth]{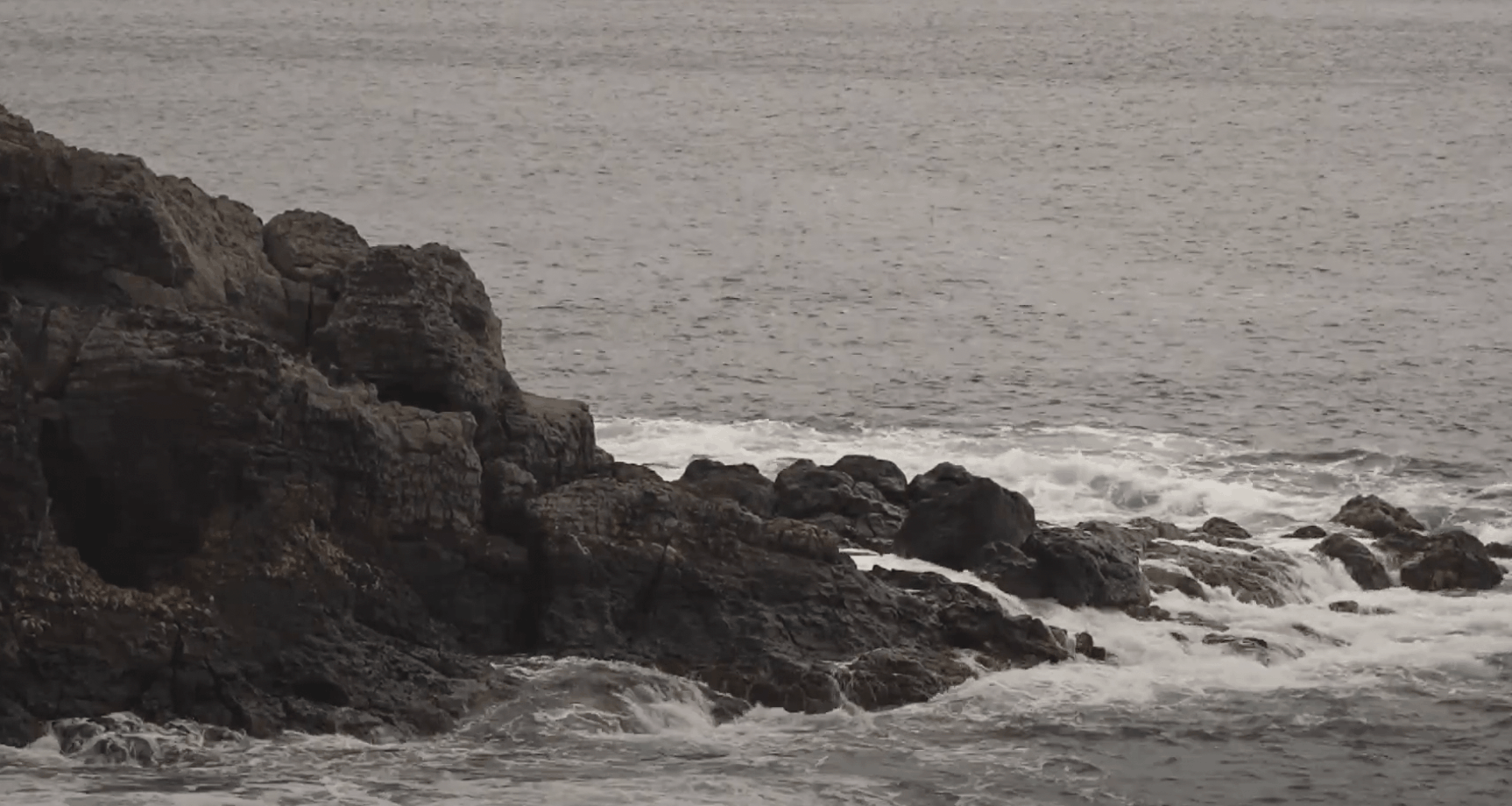} \includegraphics[width=0.07\linewidth]{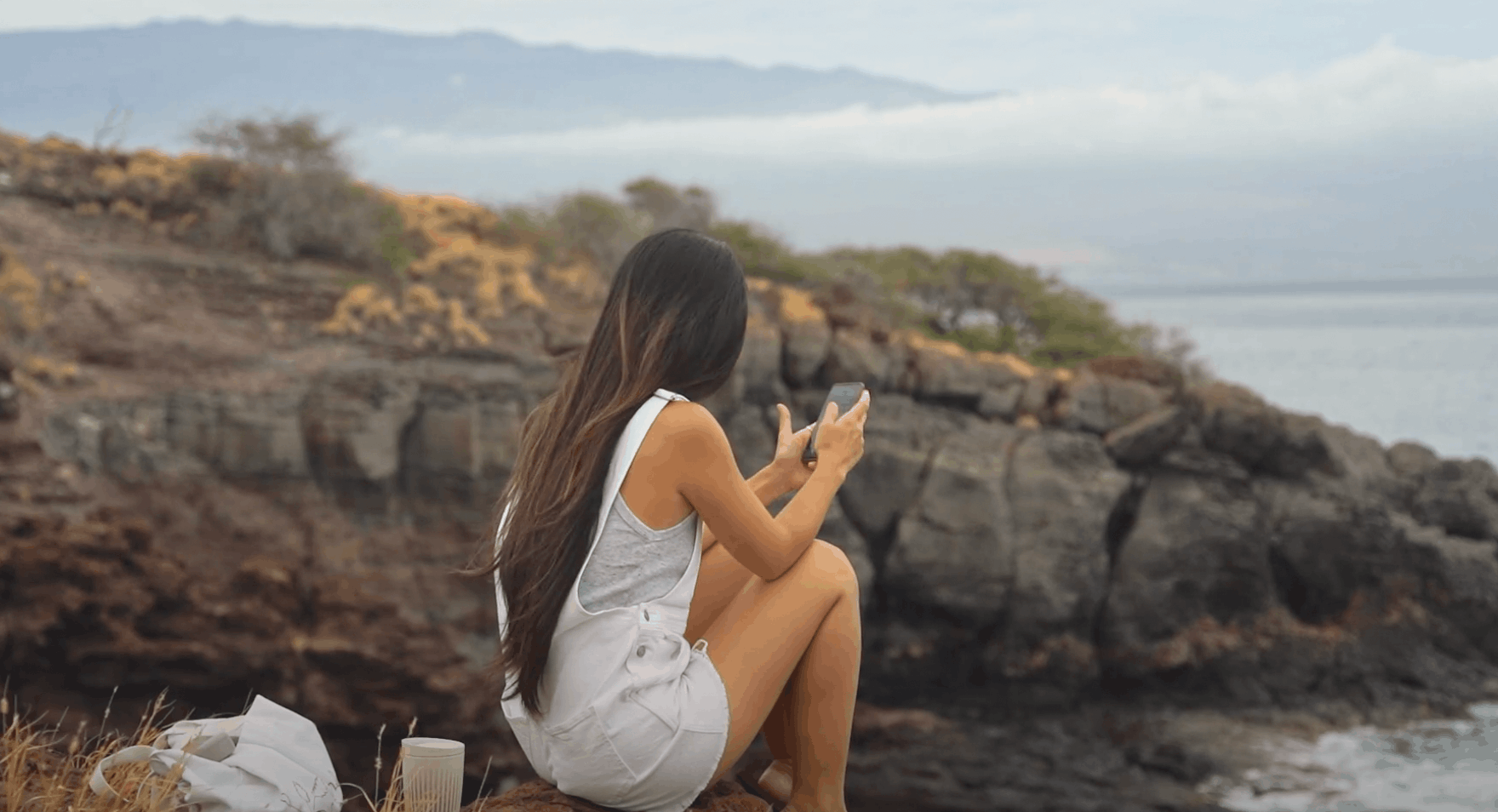}}              & MG                                           & During which frames in the video can we observe ''\textit{woman in long white dress walking up a hillside path}``? & from frame 1 to frame 4                   \\ \cline{2-4}
     \multirow{3}{*} {\includegraphics[width=0.07\linewidth]{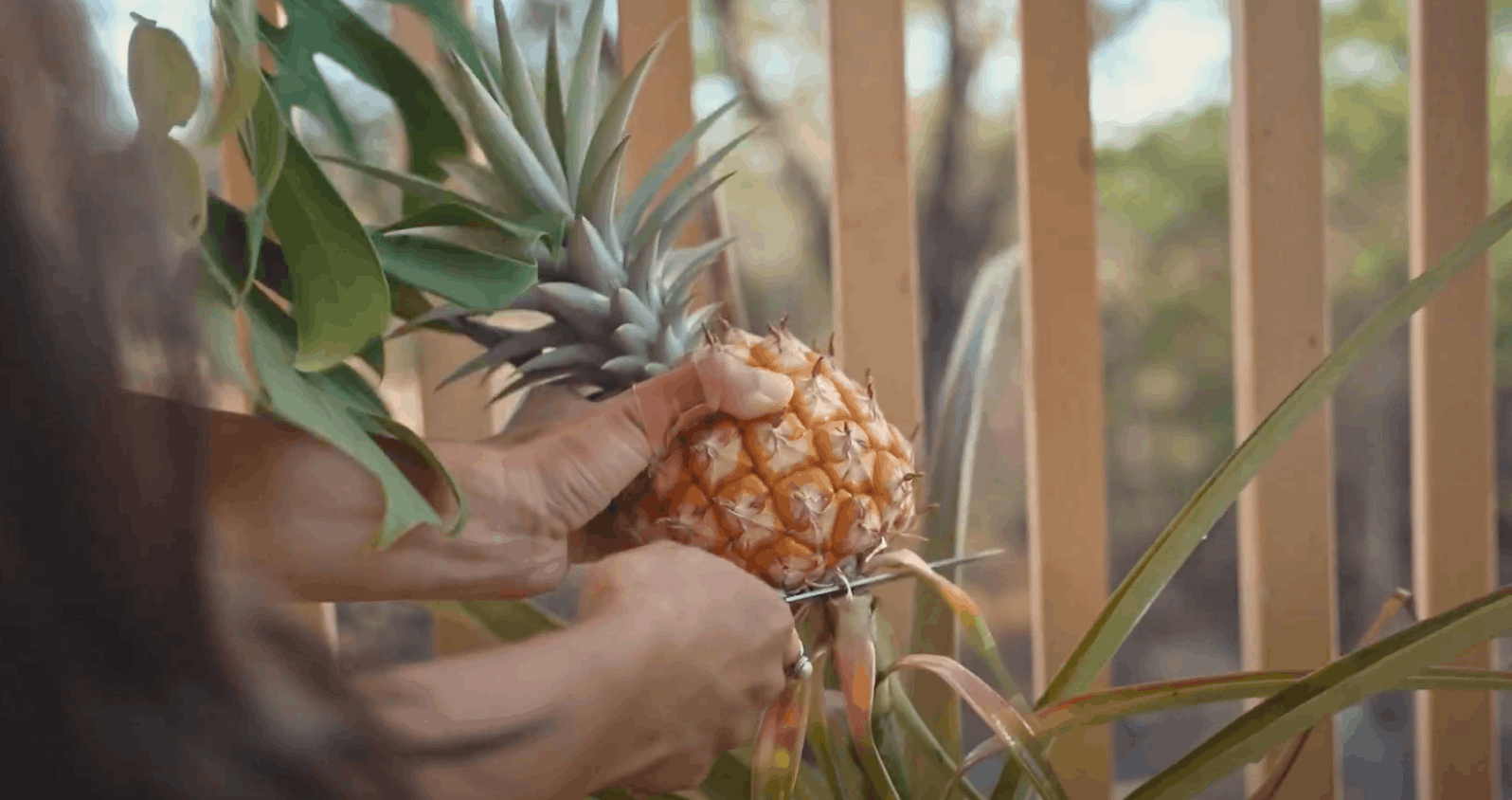} \includegraphics[width=0.07\linewidth]{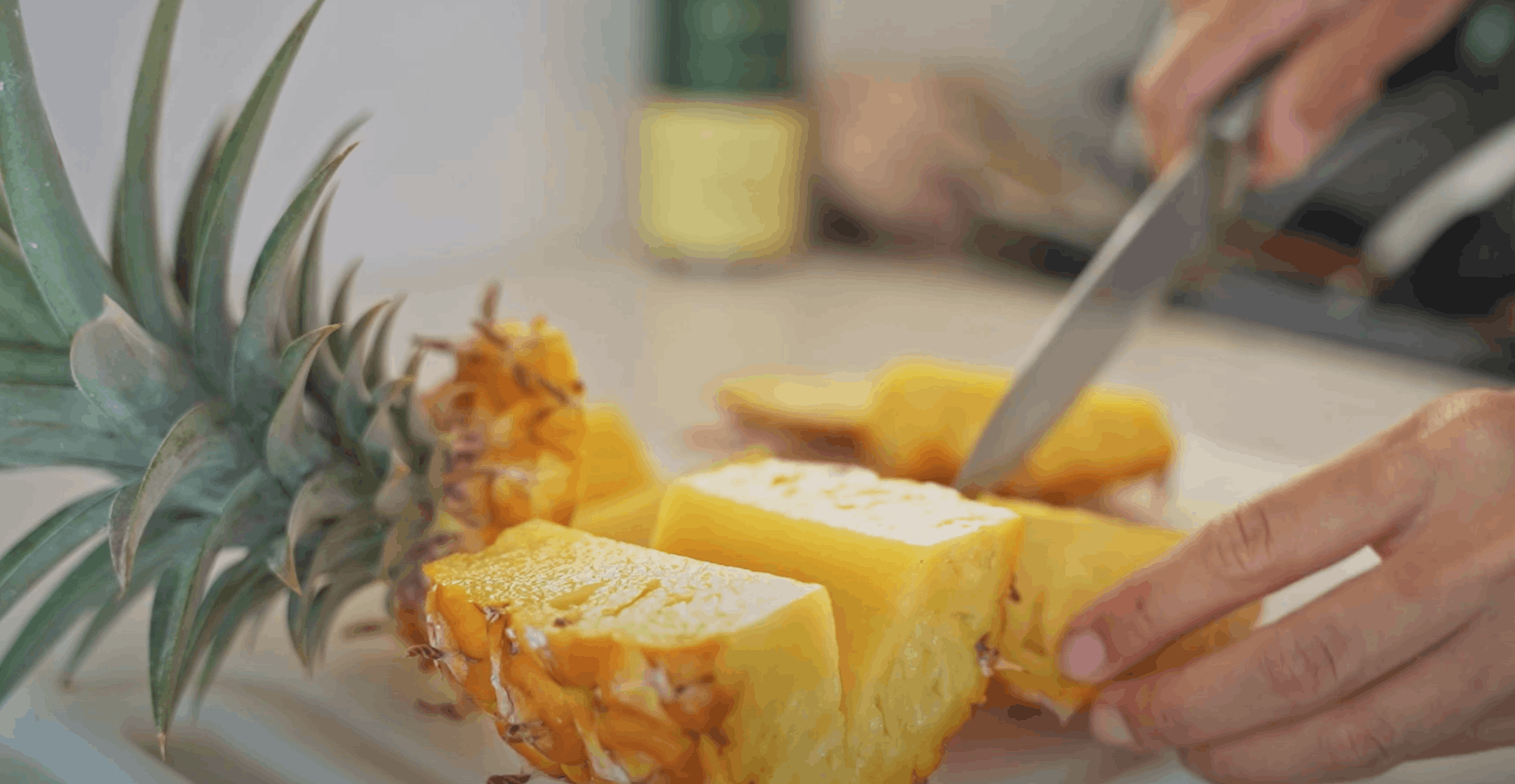}
\includegraphics[width=0.07\linewidth]{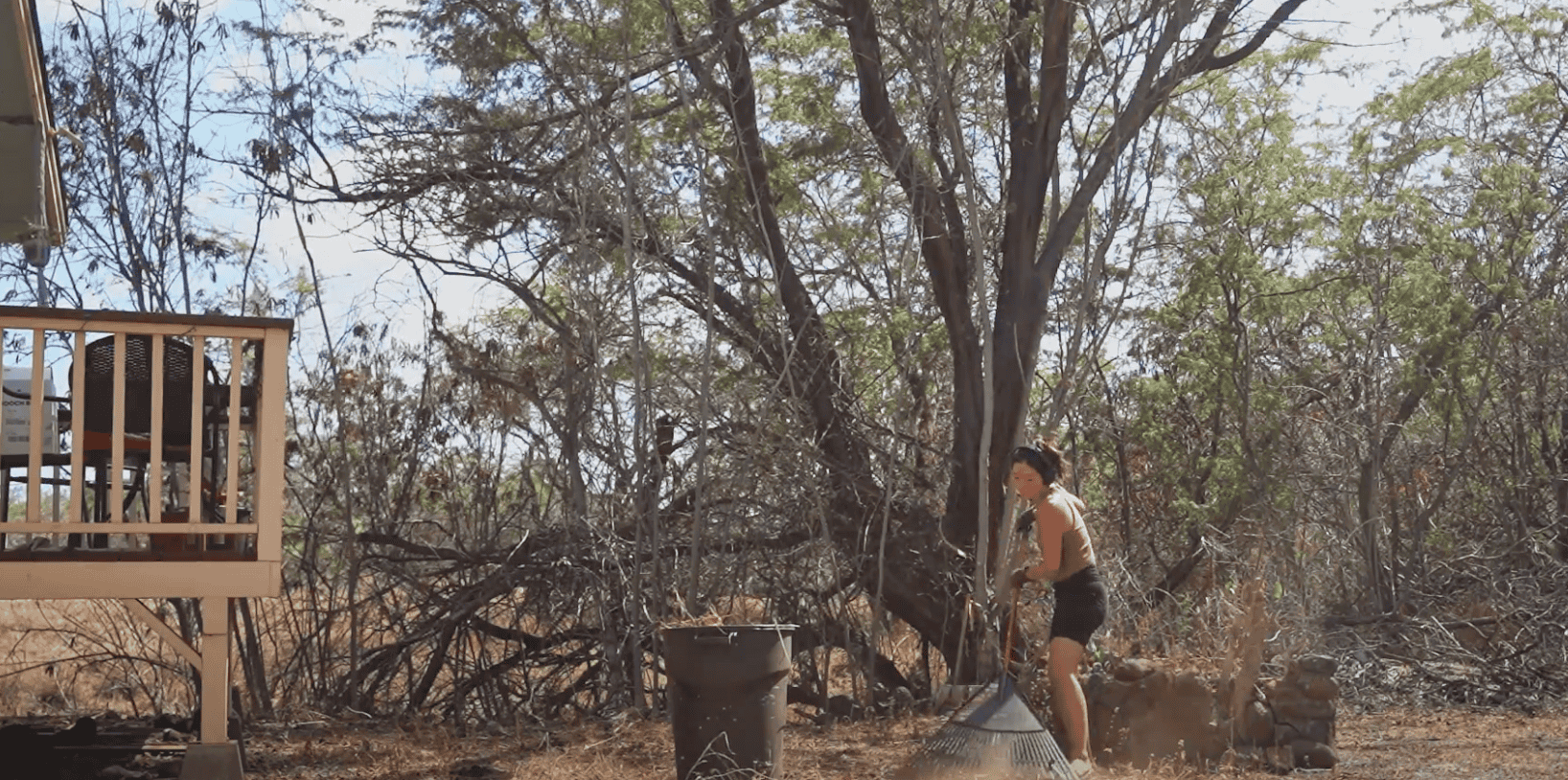} \includegraphics[width=0.07\linewidth]{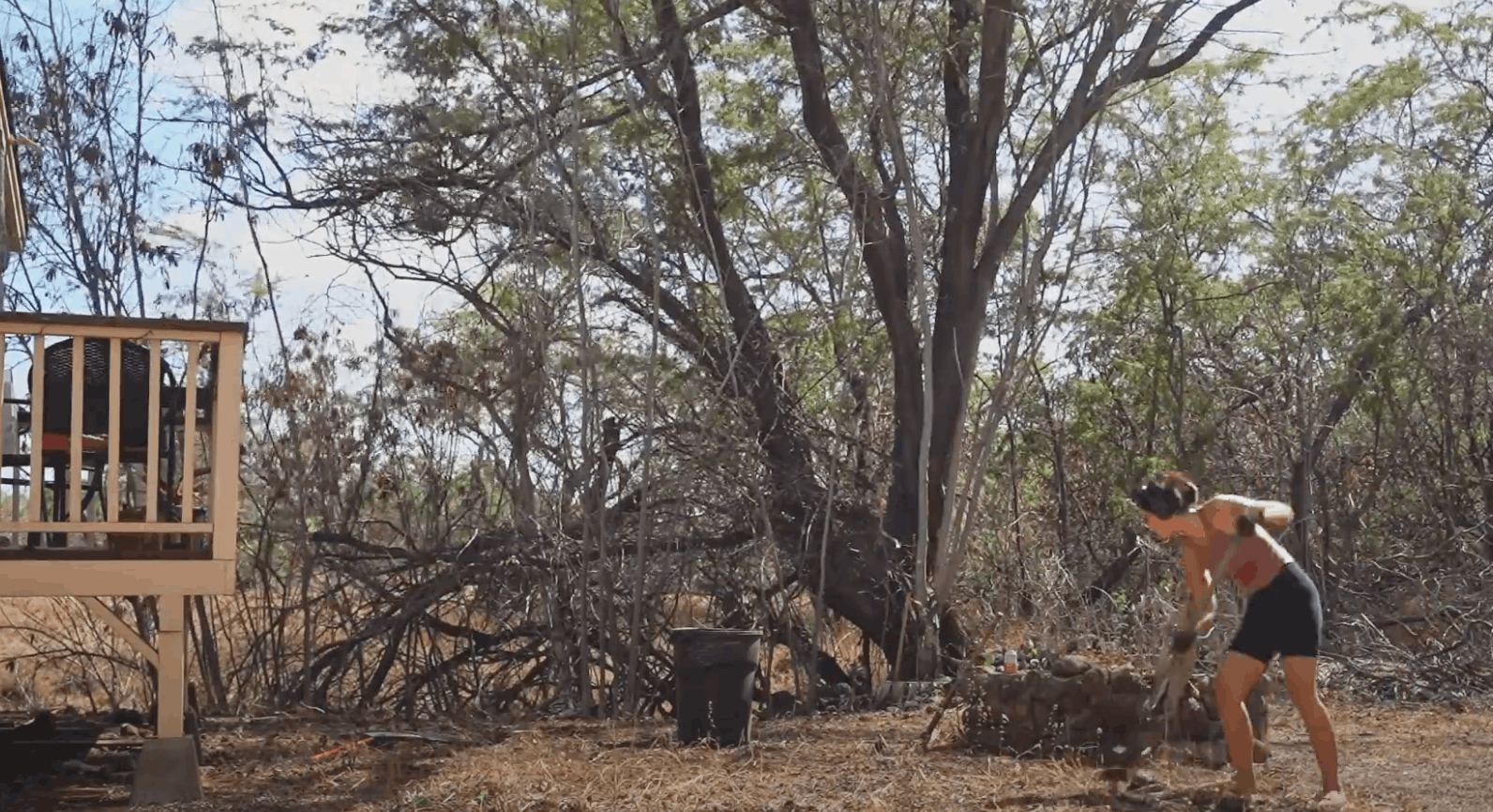}}             & DC                                           & Can you give me a breakdown of the occurrences at different timestamps in the video?                     & woman in long white dress walking up a hillside path, from 1 to 4. a woman sitting on the beach with long hair, from 5 to 8. [...] \\ \cline{2-4}
\hline
\end{tabular}}
\end{table}}

Based on the results presented in Table \ref{tab:step2_temporal_oriented_training_7b} and \ref{tab:step2_temporal_oriented_training_13b}, we observe that all temporal-oriented training schemes enhance the temporal understanding capabilities of LVLMs. Notably, the 13B-LVLM shows a more pronounced improvement, particularly when trained with the aggregated scheme VC+MC+MG+DC. This indicates significant untapped potential for even larger LVLMs, especially those exceeeding the 20B parameter scale. Due to computational constraints, we leave the exploration of such large-scale models to future work. 

\textit{Takeaway 2: For the remaining experiments, we add an additional temporal-oriented training stage and use VC, MC, MG, and DC as training schemes.}

\subsection*{Step 3: Memory Bank for Video Representations}
{\renewcommand{\arraystretch}{1.2}
\begin{table}[t]
\centering
\caption{Effect of Memory Bank on 7B-LVLM}
\label{tab:step3_memory_bank_7b}
\resizebox{0.75\linewidth}{!}{
\begin{tabular}{l|c|c|c|c|c|c}
\toprule
\rowcolor{HeaderBlue}
\textbf{Memory Bank ($B$)} & \textbf{MSRVTT} & \textbf{MSVD} & \textbf{ActivityNet-QA} & \textbf{Breakfast} & \textbf{COIN} & \textbf{LVU} \\ 
\midrule
$B = 0$  & 54.5 & 66.4 & 51.4 & 93.7 & 94.1 & 65.5 \\
$B = 10$ & 56.2 & 68.2 & 51.9 & 93.9 & 94.1 & 66.0 \\
$B = 20$ & 60.7 & 72.5 & 52.5 & 94.3 & 94.7 & 68.1 \\
$B = 30$ & 60.7 & 72.5 & 52.5 & 94.7 & 94.6 & 68.1 \\
$B = 40$ & 60.7 & 72.6 & 52.5 & 94.9 & 94.8 & 68.2 \\
$B = 50$ & 60.8 & 72.6 & 52.6 & 94.9 & 94.9 & 68.2 \\
\rowcolor{BestRow}
$B = 60$ & \textbf{60.8} & \textbf{72.6} & \textbf{52.6} & \textbf{95.0} & \textbf{95.0} & \textbf{68.3} \\ 
\bottomrule
\end{tabular}}
\end{table}}

{\renewcommand{\arraystretch}{1.2}
\begin{table}[t]
\centering
\caption{Effect of Memory Bank on 13B-LVLM}
\label{tab:step3_memory_bank_13b}
\resizebox{0.75\linewidth}{!}{
\begin{tabular}{l|c|c|c|c|c|c}
\toprule
\rowcolor{HeaderBlue}
\textbf{Memory Bank ($B$)} & \textbf{MSRVTT} & \textbf{MSVD} & \textbf{ActivityNet-QA} & \textbf{Breakfast} & \textbf{COIN} & \textbf{LVU} \\ 
\midrule
$B = 0$  & 62.7 & 75.8 & 54.5 & 95.1 & 94.7 & 72.8 \\
$B = 10$ & 63.3 & 75.9 & 54.9 & 95.3 & 94.9 & 73.1 \\
$B = 20$ & 63.5 & 76.3 & 55.2 & 96.0 & 95.0 & 73.4 \\
$B = 30$ & 63.5 & 76.3 & 55.4 & 96.5 & 95.6 & 73.9 \\
\rowcolor{BestRow}
$B = 40$ & \textbf{64.4} & \textbf{77.5} & \textbf{55.8} & \textbf{97.6} & \textbf{96.8} & \textbf{74.7} \\
$B = 50$ & 63.9 & 77.0 & 55.6 & 96.8 & 96.0 & 74.6 \\
$B = 60$ & 63.5 & 76.5 & 55.5 & 96.7 & 96.0 & 74.3 \\ 
\bottomrule
\end{tabular}}
\end{table}}

\noindent Building upon the model developed in Step 2, we further investigate how the LVLM processes video inputs. A straightforward approach involves encoding visual frames or patches and concatenating their representations along the temporal axis. However, the limited context length limit of the LVLM, coupled with GPU memory constraints, restricts the number of video frames that can be processed simultaneously. An alternative strategy is to apply temporal pooling \citep{maaz2023video, xu2024pllava}, but as demonstrated in our Step 1 analysis, this leads to suboptimal performance. Instead, we propose a different approach, \textit{i.e.} processing video frames sequentially and storing their features in a memory bank. We conduct an ablation study on the size of the memory bank $B \in \{10, 20, 30, 40, 50, 60\}$ and present our findings in Table \ref{tab:step3_memory_bank_7b} and \ref{tab:step3_memory_bank_13b}.

Based on these results, we find that incorporating a memory bank is an effective strategy, consistently outperforming standard pooling methods. Randomly sampling a fixed number of frames also proves suboptimal, particularly for long-term temporal understanding, as the sampled frames might fail to capture critical video context. Lastly, we note that increasing the memory bank size yields more significant improvement for the 13B-LVLM than the 7B-LVLM. This indicates that larger-scale models possess greater capacity to absorb and utilize richer video information. 

\textit{Takeaway 3: For our remaining experiments, we add a memory bank for video encoding.}

\subsection*{Step 4: Mixture-of-Experts for Q-Former}
\noindent Building upon Step 3, we next explore strategies to enhance the capacity of the vision-language interface, which plays a critical role in conveying video information to the LLM. Given that naively adding randomly initialized layers tends to yield suboptimal performance---as demonstrated in Step 1---we turn to the mixture-of-experts (MoE) approach. An MoE module consists of a router and a set of experts, where each expert is a feedforward network. The router typically comprises a linear projection followed by a gating function, \textit{e.g.} ReLU or Softmax, to compute the probabilities for routing a query token to specific experts. When a token encounters the MoE, the router selects a subset of experts to process the token, and their outputs are combined additively. This technique allows us to expand the parameter capacity of the Q-Former while keeping computational cost and latency manageable, as the model activates only a fraction of the total parameters for each token.

The exploration of MoE has remained scarce for LVLMs, especially for the vision-language interface, even though it has been investigated extensively in LLMs \citep{cai2024survey}. In our work, we will experiment with the following categories of MoE:
\begin{itemize}
    \item \textbf{Dense MoE:} the dense MoE activates all expert networks during each iteration. Based on the probability that the router produces for each expert, the outputs for an input token will be aggregated accordingly. 

    \item \textbf{Sparse MoE:} to reduce computational overhead, we can activate only a subset of experts during each forward pass. To achieve this sparsity, we can compute a weighted sum of the expert outputs from only the top-$k$ experts, rather than combining the outputs from all experts. In our work, we experiment with top-$k$ where $k = 1$ or $k = 2$.
\end{itemize}

For each type of MoE, we ablate the number of experts $E \in \{2, 4, 8\}$. In addition to Q-Former, we also add MoE to LLM to comprehensively study its effect on the LVLM.

{\renewcommand{\arraystretch}{1.2}
\begin{table}[t]
\centering
\caption{Effect of Mixture-of-Experts (MoE) on 7B-LVLM}
\label{tab:step4_moe_lvlm_7b}
\resizebox{0.9\linewidth}{!}{
\begin{tabular}{llc|ccccccc}
\toprule
\rowcolor{HeaderBlue}
\textbf{Position of MoE} & \multicolumn{1}{c}{\textbf{Type}} & \textbf{Experts ($E$)} & \textbf{MSRVTT} & \textbf{MSVD} & \textbf{ActivityNet-QA} & \textbf{Breakfast} & \textbf{COIN} & \textbf{LVU} \\ 
\midrule
Q-Former & Sparse & 2 & 63.9 & 76.9 & 56.3 & 95.6 & 95.4 & 72.9 \\
         &        & 4 & 64.1 & 77.6 & 56.7 & 96.3 & 96.1 & 73.7 \\
         &        & 8 & \textbf{65.0} & \textbf{78.1} & \textbf{57.3} & \textbf{97.0} & \textbf{96.6} & \textbf{74.5} \\
\rowcolor{GrayRow}
         & Dense  & 2 & 63.9 & 76.4 & 55.7 & 95.2 & 94.9 & 72.3 \\
\rowcolor{GrayRow}
         &        & 4 & 64.4 & 76.8 & 56.6 & 96.0 & 95.8 & 73.0 \\
\rowcolor{GrayRow}
         &        & 8 & 64.9 & 77.3 & 57.0 & 96.1 & 96.2 & 73.4 \\
\midrule
LLM      & Sparse & 2 & 54.9 & 72.9 & 55.4 & 93.4 & 92.0 & 72.5 \\
         &        & 4 & 54.9 & 73.5 & 55.5 & 93.6 & 92.2 & 72.7 \\
         &        & 8 & 55.4 & 73.6 & 55.9 & 93.6 & 92.5 & 72.8 \\
\rowcolor{GrayRow}
         & Dense  & 2 & 54.5 & 72.4 & 54.9 & 93.1 & 91.7 & 72.4 \\
\rowcolor{GrayRow}
         &        & 4 & 54.8 & 72.7 & 55.2 & 93.4 & 91.8 & 72.6 \\
\rowcolor{GrayRow}
         &        & 8 & 54.9 & 73.1 & 55.6 & 93.7 & 92.0 & 72.9 \\
\bottomrule
\end{tabular}}
\end{table}}

{\renewcommand{\arraystretch}{1.2}
\begin{table}[h!]
\centering
\caption{Effect of Mixture-of-Experts (MoE) on 13B-LVLM}
\label{tab:step4_moe_lvlm_13b}
\resizebox{0.85\linewidth}{!}{
\begin{tabular}{llc|ccccccc}
\toprule
\rowcolor{HeaderBlue}
\textbf{Position of MoE} & \multicolumn{1}{c}{\textbf{Type}} & \textbf{Experts ($E$)} & \textbf{MSRVTT} & \textbf{MSVD} & \textbf{ActivityNet-QA} & \textbf{Breakfast} & \textbf{COIN} & \textbf{LVU} \\ 
\midrule
Q-Former & Sparse & 2 & 65.2 & 78.4 & 57.2 & 97.6 & 97.5 & 75.5 \\
         &        & 4 & 65.6 & 78.8 & 57.6 & 97.8 & 97.7 & 75.9 \\
         &        & 8 & \textbf{66.7} & \textbf{79.5} & \textbf{58.3} & \textbf{98.5} & \textbf{97.8} & \textbf{76.1} \\
\rowcolor{GrayRow}
         & Dense  & 2 & 65.0 & 77.3 & 56.1 & 96.4 & 96.5 & 74.7 \\
\rowcolor{GrayRow}
         &        & 4 & 65.5 & 77.7 & 56.8 & 96.6 & 96.8 & 74.9 \\
\rowcolor{GrayRow}
         &        & 8 & 66.2 & 78.3 & 57.3 & 97.6 & 97.0 & 75.7 \\
\midrule
LLM      & Sparse & 2 & 59.3 & 73.3 & 53.4 & 91.9 & 92.9 & 69.1 \\
         &        & 4 & 59.8 & 73.6 & 53.7 & 93.1 & 93.6 & 69.2 \\
         &        & 8 & 60.2 & 73.7 & 54.1 & 93.2 & 94.2 & 69.6 \\
\rowcolor{GrayRow}
         & Dense  & 2 & 58.7 & 72.7 & 52.9 & 92.2 & 92.3 & 71.9 \\
\rowcolor{GrayRow}
         &        & 4 & 59.0 & 72.9 & 53.0 & 92.3 & 92.5 & 70.0 \\
\rowcolor{GrayRow}
         &        & 8 & 59.4 & 73.7 & 53.9 & 92.3 & 93.1 & 70.4 \\
\bottomrule
\end{tabular}}
\end{table}}

Based on the results in Table \ref{tab:step4_moe_lvlm_7b} and \ref{tab:step4_moe_lvlm_13b}, we observe that integrating MoE into Q-Former leads to a substantial boost in video understanding performance. Moreover, we note that both sparse and dense MoE categories bring improvement, with sparse MoE being slightly more effective. We hypothesize that sparse MoE provides a higher degree of specialization for LVLM to handle specific types of temporal circumstances. Perhaps unsurprisingly, scaling up MoE with more experts puts more significant impact to the 13B-LVLM than the 7B-LVLM, which implies further potential for LVLM in the upscaling direction. On the other hand, adding MoE to LLM degrades the performance. This indicates that MoEs might tamper with the pre-trained knowledge in LLM.

\textit{Final takeway: Our final scaled-up temporal-oriented LVLM improves the initial LVLM baseline by 10.1\% and 5.1\% in terms of the 7B and 13B variant, respectively.}
\section{Experimental Results}
\label{sect:experimental_results}
\noindent We validate our temporal-oriented recipe on two popular video understanding tasks, \textit{i.e.} video question answering and video captioning. All of our experiments are conducted using 8 H100 GPUs. 

\noindent\textbf{Video question answering.} We compare our results with existing methods on six datasets MSRVTT \citep{xu2016msr}, MSVD \citep{chen2011collecting}, ActivityNet-QA \citep{krishna2017dense}, Breakfast \citep{kuehne2014language}, COIN \citep{tang2019coin}, and LVU \citep{wu2021towards} in Table \ref{tab:experimental_results}. Our method substantially outperforms previous approaches on multiple datasets, achieving average accuracies of 66.7\% (\textbf{+18.2\%}), 79.5\% (\textbf{+18.9\%}), 58.3\% (\textbf{+8.5\%}), 98.5\% (\textbf{+5.5\%}), 97.8\% (\textbf{+4.6\%}), and 76.1\% (\textbf{+13.1\%}) on MSRVTT, MSVD, ActivityNet-QA, Breakfast, COIN, and LVU, respectively. 

\noindent\textbf{Video captioning.} We present our results for the video captioning task on MSRVTT \citep{xu2016msr} and MSVD \citep{chen2011collecting}. Our results indicate that we significantly improve upon previous works by a large margin. Particularly, we outperform existing methods by \textbf{20.8\%} on MSRVTT and \textbf{21.0\%} on MSVD.

{\renewcommand{\arraystretch}{1.2}
\begin{table}[t]
\centering
\caption{Comparison with existing methods on Video Question Answering (VideoQA) and Video Captioning tasks. The best results are in \textbf{bold}, and the second-best are \underline{underlined}.}
\label{tab:experimental_results}
\resizebox{0.85\linewidth}{!}{
\begin{tabular}{l|cccccc|cc}
\toprule
\rowcolor{HeaderBlue}
\multicolumn{1}{c|}{\textbf{Method}}  & \multicolumn{6}{c|}{\textbf{VideoQA}} & \multicolumn{2}{c}{\textbf{Video Captioning}} \\
\rowcolor{HeaderBlue}
& \textbf{MSRVTT} & \textbf{MSVD} & \textbf{ActivityNet-QA} & \textbf{Breakfast} & \textbf{COIN} & \textbf{LVU} & \textbf{MSRVTT} & \textbf{MSVD} \\ 
\midrule
MA-LMM \citep{he2024ma}             & 48.5 & 60.6 & 49.8 & 93.0 & 93.2 & 63.0 & 43.3 & 49.1 \\
VALLEY \citep{luo2023valley}        & 50.8 & 69.2 & 44.9 & 83.8 & 84.0 & 56.8 & 39.0 & 44.3 \\
LLaMA-VID \citep{li2024llama}       & 58.9 & 70.0 & 47.5 & 88.7 & 88.9 & 60.1 & 41.3 & 46.8 \\
VideoChat2 \citep{li2024mvbench}    & 54.1 & 70.0 & 49.1 & 91.7 & 91.9 & 62.1 & 42.7 & 48.4 \\
Video-ChatGPT \citep{maaz2023video} & 49.3 & 64.9 & 35.2 & 65.2 & 65.9 & 44.5 & 30.6 & 34.7 \\
Video-LLaVA \citep{lin2023video}    & 59.2 & 70.7 & 45.3 & 84.3 & 84.8 & 57.3 & 39.4 & 44.7 \\
GPT4Video \citep{wang2024gpt4video} & 49.8 & 66.3 & 48.7 & 90.9 & 91.1 & 61.6 & 42.3 & 48.0 \\
7B-PLLaVA \citep{xu2024pllava}      & 62.0 & 76.6 & 56.3 & 95.1 & 85.4 & 71.2 & 49.0 & 55.5 \\
13B-PLLaVA \citep{xu2024pllava}     & 63.2 & 75.7 & 56.3 & 95.4 & 86.7 & 72.9 & 49.5 & 58.6 \\
ST-LLM \citep{liu2024st}            & 63.2 & 74.6 & 50.9 & 95.1 & 95.3 & 64.4 & 44.3 & 50.2 \\
Chat-UniVi \citep{jin2024chat}      & 54.6 & 65.0 & 45.8 & 84.5 & 85.7 & 57.9 & 39.8 & 45.2 \\
\midrule
\underline{7B-LVLM (Ours)}          & \underline{65.0} & \underline{78.1} & \underline{57.3} & \underline{97.0} & \underline{96.6} & \underline{74.5} & \underline{50.9} & \underline{59.0} \\
\rowcolor{GrayRow}
\textbf{13B-LVLM (Ours)}            & \textbf{66.7} & \textbf{79.5} & \textbf{58.3} & \textbf{98.5} & \textbf{97.8} & \textbf{76.1} & \textbf{52.3} & \textbf{59.4} \\
\bottomrule
\end{tabular}}
\end{table}}
\section{Summary}
\noindent In this work, we highlight the critical role of temporal modeling in the design of modern Large Vision-Language Models (LVLMs). Throughout extensive investigation, we discover that key components, including query transformer (Q-Former), temporal-oriented training schemes, memory bank, and MoE augmentation for Q-Former, are pivotal for effective video understanding with LVLMs. Our empirical findings culminate in a step-by-step, temporal-oriented recipe for constructing effective temporal modeling capacity in LVLM. Compared with existing LVLMs, our proposed approach achieves superior performance across a broad range of standard video understanding datasets. Notably, the benefits of our recipe become more pronounced for larger-scale LVLMs, underscoring the potential of explicitly incorporating temporal modeling into large-scale architectures. 
\chapter{Conclusion}
\label{chapter:conclusion}
\bigskip
\noindent This chapter concludes the thesis and provides a retrospective summary of the proposed contributions. In this thesis, we investigate temporal perspectives to advance a series of video understanding tasks, including video question answering, text-video retrieval, temporal grounding, and temporal panoptic scene graph generation. With respect to these perspectives, we propose novel building blocks and training frameworks for pushing the model limit, along with achieving effective and efficient capture of temporal relations within videos. Last but not least, we proceed to elaborate possible future research directions.

\section{Overall Summary}
\noindent Video understanding plays an important role in many application domains, spanning from entertainment to security areas. Nevertheless, there has not been a full-fledged effort to comprehensively investigate model architecture, training strategy, and data aspects of video understanding. Therefore, \textbf{Chapter \ref{ch0:literature_review}} constructs a multi-faceted literature review to connect aspects of video understanding. Particularly, we summarize the key tasks and discuss common challenges, \textit{i.e.} intra-modal and cross-modal interaction, cross-domain adaptation, and data preparation. Then, we construct a lucid taxonomy to categorize existing works from three perspectives according to the three discussed challenges: model architecture, model training, and data perspective. 

Based on the insights from our literature review, we recognize the existing limit of the field, and proceed to push the limit by automatically annotating a vast set of videos in \textbf{Chapter \ref{ch1:mama}}. Our framework encompasses a dynamic regularization strategy to control model attention during training, thus mitigating the noisy effect of automatically annotated samples. We outperform previous state-of-the-art methods on standard video question answering and text-video retrieval benchmarks. 

Despite the effectiveness of our automatic annotating framework, in practice it might still be infeasible to obtain large-scale video training data, \textit{e.g.} in security- or privacy-sensitive domains. Hence, there is a need to devise an efficient training strategy to work productively in such low-resource scenarios. As such, \textbf{Chapter \ref{ch2:read}} introduces lightweight adapters with incorporated recurrent computation layers and only update them during fine-tuning. Through the use of video-language alignment objective, we motivate task-related information to flow through our adapter modules. Our framework significantly outperforms existing fine-tuning strategies on temporal grounding and video summarization tasks in low-resource settings. 

Then \textbf{Chapter \ref{ch3:global}} studies and presents three temporal inductive biases for video understanding, \textit{i.e.} global video semantics, motion-aware contrastive learning, and multi-scale contrastive learning. We show that global semantics over densely sampled video frames brings about significant benefits to long-term video question answering, which has recently suffered from information loss due to downsampling video frames to adapt to high computational cost. In addition to global semantics, \textbf{Chapter \ref{ch4:motionpvsg}} and \textbf{Chapter \ref{ch5:mstg}} demonstrate interrelated semantics among video elements, invoked by motion-aware and multi-scale contrastive learning, can advance temporal expressiveness in video representations. 

Finally, \textbf{Chapter \ref{ch6:recipe}} conduct an encyclopedic empirical study to demystify essential components for temporal understanding in large vision-language models (LVLMs). We discover that the vision-language interface that bridges visual encoder and large language model plays a fundamental role in encoding temporal inductive biases in LVLMs. Building on this insight, we propose a temporal-oriented recipe to progressively expand an LVLM through four key steps: (i) choosing a QueryFormer architecture for the vision-language interface, (ii) add temporal-oriented training schemes, (iii) integrate temporal memory bank, and (iv) inject mixtures-of-experts to upscale the interface. We contend that our recipe substantially advances a model to achieve impressive performance on various video understanding benchmarks.

\section{Future Directions and Challenges}
\noindent This section provides several promising future directions that extend the contributions of this thesis. Each direction is motivated by the overarching objective of advancing video understanding through more efficient (RO1), effective (RO2), and capacity-enhancing (RO3) approaches to capturing temporal relations. While there might be no principled solution to any of the following, many of these future directions remain as interesting goal posts for AI research in video understanding.

\subsection{Streaming Video Understanding (RO1, RO2)}
\noindent Most prior studies assume an offline setting, where the model has full access to a video. Streaming video understanding introduces unique challenges, \textit{i.e.} models must operate without access to future frames, make predictions in real-time, and adapt to dynamically arriving information. This setting directly tests a model’s efficiency (RO1) and effectiveness (RO2) in capturing temporal relations under resource and time constraints. Recent works have started pushing boundaries of video understanding in this streaming setting \citep{qian2024streaming, chen2024videollm}, but further advances are necessary to achieve robust, real-time temporal reasoning in practice.

\subsection{Multi-Video Reasoning and Understanding (RO2, RO3)}
\noindent While current video understanding systems typically operate on single video inputs, many real-world scenarios demand reasoning across videos \citep{panda2017diversity}. Extending temporal-oriented inductive biases to this setting could enable models to infer relations among movie episodes, cross-video narratives, or complementary sources of evidence. Although multi-scale contrastive learning already acknowledges relations among video clips, there remains significant potential for improvement, particularly when scaling to longer or interdependent clips. Specialized modules and training strategies for multi-video reasoning represent an open challenge with growing interest \citep{messaoud2021deepqamvs, ansari2023multi, zhu2024adastreamer}.

\subsection{Probing Models for Temporal Understanding (RO3)}
\noindent A key challenge in advancing video understanding lies in interpretability. Current video models remain opaque, leaving open questions about how they infer temporal relationships between objects, actions, and events. Developing probing tools to reveal whether models form coherent chronologies or align with human-like temporal intuition might be crucial for uncovering temporal reasoning capabilities of future models. Further progress will not only improve trustworthiness in safety-critical applications but also guide the design of models that can faithfully capture complex temporal dependencies.

\subsection{Cost-Accuracy Tradeoff for Video Understanding (RO1)}

\noindent Deploying video understanding models in resource-constrained environments raises the issue of balancing performance with computational cost. Models shall remain efficient enough to meet real-time demands without sacrificing accuracy. There is always a need to advance RO1, or to discover better Pareto frontier for cost-accuracy tradeoff of video understanding models. This entails designing parameter-efficient modules, leveraging dynamic inference, and rethinking training strategies to optimize both computational and temporal understanding capabilities.

\section{Reflections}
\noindent This thesis is the culmination of rigorous research, from constructive brainstorming to meticulous experimental design. Throughout this journey, I have learned countless lessons. Here, I would like to reflect on the most memorable insights in the hope that they can guide and inspire future researchers. 

\subsection{Choosing a Steady Research Direction is Important}
\noindent First, I want to emphasize the value of choosing a steady research direction to pursue. As more researchers enter the field and powerful large language models (LLMs) lower the barrier to publication, the publication task has become significantly less difficult than in the past. One interesting observation is the total number of submissions grows each year but the acceptance rate has remained stable. Consequently, more papers are accepted into top conferences. While finding the ``\textit{right}'' topic can indeed lead to a successful submission, this short-term strategy often yields a portfolio of disparate and shallow papers. Such breadth may sustain one's career in the near term, but it prevents the emergence of a signature research identity. 

Thanks to my supervisor's guidance, I have learned this lesson during my PhD. Although I developed a knack for identifying promising problems, I struggled to commit to a single, unifying theme. One of the reasons is my research domain, \textit{i.e.} video understanding, spans various subareas that one can easily find a straightforward but not fundamental problem to solve. Eventually, I recognized temporal relations---what truly distinguish video from images--- remained underexplored. By focusing on modeling these relations, I found a coherent thread that has guided my works and underpins robust video-understanding systems.

Based on my experience, I recommend that anyone serious about research invest time up front to fine a lasting, focused direction. This direction not only prevents one from drifting between unrelated projects, but also forms a cohesive system of contributions which promote their impact to the community.

\subsection{Scalability is important but we still need principled research}
\noindent Second, I would like to highlight a contentious trait of deep learning models, \textit{i.e.} scalability. Recently, deep learning models have achieved impressive performance with their increased model size or dataset scale \citep{hestness2017deep, hoffmann2022training}. However, not all of the scholars are attentive towards scalability-oriented research whose aim is to investigate effective methods to scale up model or data to push the performance boundary. Partly, this reflects a disparity in compute resources between industry and academic labs \citep{togelius2023choose}, which limits many researchers' ability to train larger models. Additionally, some scholars regard scalability as self-evident: if we enlarge the model or data, the performance will automatically follow, so they prefer to investigate more subtle, foundational principles of deep learning.

From my side and based on my experience, I would like not to defend or argue against either of these views. Indeed, Chapter \ref{ch1:mama} and Chapter \ref{ch6:recipe} demonstrate that at least for video understanding, naive scaling often falls short. Particularly, Chapter \ref{ch1:mama} shows how automatic data-scaling frameworks can introduce noise and degrade learning. To counteract this, I propose a subtractive angular margin contrastive learning approach, which we prove theoretically provides a regularizing effect that mitigates noisy effects. Overall, I believe future research should continue exploring scale while simultaneously developing principled methods to resolve fundamental weaknesses of upscaled frameworks. 

\subsection{Academia versus Industry: Does Industry have more resources than Academia in all cases?}

\noindent Finally, I would like to share some reflections on the differences between academia and industry. It is often said that the distinction between the two mirrors the gap between theory and practice. I believe that this distinction is multi-faceted, and different people from different fields can experience this in different ways. 

In recent years, tech heavyweights have grown immensely influential, aided by sophisticated marketing strategies that promote the perception of a significant resource gap between academic and industry research. These narrative often suggest that industry labs operate with minimal constraints. However, this is a misconception. In reality, industry also faces its own set of limitations. 

For example, cost is a persistent constraint in most companies. Unless you are a tech giant like OpenAI or Google, or a startup backed by a wealthy and generous angel investor, minimizing operational expenses is essential for long-term sustainability. In fact, even large companies are mindful of costs. It is said one team at Google runs a weekly cron job to track infrastructure spending per employee, ranks the results, and emails the top spenders weekly as a cost-awareness measure. During my time in industry, I also had to adapt to various constraints. On one occasion, I developed a clustering algorithm to re-identify individuals by comparing new inputs to a database. However, when the deployment phase began, we discovered that the backend team’s database library lacked support for the optimization algorithms my algorithm required. As a result, we had to fall back on a simpler approach based on naive vector similarity and thresholding. Experiences like these illustrate that industry is not exempt from limitations---they simply operate under a different set of them.

Now, I hope that it is clear both industry and academia operate under constraints. What I suggest that is we, as researchers, make a more conscious effort to identify and communicate these constraints early in the project lifecycle or during discussions. In university settings, I often notice an emphasis on ideas and contributions, with relatively less attention paid to practical considerations---such as available GPUs, storage capacity, or whether a proposed method can be implemented or published before a deadline. In this regard, I am especially grateful to Prof. See-Kiong and Prof. Anh Tuan, who have always been transparent about the resource status of their labs, including GPU acquisition plans and infrastructure updates. Such openness helps set realistic expectations and guides the direction of research more effectively.

Understanding constraints is crucial because they shape our mental models of what is feasible. They also help prevent misaligned assumptions---whether between students and supervisors, or between principal investigators and funding agencies. In fact, by being acutely aware of the data constraints in video understanding, I was able to propose the Meta-optimized Angular MArgin framework in Chapter \ref{ch1:mama} to elevate the data scale, and the recurrent adapter in Chapter \ref{ch2:read} tailored for low-resource video understanding. 

In conclusion, I recommend that future research discussions include a more deliberate and structured conversation about constraints and feasibility. Doing so will foster more grounded, efficient, and collaborative research practices.

\bibliographystyle{acl_natbib}
\bibliography{references}

\end{document}